\documentclass[10pt,final,journal,twocolumn,singlespaced]{IEEEtran}

\usepackage{algorithm}
\usepackage{algpseudocode}
\usepackage{amsmath}
\usepackage{amssymb}
\usepackage{bm}
\usepackage{booktabs}
\usepackage{color}
\usepackage{cite}
\usepackage{epsfig}
\usepackage{epstopdf}
\usepackage{subfigure}
\usepackage{colortbl}
\usepackage{arydshln}
\usepackage{float}
\usepackage{graphicx}
\usepackage{multirow}
\usepackage[square, comma, sort&compress, numbers]{natbib}
\usepackage{makecell} 
\usepackage{url}  
\usepackage{pifont}
\usepackage{flafter}
\usepackage{soul}
\usepackage{times}
\usepackage{threeparttable}
\usepackage{xcolor}

\newcommand{\argmin}{\mathop{\mathrm{arg\,min}}}

\begin{document}

\title{FastDeRain: A Novel Video Rain Streak Removal Method Using Directional Gradient Priors
 \thanks{\IEEEauthorrefmark{1} Corresponding authors. Tel.: +86 28 61831016.}
\thanks{T.-X Jiang, T.-Z. Huang, X.-L. Zhao and L.-J. Deng are with the School of Mathematical Sciences, University of Electronic Science and Technology of China, Chengdu, Sichuan 611731, P. R. China. Y. Wang is with the School of Mathematics and Statistics, Xian Jiaotong University, Xian 710049, P. R. China. E-mails: \{taixiangjiang, yao.s.wang\}@gmail.com, \{tingzhuhuang, liangjian1987112\}@126.com, xlzhao122003@163.com.}
}

\author{Tai-Xiang Jiang,
Ting-Zhu Huang\IEEEauthorrefmark{1},
Xi-Le Zhao\IEEEauthorrefmark{1},
Liang-Jian Deng
and Yao Wang}

\IEEEpeerreviewmaketitle
\maketitle

\begin{abstract}
   Rain streaks removal is an important issue in outdoor vision systems and has recently been investigated extensively.
   In this paper, we propose a novel video rain streak removal approach FastDeRain, which fully considers the discriminative characteristics of rain streaks and the clean video in the gradient domain.
   Specifically, on the one hand, rain streaks are sparse and smooth along the direction of the raindrops, whereas on the other hand, clean videos exhibit piecewise smoothness along the rain-perpendicular direction and continuity along the temporal direction.
   Theses smoothness and continuity results in the sparse distribution in the different directional gradient domain, respectively.
   Thus, we minimize 1) the $\ell_1$ norm to enhance the sparsity of the underlying rain streaks, 2) two $\ell_1$ norm of unidirectional Total Variation (TV) regularizers to guarantee the anisotropic spatial smoothness, and 3) an $\ell_1$ norm of the time-directional difference operator to characterize the temporal continuity.
   A split augmented Lagrangian shrinkage algorithm (SALSA) based algorithm is designed to solve the proposed minimization model.
   Experiments conducted on synthetic and real data demonstrate the effectiveness and efficiency of the proposed method.
   According to comprehensive quantitative performance measures, our approach outperforms other state-of-the-art methods, especially on account of the running time.
   The code of FastDeRain can be downloaded at \url{https://github.com/TaiXiangJiang/FastDeRain}.
\end{abstract}


\begin{IEEEkeywords}
video rain streak removal,
unidirectional total variation,
split augmented Lagrangian shrinkage algorithm (SALSA) .
\end{IEEEkeywords}

\section{Introduction}\label{sec:Intro}

\begin{figure}[!t]
 \begin{center}
  \includegraphics[width=0.32\linewidth]{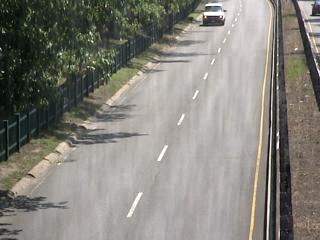}
  \includegraphics[width=0.32\linewidth]{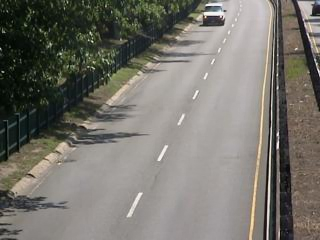}
  \includegraphics[width=0.32\linewidth]{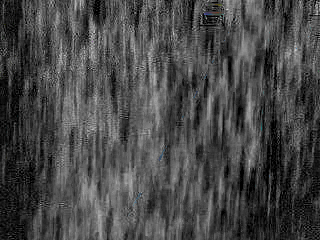}
 \end{center}
 \caption{A frame of a rainy video (left), the rain streaks removal result by the proposed method FastDeRain (middle) and the extracted rain streaks (right). The pixel values of the rain streaks are scaled for better visualization.}
 \label{Example}
\end{figure}

\IEEEPARstart{O}{utdoor} vision systems are frequently affected by bad weather conditions, one of which is the rain.
Raindrops usually introduce bright streaks into the acquired images or videos, because of their scattering of light into complementary metal--oxide--semiconductor cameras and their high velocities.
Moreover, rain streaks also interfere with nearby pixels because of their specular highlights, scattering, and blurring effects \cite{li2016rain}.
This undesirable interference will degrade the performance of various computer vision algorithms \cite{Bouwmans2014Traditional},
such as event detection \cite{shehata2008video}, object detection \cite{zhang2017bayesian}, tracking \cite{ma2017saliency}, recognition \cite{garg2007vision}, and scene analysis \cite{itti1998model}.
Therefore, the removal of rain streaks is an essential task, which has recently received considerable attention.

Numerous methods have been proposed to improve the visibility of images/videos captured with rain streak interference \cite{kang2012automatic,sun2014exploiting,chen2013generalized,chen2014rain,chen2014visual,luo2015removing,son2016rain,li2017single,Zhu_2017_ICCV,chen2017error,gu2017joint,chang2017transformed,deng2018directional,du2018single,wang2017hierarchical,ren2018simultaneous,eigen2013restoring,Yang_2017_CVPR,fu2017clearing,fu2017removing,zhang2017image,qian2017attentive,li2018fast,zhang2018density,
garg2004detection,tripathi2014removal,kim2015video,santhaseelan2015utilizing,you2016adherent,Jiang_2017_CVPR, ren2017video,Wei_2017_ICCV,li2018video,Chen_2018_CVPR,liu2018erase}.
They can be classified into two categories: multiple-images/videos based techniques and single-image based approaches.
Fig. \ref{Example} exhibits an example of video rain streaks removal.
Without loss of generality, in this paper, we use ``background'' to denote the rain-free content of the data.

For the single-image de-raining task, Kang {\em et al}. \cite{kang2012automatic} decomposed a rainy image into low-frequency (LF) and high-frequency (HF) components using a bilateral filter and then performed morphological component analysis (MCA)-based dictionary learning and sparse coding to separate the rain streaks in the HF component.
To alleviate the loss of the details when learning HF image bases, Sun {\em et al}. \cite{sun2014exploiting} tactfully exploited the structural similarity of the derived HF image bases.
Chen {\em et al}. \cite{chen2013generalized} considered the similar and repeated patterns of the rain streaks and the smoothness of the background.
Sparse coding and dictionary learning were adopted in \cite{chen2014visual,luo2015removing,son2016rain}. In their results, the details of backgrounds were well preserved.
The recent work by Li {\em et al}. \cite{li2017single} was the first to utilize Gaussian mixture model (GMM) patch priors for rain streak removal, with the ability to account for rain streaks of different orientations and scales.
Zhu {\em et al}. \cite{Zhu_2017_ICCV} proposed a joint bi-layer optimization method progressively separate rain streaks from background details, in which the gradient statistics are analyzed.
Meanwhile, the directional property of rain streaks received a lot of attention in \cite{chang2017transformed,deng2018directional,du2018single} and these methods achieved promising performances.
Ren {\em et al.} \cite{ren2018simultaneous} removed the rain streaks from the image recovery perspective.
Wang {\em et al.} \cite{wang2017hierarchical} took advantage the image decomposition and dictionary learning.
The recently developed deep learning technique was also applied to the single image rain streaks removal task, and excellent results were obtained \cite{eigen2013restoring,Yang_2017_CVPR,fu2017clearing,fu2017removing,zhang2017image,qian2017attentive,li2018fast,zhang2018density}.

For the video rain streaks removal, Garg {\em et al}. \cite{garg2004detection} firstly raised a video rain streaks removal method with comprehensive analysis of the visual effects of the rain on an imaging system.
Since then, many approaches have been proposed for the video rain streaks task and obtained good rain removing performance in videos with different rain circumstances.
Comprehensive early existing video-based methods are summarized in \cite{tripathi2014removal}.
Chen {\em et al.} \cite{chen2014rain} took account of the highly dynamic scenes.
Whereafter, Kim {\em et al}. \cite{kim2015video} considered the temporal correlation of rain streaks and the low-rank nature of clean videos. 
Santhaseelan {\em et al.} \cite{santhaseelan2015utilizing} detected and removed the rain streaks based on phase congruency features.
You {\em et al.} \cite{you2016adherent} dealt with the situations where the raindrops are adhered to the windscreen or the window glass.
In \cite{Jiang_2017_CVPR}, a novel tensor-based video rain streak removal approach was proposed considering the directional property.
Ren {\em et al.} \cite{ren2017video} handled the video desnowing and deraining task based on matrix decomposition.
The rain streaks and the clean background were stochastically modeled as a mixture of Gaussians by Wei {\em et al.} \cite{Wei_2017_ICCV} while
Li {\em et al.} \cite{li2018video} utilized the multiscale convolutional sparse coding.
For the video rain streaks removal, the deep learning based methods also started to reveal their effectiveness \cite{Chen_2018_CVPR,liu2018erase}.

In general, the observation model for a rainy image is formulated as $\mathbf O= \mathbf B + \mathbf R$ \cite{li2016rain}, which can be generalized to the video case as: $\mathbf{\mathcal{O}}=\mathbf{\mathcal{B}}+\mathbf{\mathcal{R}}$, where $\mathbf{\mathcal{O}}$, $\mathbf{\mathcal{B}},\text{ and } \mathbf{\mathcal{R}}\in \mathbb{R}^{m\times n\times t}$ are three 3-mode tensors representing the observed rainy video, the unknown rain-free video and the rain streaks, respectively.
When considering the noise or error, the observation model is modified as $\mathbf{\mathcal{O}}=\mathbf{\mathcal{B}}+\mathbf{\mathcal{R}}+\mathbf{\mathcal{N}}$, where $\mathbf{\mathcal{N}}$ is the noise or error term.
The goal of video rain streak removal is to distinguish the clean video $\mathbf{\mathcal{B}}$ and the rain streaks $\mathbf{\mathcal{R}}$ from an input rainy video $\mathbf{\mathcal{O}}$.
This is an ill-posed inverse problem, which can be handled by imposing prior information.
Therefore, from this point of view, the most significant issues are the rational extraction and sufficient utilization of the prior knowledge, which is helpful to wipe off the rain streaks and reconstruct the rain-free video.
In this paper, we mainly focus on the discriminative characteristics of rain streaks and background in different directional gradient domains.

\begin{figure}[!htbp]
 \begin{center}
 \includegraphics[width=0.98\linewidth]{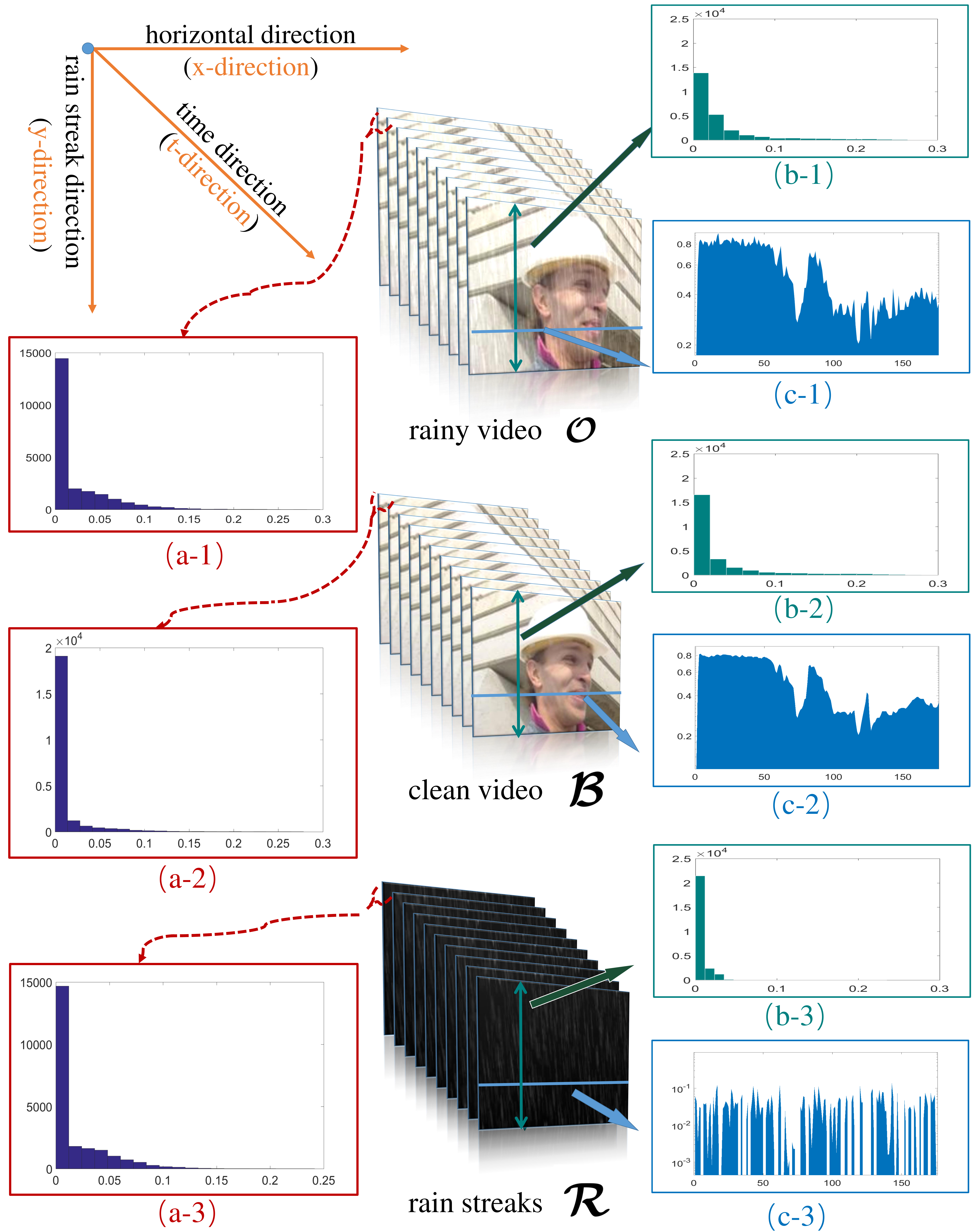}
 \end{center}
 \caption{From left to right: the histograms of temporal gradient of the rainy video (a-1), the clean video (a-2) and the isolated rain streaks (a-3), respectively;  several example frames from the rainy video, the clean video and the isolated rain streaks; and the histograms of the vertical gradient (b-1,2,3) and the intensities along a row (c-1,2,3) in the rainy video, the clean video and the isolated rain streaks, respectively.}
 \label{Motivation}
\end{figure}

From the temporal perspective, the clean video is continuous along the time direction, while the rain streaks do not share this property \cite{kim2015video,starik2003simulation,Wei_2017_ICCV}.
As observed in Fig. \ref{Motivation}, the time-directional gradient of the rain-free video (a-2) exhibits a different histogram compared with those of the rainy video (a-1) and the rain streaks (a-3).
The temporal gradient of the clean video is much sparser and it is corresponding to the temporal continuity of the clean video.
Therefore, we intend to minimize $\|\nabla_t\mathcal{B}\|_1$, where $\nabla_t$ is the temporal differential operator.

From the spatial perspective, it has been widely recognized that natural images are largely piecewise smooth and their gradient
fields are typically sparse \cite{guo2015generalized,jiang2016video}.
Many aforementioned de-rain methods take the spatial gradient into consideration and use the total variation (TV) to depict the property of the rain-free part \cite{li2016rain,chen2013generalized}.
However, the effects of the rain streaks on the vertical gradient and horizontal gradient are different. This phenomenon was likewise noticed in \cite{chang2017transformed,deng2018directional,du2018single}.
Initially, for the sake of convenience, we assume that rain streaks are approximately vertical.
The impact of the vertical rain streaks on the vertical gradient is limited.
The subfigures (b-1,2,3) in Fig. \ref{Motivation} reveal that the vertical gradient of rain streaks are much sparser than those of the clean video and the rainy video.
Nonetheless, the vertical rain streaks severely disrupt the horizontal piecewise smoothness.
As exhibited in Fig. \ref{Motivation} (c-1,2,3), the pixel intensity is piecewise smooth only in (c-2), whereas burrs frequently appear in (c-1) and (c-3).
Therefore, we intend to minimize $\|\nabla_1\mathcal{R}\|_1$ and $\|\nabla_2\mathcal{B}\|_1$, where $\nabla_1$ and $\nabla_2$ are respectively the vertical difference (or say vertical unidirectional TV \cite{chang2014simultaneous,chang2016remote,dou2018directional}) operator and horizontal difference (or say horizontal unidirectional TV) operator.

Given a real rainfall-affected scene, without the wind, the raindrops generally fall from top to bottom.
Meanwhile, when not very windy, the angles between rain streaks and the vertical direction are usually not very large.
Therefore, the rain streak direction can be approximated as the vertical direction, {\em i.e.} the mode-1 (column) direction of the video tensor.
Actually, this assumption is reasonable for parts of the rainy sceneries.
For the rain streaks that are oblique (or say far from being vertical), directly utilizing the directional property is very difficult for the digital video data, which are cubes of distinct numbers.
To cope with this difficulty, in Sec. \ref{strategies}, we would design the shift strategy, based on our automatical rain streaks' direction detection method.

The contributions of this paper include three aspects.
 \begin{itemize}
 \item  We propose a video rain streaks removal model, which fully considers the discriminative prior knowledge of the rain streaks and the clean video.
\item  We design a split augmented Lagrangian shrinkage algorithm (SALSA) based algorithm to efficiently and effectively solve the proposed minimization model. The convergence of our algorithm is theoretically guaranteed. Meanwhile, the implementation on the graphics processing unit (GPU) device further accelerates our method.
\item To demonstrate the efficacy and the superior performance of the proposed algorithm in comparison with state-of-the-art alternatives, extensive experiments both on the synthetic data and the real-world rainy videos are conducted.
\end{itemize}

This work is an extension of the material published in \cite{Jiang_2017_CVPR}.
The new material is the following:
a) the proposed rain streaks removal model is improved 
and herein introduced in more technical details;
b) we explicitly use the split augmented Lagrangian shrinkage algorithm to solve the proposed model;
c) to make the proposed method more applicable, we design an automatical rain streaks' direction detecting method and provide the shift strategy to deal with oblique rain streaks;
d) in our experiments, we re-simulate the rain streaks for the synthetic data, using two different techniques and considering the rain streaks not very vertical;
e) three recent state-of-the-art methods \cite{fu2017removing,Wei_2017_ICCV,li2018video} are brought into comparison.

The paper organized as follows.
Section \ref{sec:Pre} gives the preliminary on the tensor notations.
In Section \ref{sec:Mod}, the formulation of our model is presented along with a SALSA solver.
Experimental results are reported in Section \ref{sec:Exp}.
Finally, we draw some conclusions in Section \ref{sec:Con}.

\section{Notation and preliminaries}\label{sec:Pre}
\begin{table}[htbp]
\renewcommand\arraystretch{1.5}
\caption{Tensor notations}
 \begin{tabular}{c p{0.66\columnwidth}}
  \toprule
Notation &  Explanation \\
  \midrule
$\mathbf{\mathcal{X}},\mathbf{X},\mathbf{x},x$
                                & Tensor, matrix, vector, scalar.\\
\multirow{2}{*}{$\mathbf{x}(:i_2i_3\cdots i_N)$}
                                & A \textbf{fiber} of a tensor $\mathbf{\mathcal{X}}$, defined by fixing every index but one.\\
\multirow{2}{*}{$\mathbf{X}(::i_3\cdots i_N)$}
                                & A \textbf{slice} of a tensor $\mathbf{\mathcal{X}}$, defined by fixing all but two indices.\\
\multirow{2}{*}{$\langle\mathbf{\mathcal{X}},\mathbf{\mathcal{Y}}\rangle$}
                                & The \textbf{inner product} of two same-sized tensors $\mathbf{\mathcal{X}}$ and $\mathbf{\mathcal{Y}}$.\\
$\left\|\mathbf{\mathcal{X}}\right\|_{F}$
                                & The \textbf{Frobenius norm} of a tensor $\mathbf{\mathcal{X}}$.\\
\bottomrule
 \end{tabular}
 \end{table}
Following \cite{jiang2018matrix,li2018fusing,ji2018nonlocal}, we use lower-case letters for vectors, e.g., $\mathbf a$;
upper-case letters for matrices, e.g., $\mathbf A$; and calligraphic letters for tensors, e.g., $\mathbf{\mathcal{A}}$.
An $N$-mode tensor is defined as $\mathbf{\mathcal{X}}\in \mathbb{R}^{I_{1}\times I_2\times\dots\times I_{N}}$, and $x_{i_{1},i_2,\cdots, i_{N}}$ denotes its $(i_{1},i_2,\cdots,i_{N})$-th component.


A \textbf{fiber} of a tensor is defined by fixing every index but one.
A third-order tensor has column, row, and tube fibers, denoted by $\mathbf x_{:jk}$, $\mathbf x_{i:k}$, and $\mathbf x_{ij:}$, respectively.
When extracted from their tensors, fibers are always assumed to be oriented as column vectors.
%

A \textbf{slice} is a two-dimensional section of a tensor, defined by fixing all but two indices.
The horizontal, lateral, and frontal slides of a third-order tensor $\mathbf{\mathcal{X}}$ are denoted by $\mathbf X_{i::}$, $\mathbf X_{:j:}$, and $\mathbf X_{::k}$, respectively.
Alternatively, the $k$-th frontal slice of a third-order tensor, $\mathbf X_{::k}$, may be denoted more compactly by $\mathbf X_k$.


The \textbf{inner product} of two same-sized tensors $\mathbf{\mathcal{X}}$ and $\mathbf{\mathcal{Y}}$ is defined as $\langle\mathbf{\mathcal{X}},\mathbf{\mathcal{Y}}\rangle:=\sum\limits_{i_{1},i_{2},\cdots,i_{N}}x_{i_{1}i_2\cdots i_{N}}\cdot y_{i_{1}i_2\cdots i_{N}}$.
The corresponding norm (\textbf{Frobenius norm}) is then defined as $\left\|\mathbf{\mathcal{X}}\right\|_{F}:=\sqrt{\langle\mathbf{\mathcal{X}},\mathbf{\mathcal{X}}\rangle}$.


Please refer to \cite{kolda2009tensor} for a more extensive overview.

\section{Main results}\label{sec:Mod}
\subsection{Problem formulation}
As mentioned before, a rainy video $\mathbf{\mathcal{O}}\in \mathbb{R}^{m\times n\times t}$ can be modeled as a linear superposition:%
\begin{equation}
\mathbf{\mathcal{O}}=\mathbf{\mathcal{B}}+\mathbf{\mathcal{R}}+\mathcal{N},
\label{obnoise}
\end{equation}
where $\mathbf{\mathcal{O}},\mathbf{\mathcal{B}}, \mathbf{\mathcal{R}}\text{ and }\mathcal{N}\in \mathbb{R}^{m\times n\times t}$ are four 3-mode tensors representing the observed rainy video, the unknown rain-free video, the rain streaks and the noise (or error) term, respectively.

Our goal is to decompose the rain-free video $\mathbf{\mathcal{B}}$ and the rain streaks $\mathbf{\mathcal{R}}$ from an input rainy video $\mathbf{\mathcal{O}}$.
To solve this ill-posed inverse problem, we need to analyze the prior information for both $\mathbf{\mathcal{B}}$ and $\mathbf{\mathcal{R}}$ and then introduce corresponding regularizers, which will be discussed in the next subsection.

\subsection{Priors and regularizers}\label{Sec:Prior}
In this subsection, we continue the discussion on the prior knowledge with the assumption that rain streaks are approximately vertical.

\paragraph{Sparsity of rain streaks}
When the rain is light, the rain streaks can naturally be considered as being sparse.
To boost the sparsity of rain streaks, minimizing the $\ell_1$ norm of the rain streaks $\mathcal{R}$ is an ideal option.
When the rain is very heavy, it seems that this regularization is not proper.
However, when the rain is extremely heavy, it is very difficult or even impossible to recover the rain-free part because of the huge loss of the reliable information.
The rainy scenarios discussed in this paper are not that extreme, and we assume that the rain streaks always maintain lower energy than the background clean videos.
Therefore, when the rain streaks are dense, the $\ell_1$ norm can be viewed as a role to restrain the magnitude of the rain streaks.
Meanwhile, in our model, other regularization terms would also contribute to distinguishing the rain streaks.
Thus, we can tackle the heavy raining scenarios by tuning the parameter of the sparsity term so as to reduce its effect.

\paragraph{The horizontal direction}
In Fig. \ref{Motivation}, (c-1,2,3) show the pixel intensities along a fixed row of the rainy video, the clean video and the rain streaks, respectively.
It is obvious that the variation of the pixel intensity is piecewise smooth only in (c-2), whereas burrs frequently appear in (c-1) and (c-3).
Therefore, a horizontal unidirectional TV regularizer is a suitable candidate for $\mathbf{\mathcal{B}}$.

\paragraph{The vertical direction}
It can be seen from Fig. \ref{Motivation} that (b-3), which is the histogram of the intensity of the vertical gradient in a rain-streak frame, exhibits a distinct distribution with respect to (c-1) and (c-2).
The long-tailed distributions in (c-1) and (c-3) indicate that the minimization of the $l_1$ norm of $\bm\nabla_1\mathcal{R}$ would help to distinguish the rain streaks.


\paragraph{The temporal direction}
From the first column of Fig. \ref{Motivation}, it can be observed that clean videos exhibit the continuity along the time axis.
Sub-figures (a-1,2,3), which present the histograms of the magnitudes in the temporal directional gradient, illustrate that the clean video's temporal gradients consist of more zero values and smaller non-zero values, whereas those of the rainy video and rain streaks tend to be long-tailed.
Therefore, it is natural to minimize the $l_1$ norm of the temporal gradient of the clean video $\mathbf{\mathcal{B}}$.
By the way, the low-rank regularization used in \cite{Jiang_2017_CVPR} is discarded since that the low-rank assumption is not reasonable for the videos captured by dynamic cameras and the rain streaks, which always share the repetitive patterns, can occasionally be more low-rank than the background along the spatial directions.

\subsection{The proposed model}\label{formulation}


Generally, there is an angle between the vertical direction and the real falling direction of the raindrops.
The rain streaks pictured in Fig. \ref{Motivation} are not strictly vertical and there is a 5-degree angle between the rain streaks and the y-axis.
In other words, the prior knowledge discussed above are still valid when this angle is small.
Large-angle cases would be discussed in Sec. \ref{strategies}).
Therefore, the rain streak direction is referred to as the vertical direction corresponding to the y-axis, whereas the rain-perpendicular direction is referred to as the horizontal direction corresponding to the x-axis.
Thus, as a summary of the discussion of the priors and regularizers, our model can be compactly formulated as follows:

\begin{equation}
\begin{aligned}
\underset{\mathbf{\mathcal{B}},\mathbf{\mathcal{R}}}{\min}\
&\alpha_1\|\bm\nabla_{1}\mathbf{\mathcal{R}}\|_1
+\alpha_2\|\mathbf{\mathcal{R}}\|_1
+\alpha_3\|\bm\nabla_{2}\mathbf{\mathcal{B}}\|_1\\
&+\alpha_4\|\bm\nabla_t\mathbf{\mathcal{B}}\|_1+\frac{1}{2}\|\mathcal{O}-(\mathcal{B}+\mathcal{R})\|_F^2\\
\text{s.t.}\ &\mathbf{\mathcal{O}}\geqslant\mathbf{\mathcal{B}}\geqslant \mathbf0,\ \mathbf{\mathcal{O}}\geqslant\mathbf{\mathcal{R}}\geqslant \mathbf0,
\end{aligned}
\label{mainModel2}
\end{equation}
where $\nabla_{1}$, $\nabla_{2}$ and $\nabla_{t}$ are the vertical, horizontal and temporal differential operators, respectively.
$\nabla_{1}$ and $\nabla_{2}$ are also written as $\nabla_{y}$ and $\nabla_{x}$ in \cite{chang2017transformed,Jiang_2017_CVPR}.
An efficient algorithm is proposed in the following subsection to solve (\ref{mainModel2}).


\subsection{Optimization}

Since the proposed model (\ref{mainModel2}) is concise and convex, many state-of-the-art solvers are available to solve it.
Here, we apply the ADMM \cite{Boyd2011Distributed}, which has been proved an effective strategy for solving
large scale optimization problems \cite{jiang2017novel2,Zhao2013TGRS,zhao2014new}.
More specifically, we adopt SALSA \cite{afonso2011augmented}.

After introducing four auxiliary tensors the proposed model (\ref{mainModel2}) is reformulated as the following equivalent constrained problem:
\begin{equation}
\begin{array}{rl}
\min\limits_{\mathbf{\mathcal{B}},\mathbf{\mathcal{V}}_i,\mathbf{\mathcal{D}}_i}\  &\alpha_1\|\mathbf{\mathcal{V}}_1\|_1+ \alpha_2 \|\mathbf{\mathcal{V}}_2\|_1+\alpha_3\|\mathbf{\mathcal{V}}_3\|_1 +\alpha_4\|\mathbf{\mathcal{V}}_4\|_1 \\
&+\frac{1}{2}\|\mathcal{O}-(\mathcal{B}+\mathcal{R})\|_F^2\\
\text{s.t.}\quad&
\mathcal{V}_1 = \bm\nabla_1(\mathcal{R}),\ \mathcal{V}_2 = \mathcal{R},\ \mathcal{V}_3 = \bm\nabla_2(\mathcal{B}),\\
&\mathcal{V}_4 = \bm\nabla_t(\mathcal{B}),\ \mathcal{O}\geqslant\mathcal{B}\geqslant \mathbf0,\ \mathcal{O}\geqslant\mathcal{R}\geqslant \mathbf0
\end{array} \label{ADM}
\end{equation}
where $\mathbf{\mathcal{V}}_i\ \in\mathbb{R}^{m\times n\times t}$ $(i = 1,2,3,4)$.

Then, the augmented Lagrangian function of (\ref{ADM}) is
\begin{equation*}
\begin{aligned}
L_{\mu}& (\mathbf{\mathcal{B,R}},\mathbf{\mathcal{V}}_i,\mathbf{\mathcal{D}}_i)= \frac{1}{2}\|\mathcal{O-B-R}\|_F^2+ \alpha_1\|\mathbf{\mathcal{V}}_1\|_1+\alpha_2 \|\mathbf{\mathcal{V}}_2\|_1\\
& +\alpha_3\|\mathbf{\mathcal{V}}_3\|_1+ \alpha_4\|\mathbf{\mathcal{V}}_4\|_1+ \frac{\mu}{2}\|\bm\nabla_1\mathcal{R}-\mathcal{V}_1-\mathcal{D}_1\|_F^2 \\ &+\frac{\mu}{2}\|\mathcal{R}-\mathcal{V}_2-\mathcal{D}_2\|_F^2+ \frac{\mu}{2}\|\bm\nabla_2\mathcal{B}-\mathcal{V}_3-\mathcal{D}_3\|_F^2 \\
&+ \frac{\mu}{2}\|\bm\nabla_t\mathcal{B}-\mathcal{V}_4-\mathcal{D}_4\|_F^2,
\end{aligned}
\label{ALF1}
\end{equation*}
where the $\mathcal{D}_i$s $(i=1,2,3,4)$ are the scaled Lagrange multipliers and the $\mu$ is a positive scalar.


\paragraph{$\mathcal{V}_i$ sub-problems}
For $i = 1,2,3,4$, the $\mathbf{\mathcal{V}}_i$ sub-problem can be written as a equivalent problem:

\begin{equation*}
\begin{aligned}
\mathbf{\mathcal{V}}_i^+=\underset{\mathbf{\mathcal{V}}_i}{\arg\min}\quad &\alpha_i\|\mathbf{\mathcal{V}}_i\|_1+\frac{\mu}{2}\|\mathbf{\mathcal{A}}_i-\mathbf{\mathcal{V}}_i\|_F^2.\\
\end{aligned}
\end{equation*}
Such a problem has a closed-form solution, obtained through soft thresholding:
\begin{equation*}
\mathbf{\mathcal{V}}_i^+=\bm{\mathcal{S}}_{\frac{\alpha_i}{\mu}}\left(\mathbf{\mathcal{A}}_i\right).
\end{equation*}
Here, the tensor non-negative \textbf{soft-thresholding operator} $\bm{\mathcal{S}}_{v} (\cdot )$ is defined as
\begin{equation*}
\bm{\mathcal{S}}_{v}(\mathbf{\mathcal{A}})= \bar{\mathbf{\mathcal{A}}}\\
\end{equation*}
with
\begin{equation*}
 \bar{a}_{i_{1}i_2\cdots i_{N}}=\left\{
\begin{aligned}
   &a_{i_{1}i_2\cdots i_{N}}-v,& \quad &a_{i_{1}i_2\cdots i_{N}}>v, \\
   &0,& \quad &\text{otherwise}. \\
   \end{aligned}
   \right.
\end{equation*}

Therefore, $\mathbf{\mathcal{V}}_i$ $(i =1,2,3,4)$ can respectively be updated as follows:
\begin{equation}
\left\{
\begin{aligned}
\mathcal{V}_1^{(t+1)} &= \bm{\mathcal{S}}_{\frac{\alpha_1}{\mu}}\left( \bm\nabla_1\mathcal{R}-\mathcal{D}_1 \right),\\
\mathcal{V}_2^{(t+1)} &= \bm{\mathcal{S}}_{\frac{\alpha_2}{\mu}}\left( \mathcal{R}-\mathcal{D}_2 \right),\\
\mathcal{V}_3^{(t+1)} &= \bm{\mathcal{S}}_{\frac{\alpha_3}{\mu}}\left( \bm\nabla_2\mathcal{B}-\mathcal{D}_3 \right),\\
\mathcal{V}_4^{(t+1)} &= \bm{\mathcal{S}}_{\frac{\alpha_4}{\mu}}\left( \bm\nabla_t\mathcal{B}-\mathcal{D}_4 \right).\\
\end{aligned}\right.
\label{vi_update}
\end{equation}
The time complexity of each sub-problem above is $O(mnt)$.


\paragraph{$\mathbf{\mathcal{B}}$ and $\mathbf{\mathcal{R}}$ sub-problems}
$\mathbf{\mathcal{B}}$ and $\mathbf{\mathcal{R}}$ sub-problems are least-squares problems:
\begin{equation*}
\begin{aligned}
\mathbf{\mathcal{B}}^+
=\argmin\limits_{\mathcal{O}\leq\mathcal{B}\leq \mathbf 0}\  &\frac{1}{2}\|\mathcal{O-B-R}\|_F^2+\frac{\mu}{2}\|\bm\nabla_2\mathcal{B}-\mathcal{V}_3-\mathcal{D}_3\|_F^2 \\
&+\frac{\mu}{2}\|\bm\nabla_t\mathcal{B}-\mathcal{V}_4-\mathcal{D}_4\|_F^2,\\
\mathbf{\mathcal{R}}^+
=\argmin\limits_{\mathcal{O}\leq\mathcal{R}\leq \mathbf 0}\  &\frac{1}{2}\|\mathcal{O-B-R}\|_F^2+\frac{\mu}{2}\|\bm\nabla_1\mathcal{R}-\mathcal{V}_1-\mathcal{D}_1\|_F^2\\
&+\frac{\mu}{2}\|\mathcal{R}-\mathcal{V}_2-\mathcal{D}_2\|_F^2.\\
\end{aligned}
\end{equation*}
Then, we have
\begin{equation}
\begin{aligned}
\mathbf{\mathcal{B}}^+
=& \frac{\mathcal{O-R}+\mu\bm\nabla_2^\top(\mathcal{V}_3-\mathcal{D}_3) +\mu\bm\nabla_t^\top(\mathcal{V}_4-\mathcal{D}_4)} {\bm 1 + \mu\bm\nabla_2^\top\bm\nabla_2 +\mu\bm\nabla_t^\top\bm\nabla_t}\\
\mathbf{\mathcal{R}}^+
=&\frac{\mathcal{O-B}+\mu\bm\nabla_1^\top(\mathcal{V}_1-\mathcal{D}_1) +\mu(\mathcal{V}_2-\mathcal{D}_2)}{\bm 1 + \mu\bm\nabla_1^\top\bm\nabla_1+\mu }\\
\end{aligned}
\label{br_update}
\end{equation}
We adopt the fast Fourier transform (FFT) for fast calculation when updating $\mathbf{\mathcal{B}}$ and $\mathbf{\mathcal{R}}$.
Meanwhile, the elements in $\mathbf{\mathcal{B}}^{(t+1)}$ and $\mathbf{\mathcal{R}}^{(t+1)}$that are smaller than 0 or larger than the corresponding elements in $\mathbf{\mathcal{O}}$ will be shrunk.
The time complexity of updating $\mathbf{\mathcal{B}}$ (or $\mathbf{\mathcal{R}}$) is $O(mnt\cdot\text{log}(mnt))$.



\paragraph{Multipliers updating}
The Lagrange multipliers $\mathcal{D}_i$s ($i = 1,2,3,4$) can be updated as follows:
\begin{equation}
\left\{
\begin{aligned}
&\mathcal{D}_1 = \mathcal{D}_1+ \bm\nabla_1\mathcal{R}-\mathcal{V}_1\\
&\mathcal{D}_2 = \mathcal{D}_2+ \mathcal{R}-\mathcal{V}_2\\
&\mathcal{D}_3 = \mathcal{D}_3+ \bm\nabla_2\mathcal{B}-\mathcal{V}_3\\
&\mathcal{D}_4 = \mathcal{D}_4+ \bm\nabla_t\mathcal{B}-\mathcal{V}_4
\end{aligned}\right.
\label{multipliers_update}
\end{equation}

The proposed algorithm for video rain streak removal is denoted as ``FastDeRain'' and summarized in Algorithm \ref{alg}.
For a video with dimensions of $m\times n\times t$, the time complexity of the proposed algorithm is proportional to $O\left(mnt\log(mnt)\right)$.
\begin{algorithm}[htp]
\renewcommand\arraystretch{1.2}
\caption[Caption for LOF]{FastDeRain}
\begin{algorithmic}[1]
\renewcommand{\algorithmicrequire}{\textbf{Input:}} 
\Require
The rainy video $\mathbf{\mathcal{O}}$;
\renewcommand{\algorithmicrequire}{\textbf{Initialization:}} 
\Require $\mathbf{\mathcal{B}}^{(0)}=\mathbf{\mathcal{O}}$, $\mathbf{\mathcal{O}}=\mathbf 0$
\While {not converged}
\State Update $\mathbf{\mathcal{V}}_i$ $(i = 1,2,3,4)$ via Eq. (\ref{vi_update});
\State Update $\mathbf{\mathcal{B}}$ and $\mathbf{\mathcal{R}}$ via (\ref{br_update});
\State Update $\mathbf{\mathcal{D}}_i$ $(i = 1,2,3,4)$ via Eq. (\ref{multipliers_update});
\EndWhile
\renewcommand{\algorithmicrequire}{\textbf{Output:}}
\Require The estimates of the rain-free video $\mathbf{\mathcal{B}}$ and the rain streaks $\mathcal{R}$.
\end{algorithmic}
\label{alg}
\end{algorithm}

\subsection{Discussion of the oblique rain streaks}\label{strategies}
As we know that, in a real rainfall-affected scene, the rain streaks are not always vertical.
Thus, the directional property we utilized in our model is a double-edged sword when dealing with digital videos.
In this subsection, we design an automatical rain streaks' angle detection method, and based on it, we propose the shift strategy to deal with rain streaks not vertical.

\setcounter{paragraph}{0}
\paragraph{Rain streaks direction detection}

Before starting our strategy, one important issue is how to automatically detect the direction of the rain streaks.
Based on our analysis of the prior knowledge, it's not difficult to come up with a simple and effective method to detect the direction.
In this subsection, we assume that the rain streaks are in the same direction and the angle between rain streaks and the vertical direction are denoted as $\theta$.
For a rainy video $\mathcal{O}\in\mathbb{R}^{m\times n\times t}$, our method consists of three steps:\\
\indent 1) Filter the horizontal slices of the rainy video with a $3\times 3$ median filter, {\em i.e.}, for $i = 1,2,\cdots,m$,
$\widehat{\mathcal{O}}(i,:,:) = \text{med}(\mathcal{O}(i,:,:))$, and obtain $\mathcal{R}_0 = \mathcal{O}-\widehat{\mathcal{O}}$.\\
\indent 2) Rotate each frame of $\mathcal{R}_0$ with $\theta_i = i^{\circ}$, and obtain $\mathcal{R}_0^{\theta_i}$ $(i = 0,1,\cdots,t)$.\\
\indent 3) For each $\mathcal{R}_0^{\theta_i}$, denote $y_i = \|\nabla_1\mathcal{R}_0^{\theta_i}\|_1$, then the detected rain streaks angle $\hat{\theta} = \argmin_{\theta_i}{y_i}$.

\begin{figure}[htp]
\centering
\includegraphics[width=0.66\linewidth]{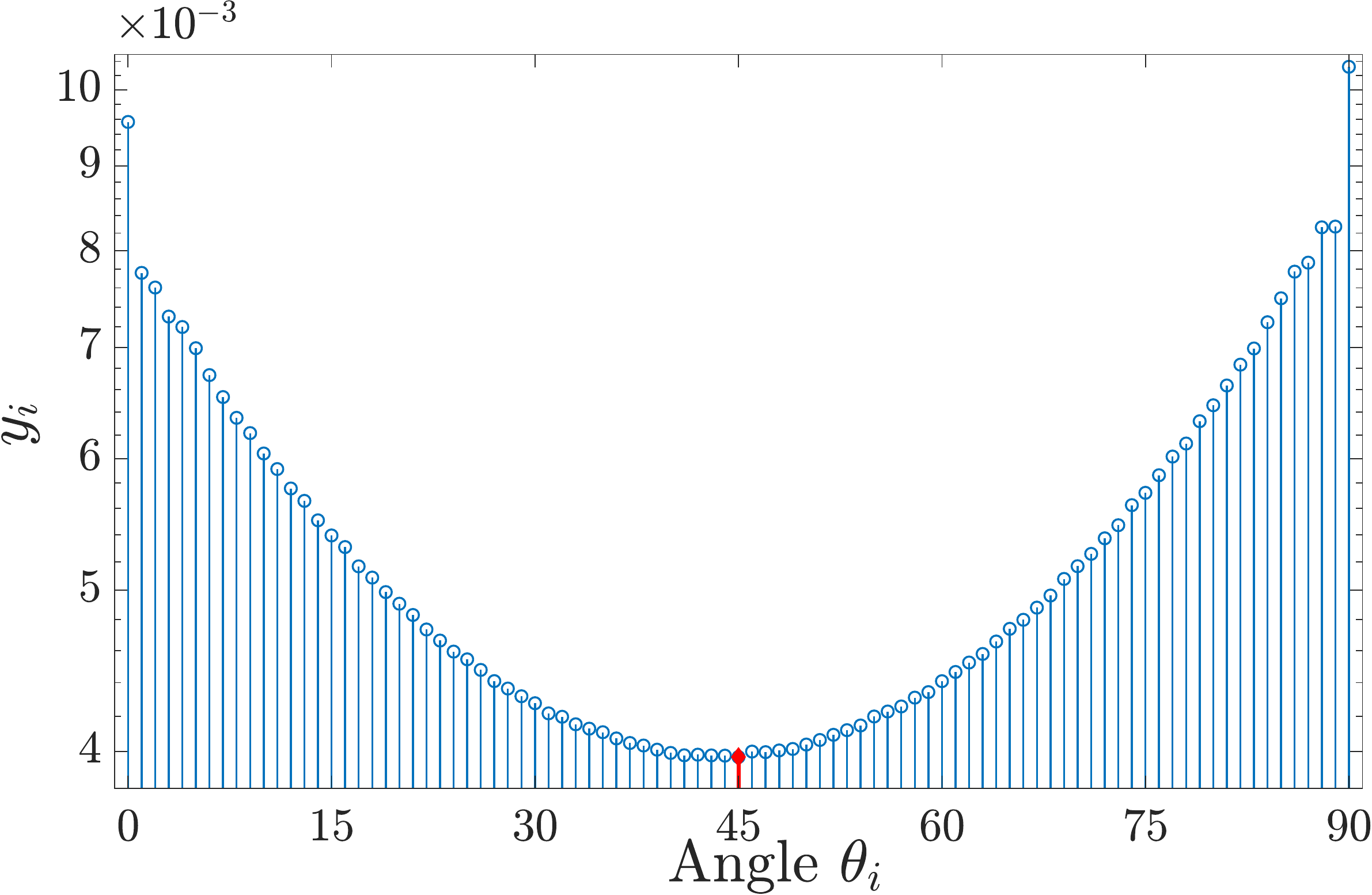}
\caption{The magnitude of $y_i$s with respect to $\theta_i$s. }
\label{detection}
\end{figure}

Fig. \ref{detection} shows an example of our detection method, where the rain streaks are simulated with angle $45^{\circ}$ and the detection result (labeled red) is exactly 45$^\circ$.
Actually, the $y_i$s are very low when $\theta_i$ is close to $45^{\circ}$, according with the discussion in \ref{Sec:Prior}.
Generally, the angle between the rain streaks and the vertical direction distributes in $\left(-90^\circ,90^\circ\right)$.
If the angle $\hat{\theta}\in\left(-90^\circ,0^\circ\right)$, we can restrict it to the range of $\left(0^\circ,90^\circ\right)$ by the left-right flipping of each frame.
If the angle $\hat{\theta}\in\left(45^\circ,90^\circ\right)$, we can restrict it to the range of $\left(0^\circ,45^\circ\right)$ by transposing (i.e. interchanging the rows and columns of a given matrix) each frame.
To save space, we only discuss the situations where $\hat{\theta}\in[0^\circ,45^\circ]$ in the following.

\begin{figure}[htp]
\centering
\includegraphics[width=0.99\linewidth]{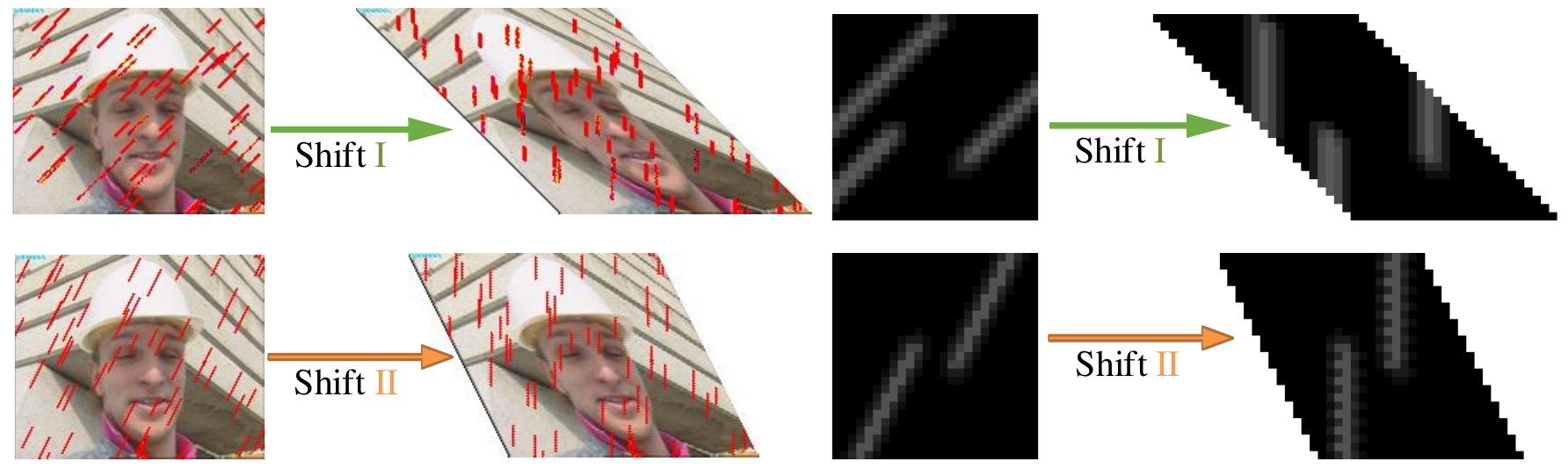}
\caption{Illustrations of the shift I and the shift II operations. For better visualization, the rain streaks in the left part are roughly labeled with the red color, while the pixel values of the rain streaks images in the right are scaled.}
\label{shiftoperation}
\end{figure}
\paragraph{The shift strategy}
When the detected angle $\hat\theta\in\left[15^\circ,45^\circ\right]$, we apply the shift strategy, which consists of two shifting operations, as shown in Fig. \ref{shiftoperation}, for different situations.
The two shift operations are detailed as follows:

\hspace{0.1cm}\textbf{Shift I} \ If $\hat\theta\in\left[35^\circ,45^\circ\right]$, for each frame $\mathbf{O}_{::k}$, we slide the $i$-th row $(i-1)$ pixel(s) to the right.

\hspace{0.1cm}\textbf{Shift II} If $\hat\theta\in\left[15^\circ,35^\circ\right)$, for each frame $\mathbf{O}_{::k}$, we slide the $i$-th row $\lfloor\frac{(i-1)}{2}\rfloor$\footnote{$\lfloor x\rfloor$ denotes the rounding the $x$ to the nearest integers towards minus infinity.} pixel(s) to the right.

\begin{figure*}[t]
    \centering\color{red}
    \includegraphics[width=0.81\linewidth]{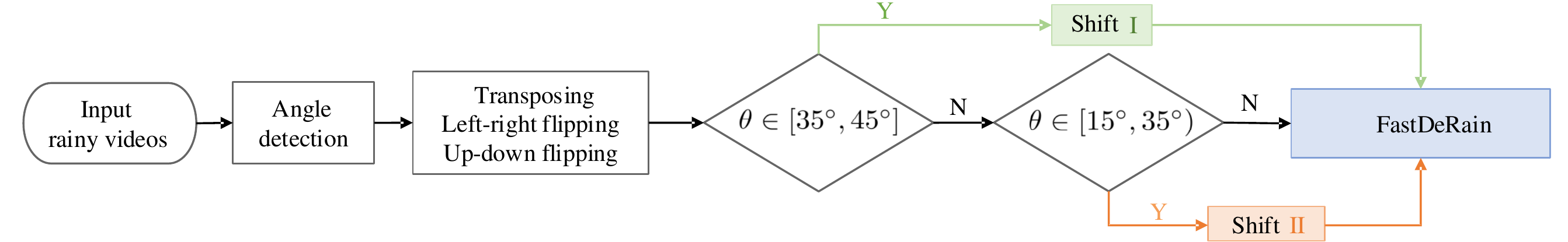}
\caption{The flowchart of the dealing with rainy videos with the rain streaks of different directions.}
\label{flowchart}
\end{figure*}

Different from the rotation strategy recommended in \cite{Jiang_2017_CVPR}, the core idea of the shift strategy is to rationally slide the rows of the rainy frames and make the rain streaks being approximately vertical without any degradation caused by interpolation
Meanwhile, it is notable that these shifting operations wouldn't affect the prior knowledges mentioned in \ref{Sec:Prior}.
After shifting, the rain streaks is close to being vertical, and we can apply the algorithm \ref{alg}.
Finally, the result would be shifted back.
The flowchart of applying our FastDeRain with the shift strategy is shown in Fig. \ref{flowchart}.

%



\section{Experimental results}\label{sec:Exp}
\setcounter{paragraph}{0}
In this section, we evaluate the performance of the proposed algorithm on synthetic data and real-world rainy videos.

\paragraph{Implementation details}
Throughout our experiments, color videos with dimensions of ${m\times n\times3\times t}$ are transformed into the YUV format.
YUV is a color space that is often used as part of a color image pipeline. Y stands for the luma component (the brightness), and U and V are the chrominance (color) components\footnote{\url{https://en.wikipedia.org/wiki/YUV}}.
We apply our method only to the Y channel with the dimension of ${m\times n\times t}$.
The exhibited rain streaks are scaled for better visualization.

Since that the graphics processing unit (GPU) device is able to speed up the large-scale computing, we implement our method on the platform of Windows
10 and Matlab (R2017a) with an Intel(R) Core(TM) i5-4590 CPU at 3.30GHz, 16 GB RAM, and a GTX1080 GPU.
The involved operations in algorithm \ref{alg} is convenient to be implemented on the GPU device \cite{MatGpu}.
If we conduct our algorithm on the CPU, the running time for dealing with a video of size $240\times320\times3\times100$ is about 23 seconds, while 7 seconds on the GPU device.
Meanwhile, Fu {\em et al}.'s method \cite{fu2017removing} can also be accelerated by the GPU device, from 38 seconds on the CPU to 24 seconds on the GPU, dealing with a video of size $240\times320\times3\times100$.
Thus, we only report the GPU running time of FastDeRain and Fu {\em et al.}'s method in this section.

\begin{figure}[t!] 
\centering\scriptsize\renewcommand\arraystretch{1}
\setlength{\tabcolsep}{1pt}
\begin{tabular}{cccccc}
Rainy &TCL& DDN & FastDeRain&GT\\
\includegraphics[width=0.19\linewidth]{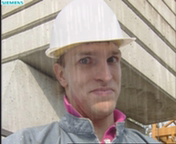} &
\includegraphics[width=0.19\linewidth]{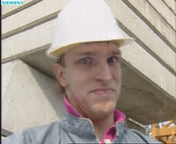} &
\includegraphics[width=0.19\linewidth]{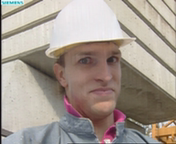} &
\includegraphics[width=0.19\linewidth]{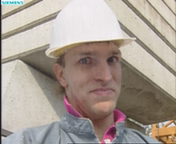} &
\includegraphics[width=0.19\linewidth]{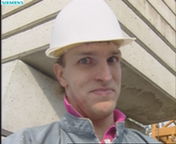} \\
&\includegraphics[width=0.19\linewidth]{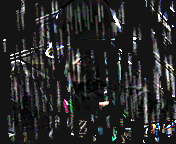} &
\includegraphics[width=0.19\linewidth]{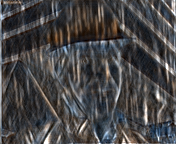} &
\includegraphics[width=0.19\linewidth]{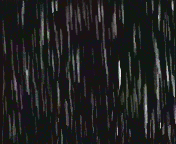} &
\includegraphics[width=0.19\linewidth]{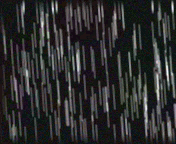} \\
\includegraphics[width=0.19\linewidth]{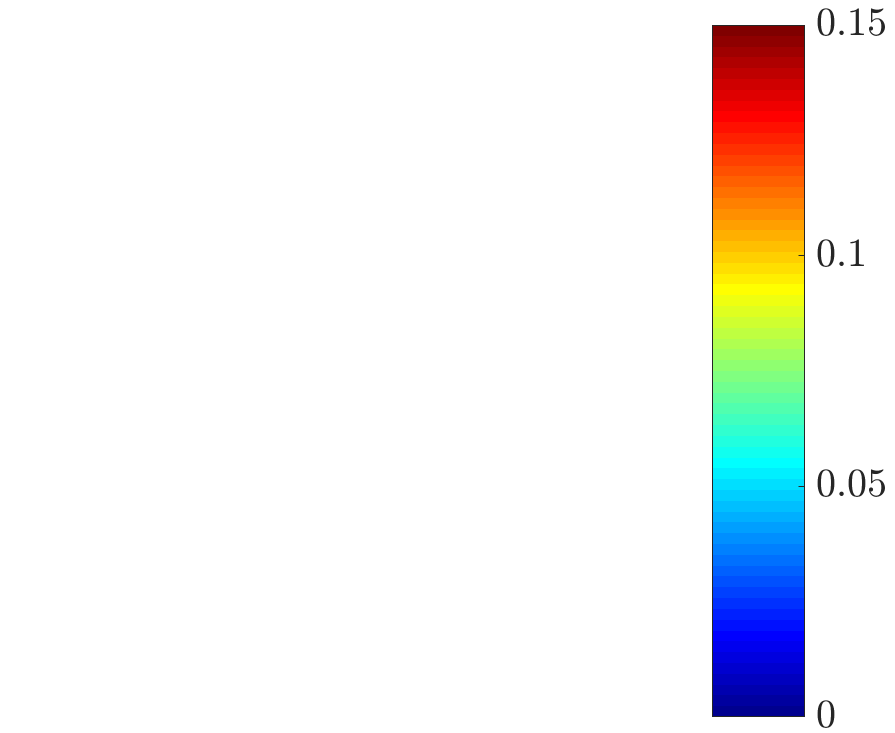}&
\includegraphics[width=0.19\linewidth]{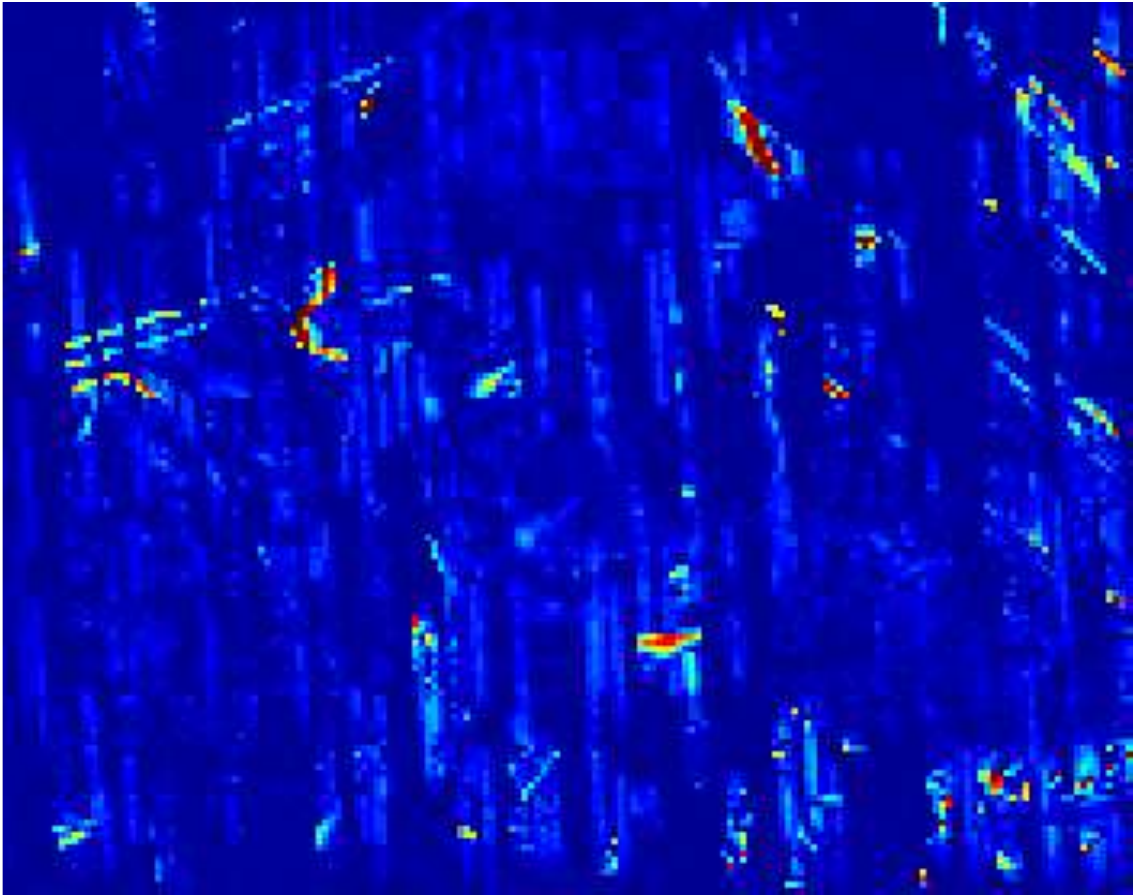}&
\includegraphics[width=0.19\linewidth]{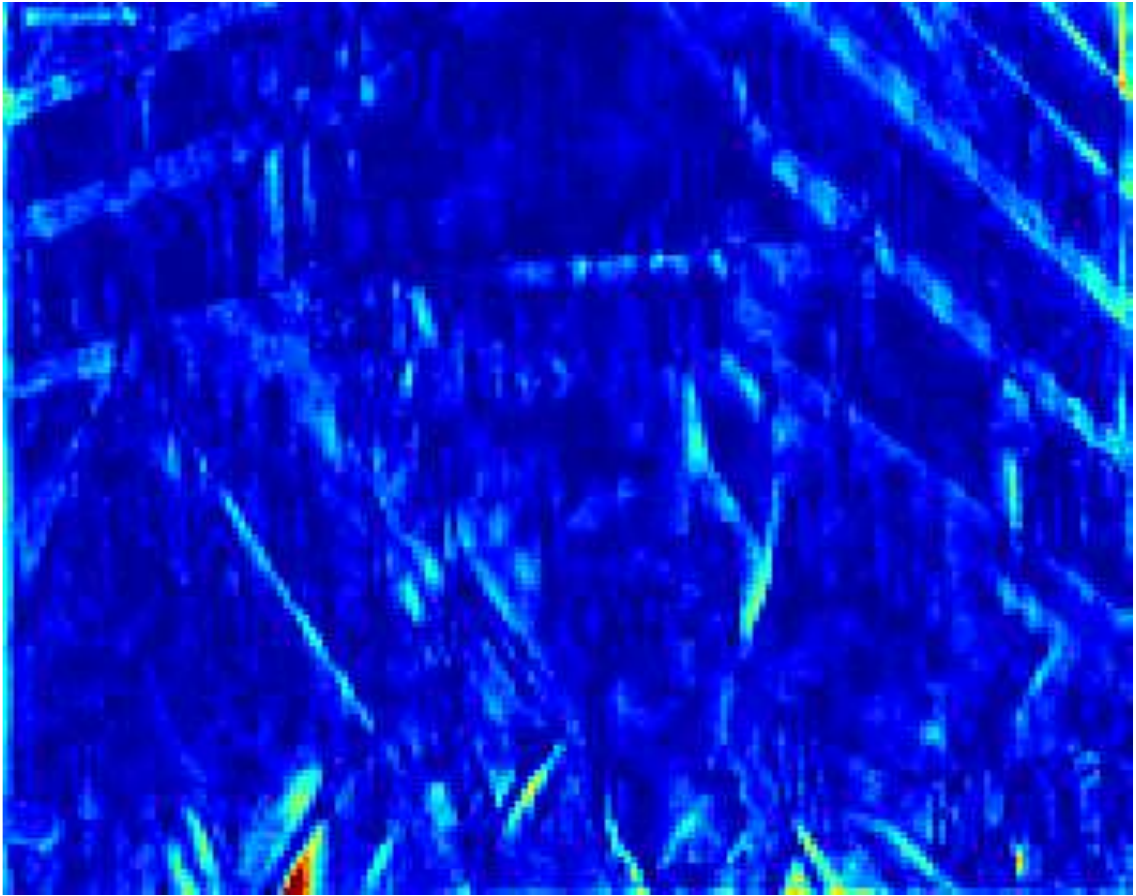}&
\includegraphics[width=0.19\linewidth]{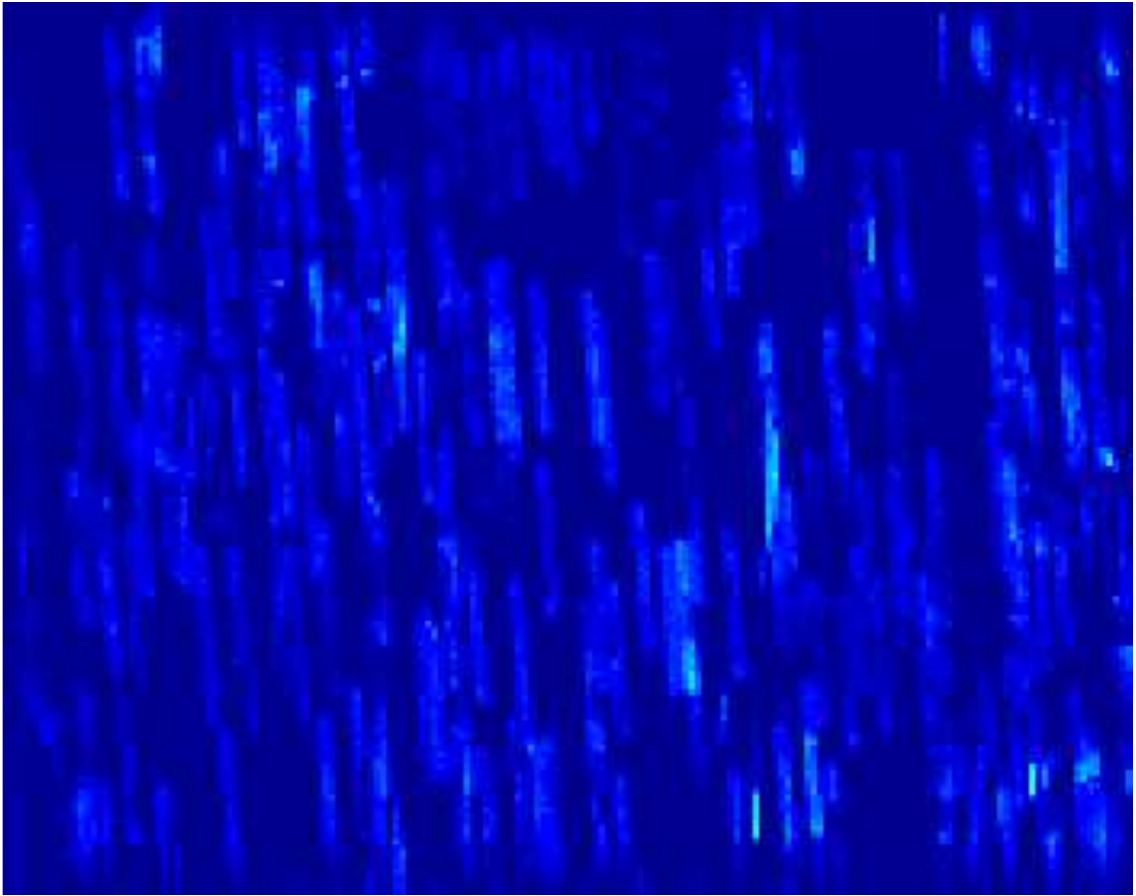} &
\includegraphics[width=0.19\linewidth]{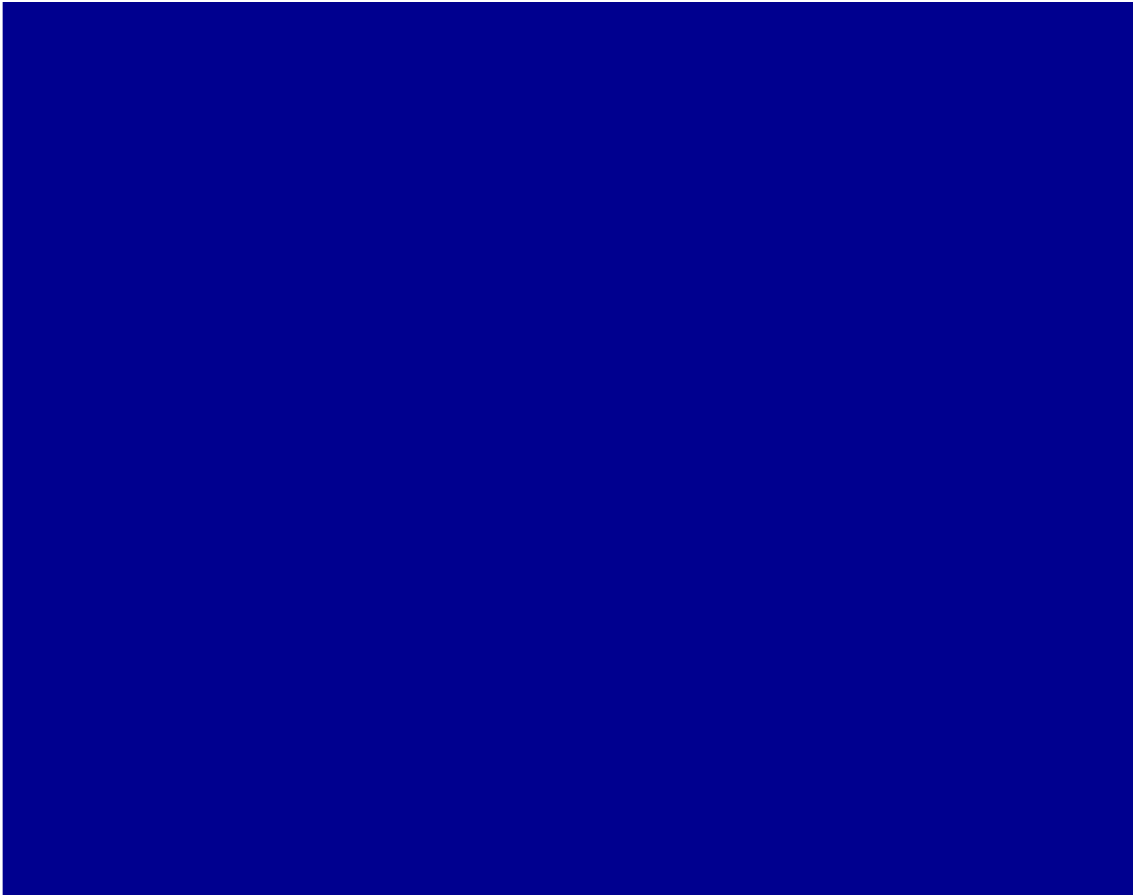} \\

Rainy &TCL & DDN & FastDeRain&GT\\
\includegraphics[width=0.19\linewidth]{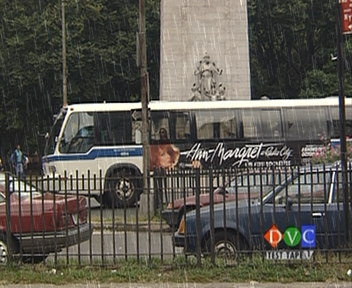} &
\includegraphics[width=0.19\linewidth]{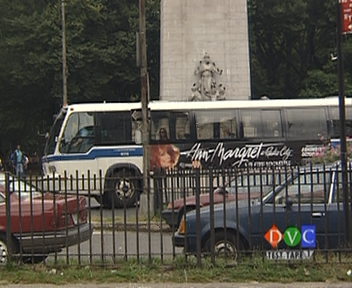} &
\includegraphics[width=0.19\linewidth]{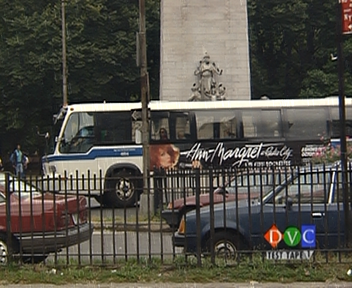} &
\includegraphics[width=0.19\linewidth]{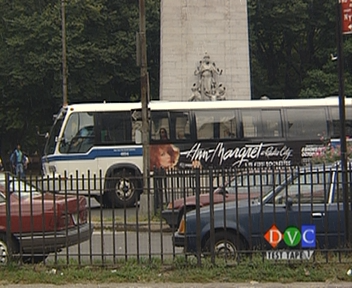} &
\includegraphics[width=0.19\linewidth]{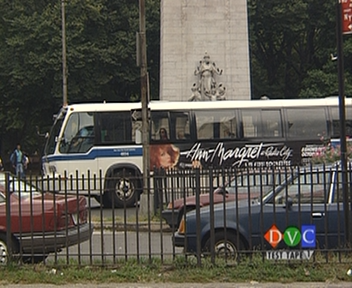} \\
&\includegraphics[width=0.19\linewidth]{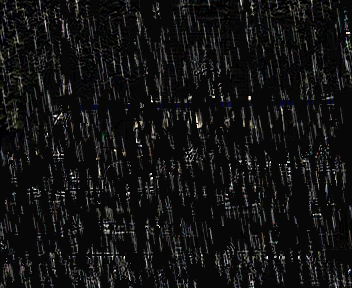} &
\includegraphics[width=0.19\linewidth]{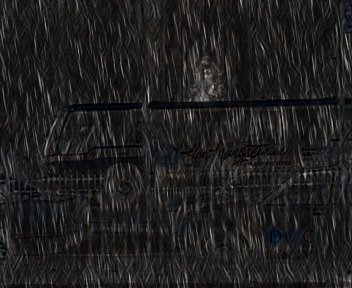} &
\includegraphics[width=0.19\linewidth]{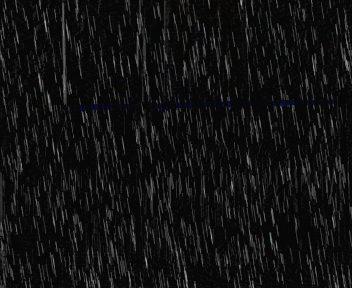} &
\includegraphics[width=0.19\linewidth]{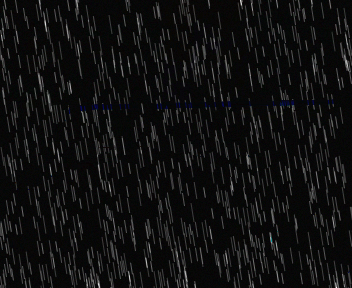} \\
\includegraphics[width=0.19\linewidth]{figs/component/bar_yuv.png}&
\includegraphics[width=0.19\linewidth]{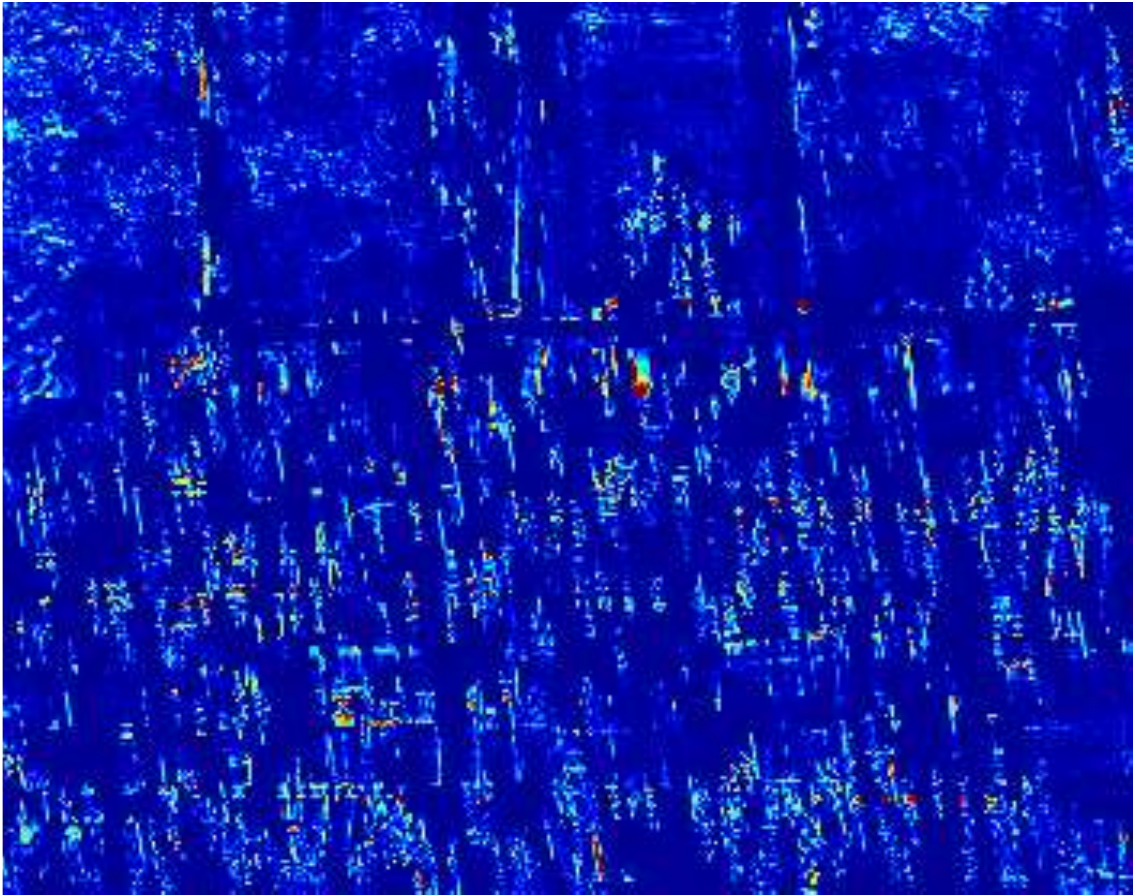}&
\includegraphics[width=0.19\linewidth]{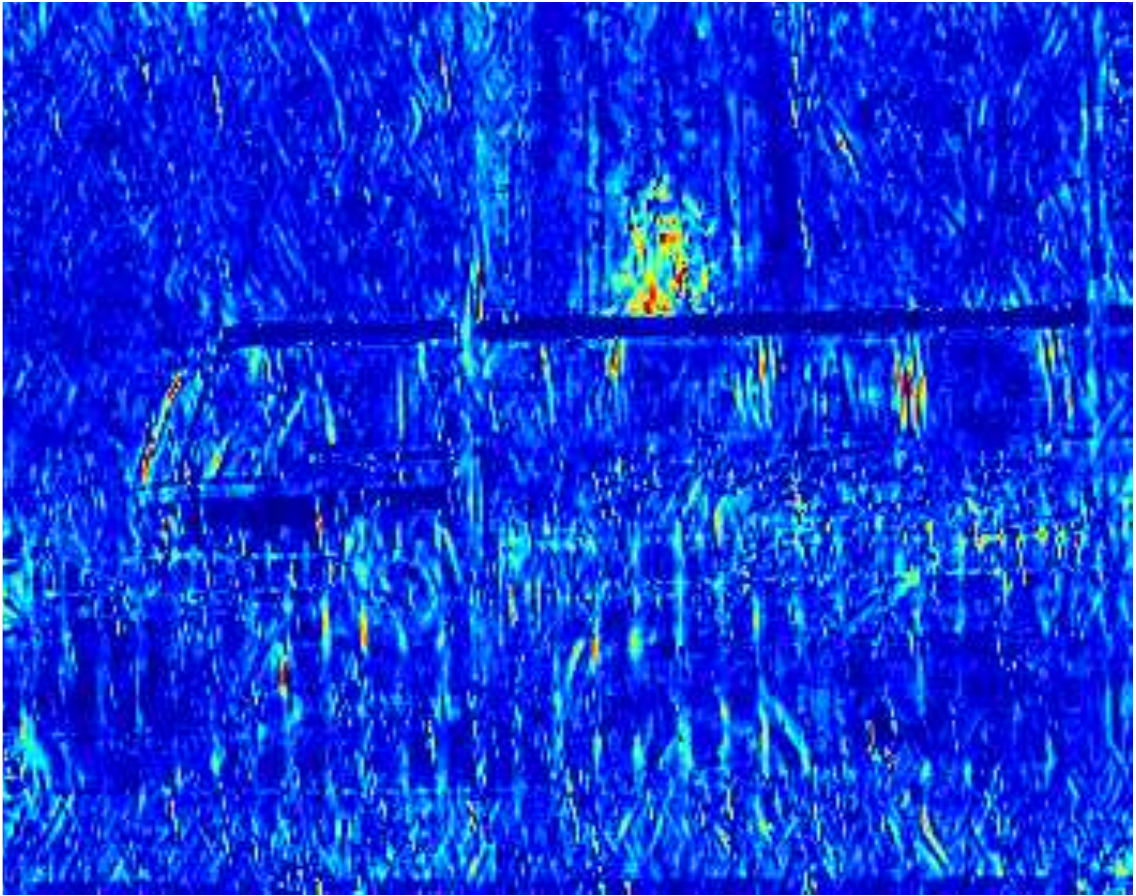}&
\includegraphics[width=0.19\linewidth]{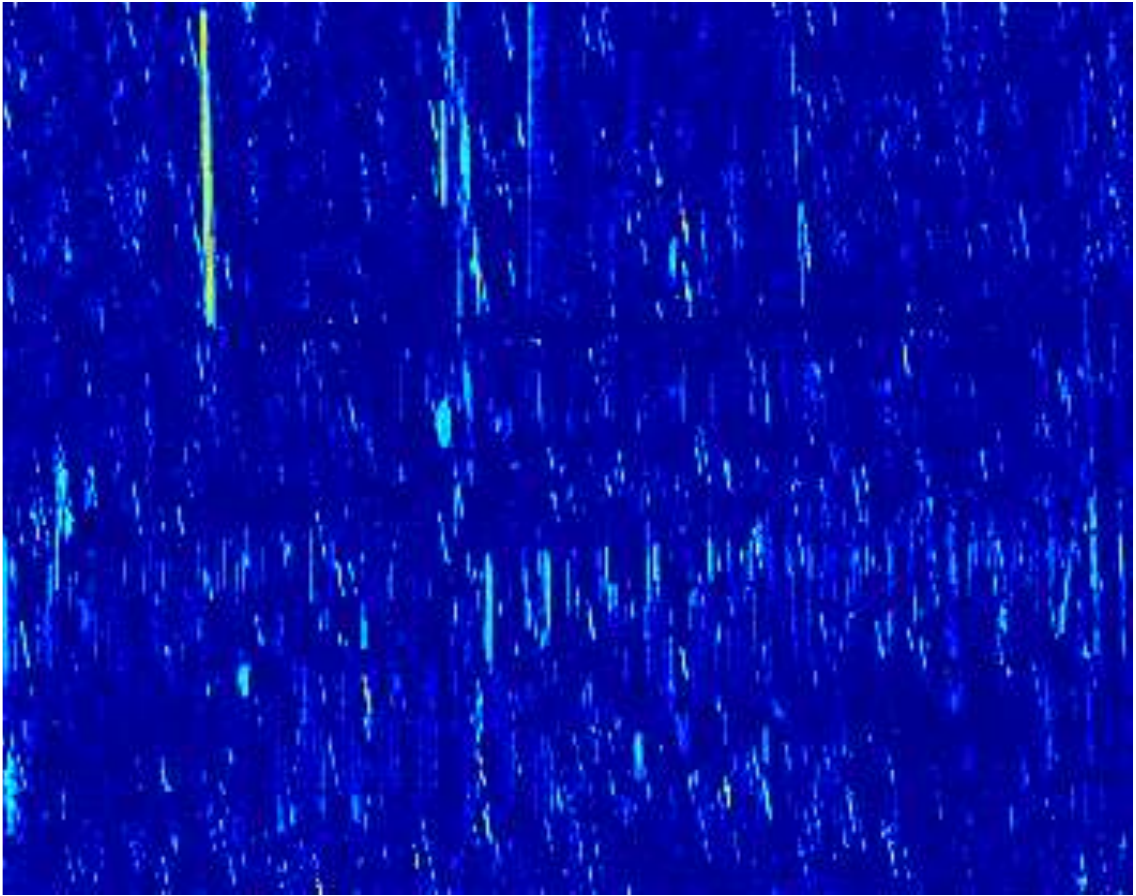} &
\includegraphics[width=0.19\linewidth]{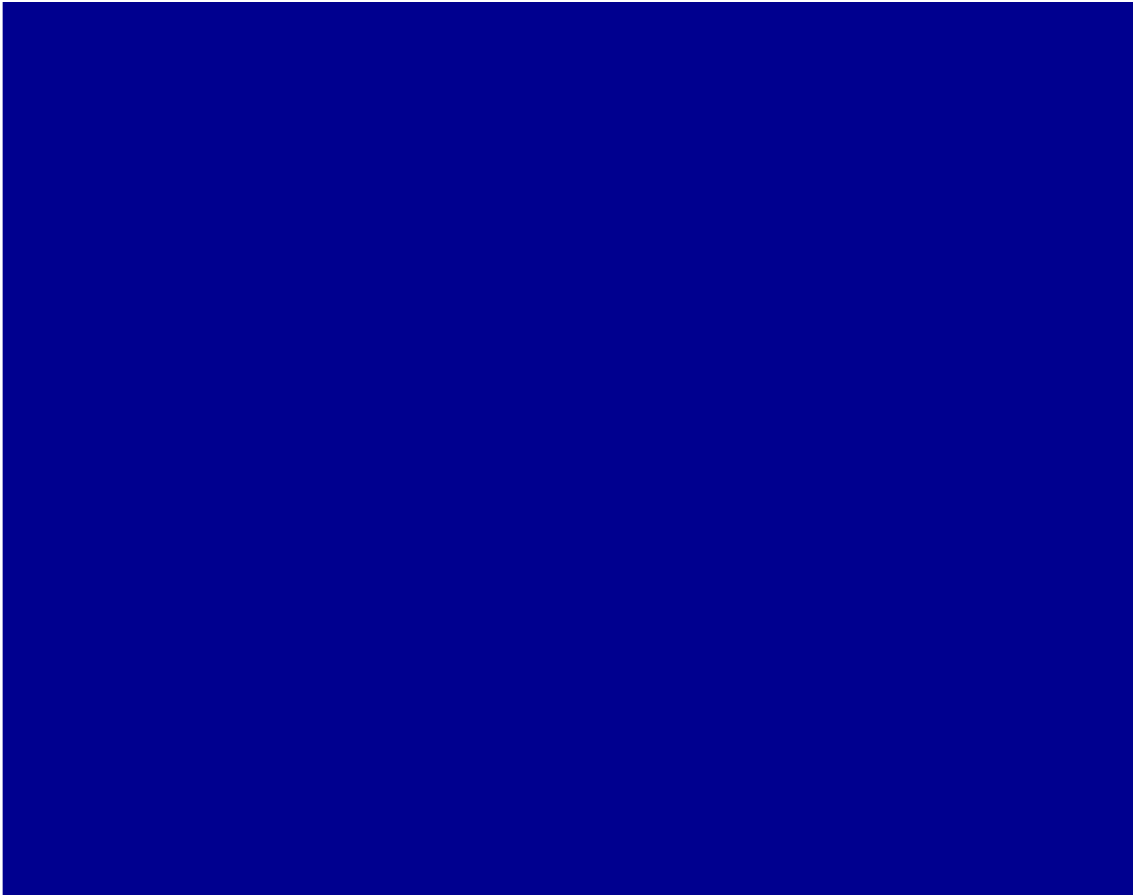} \\

Rainy &TCL & DDN & FastDeRain&GT\\
\includegraphics[width=0.19\linewidth]{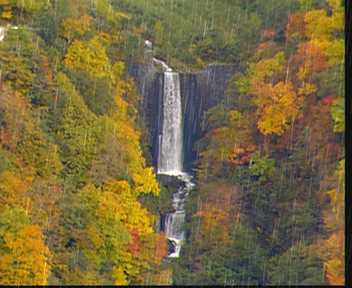} &
\includegraphics[width=0.19\linewidth]{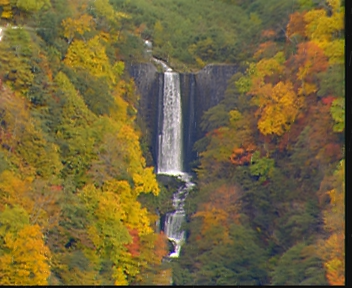} &
\includegraphics[width=0.19\linewidth]{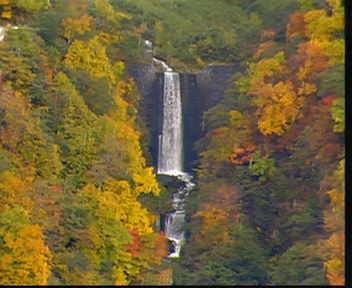} &
\includegraphics[width=0.19\linewidth]{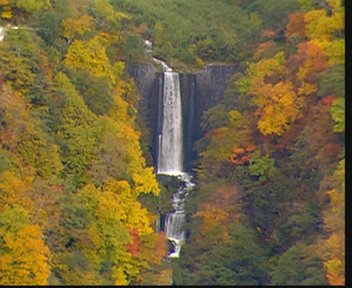} &
\includegraphics[width=0.19\linewidth]{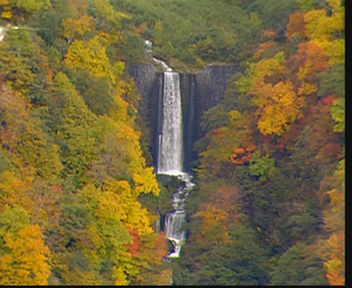} \\
&\includegraphics[width=0.19\linewidth]{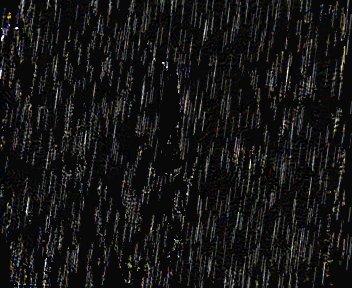} &
\includegraphics[width=0.19\linewidth]{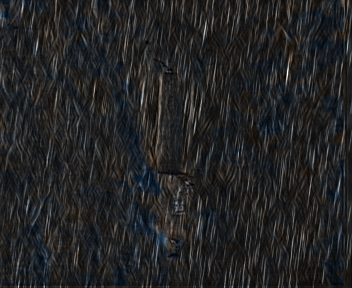} &
\includegraphics[width=0.19\linewidth]{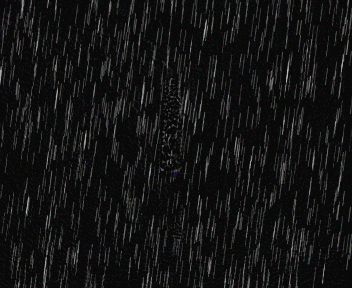} &
\includegraphics[width=0.19\linewidth]{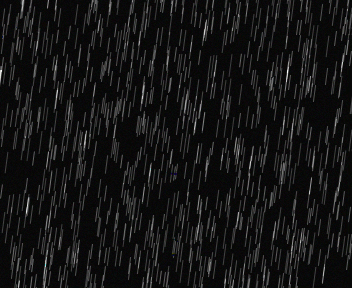} \\
\includegraphics[width=0.19\linewidth]{figs/component/bar_yuv.png}&
\includegraphics[width=0.19\linewidth]{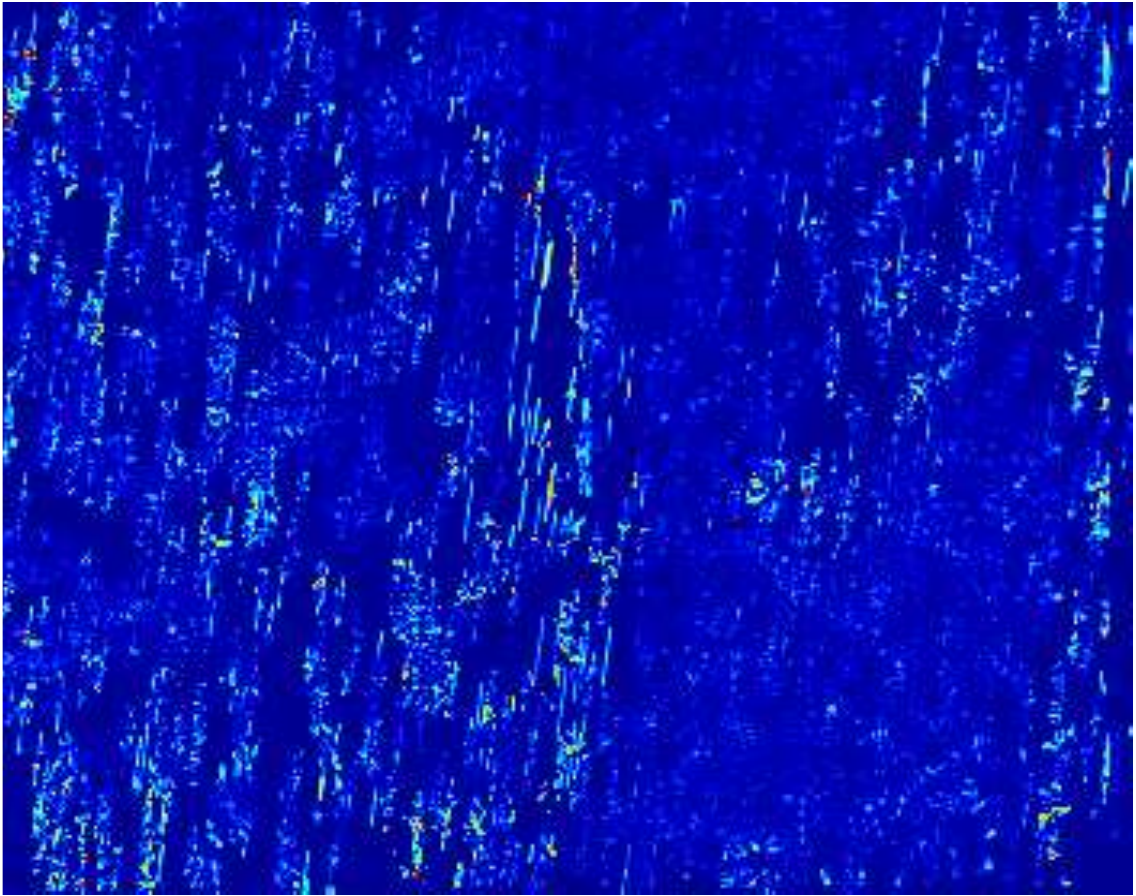}&
\includegraphics[width=0.19\linewidth]{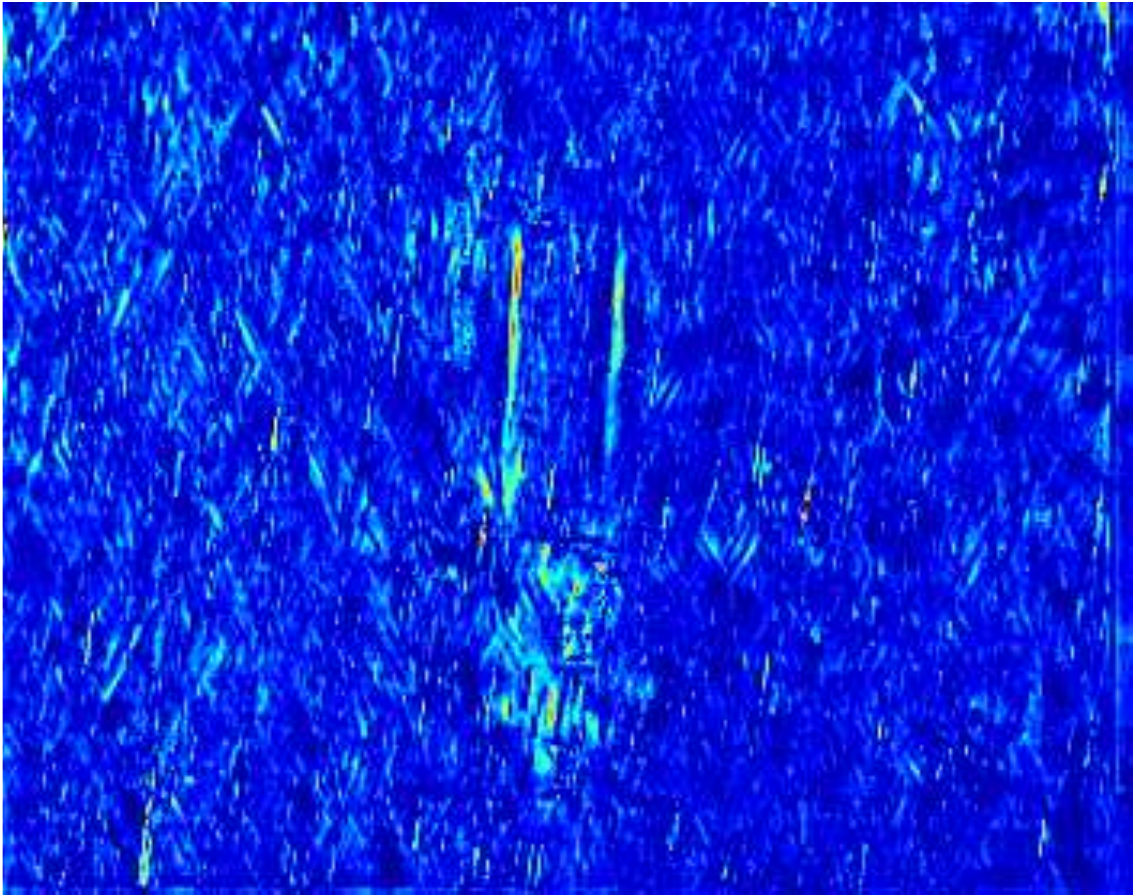}&
\includegraphics[width=0.19\linewidth]{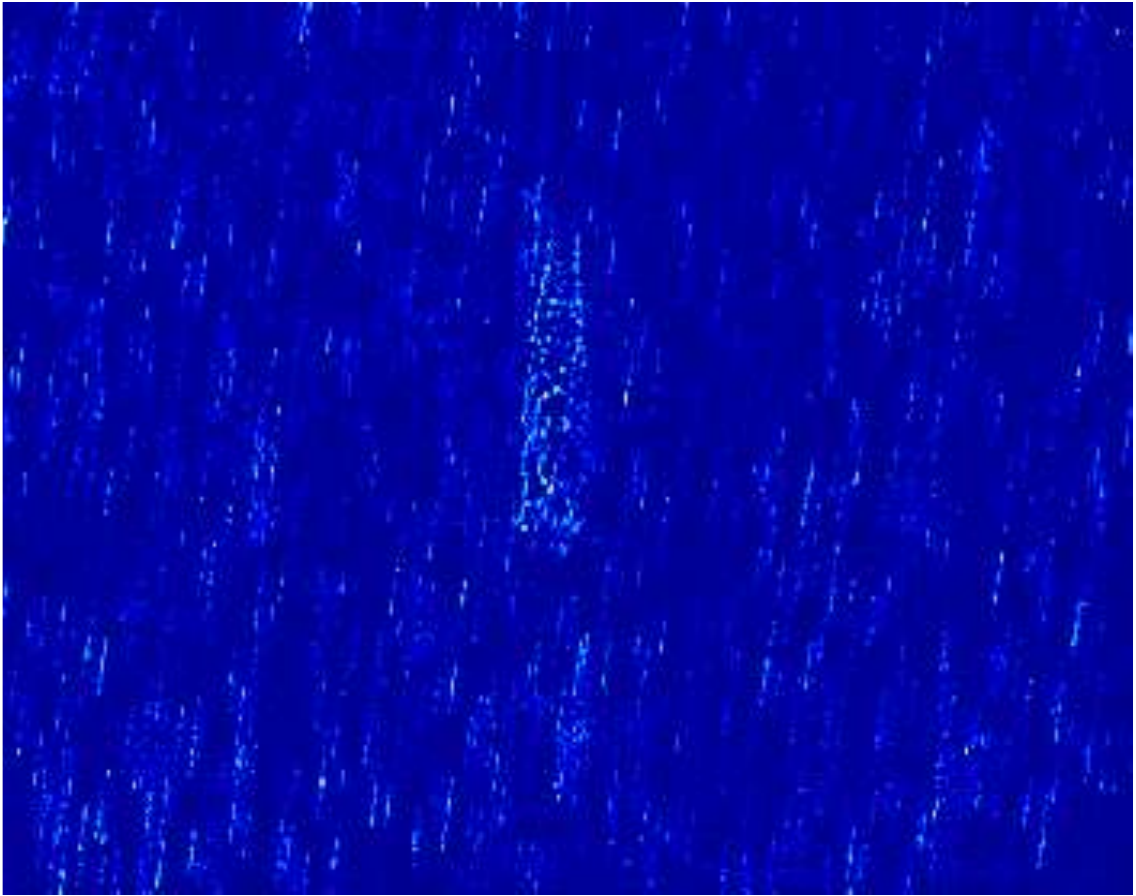} &
\includegraphics[width=0.19\linewidth]{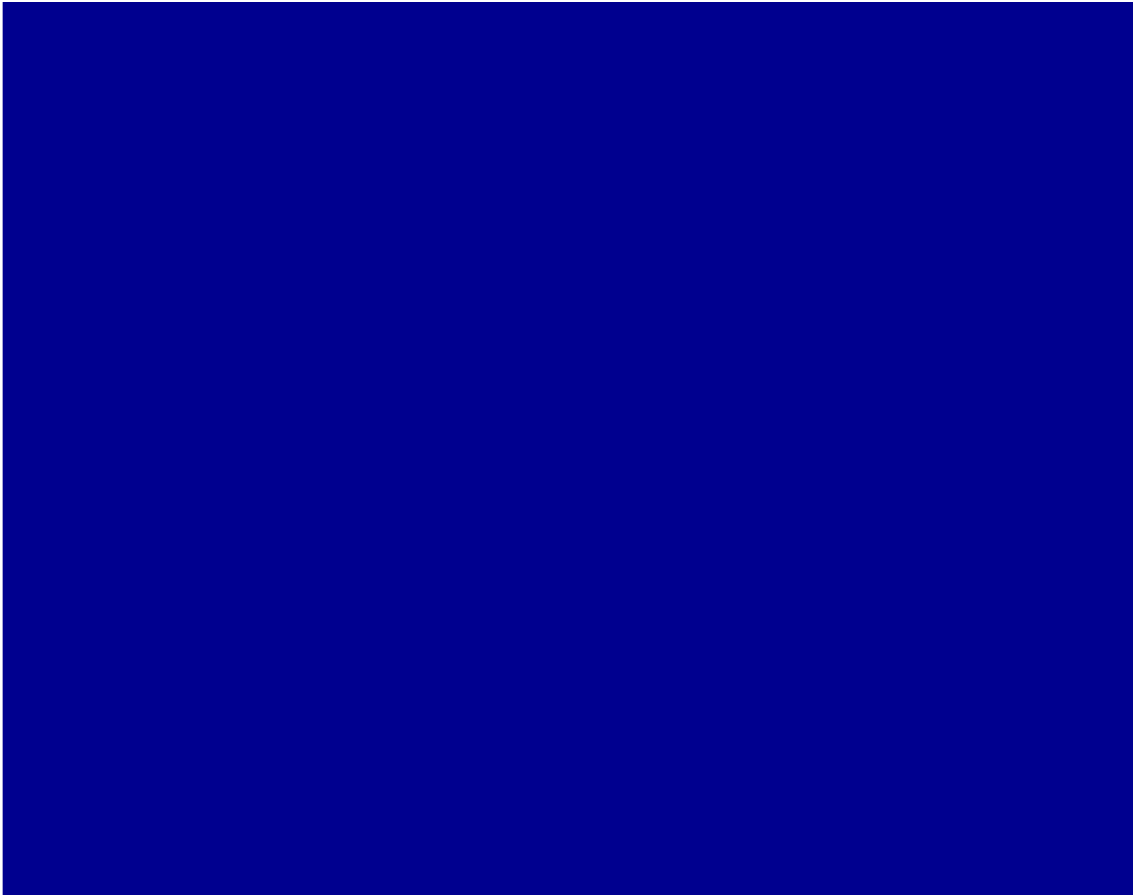} \\
\end{tabular}
\begin{tabular}{ccccccc}\scriptsize
Rainy&TCL & DDN &SE &MS-CSC  & FastDeRain&GT\\
\includegraphics[width=0.135\linewidth]{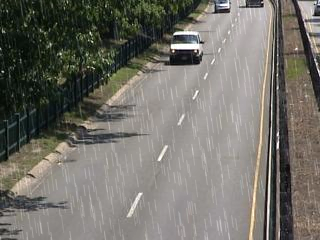} &
\includegraphics[width=0.135\linewidth]{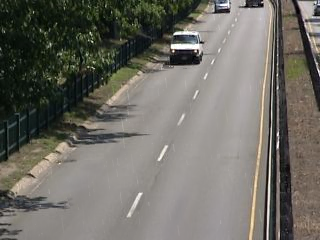} &
\includegraphics[width=0.135\linewidth]{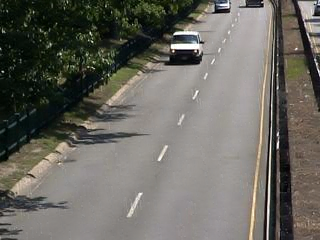} &
\includegraphics[width=0.135\linewidth]{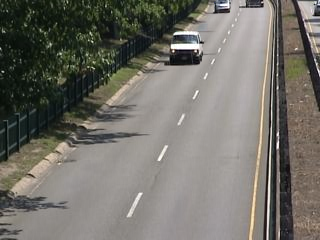} &
\includegraphics[width=0.135\linewidth]{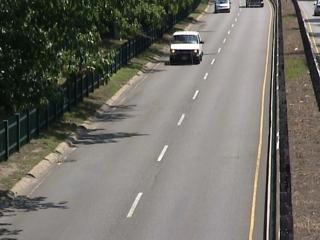} &
\includegraphics[width=0.135\linewidth]{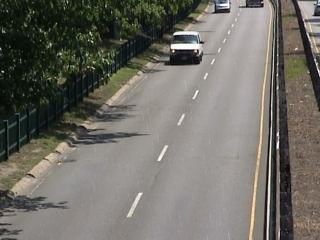} &
\includegraphics[width=0.135\linewidth]{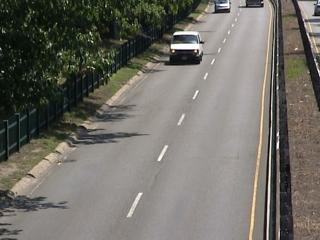} \\
&\includegraphics[width=0.135\linewidth]{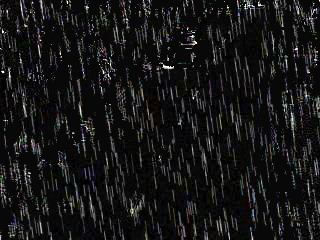} &
\includegraphics[width=0.135\linewidth]{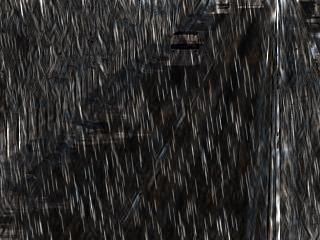} &
\includegraphics[width=0.135\linewidth]{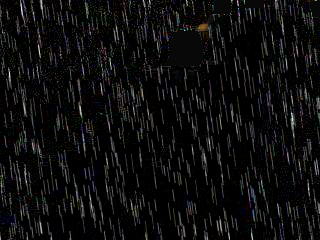} &
\includegraphics[width=0.135\linewidth]{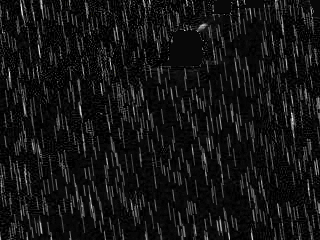} &
\includegraphics[width=0.135\linewidth]{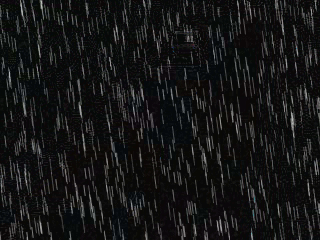} &
\includegraphics[width=0.135\linewidth]{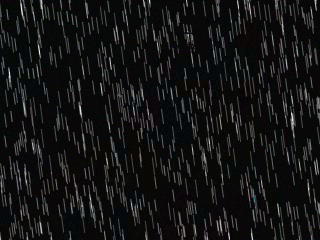} \\
\includegraphics[width=0.135\linewidth]{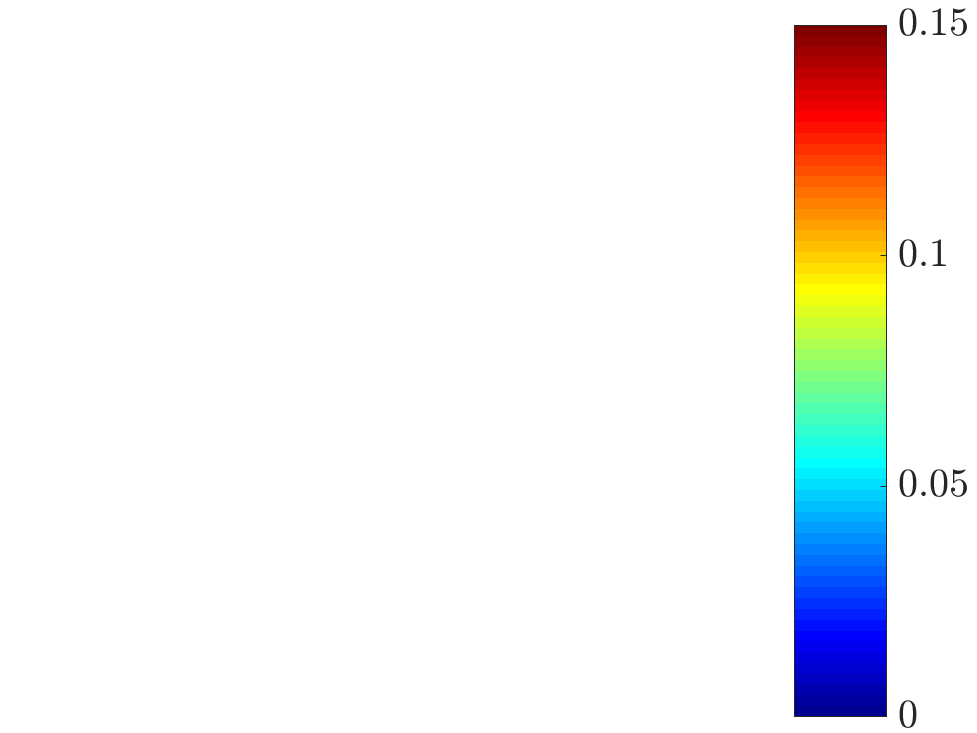}&
\includegraphics[width=0.135\linewidth]{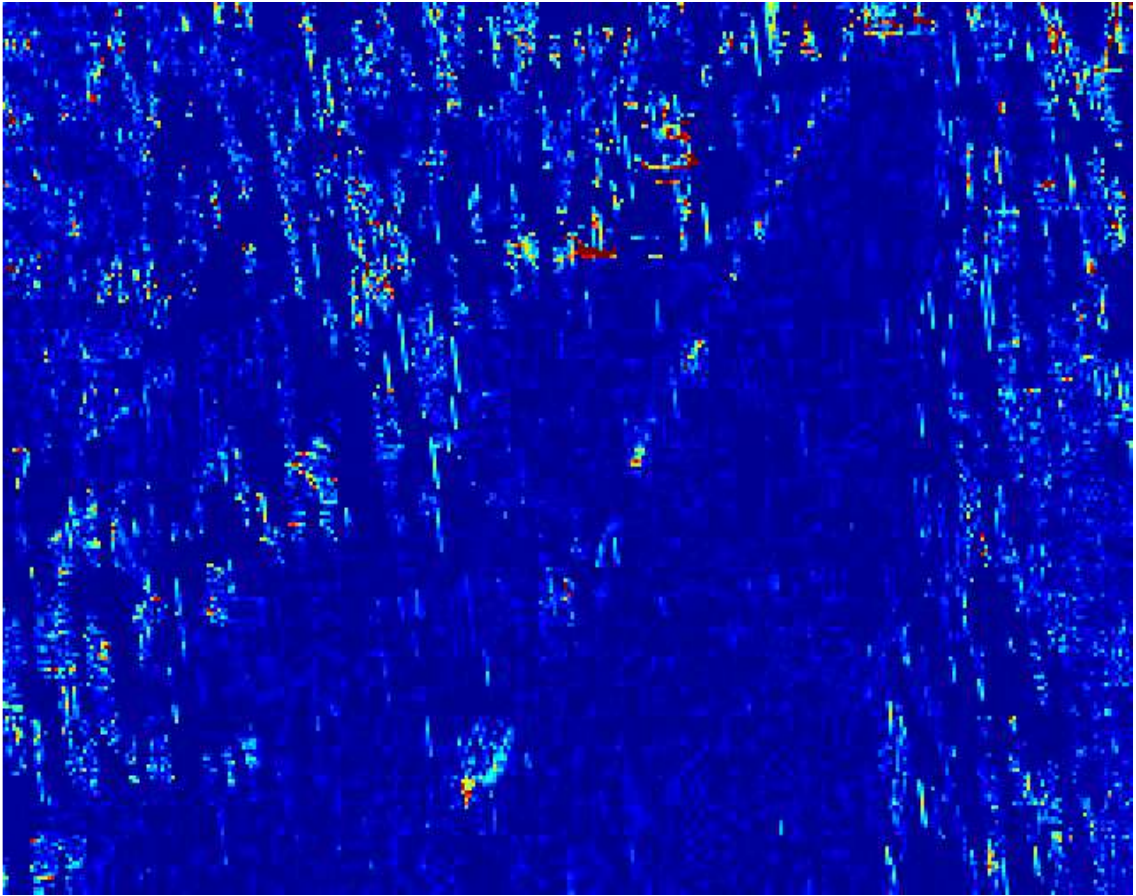}&
\includegraphics[width=0.135\linewidth]{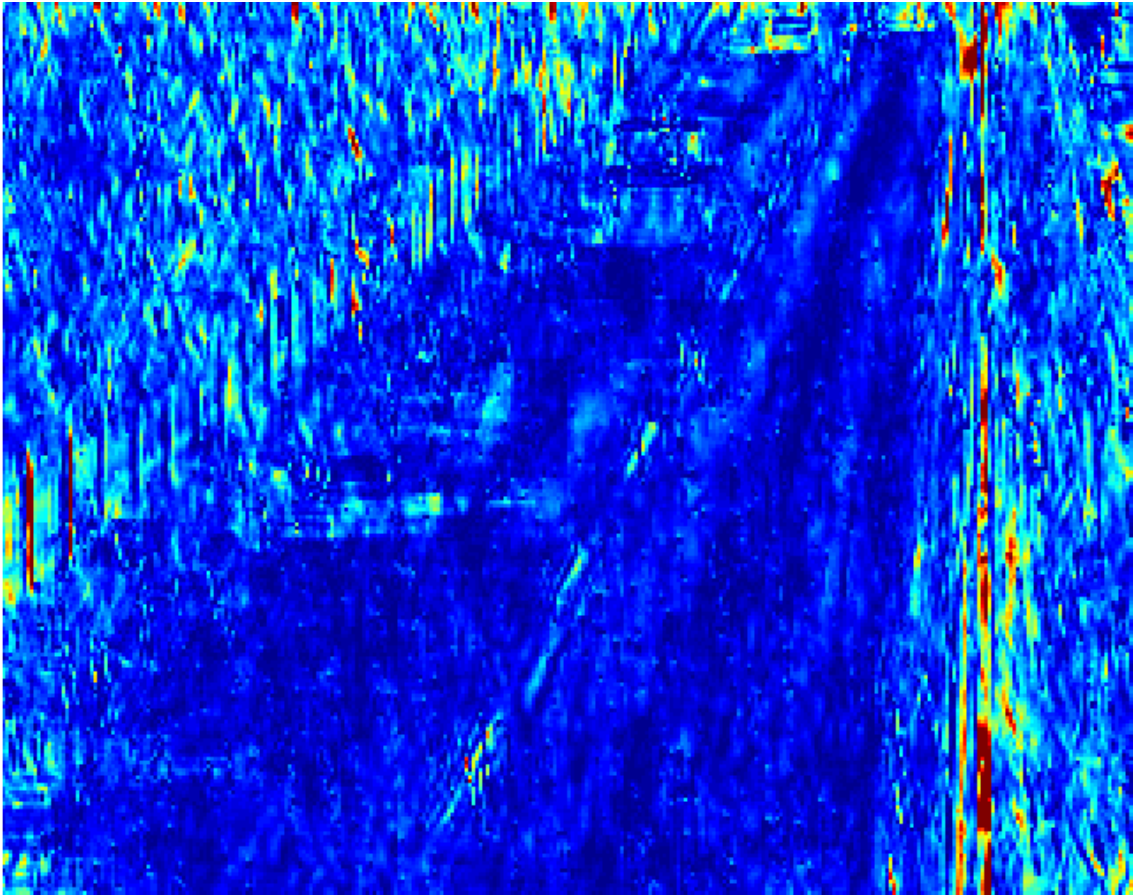}&
\includegraphics[width=0.135\linewidth]{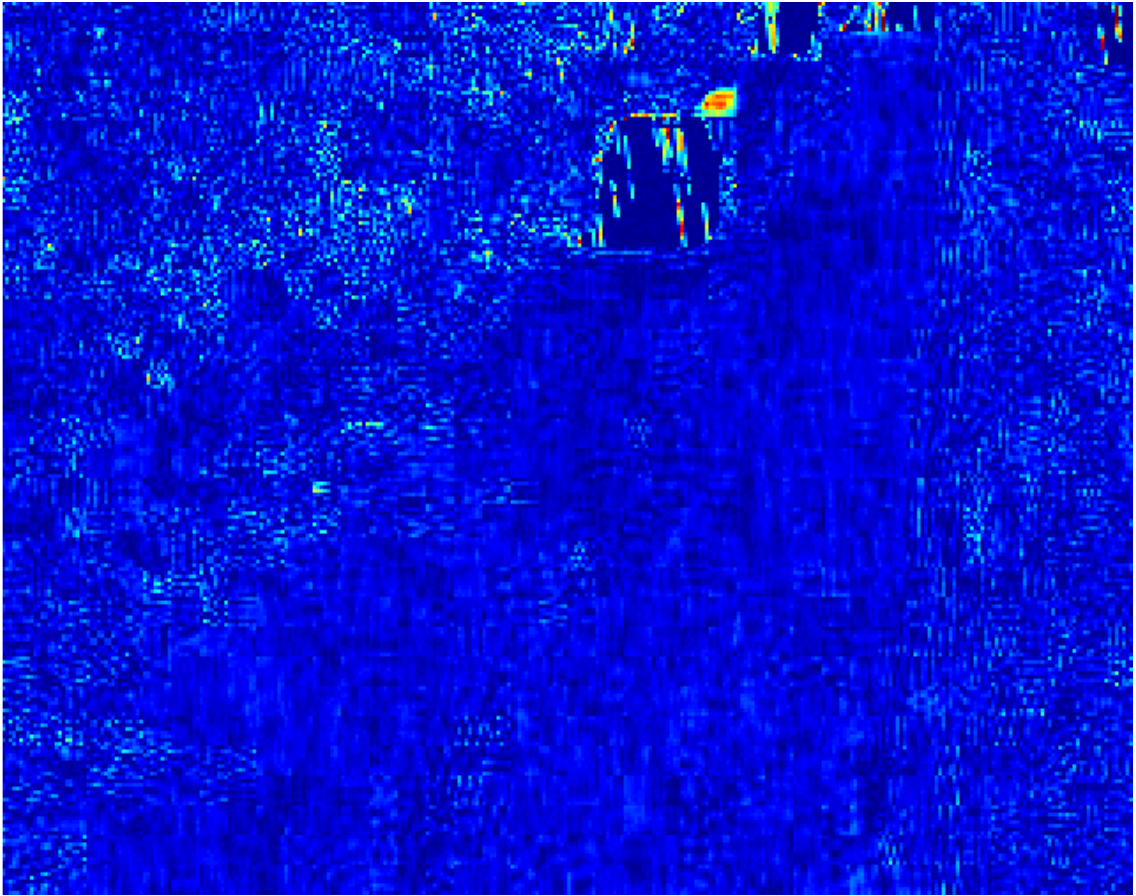}&
\includegraphics[width=0.135\linewidth]{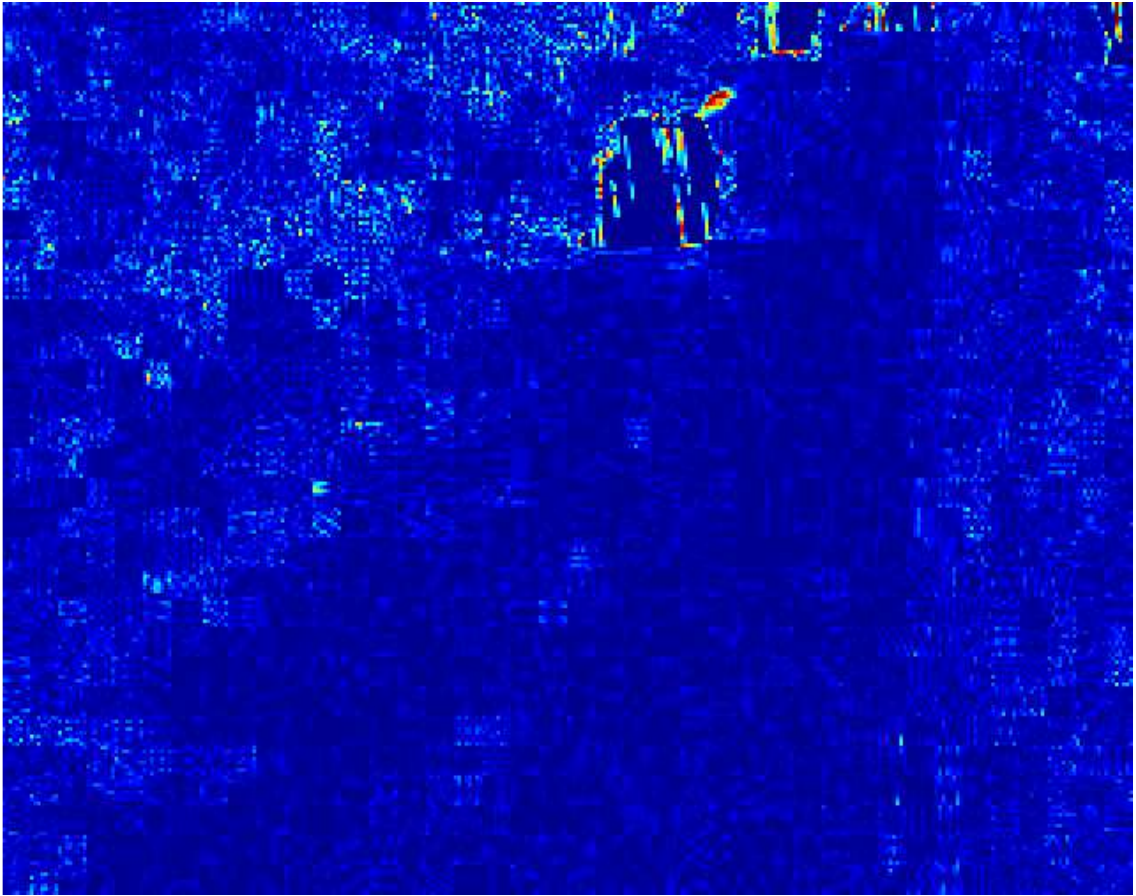}&
\includegraphics[width=0.135\linewidth]{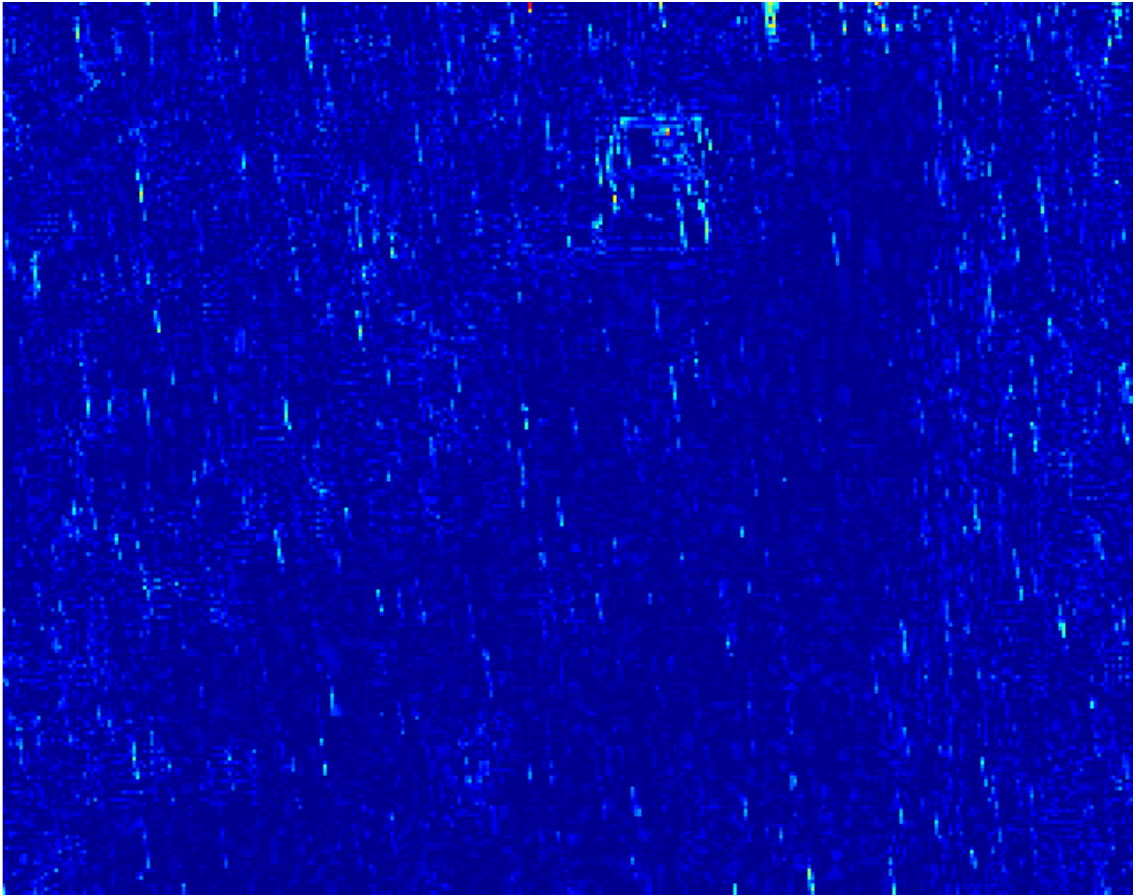} &
\includegraphics[width=0.135\linewidth]{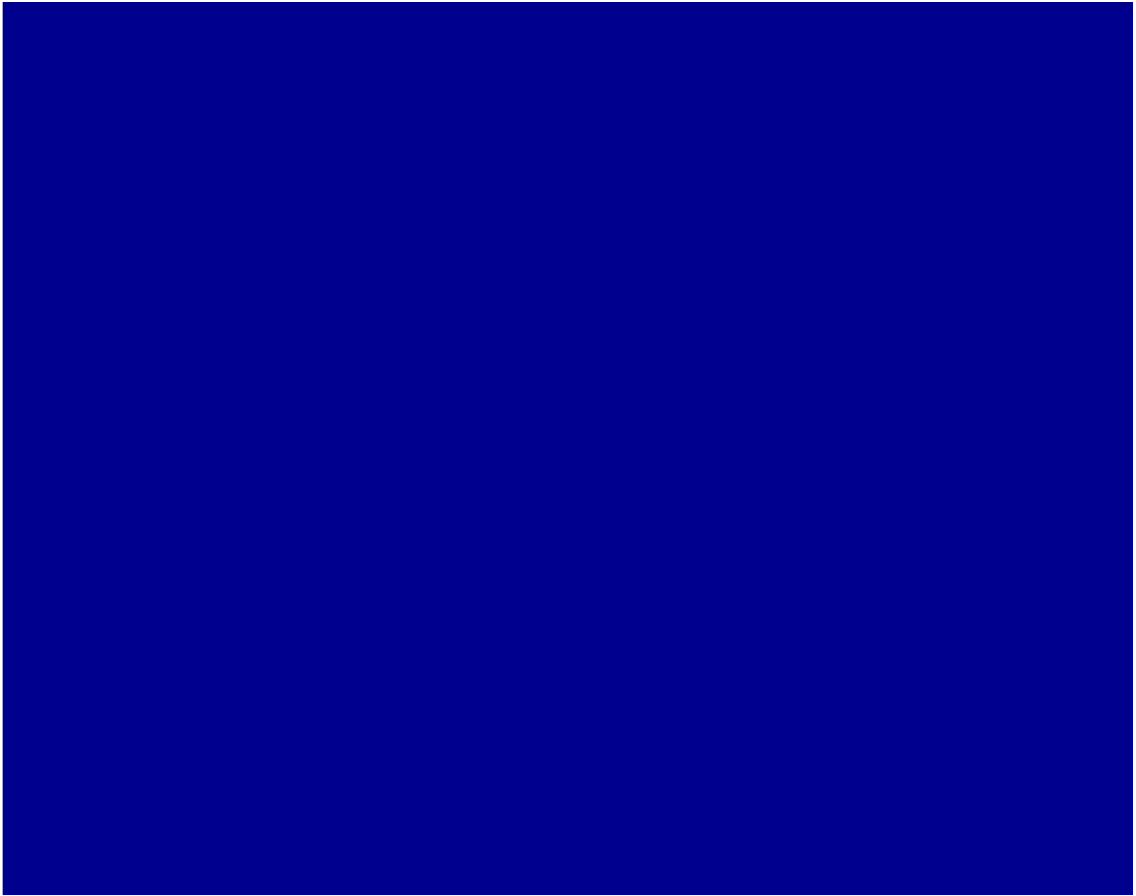} \\

\end{tabular}
\caption{The rainy frame, rain streaks removal results, extracted rain streaks and corresponding error images by different methods with synthetic rain streaks in \textbf{case 1}, respectively. The corresponding videos from top to bottom are the ``'foreman'', ''bus'', ''waterfall'' and ''highway''.
From left to right are: the rainy data (or the color bar), results by TCL \cite{kim2015video}, DDN \cite{fu2017removing}, (SE \cite{Wei_2017_ICCV}, MS-CSC \cite{li2018video},)  FastDeRain, and the ground truth (GT), respectively.}
\label{fig_case1}
\end{figure}

\paragraph{Compared methods}
To validate the effectiveness and efficiency of the proposed method, we compare our method (denoted as ``FastDeRain'') with recent state-of-the-art methods, including one single image based method, i.e.,
Fu {\em et al}.'s deep detail network (DDN) method\footnote{\url{http://smartdsp.xmu.edu.cn/xyfu.html}} \cite{fu2017removing}; and three video-based mehtods, i.e., Kim {\em et al}.'s  method using temporal correlation and low-rankness (TCL) \footnote{\url{http://mcl.korea.ac.kr/\~jhkim/deraining/deraining_code_with_example.zip}}  \cite{kim2015video}, Wei {\em et al}.'s stochastic encoding (SE) method\footnote{\url{http://gr.xjtu.edu.cn/web/dymeng}} \cite{Wei_2017_ICCV}, and Li {\em et al}.'s multiscale convolutional sparse coding (MS-CSC) method\footnote{\url{https://github.com/MinghanLi/MS-CSC-Rain-Streak-Removal}} \cite{li2018video}.
In fact, DDN is a single-image-based rain streak removal method, but their performance has already surpassed some video-based methods.
The deep learning technique shows a great vitality and an extremely wide application prospect.
Hence, the comparison with DNN is reasonable and challenging.

\setcounter{paragraph}{0}

\subsection{Synthetic data}

\begin{table}[!htb] 
\renewcommand\arraystretch{0.9}\setlength{\tabcolsep}{1.5pt}\scriptsize
\caption{Quantitative comparisons of the rain streak removal results of \cite{kim2015video}, \cite{fu2017removing}, \cite{Wei_2017_ICCV}, \cite{li2018video} and the proposed method on synthetic videos. The \textbf{best} quantitative values are in \textbf{boldface}.}
\begin{center}
\begin{threeparttable}
\begin{tabular}{cccccccccc}
\toprule
&  Video & Method & PSNR &  SSIM & FSIM &  VIF & UIQI&  GMSD &  Time \\  \midrule

\multirow{20}{*}{\rotatebox[origin=c]{90}{\textbf{Case 1}}}  &\multirow{6}{*}{\em foreman}
                        &Rainy                                   & 34.67 & 0.9541 & 0.9723 & 0.6787 & 0.8693 & 0.0524 & | \\
                        &&TCL \cite{kim2015video}                & 33.86 & 0.9612 & 0.9716 & 0.6431 & 0.8917 & 0.0400 & 1696.4 \\
                        &&DDN \cite{fu2017removing}              & 34.25 & 0.9730 & 0.9804 & 0.7253 & 0.9151 & 0.0300 & 71.2 \\
                        &&\textcolor{gray}{SE \cite{Wei_2017_ICCV}                }   & \textcolor{gray}{21.95} & \textcolor{gray}{0.6959} & \textcolor{gray}{0.7994} & \textcolor{gray}{0.3060} & \textcolor{gray}{0.4125} & \textcolor{gray}{0.1997} & \textcolor{gray}{740.8}  \\
                        &&\textcolor{gray}{MS-CSC \cite{li2018video}}     & \textcolor{gray}{26.61} & \textcolor{gray}{0.7922} & \textcolor{gray}{0.8772} & \textcolor{gray}{0.3754} & \textcolor{gray}{0.5895} & \textcolor{gray}{0.1470} & \textcolor{gray}{143.9} \\
                        &&FastDeRain                             & \bf37.57 & \bf0.9805 & \bf0.9867 & \bf0.7757 & \bf0.9364 & \bf0.0230 & \bf2.4 \\ \cmidrule{2-10}

                        &\multirow{4}{*}{``bus''}
                        &Rainy                                   & 31.01 & 0.9146 & 0.9664 & 0.6269 & 0.8800 & 0.0725 & | \\
                        &&TCL \cite{kim2015video}                & 33.06 & 0.9562 & 0.9744 & 0.6873 & 0.9329 & 0.0360 & 2429.2 \\
                        &&DDN \cite{fu2017removing}              & 31.08 & 0.9534 & 0.9714 & 0.6626 & 0.9254 & 0.0399 & 46.1 \\
                        &&FastDeRain                             & \bf35.96 & \bf0.9729 & \bf0.9849 & \bf0.7886 & \bf0.9552 & \bf0.0292 & \bf7.1 \\ \cmidrule{2-10}

                        &\multirow{4}{*}{``waterfall''}
                        &Rainy                                   & 31.63 & 0.9097 & 0.9550 & 0.5956 & 0.8834 & 0.0617 & | \\
                        &&TCL \cite{kim2015video}                & 35.57 & 0.9578 & 0.9726 & 0.7297 & 0.9426 & 0.0242 & 2338.7 \\
                        &&DDN \cite{fu2017removing}              & 32.70 & 0.9517 & 0.9677 & 0.6580 & 0.9287 & 0.0407 & 43.6 \\
                        &&FastDeRain                             & \bf40.52 & \bf0.9842 & \bf0.9900 & \bf0.8588 & \bf0.9787 & \bf0.0106 & \bf9.3 \\ \cmidrule{2-10}

                        &\multirow{6}{*}{``highway''}
                        &Rainy                                   & 30.94 & 0.8592 & 0.9411 & 0.5279 & 0.7169 & 0.0974 & | \\
                        &&TCL \cite{kim2015video}                & 34.58 & 0.9639 & 0.9728 & 0.7063 & 0.8840 & 0.0277 & 2127.3 \\
                        &&DDN \cite{fu2017removing}              & 29.59 & 0.9308 & 0.9521 & 0.6089 & 0.8074 & 0.0534 & 43.3 \\
                        &&SE \cite{Wei_2017_ICCV}                & 35.09 & 0.9730 & 0.9818 & 0.7878 & 0.9041 & 0.0127 & 656.3 \\
                        &&MS-CSC \cite{li2018video}              & 37.46 & 0.9753 & 0.9818 & 0.8173 & 0.9193 & 0.0143 & 280.5 \\
                        &&FastDeRain                             & \bf41.12 & \bf0.9829 & \bf0.9902 & \bf0.8491 & \bf0.9263 & \bf0.0117 & \bf5.2 \\ \midrule

\multirow{18}{*}{\rotatebox[origin=c]{90}{\textbf{Case 2}}}  &\multirow{4}{*}{``foreman''}
                        &Rainy                                   & 28.87 & 0.8991 & 0.9410 & 0.5535 & 0.7902 & 0.0922 & | \\
                        &&TCL \cite{kim2015video}                & 30.75 & 0.9234 & 0.9486 & 0.5078 & 0.8186 & 0.0584 & 2625.8 \\
                        &&DDN \cite{fu2017removing}              & 33.21 & 0.9526 & 0.9671 & 0.6252 & 0.8634 & 0.0494 & 66.6 \\
                        &&FastDeRain                             & \bf35.58 & \bf0.9694 & \bf0.9777 & \bf0.7314 & \bf0.9084 & \bf0.0306 & \bf3.0 \\ \cmidrule{2-10}

                        &\multirow{4}{*}{``bus''}
                        &Rainy                                   & 26.15 & 0.8238 & 0.9300 & 0.4951 & 0.7808 & 0.1150 & | \\
                        &&TCL \cite{kim2015video}                & 28.08 & 0.8669 & 0.9341 & 0.4557 & 0.8119 & 0.0838 & 3394.4 \\
                        &&DDN \cite{fu2017removing}              & 29.42 & 0.9171 & 0.9507 & 0.5468 & 0.8747 & 0.0644 & 44.9 \\
                        &&FastDeRain                             & \bf32.32 & \bf0.9375 & \bf0.9673 & \bf0.6552 & \bf0.8992 & \bf0.0496 & \bf6.7 \\ \cmidrule{2-10}

                        &\multirow{4}{*}{``waterfall''}
                        &Rainy                                   & 26.11 & 0.7827 & 0.8986 & 0.4198 & 0.7382 & 0.1096 & | \\
                        &&TCL \cite{kim2015video}                & 29.14 & 0.8457 & 0.9210 & 0.4217 & 0.8041 & 0.0796 & 2880.3 \\
                        &&DDN \cite{fu2017removing}              & 30.44 & 0.8929 & 0.9370 & 0.4882 & 0.8546 & 0.0699 & 42.0 \\
                        &&FastDeRain                             & \bf36.18 & \bf0.9556 & \bf0.9751 & \bf0.7117 & \bf0.9405 & \bf0.0246 & \bf6.5 \\ \cmidrule{2-10}

                        &\multirow{6}{*}{``highway''}
                        &Rainy                                   & 28.73 & 0.8772 & 0.9427 & 0.5320 & 0.6963 & 0.1014 & | \\
                        &&TCL \cite{kim2015video}                & 31.78 & 0.9333 & 0.9543 & 0.5481 & 0.7728 & 0.0472 & 2176.1 \\
                        &&DDN \cite{fu2017removing}              & 31.22 & 0.9407 & 0.9512 & 0.5861 & 0.7922 & 0.0569 & 43.9 \\
                        &&SE \cite{Wei_2017_ICCV}                & 30.21 & 0.9681 & 0.9819 & 0.7851 & 0.8970 & 0.0137 & 750.1 \\
                        &&MS-CSC \cite{li2018video}              & 31.79 & 0.9686 & 0.9811 & 0.7959 & 0.8970 & 0.0141 & 317.3 \\
                        &&FastDeRain                             & \bf37.99 & \bf0.9809 & \bf0.9846 & \bf0.8422 & \bf0.9219 & \bf0.0119 & \bf5.3 \\ \midrule

\multirow{18}{*}{\rotatebox[origin=c]{90}{\textbf{Case 3}}}  &\multirow{4}{*}{``foreman''}
                        &Rainy                                   & 23.75 & 0.9301 & 0.9631 & 0.6409 & 0.8355 & 0.0740 & | \\
                        &&TCL \cite{kim2015video}                & 25.13 & 0.9321 & 0.9559 & 0.5627 & 0.8430 & 0.0582 & 1991.6 \\
                        &&DDN \cite{fu2017removing}              & 26.62 & 0.9586 & 0.9735 & 0.6756 & 0.8753 & 0.0487 & 66.0\\
                        &&FastDeRain                             & \bf27.88 & \bf0.9716 & \bf0.9821 & \bf0.7483 & \bf0.9096 & \bf0.0261 & \bf4.2 \\ \cmidrule{2-10}

                        &\multirow{4}{*}{``bus''}
                        &Rainy                                   & 22.87 & 0.9101 & 0.9612 & 0.6597 & 0.8643 & 0.1067 & | \\
                        &&TCL \cite{kim2015video}                & 25.84 & 0.8965 & 0.9485 & 0.5555 & 0.8373 & 0.0813 & 2969.7 \\
                        &&DDN \cite{fu2017removing}              & 25.73 & 0.9363 & 0.9640 & 0.6434 & 0.8896 & 0.0770 & 41.3 \\
                        &&FastDeRain                             & \bf27.94 & \bf0.9611 & \bf0.9788 & \bf0.7544 & \bf0.9337 & \bf0.0447 & \bf7.0 \\  \cmidrule{2-10}

                        &\multirow{4}{*}{``waterfall''}
                        &Rainy                                   & 22.34 & 0.9235 & 0.9587 & 0.6525 & 0.9016 & 0.0682 & | \\
                        &&TCL \cite{kim2015video}                & 24.21 & 0.9226 & 0.9518 & 0.6205 & 0.9063 & 0.0463 & 2483.6 \\
                        &&DDN \cite{fu2017removing}              & 24.75 & 0.9417 & 0.9634 & 0.6533 & 0.9198 & 0.0566 & 41.1 \\
                        &&FastDeRain                             & \bf26.20 & \bf0.9701 & \bf0.9825 & \bf0.8022 & \bf0.9652 & \bf0.0240 & \bf6.7 \\ \cmidrule{2-10}

                        &\multirow{6}{*}{``highway''}
                        &Rainy                                   & 22.90 & 0.9212 & 0.9702 & 0.6611 & 0.7650 & 0.0683 & | \\
                        &&TCL \cite{kim2015video}                & 24.10 & 0.9358 & 0.9658 & 0.6437 & 0.7889 & 0.0401 & 2012.8 \\
                        &&DDN \cite{fu2017removing}              & 25.06 & 0.9362 & 0.9566 & 0.6339 & 0.7890 & 0.0551 & 41.2 \\
                        &&SE \cite{Wei_2017_ICCV}                & 23.78 & 0.9530 & 0.9805 & 0.7947 & 0.8891 & 0.0145 & 659.1 \\
                        &&MS-CSC \cite{li2018video}              & 24.19 & 0.9531 & 0.9797 & 0.8075 & 0.8903 & 0.0160 & 251.6 \\
                        &&FastDeRain                             & \bf30.17 & \bf0.9720 & \bf0.9838 & \bf0.8191 & \bf0.8951 & \bf0.0135 & \bf5.7 \\ \bottomrule
\end{tabular}
\end{threeparttable}
\end{center}
\label{QC}
\end{table}

\begin{figure}[!htb] 
\centering\scriptsize\renewcommand\arraystretch{1}
\setlength{\tabcolsep}{1pt}
\begin{tabular}{cccccc}

Rainy &TCL & DDN & FastDeRain&GT\\
\includegraphics[width=0.19\linewidth]{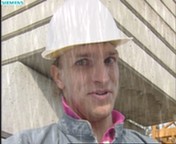} &
\includegraphics[width=0.19\linewidth]{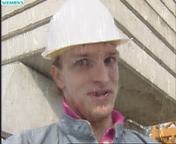} &
\includegraphics[width=0.19\linewidth]{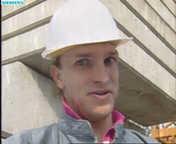} &
\includegraphics[width=0.19\linewidth]{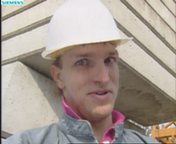} &
\includegraphics[width=0.19\linewidth]{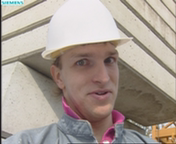} \\
&\includegraphics[width=0.19\linewidth]{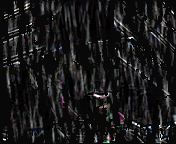} &
\includegraphics[width=0.19\linewidth]{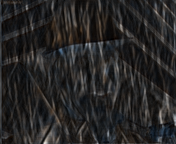} &
\includegraphics[width=0.19\linewidth]{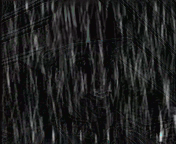} &
\includegraphics[width=0.19\linewidth]{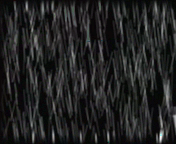} \\
\includegraphics[width=0.19\linewidth]{figs/component/bar_yuv.png}&
\includegraphics[width=0.19\linewidth]{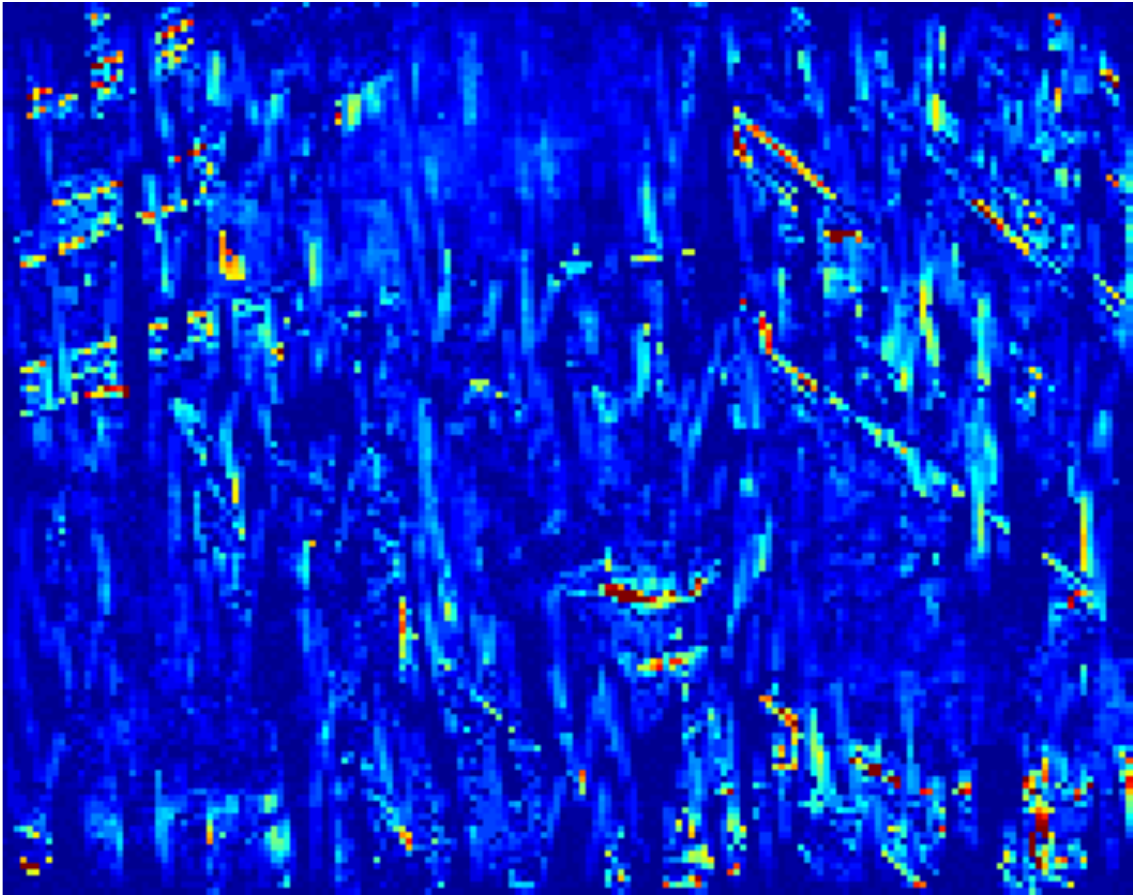}&
\includegraphics[width=0.19\linewidth]{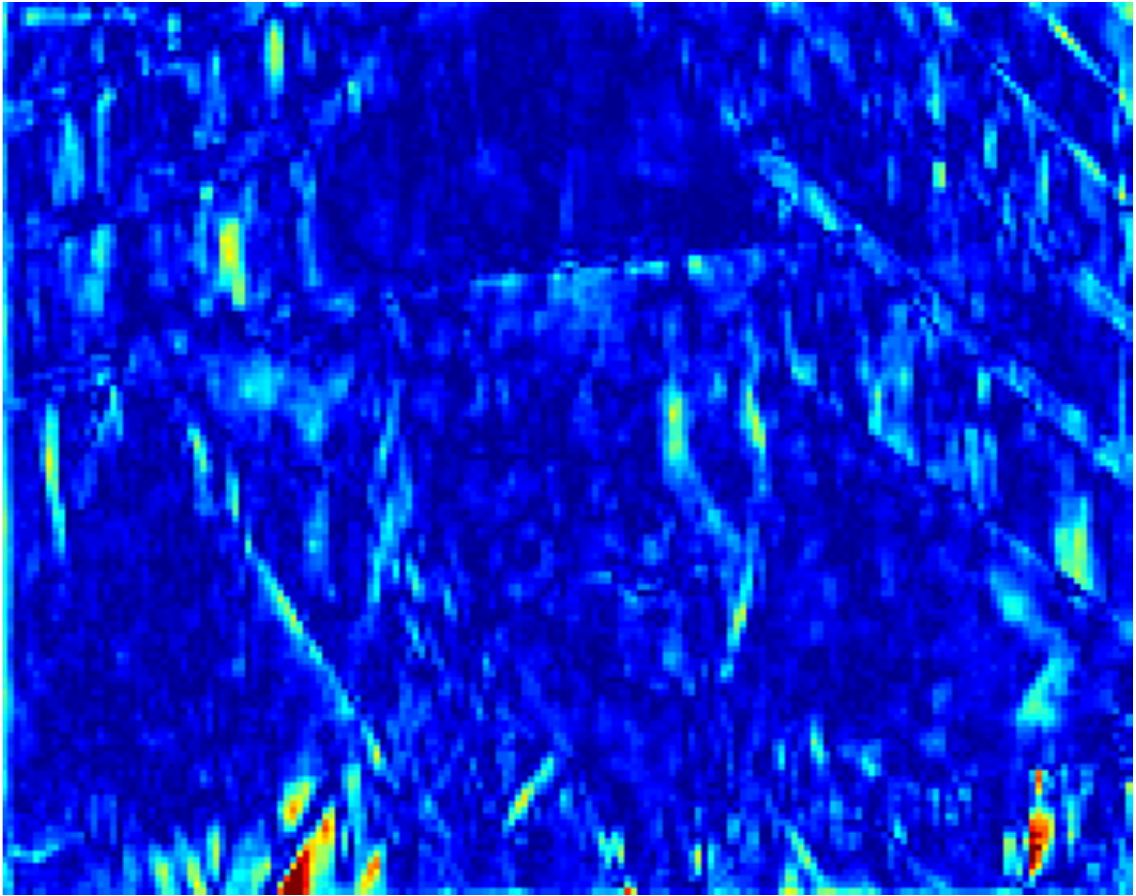}&
\includegraphics[width=0.19\linewidth]{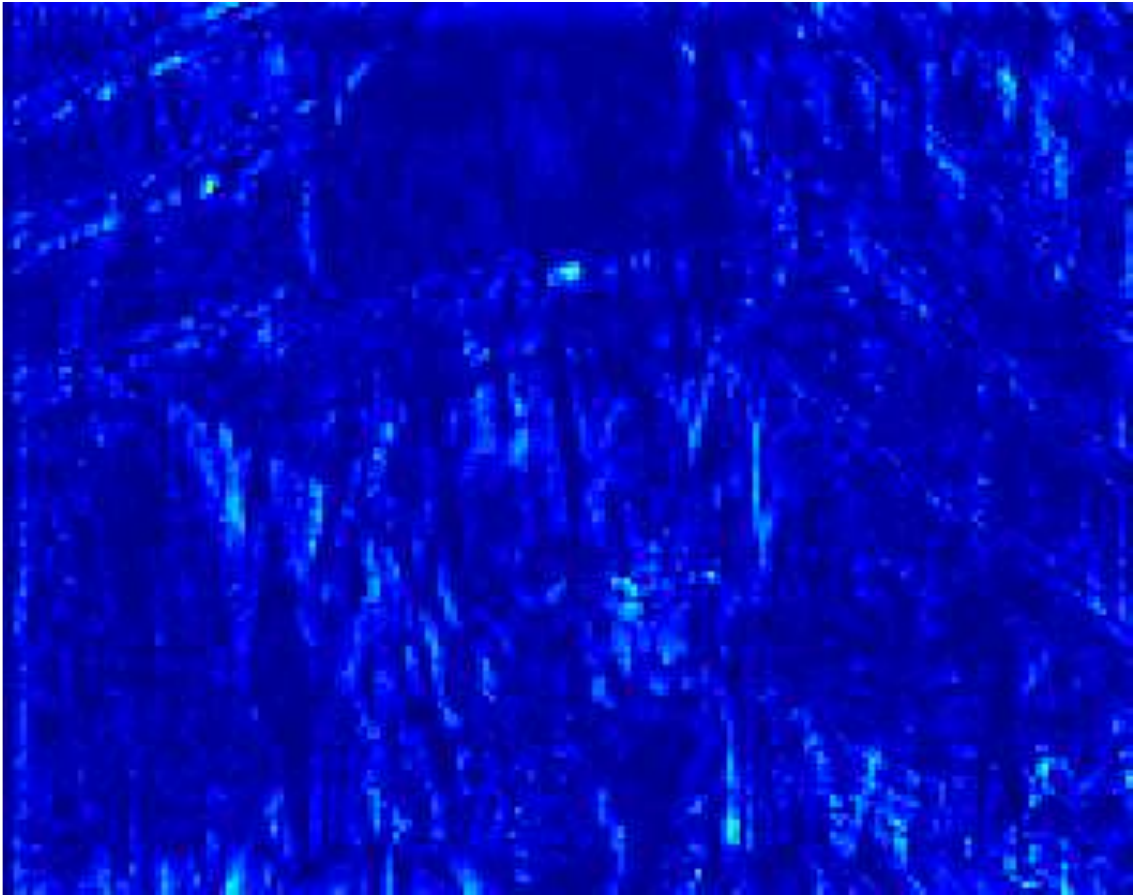} &
\includegraphics[width=0.19\linewidth]{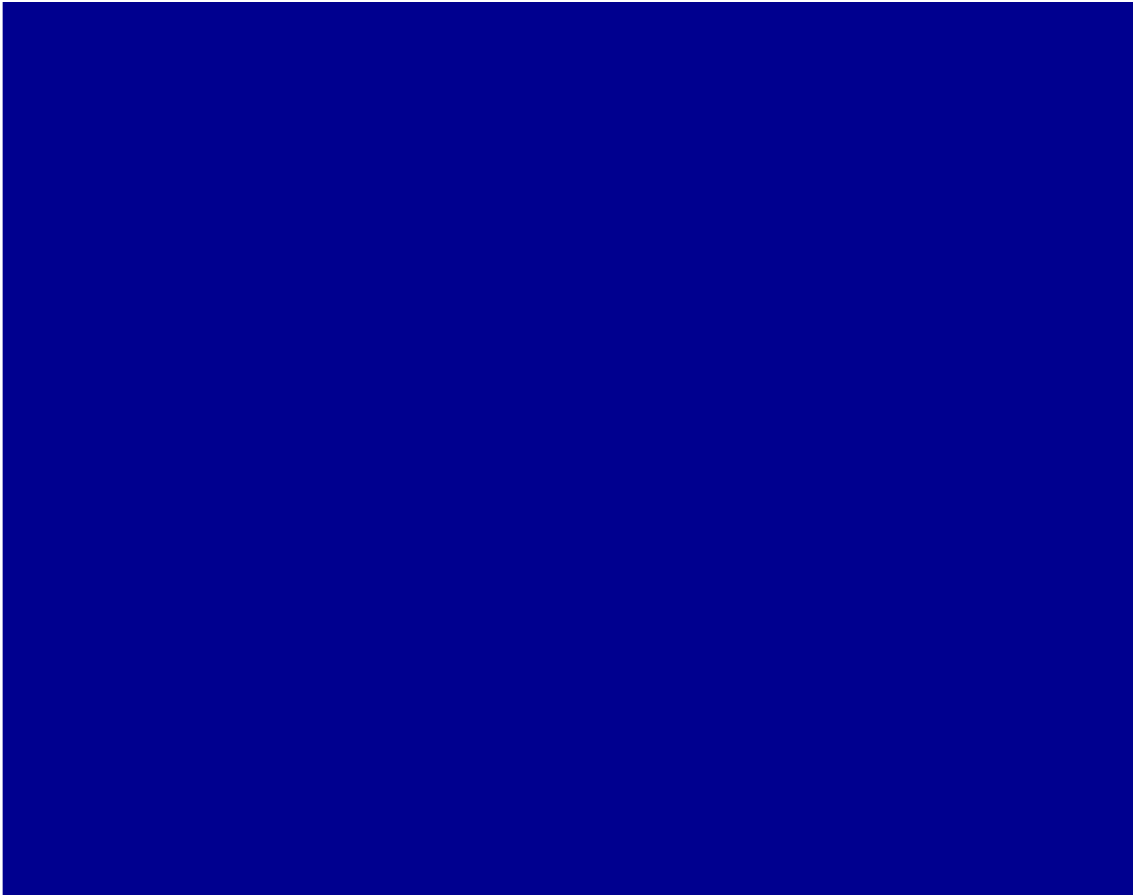} \\

Rainy &TCL & DDN & FastDeRain&GT\\
\includegraphics[width=0.19\linewidth]{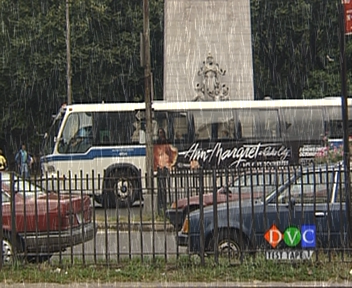} &
\includegraphics[width=0.19\linewidth]{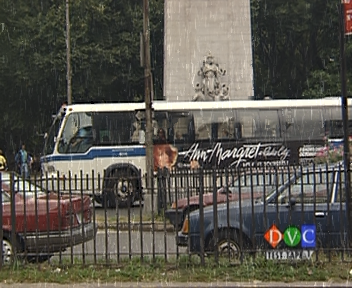} &
\includegraphics[width=0.19\linewidth]{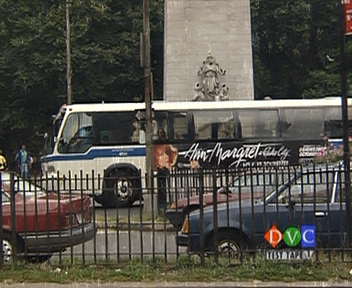} &
\includegraphics[width=0.19\linewidth]{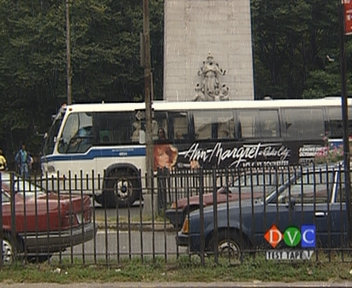} &
\includegraphics[width=0.19\linewidth]{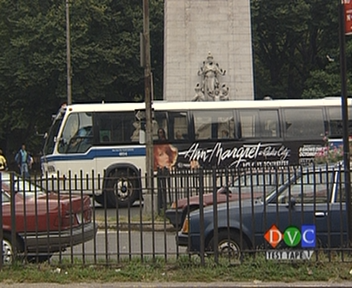} \\
&\includegraphics[width=0.19\linewidth]{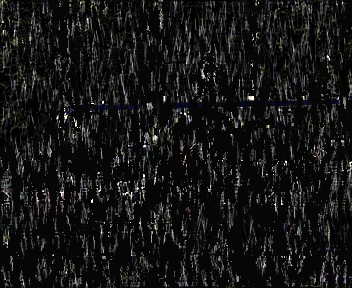} &
\includegraphics[width=0.19\linewidth]{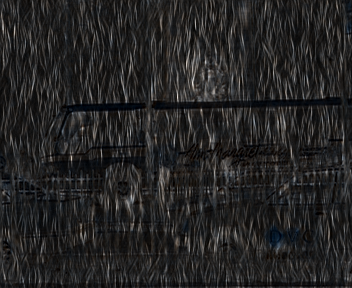} &
\includegraphics[width=0.19\linewidth]{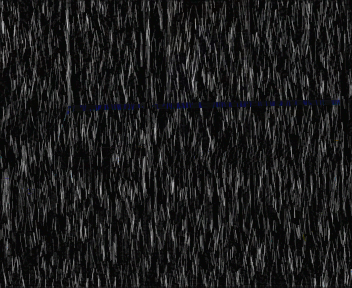} &
\includegraphics[width=0.19\linewidth]{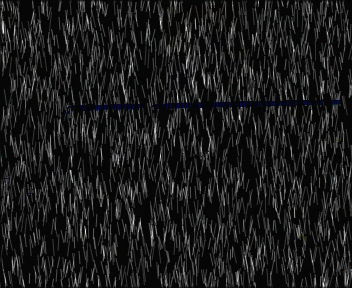} \\
\includegraphics[width=0.19\linewidth]{figs/component/bar_yuv.png}&
\includegraphics[width=0.19\linewidth]{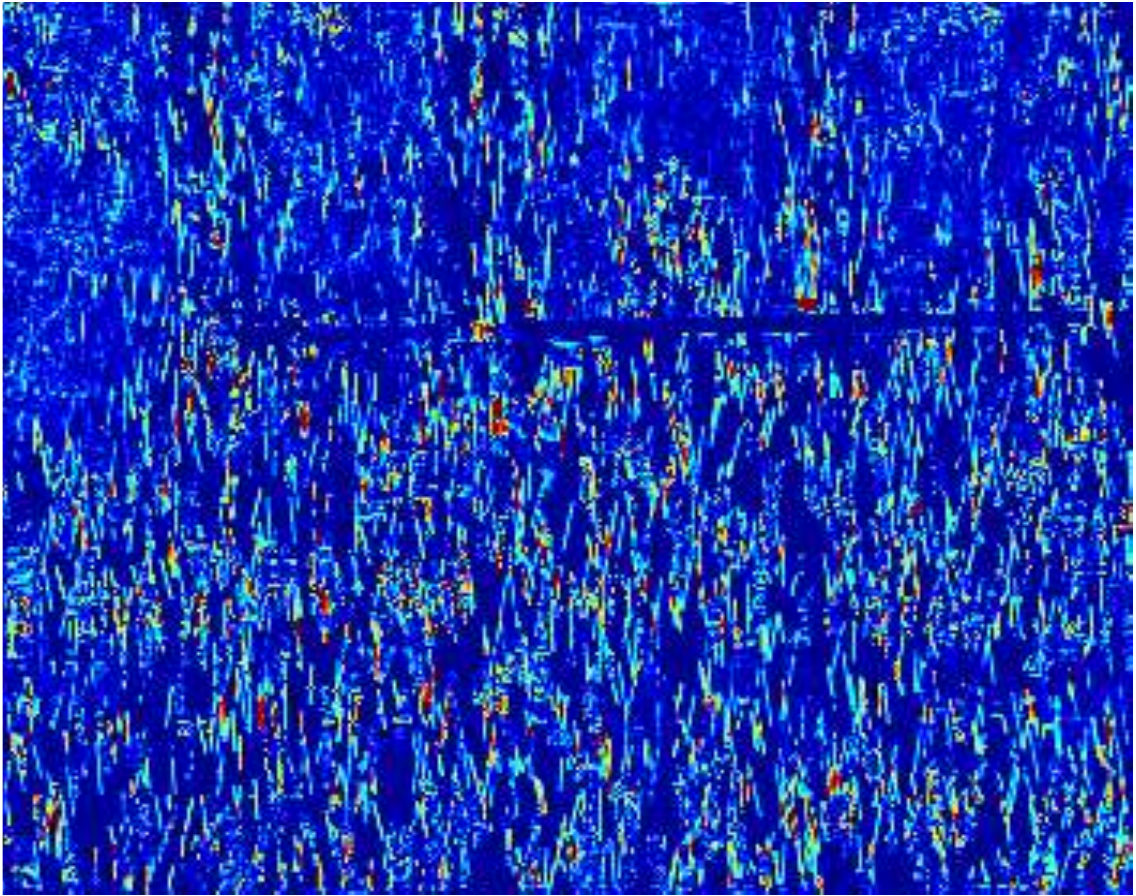}&
\includegraphics[width=0.19\linewidth]{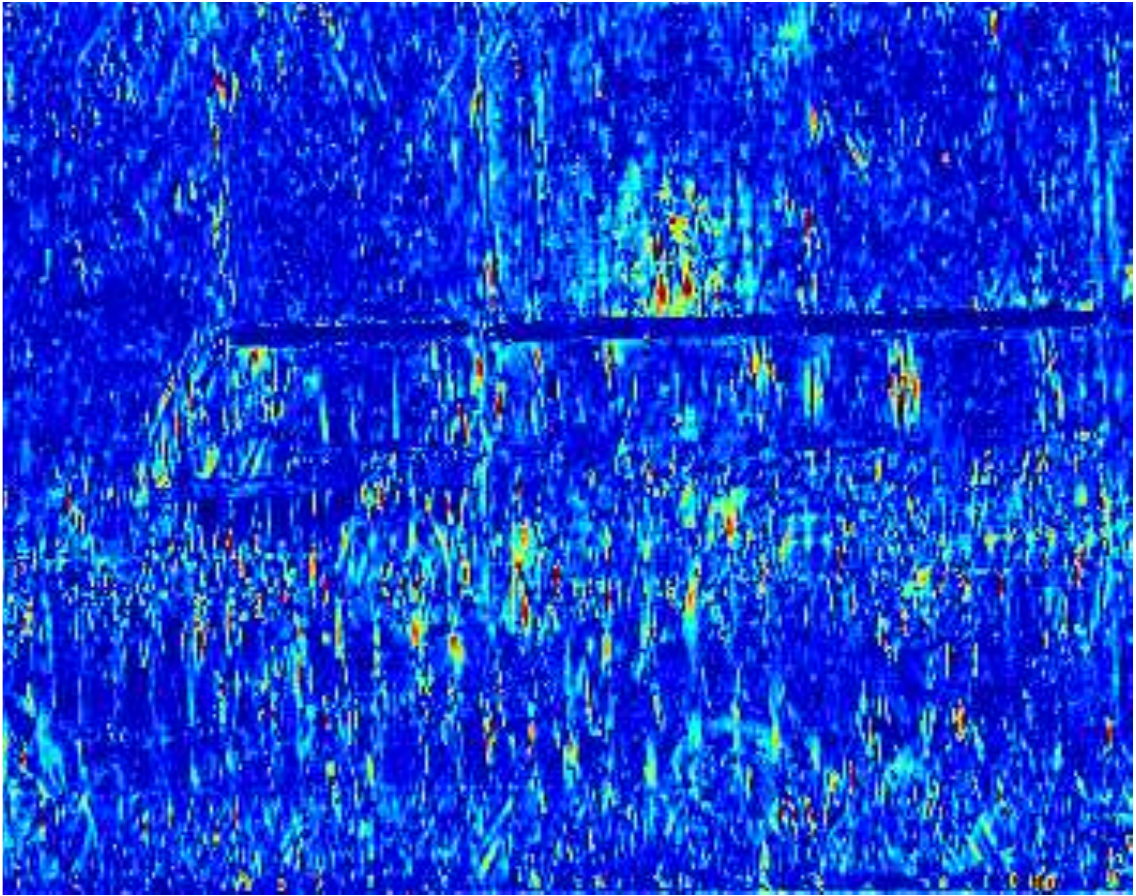}&
\includegraphics[width=0.19\linewidth]{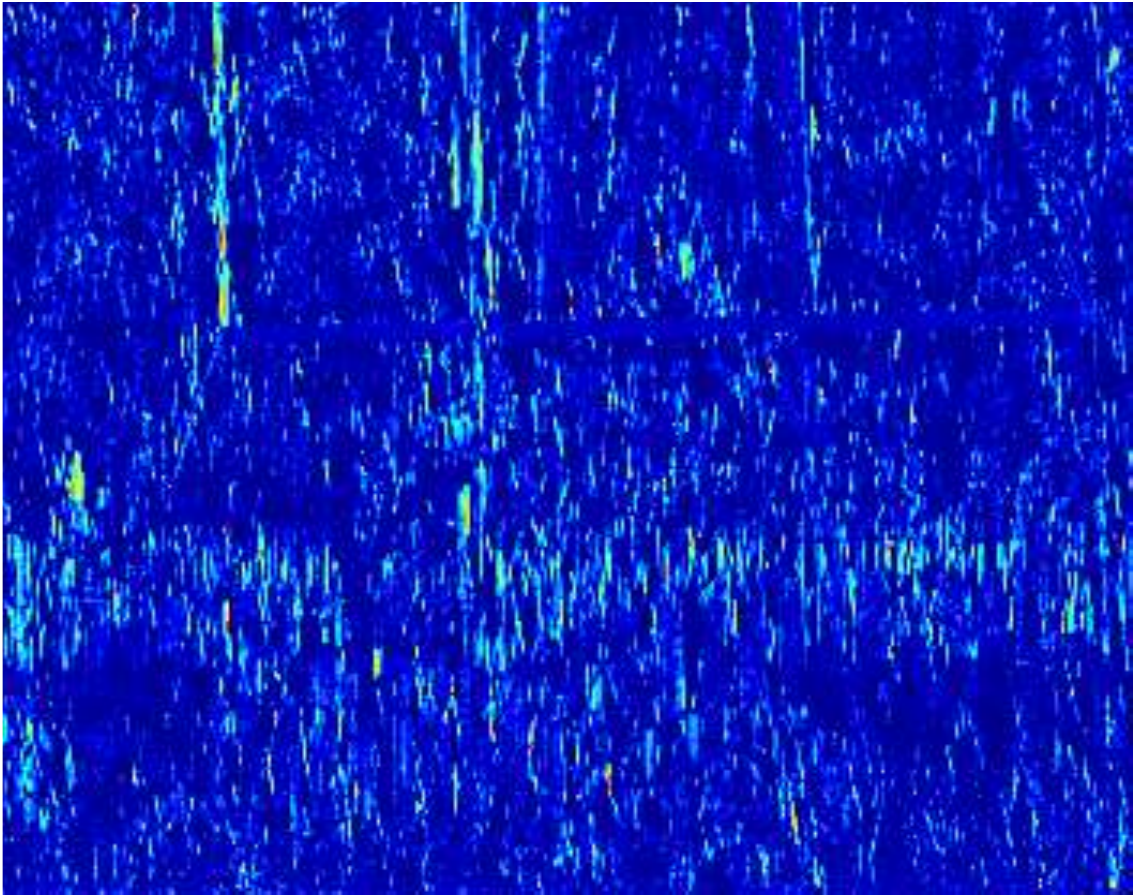} &
\includegraphics[width=0.19\linewidth]{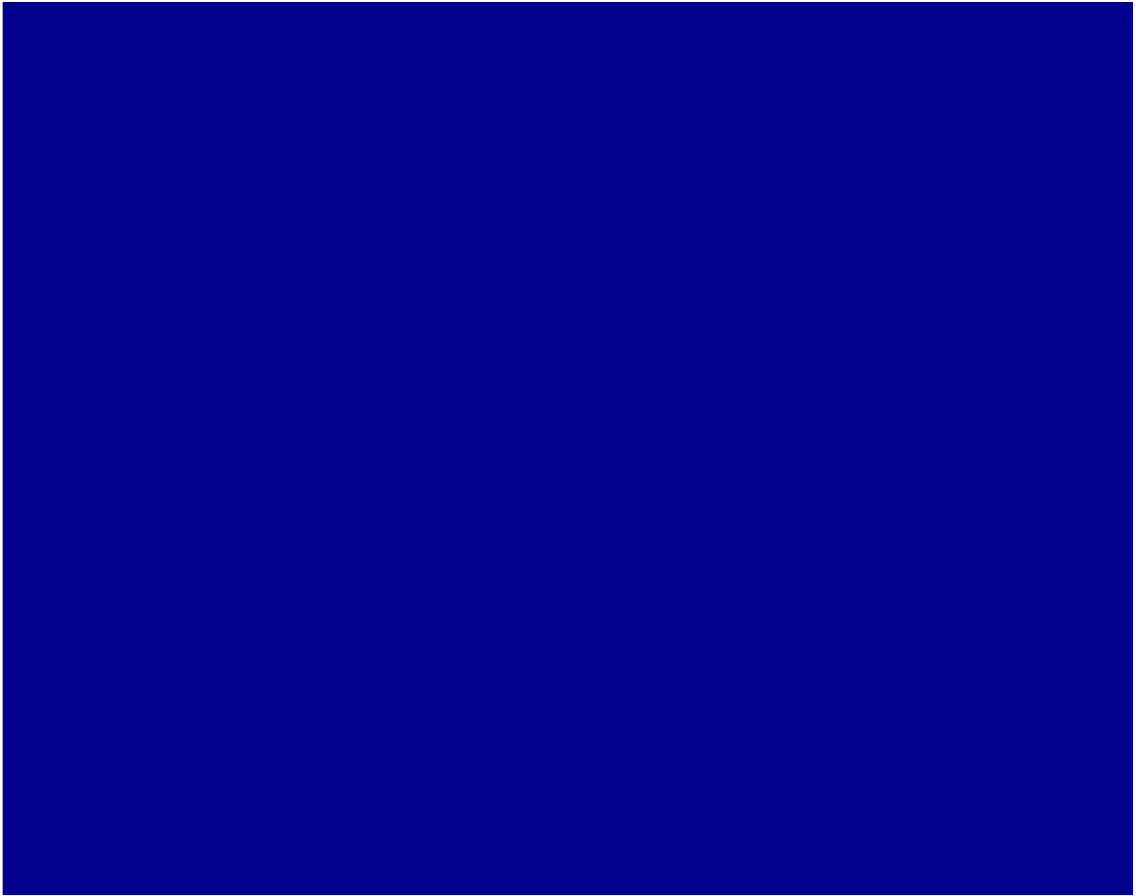} \\

Rainy &TCL & DDN & FastDeRain&GT\\
\includegraphics[width=0.19\linewidth]{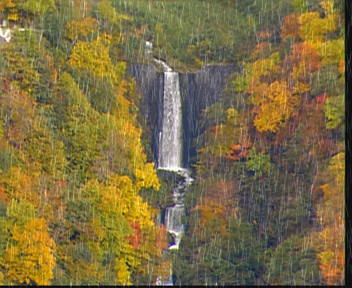} &
\includegraphics[width=0.19\linewidth]{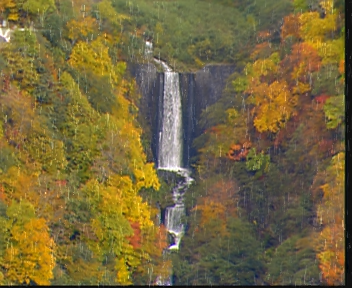} &
\includegraphics[width=0.19\linewidth]{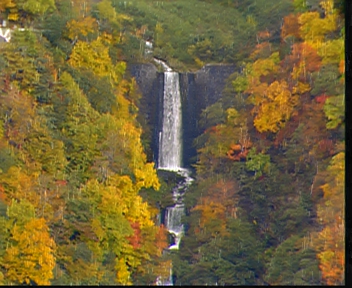} &
\includegraphics[width=0.19\linewidth]{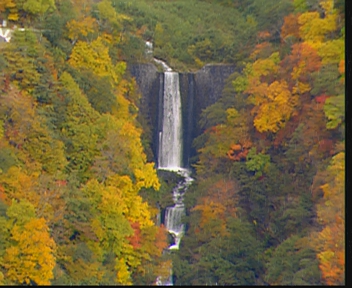} &
\includegraphics[width=0.19\linewidth]{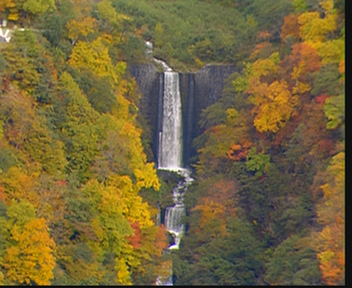} \\
&\includegraphics[width=0.19\linewidth]{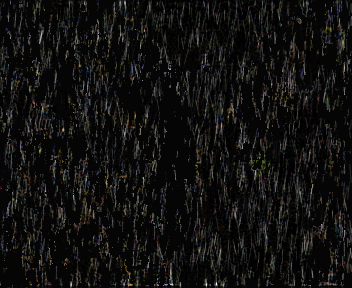} &
\includegraphics[width=0.19\linewidth]{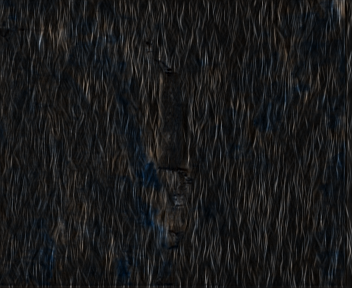} &
\includegraphics[width=0.19\linewidth]{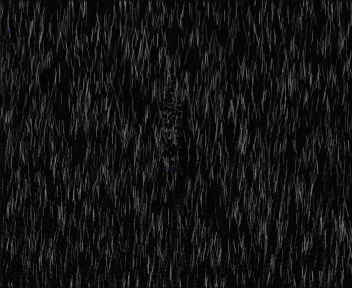} &
\includegraphics[width=0.19\linewidth]{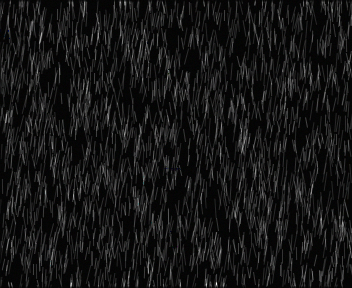} \\
\includegraphics[width=0.19\linewidth]{figs/component/bar_yuv.png}&
\includegraphics[width=0.19\linewidth]{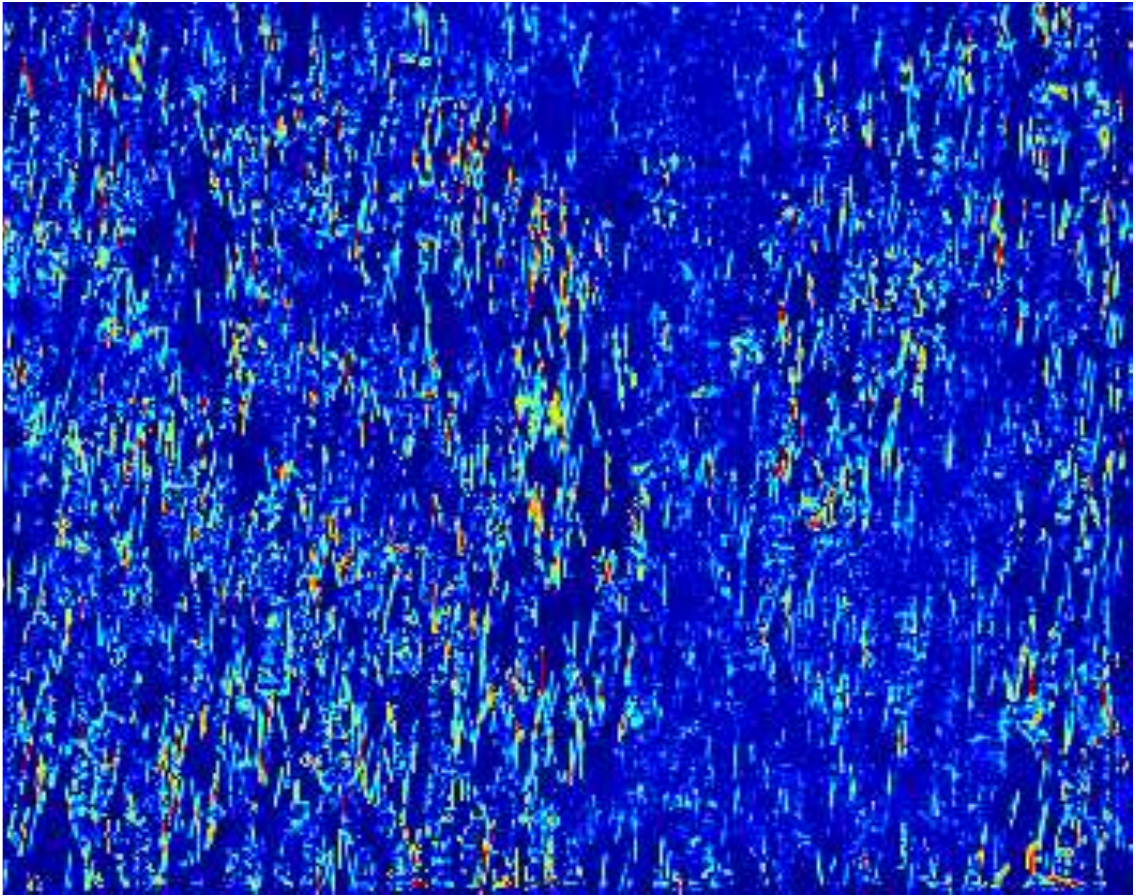}&
\includegraphics[width=0.19\linewidth]{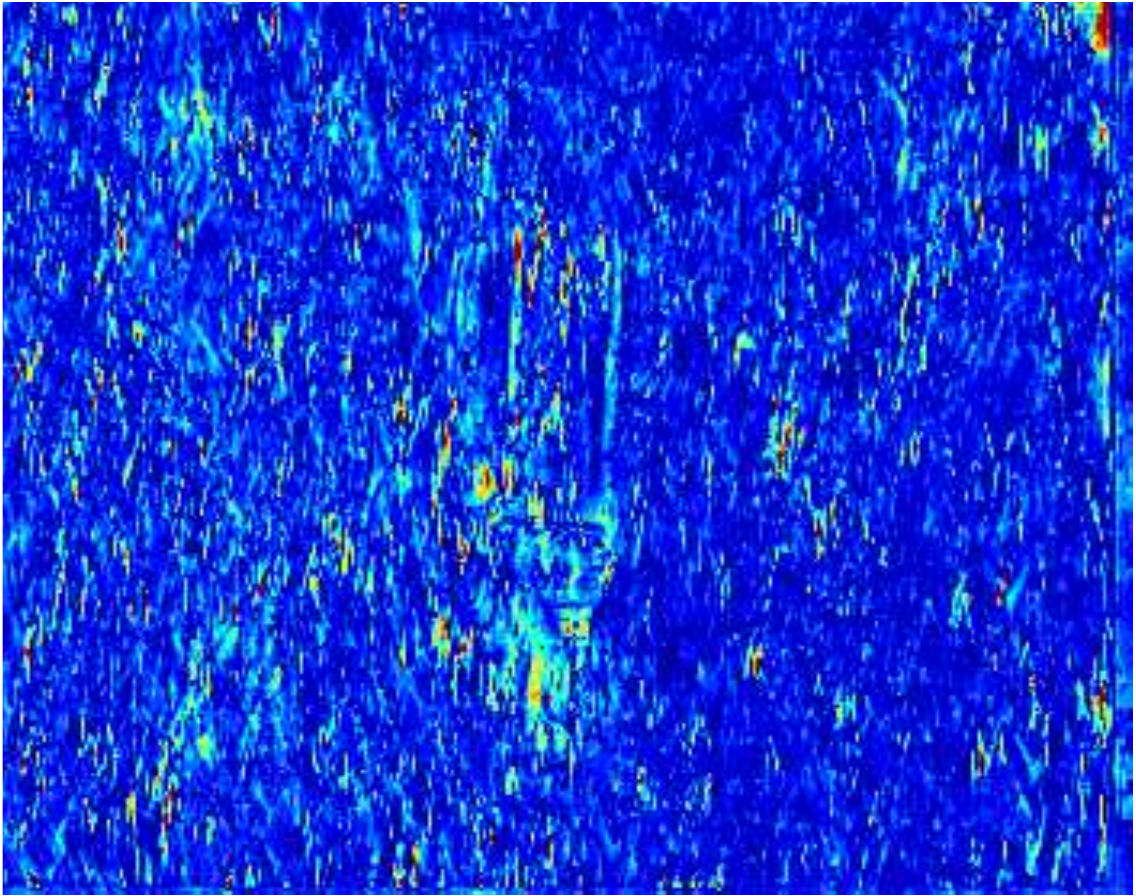}&
\includegraphics[width=0.19\linewidth]{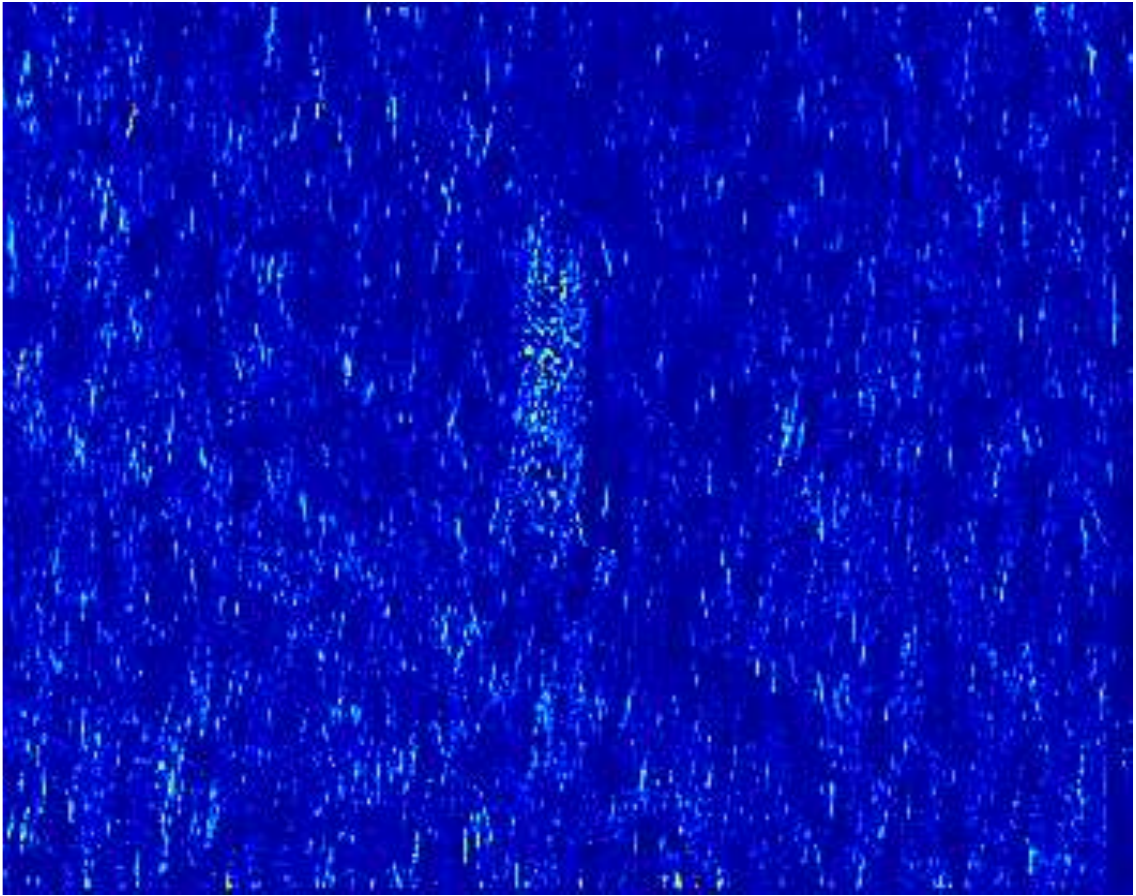} &
\includegraphics[width=0.19\linewidth]{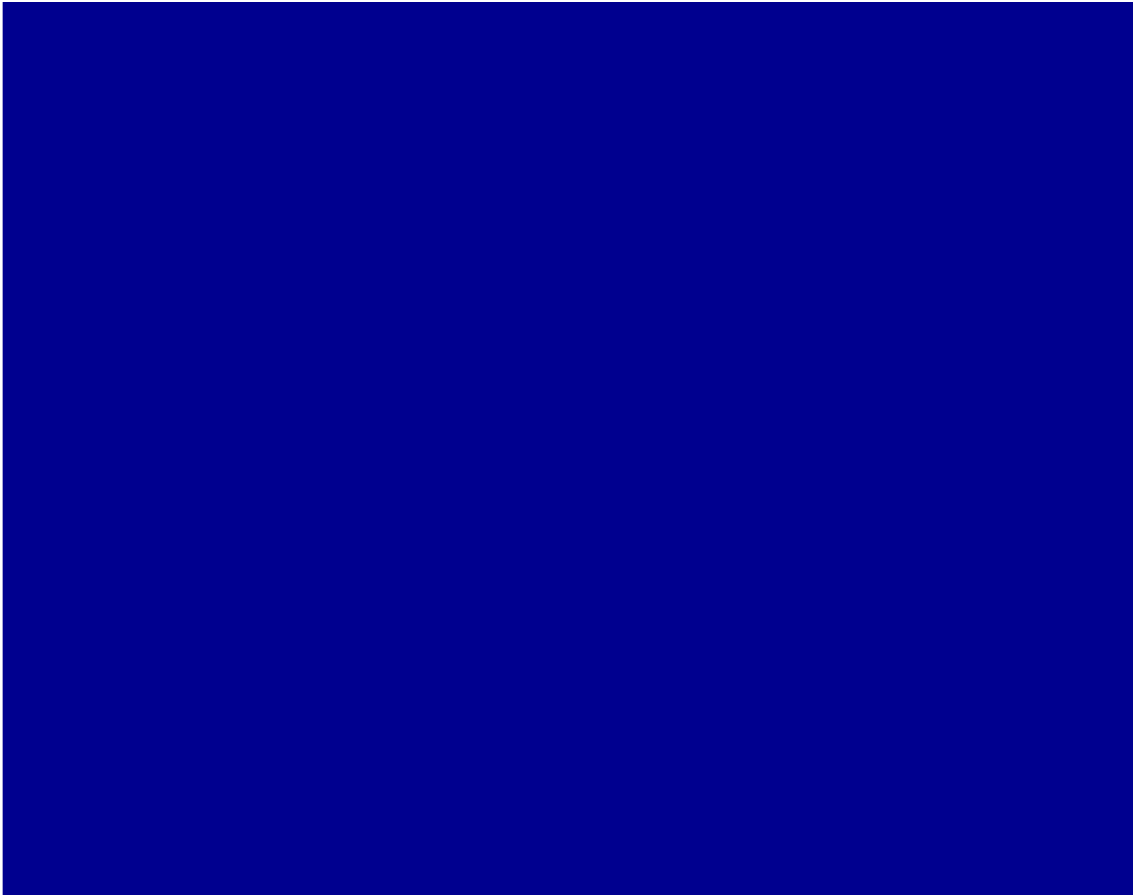} \\

\end{tabular}

\begin{tabular}{ccccccc}\scriptsize
Rainy&TCL & DDN &SE&MS-CSC& FastDeRain&GT\\
\includegraphics[width=0.135\linewidth]{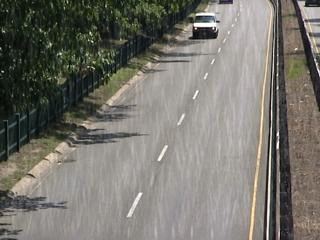} &
\includegraphics[width=0.135\linewidth]{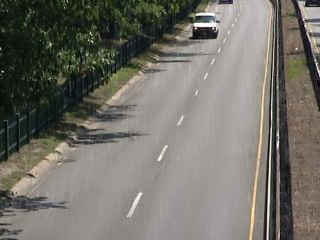} &
\includegraphics[width=0.135\linewidth]{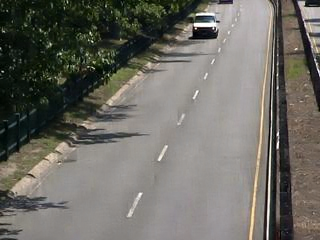} &
\includegraphics[width=0.135\linewidth]{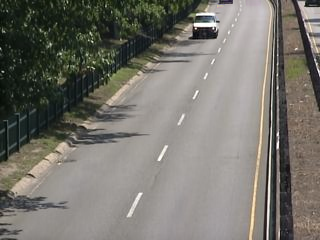} &
\includegraphics[width=0.135\linewidth]{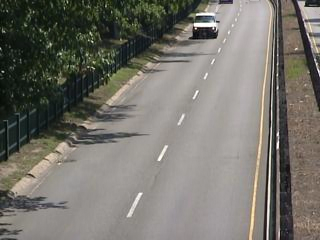} &
\includegraphics[width=0.135\linewidth]{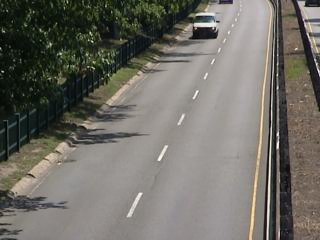} &
\includegraphics[width=0.135\linewidth]{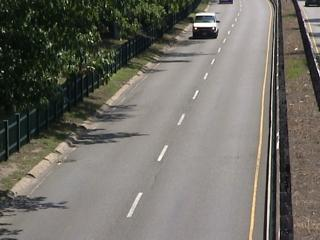} \\
&\includegraphics[width=0.135\linewidth]{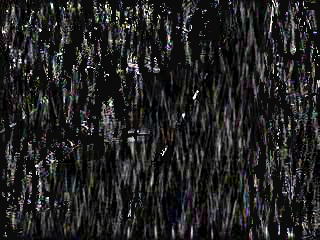} &
\includegraphics[width=0.135\linewidth]{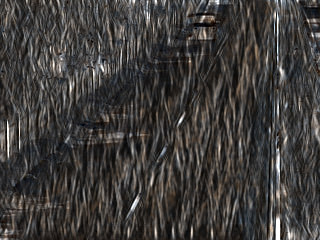} &
\includegraphics[width=0.135\linewidth]{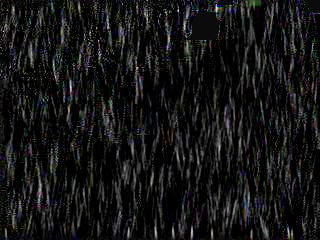} &
\includegraphics[width=0.135\linewidth]{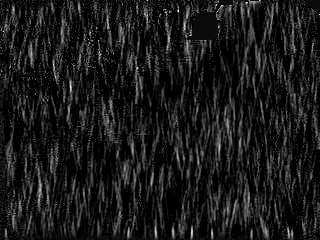} &
\includegraphics[width=0.135\linewidth]{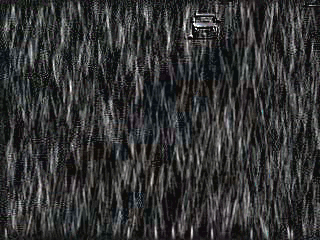} &
\includegraphics[width=0.135\linewidth]{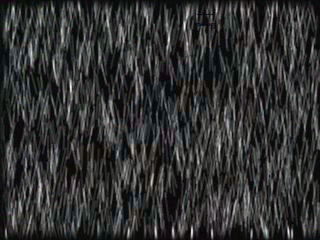} \\
\includegraphics[width=0.135\linewidth]{figs/component/bar_iccv.png}&
\includegraphics[width=0.135\linewidth]{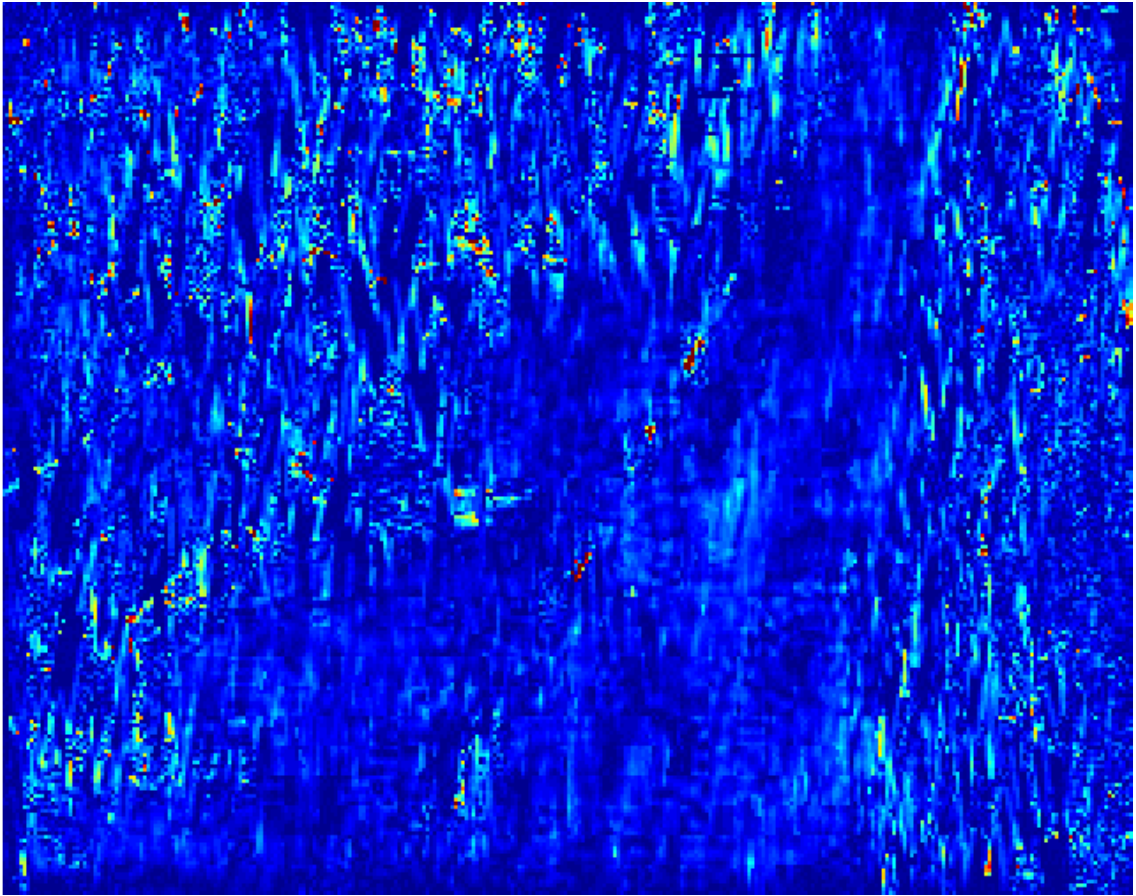}&
\includegraphics[width=0.135\linewidth]{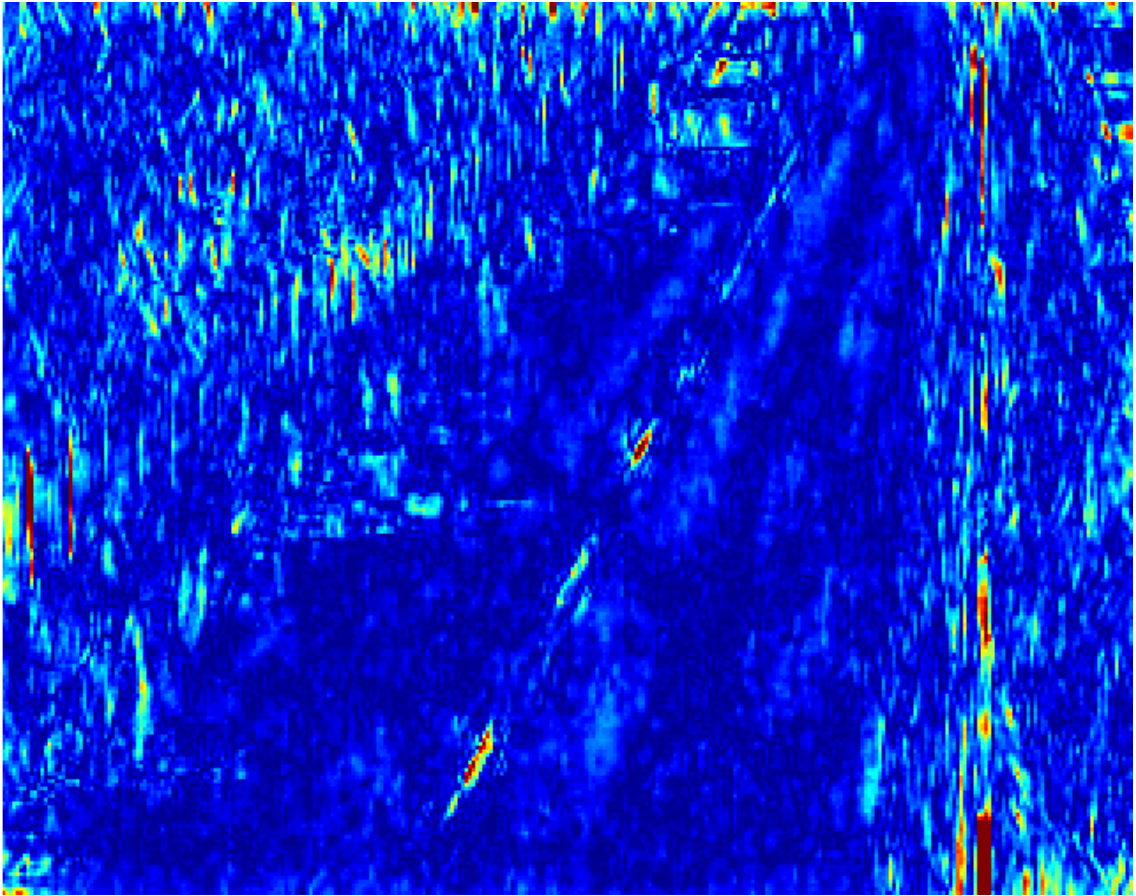}&
\includegraphics[width=0.135\linewidth]{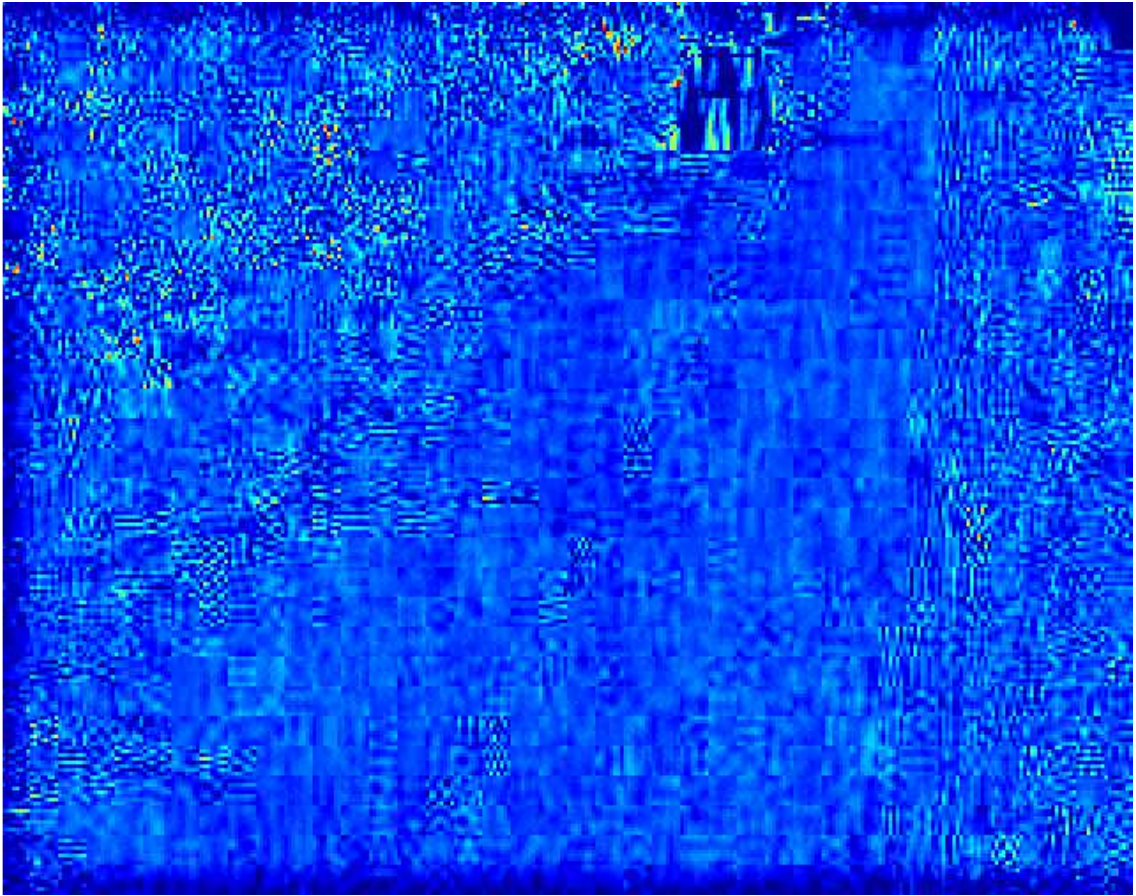}&
\includegraphics[width=0.135\linewidth]{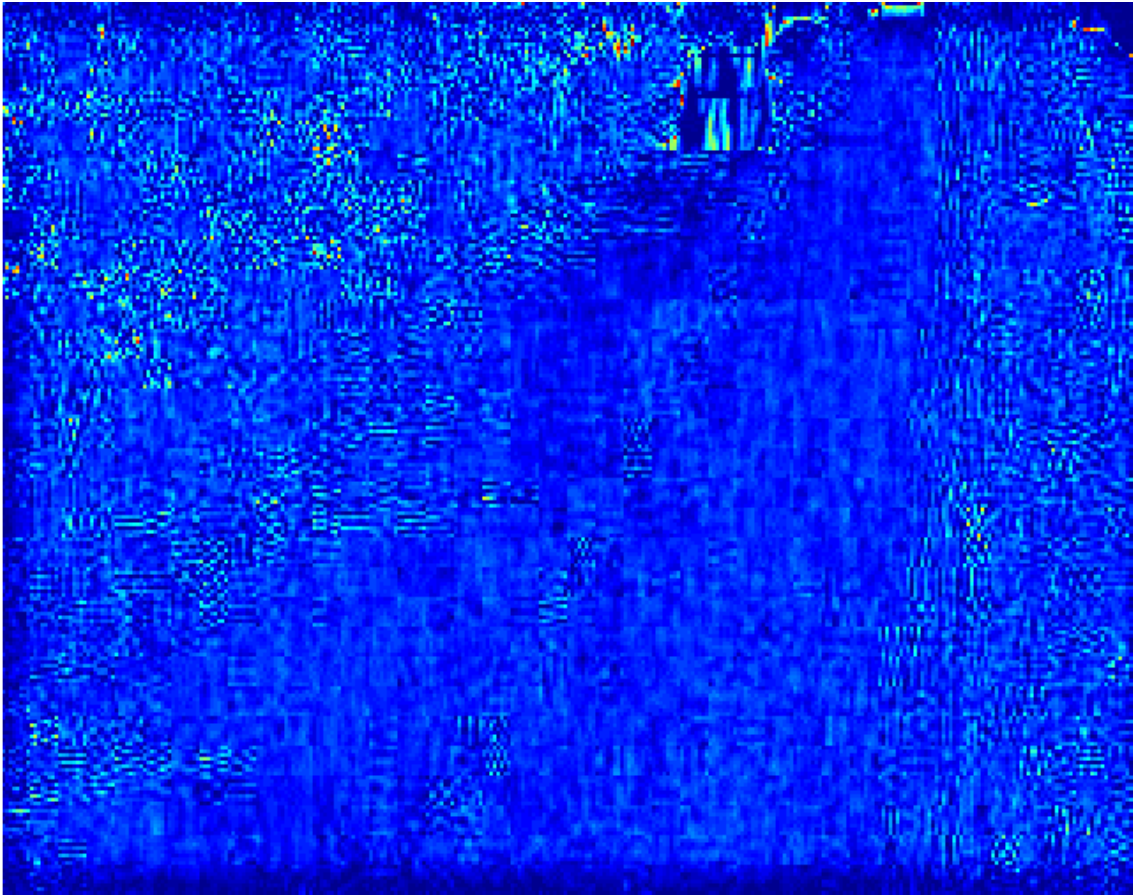}&
\includegraphics[width=0.135\linewidth]{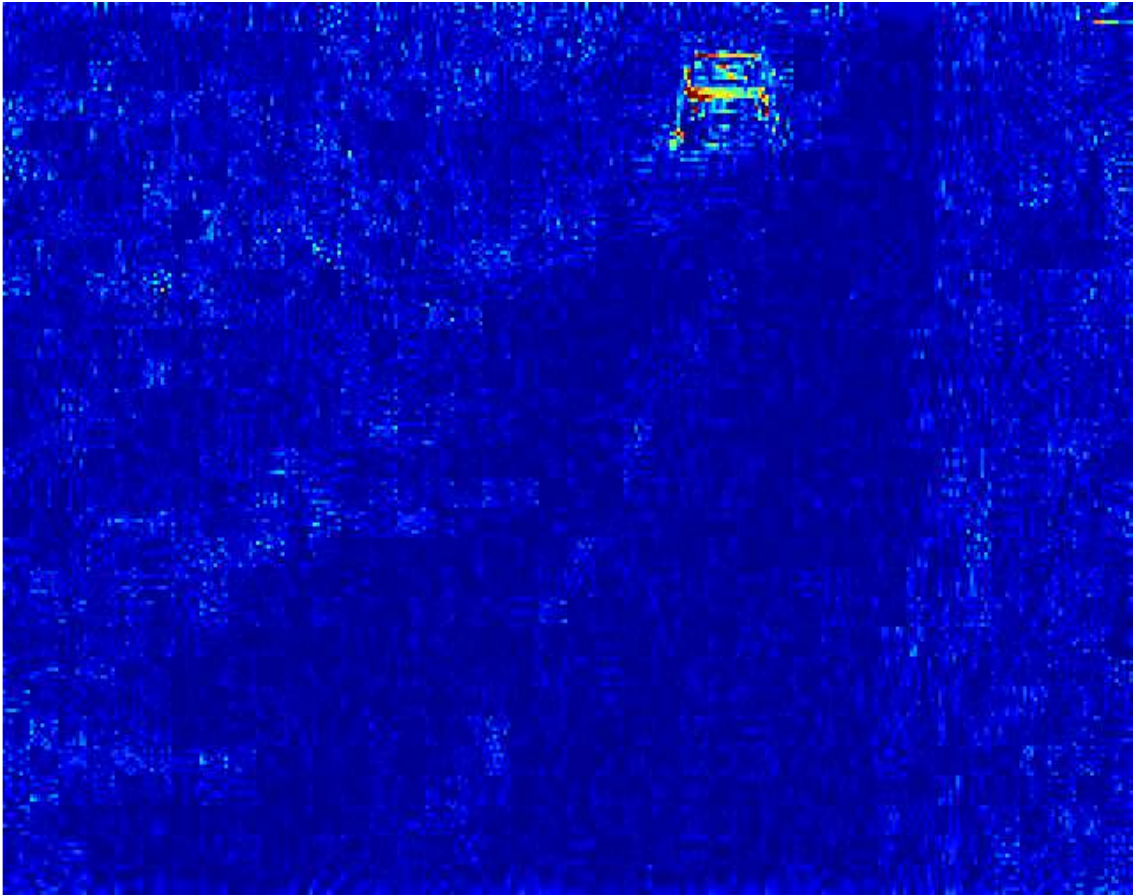} &
\includegraphics[width=0.135\linewidth]{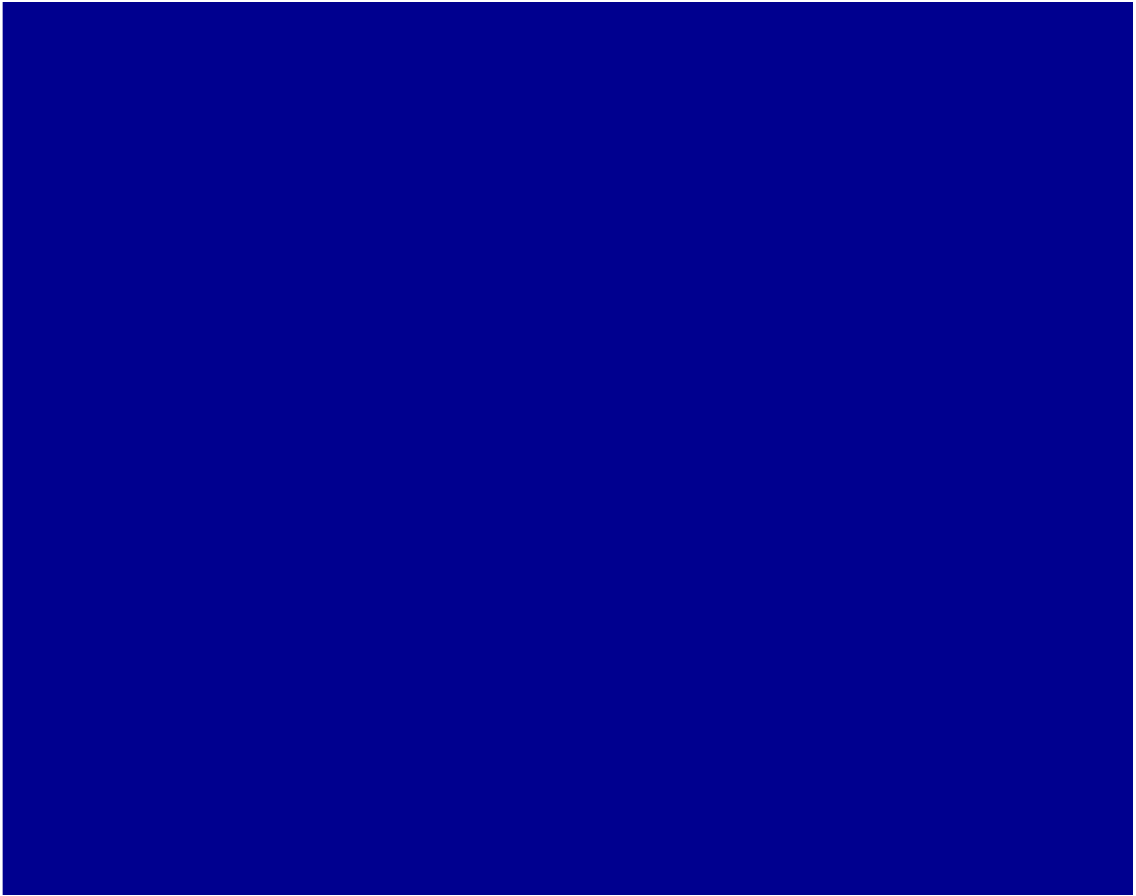} \\


\end{tabular}
\caption{The rainy frame, rain streaks removal results, extracted rain streaks and corresponding error images by different methods with synthetic rain streaks in \textbf{case 2}, respectively. The corresponding videos from top to bottom are the ``'foreman'', ''bus'', ''waterfall'' and ''highway''.
From left to right are: the rainy data (or the color bar), results by TCL \cite{kim2015video}, DDN \cite{fu2017clearing}, (SE \cite{Wei_2017_ICCV}, MS-CSC \cite{li2018video},)  FastDeRain, and the ground truth (GT), respectively.}
\label{fig_case2}
\end{figure}

\paragraph{Rain streak generation}
Adding rain streaks to a video is indeed a complex problem since there is not an existing algorithm nor a free software to accomplish it in one step.
Meanwhile, as Starik {\em et al.} pointed out in \cite{starik2003simulation} that the rain streaks can be assumed temporal independent, thus we can simulate rain streaks for each frame using the synthetic method mentioned in many recently developed single image rain streaks removal approaches \cite{kang2012automatic,luo2015removing,fu2017clearing}, {\em i.e.}, using the Photoshop software with the tutorial documents \cite{PhotoTu}.
The density of the simulated rain streaks by this method is mainly determined by the ratio of the amounts of dots (in step 8 of \cite{PhotoTu}) to the number of all the pixels, and for convenience, the ratio is denoted as $r$.
Another way to synthesize the rain streaks was proposed in \cite{Wei_2017_ICCV}, adding rain streaks taken by photographers under
black background\footnote{\url{http://www.2gei.com/video/effect/1\_rain/}}.

Referring to \cite{PhotoTu} and \cite{Wei_2017_ICCV}, we generate 3 types of rain streaks as follows:

\hspace{0.2cm}\noindent \textbf{Case 1} Rain streaks simulated referring to \cite{PhotoTu} with $r\leq0.04$. In a single frame, the rain streaks share the same angle. The fixed angles for different frames increase from $-15^\circ$ to $15^\circ$ with time;

\hspace{0.2cm}\noindent \textbf{Case 2} Rain streaks simulated referring to \cite{PhotoTu} with $r\geq0.05$. In a single frame, the rain streaks are with different angles. The angles uniformly distribute in a range $[-15^\circ,15^\circ]$;

\hspace{0.2cm}\noindent \textbf{Case 3} Rain streaks simulated referring to \cite{Wei_2017_ICCV}.

\begin{figure}[!htb] 
\scriptsize\renewcommand\arraystretch{1}
\setlength{\tabcolsep}{1pt}
\begin{tabular}{cccccc}
Rainy &TCL & DDN & FastDeRain&GT\\
\includegraphics[width=0.19\linewidth]{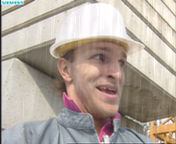} &
\includegraphics[width=0.19\linewidth]{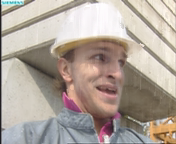} &
\includegraphics[width=0.19\linewidth]{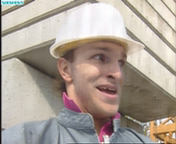} &
\includegraphics[width=0.19\linewidth]{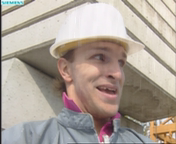} &
\includegraphics[width=0.19\linewidth]{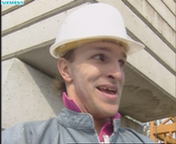} \\
&\includegraphics[width=0.19\linewidth]{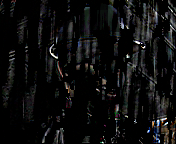} &
\includegraphics[width=0.19\linewidth]{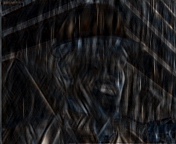} &
\includegraphics[width=0.19\linewidth]{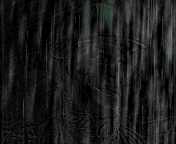} &
\includegraphics[width=0.19\linewidth]{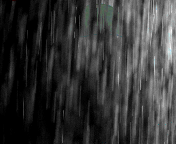} \\
\includegraphics[width=0.19\linewidth]{figs/component/bar_yuv.png}&
\includegraphics[width=0.19\linewidth]{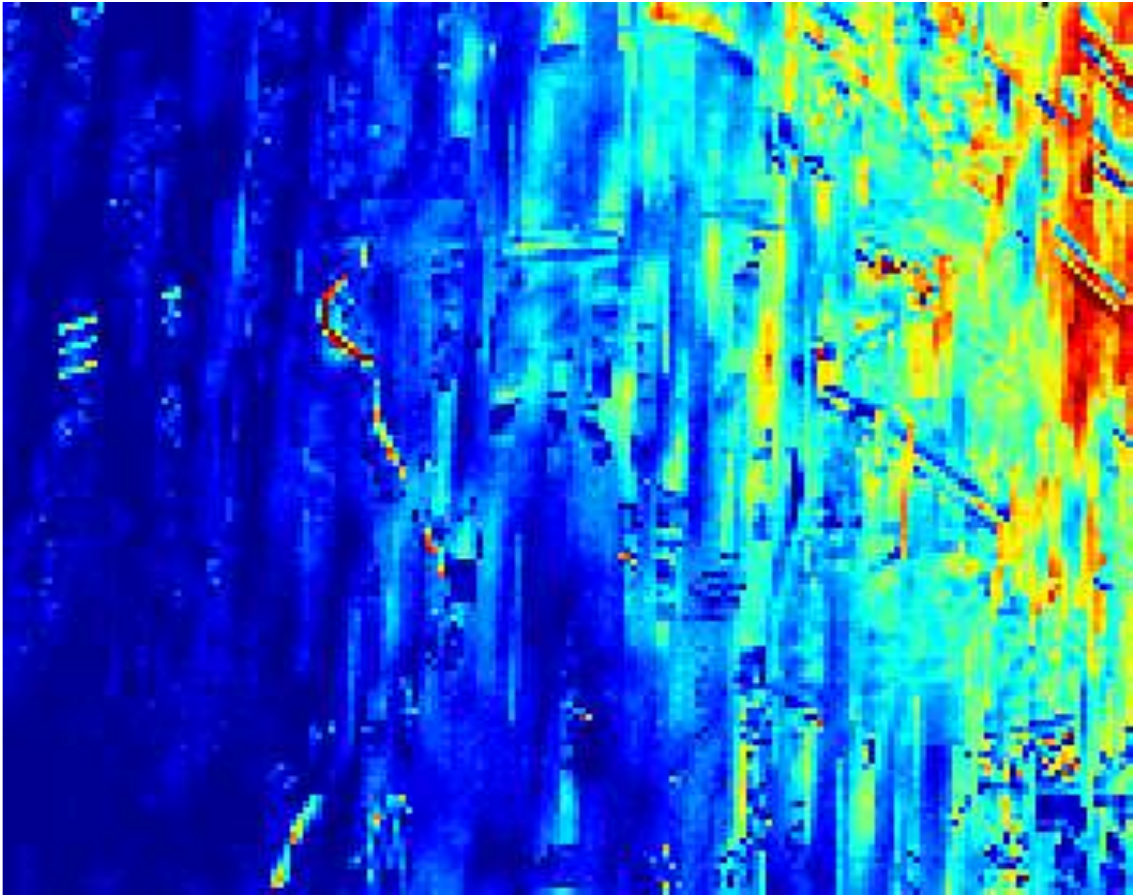}&
\includegraphics[width=0.19\linewidth]{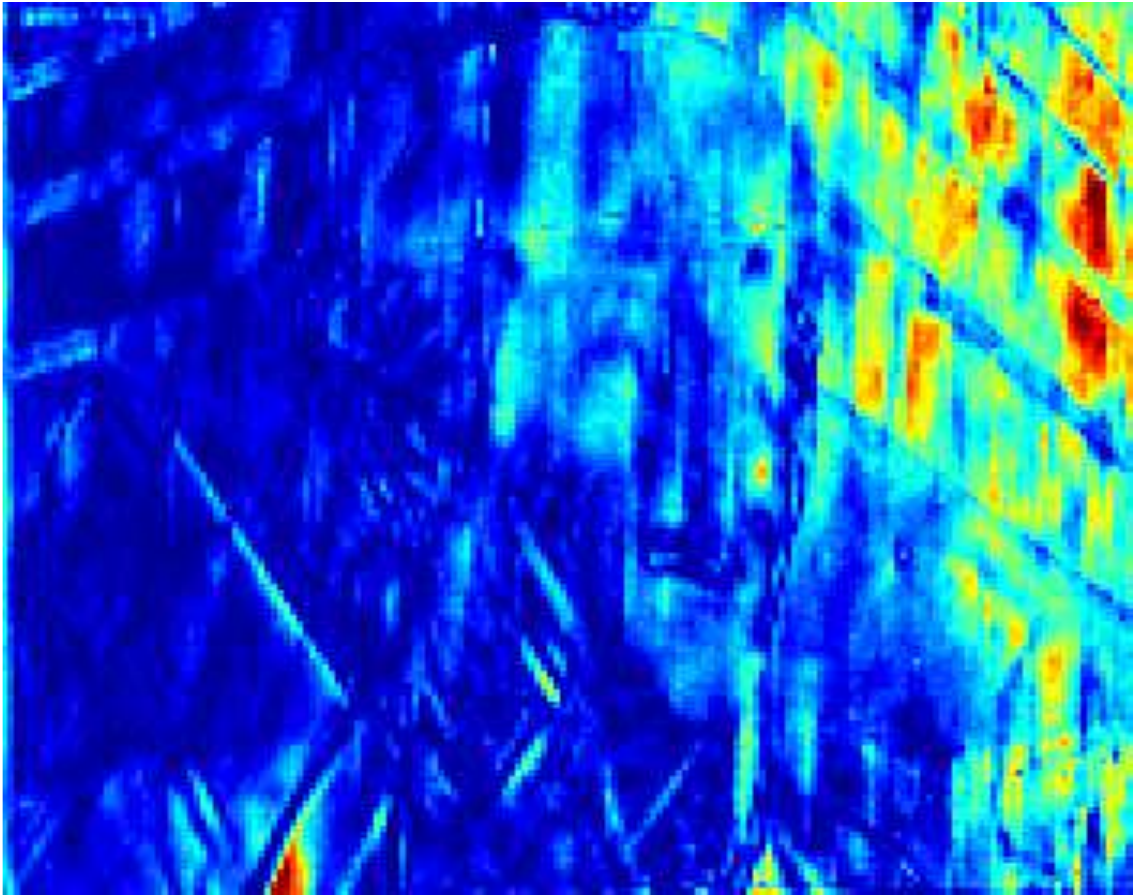}&
\includegraphics[width=0.19\linewidth]{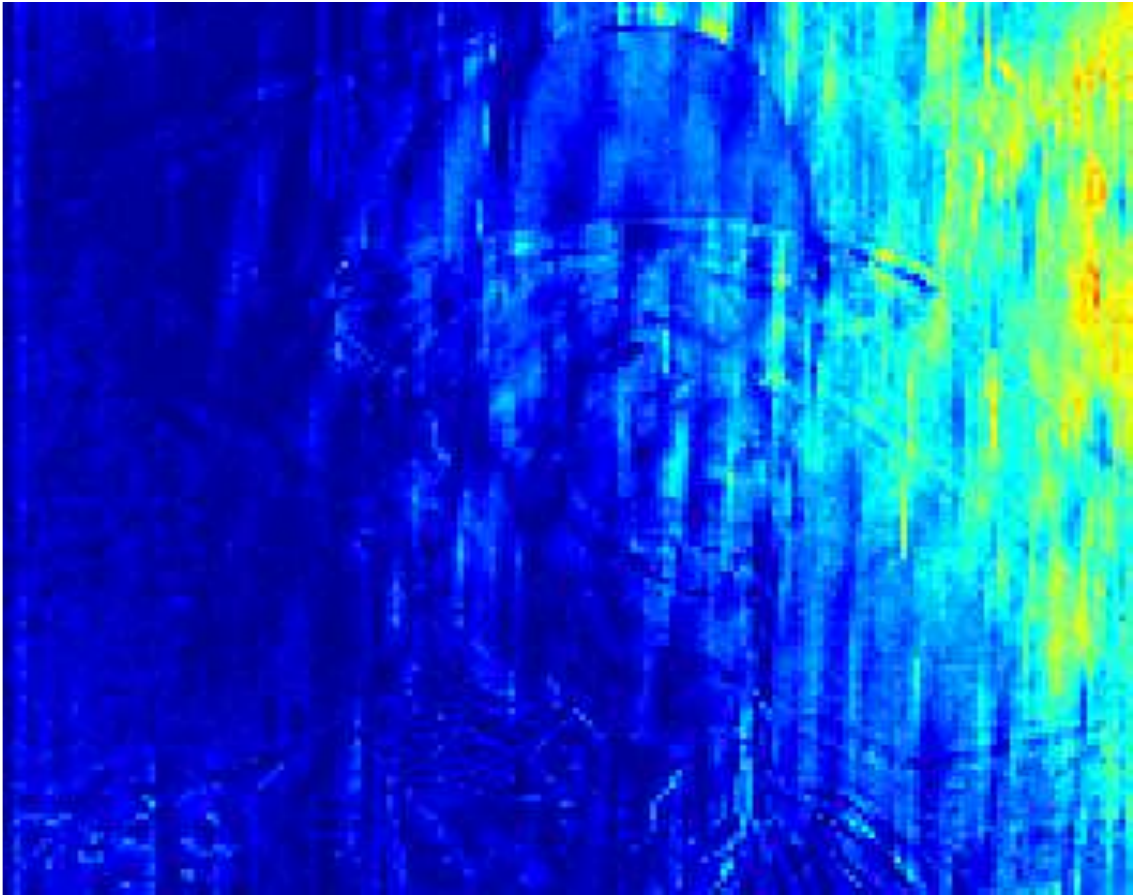} &
\includegraphics[width=0.19\linewidth]{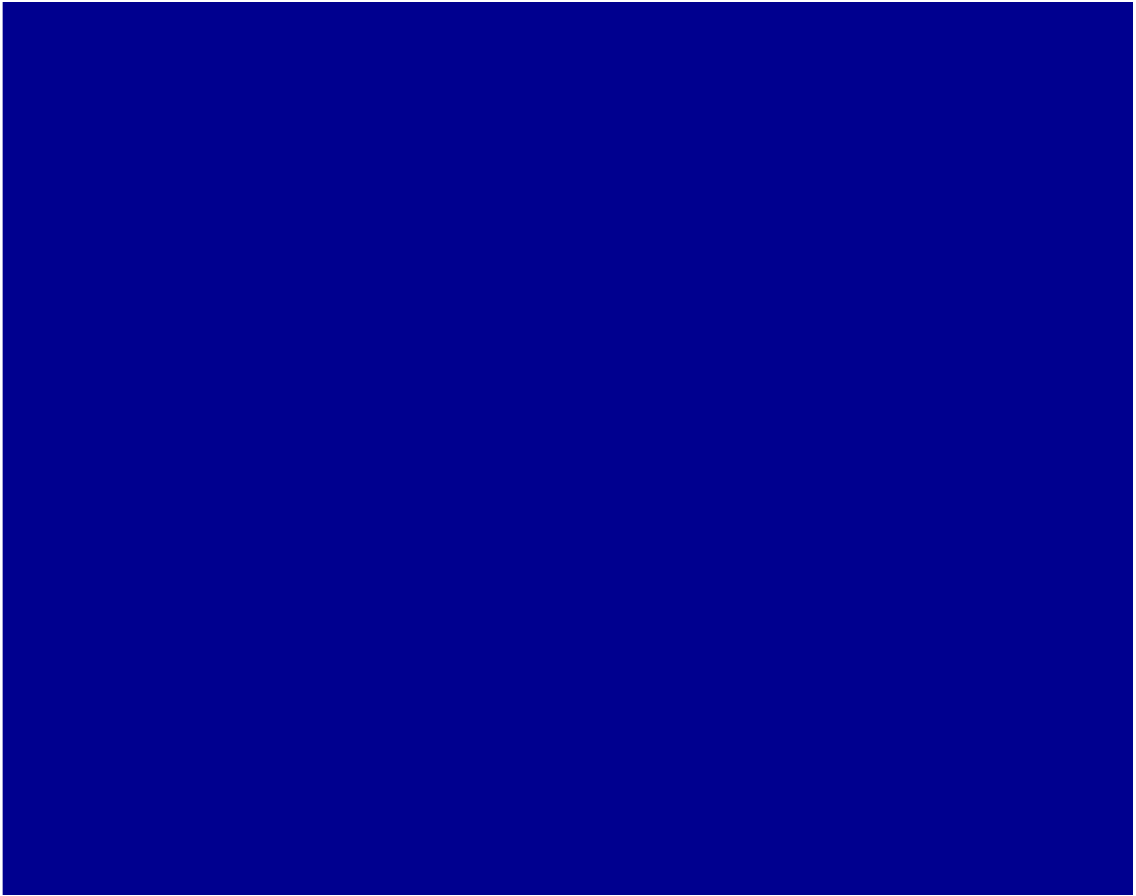} \\

Rainy &TCL & DDN & FastDeRain&GT\\
\includegraphics[width=0.19\linewidth]{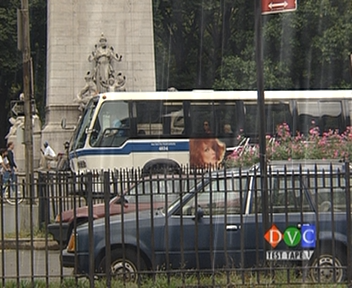} &
\includegraphics[width=0.19\linewidth]{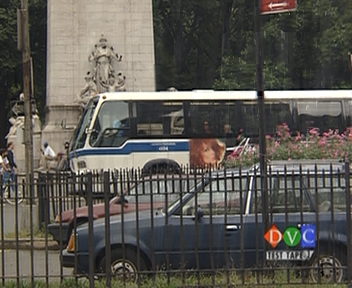} &
\includegraphics[width=0.19\linewidth]{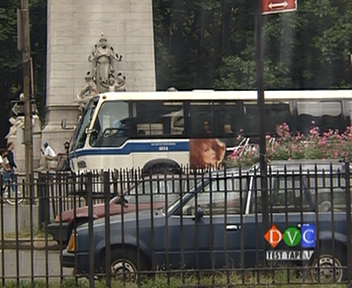} &
\includegraphics[width=0.19\linewidth]{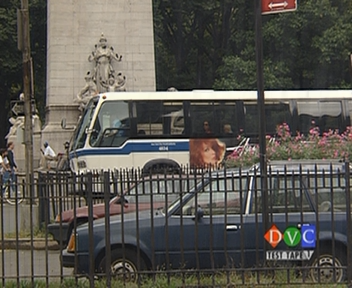} &
\includegraphics[width=0.19\linewidth]{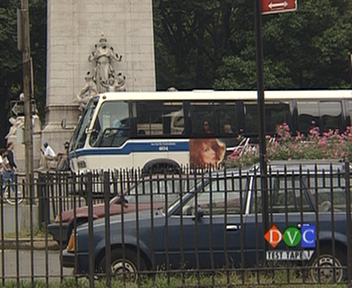} \\
&\includegraphics[width=0.19\linewidth]{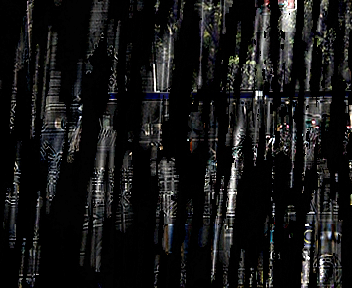} &
\includegraphics[width=0.19\linewidth]{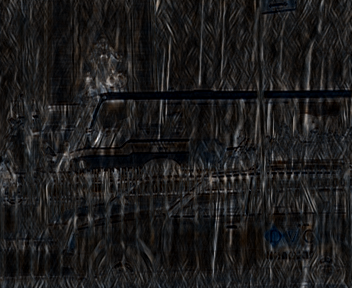} &
\includegraphics[width=0.19\linewidth]{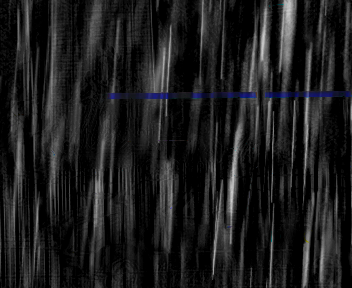} &
\includegraphics[width=0.19\linewidth]{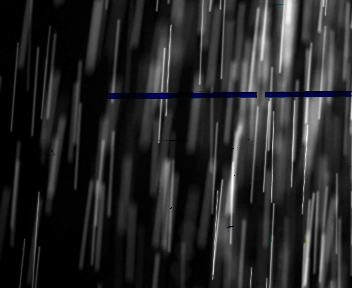} \\
\includegraphics[width=0.19\linewidth]{figs/component/bar_yuv.png}&
\includegraphics[width=0.19\linewidth]{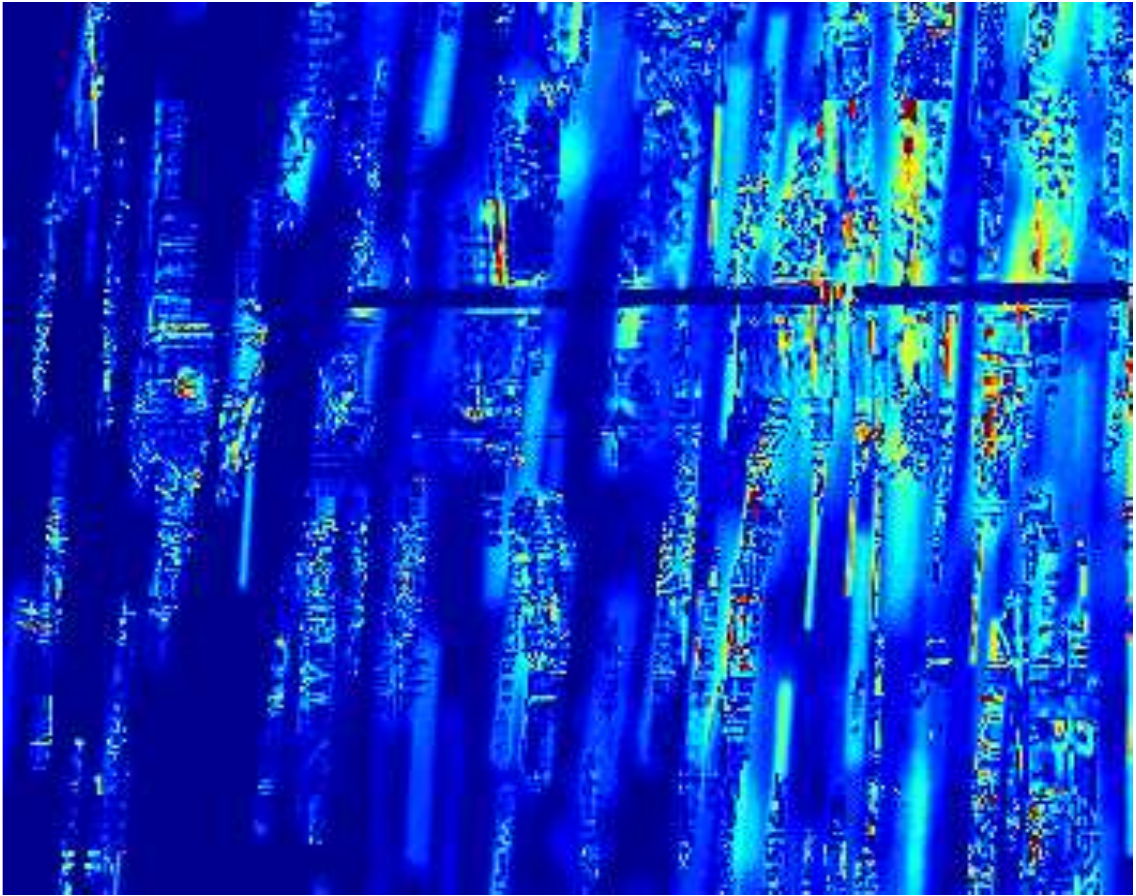}&
\includegraphics[width=0.19\linewidth]{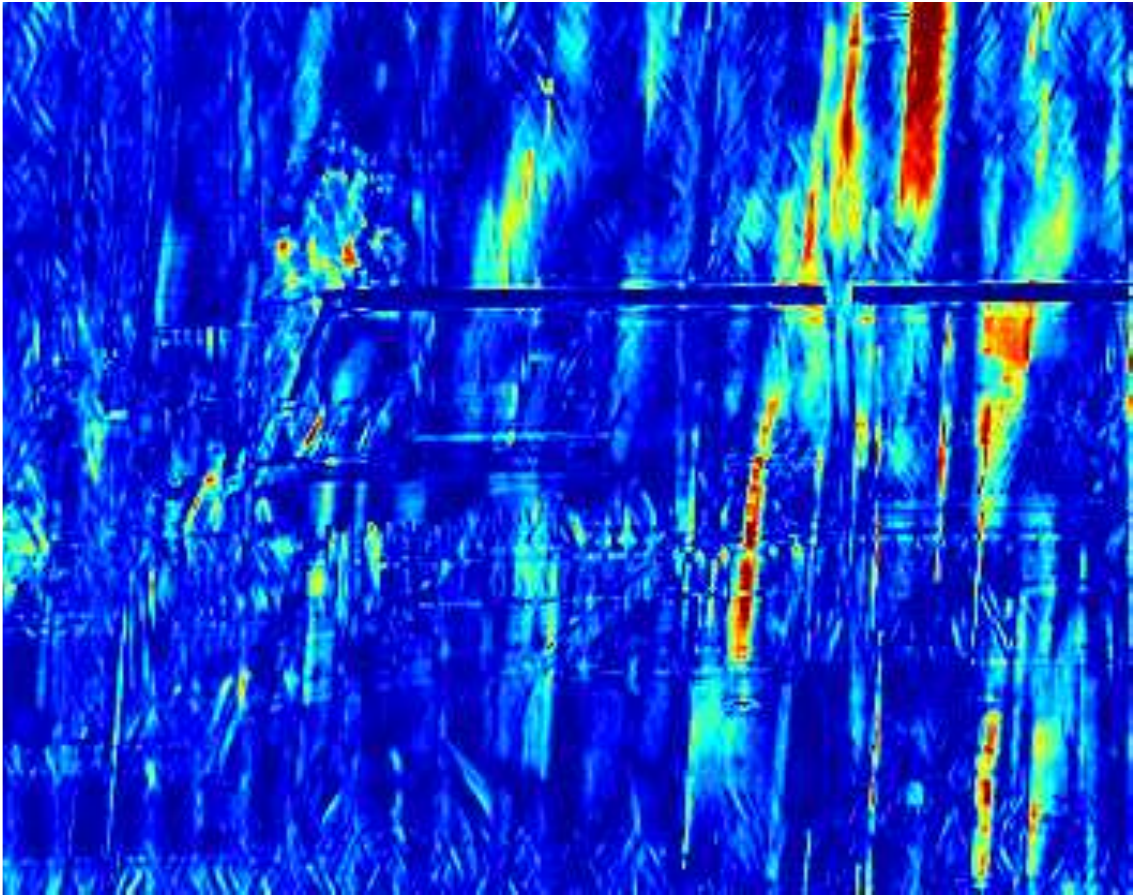}&
\includegraphics[width=0.19\linewidth]{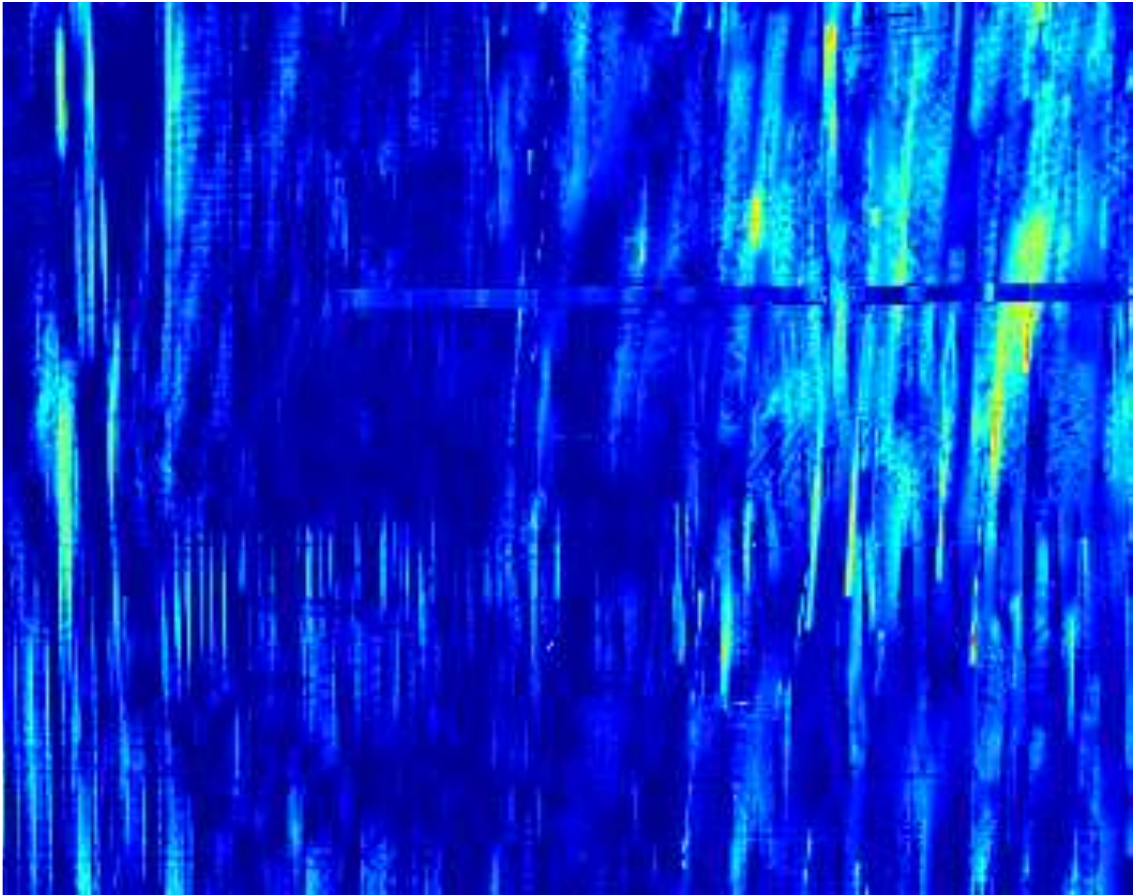} &
\includegraphics[width=0.19\linewidth]{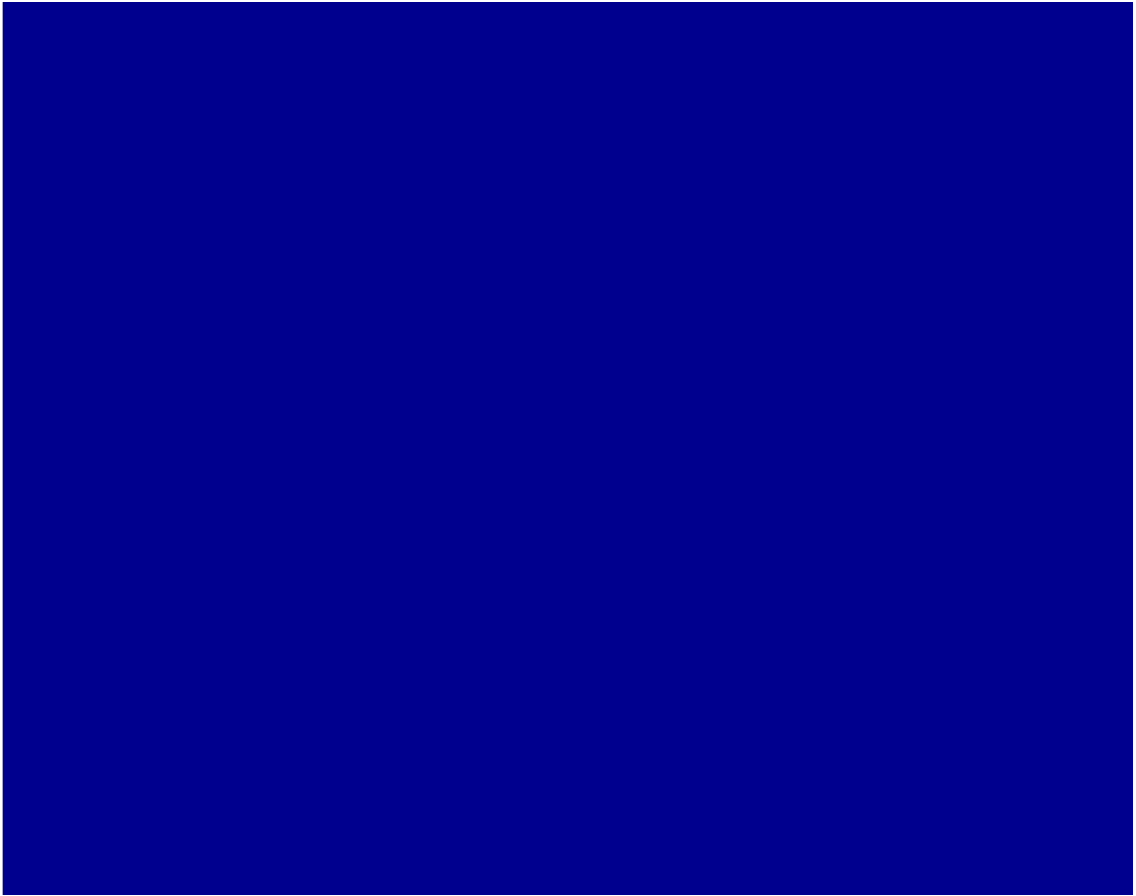} \\

Rainy &TCL & DDN & FastDeRain&GT\\
\includegraphics[width=0.19\linewidth]{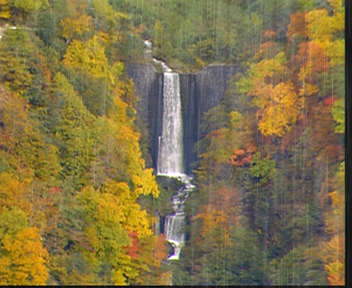} &
\includegraphics[width=0.19\linewidth]{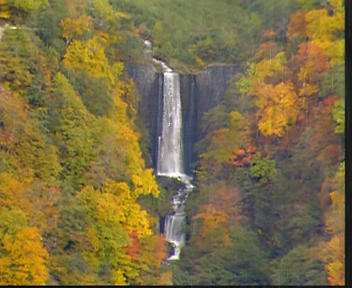} &
\includegraphics[width=0.19\linewidth]{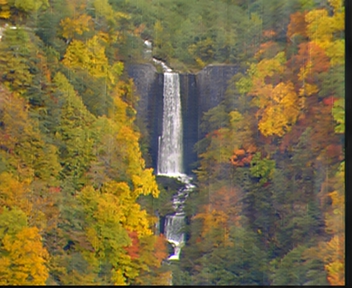} &
\includegraphics[width=0.19\linewidth]{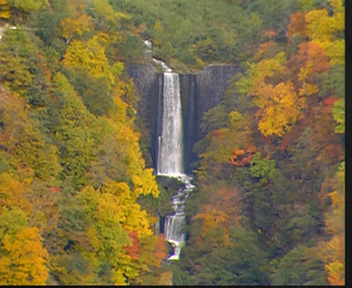} &
\includegraphics[width=0.19\linewidth]{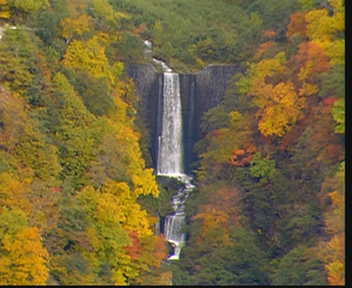} \\
&\includegraphics[width=0.19\linewidth]{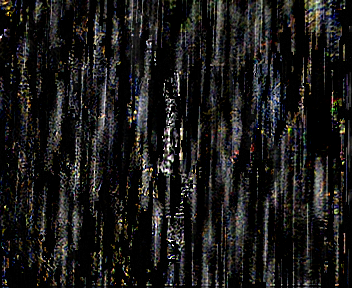} &
\includegraphics[width=0.19\linewidth]{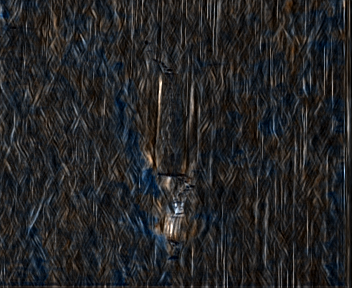} &
\includegraphics[width=0.19\linewidth]{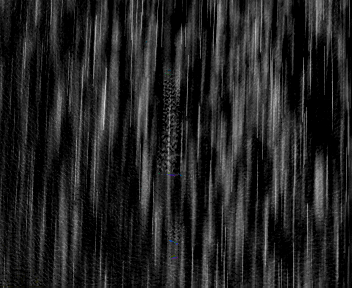} &
\includegraphics[width=0.19\linewidth]{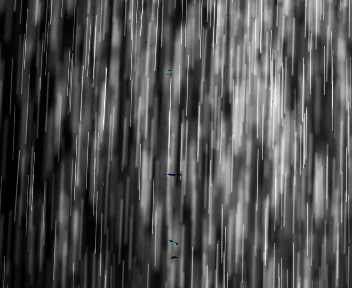} \\
\includegraphics[width=0.19\linewidth]{figs/component/bar_yuv.png}&
\includegraphics[width=0.19\linewidth]{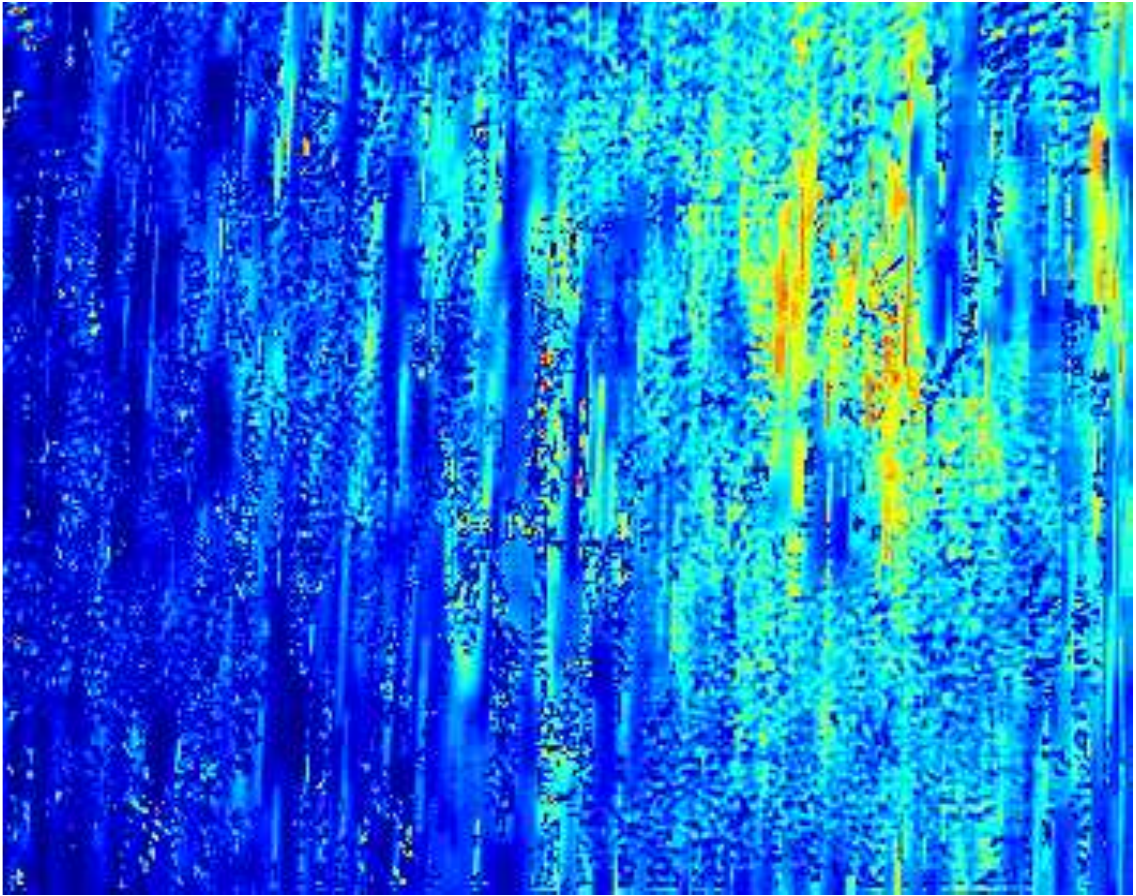}&
\includegraphics[width=0.19\linewidth]{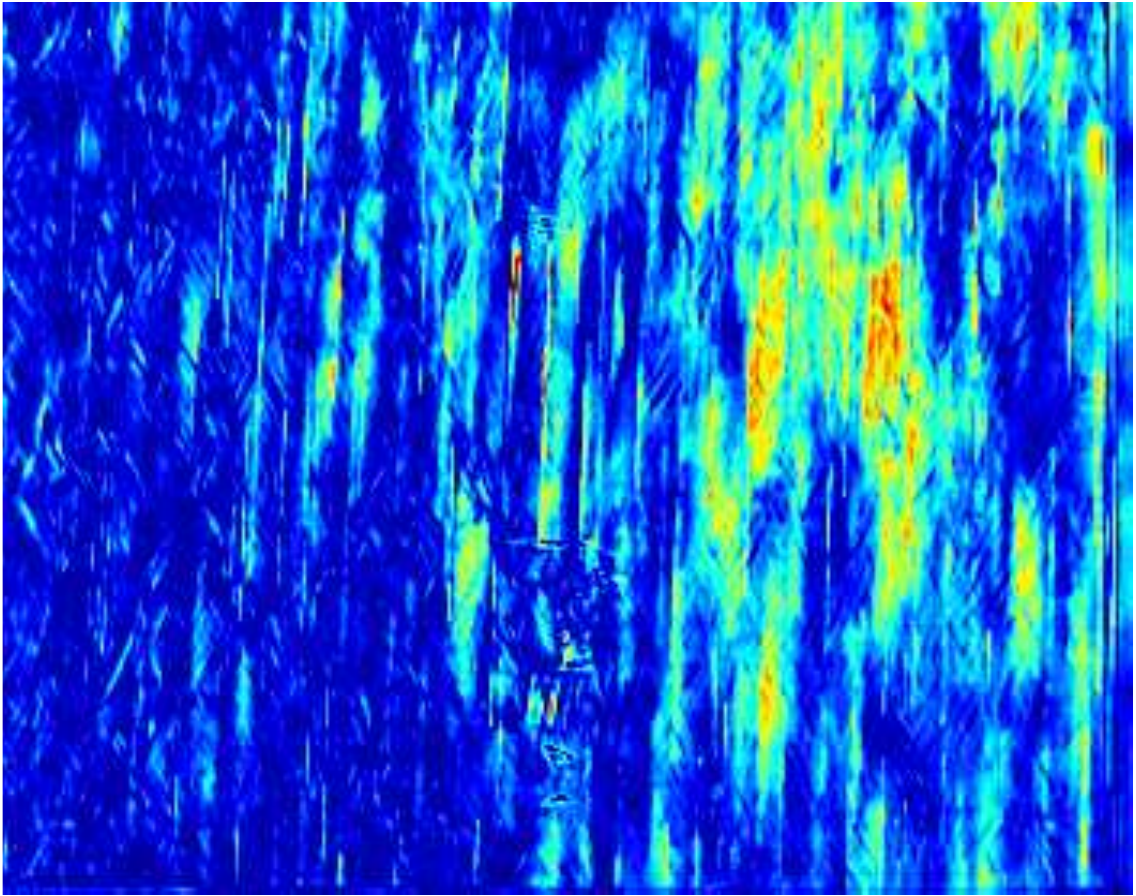}&
\includegraphics[width=0.19\linewidth]{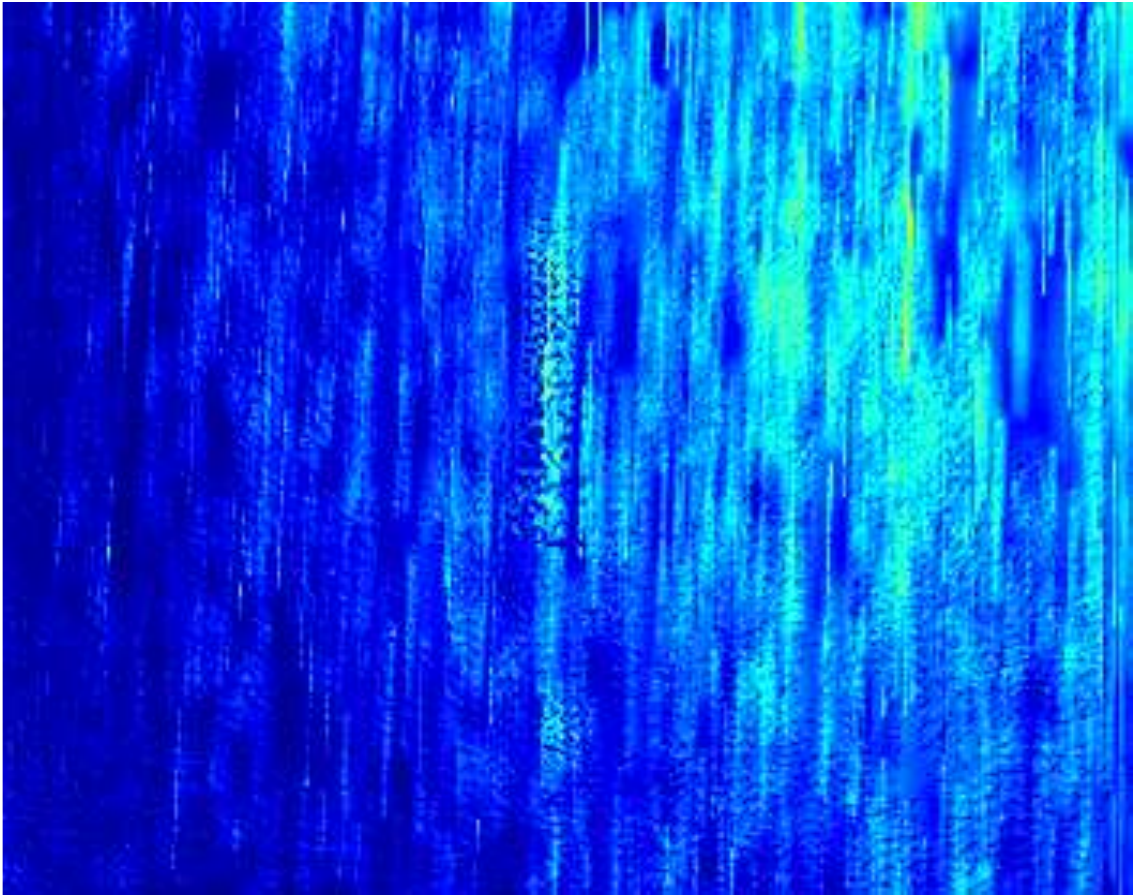} &
\includegraphics[width=0.19\linewidth]{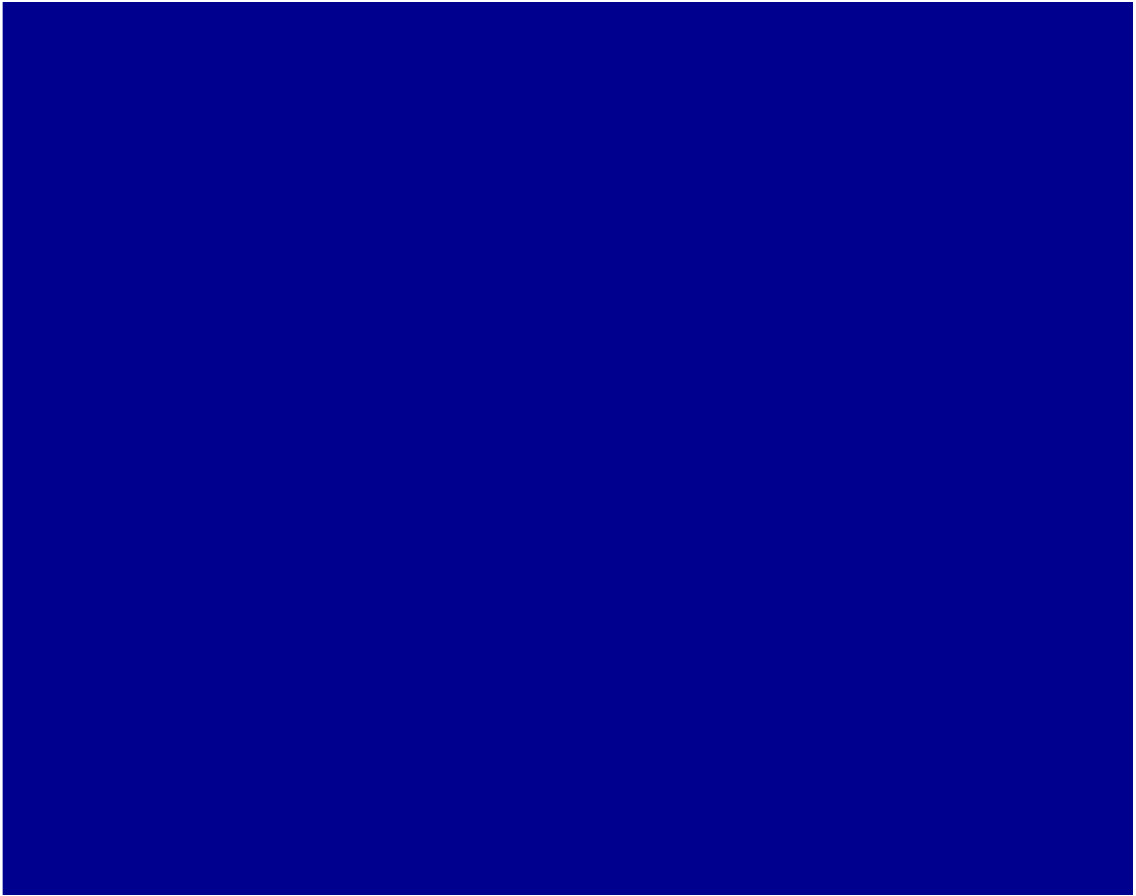} \\
\end{tabular}
\begin{tabular}{ccccccc}\scriptsize
Rainy&TCL  & DDN &SE &MS-CSC & FastDeRain&GT\\

\includegraphics[width=0.135\linewidth]{figs/case3/video_4_rainy_5.png} &
\includegraphics[width=0.135\linewidth]{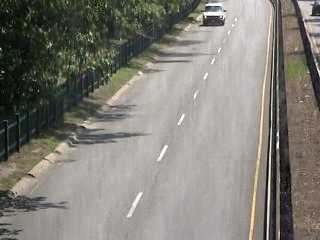} &
\includegraphics[width=0.135\linewidth]{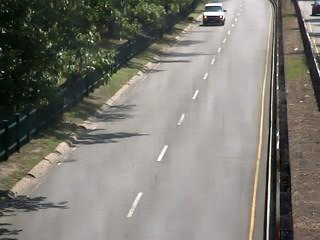} &
\includegraphics[width=0.135\linewidth]{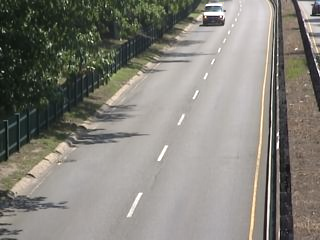} &
\includegraphics[width=0.135\linewidth]{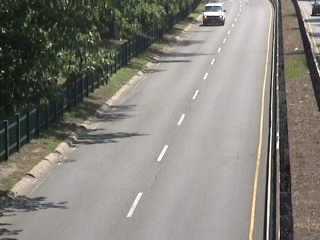} &
\includegraphics[width=0.135\linewidth]{figs/case3/video_4_oursB_5.png} &
\includegraphics[width=0.135\linewidth]{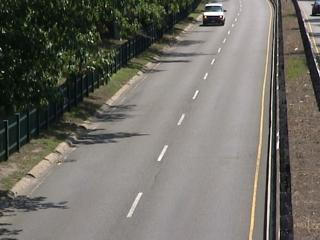} \\
&\includegraphics[width=0.135\linewidth]{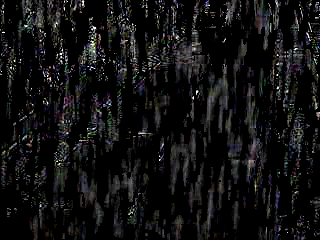} &
\includegraphics[width=0.135\linewidth]{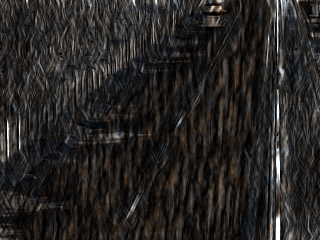} &
\includegraphics[width=0.135\linewidth]{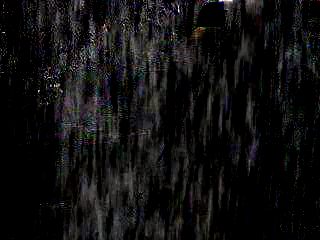} &
\includegraphics[width=0.135\linewidth]{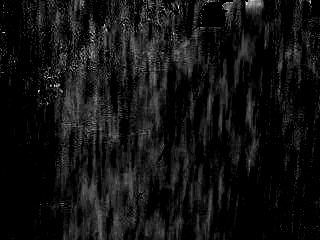} &
\includegraphics[width=0.135\linewidth]{figs/case3/video_4_oursR_5.png} &
\includegraphics[width=0.135\linewidth]{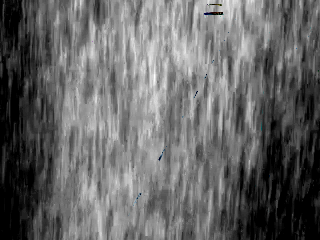} \\
\includegraphics[width=0.135\linewidth]{figs/component/bar_iccv.png}&
\includegraphics[width=0.135\linewidth]{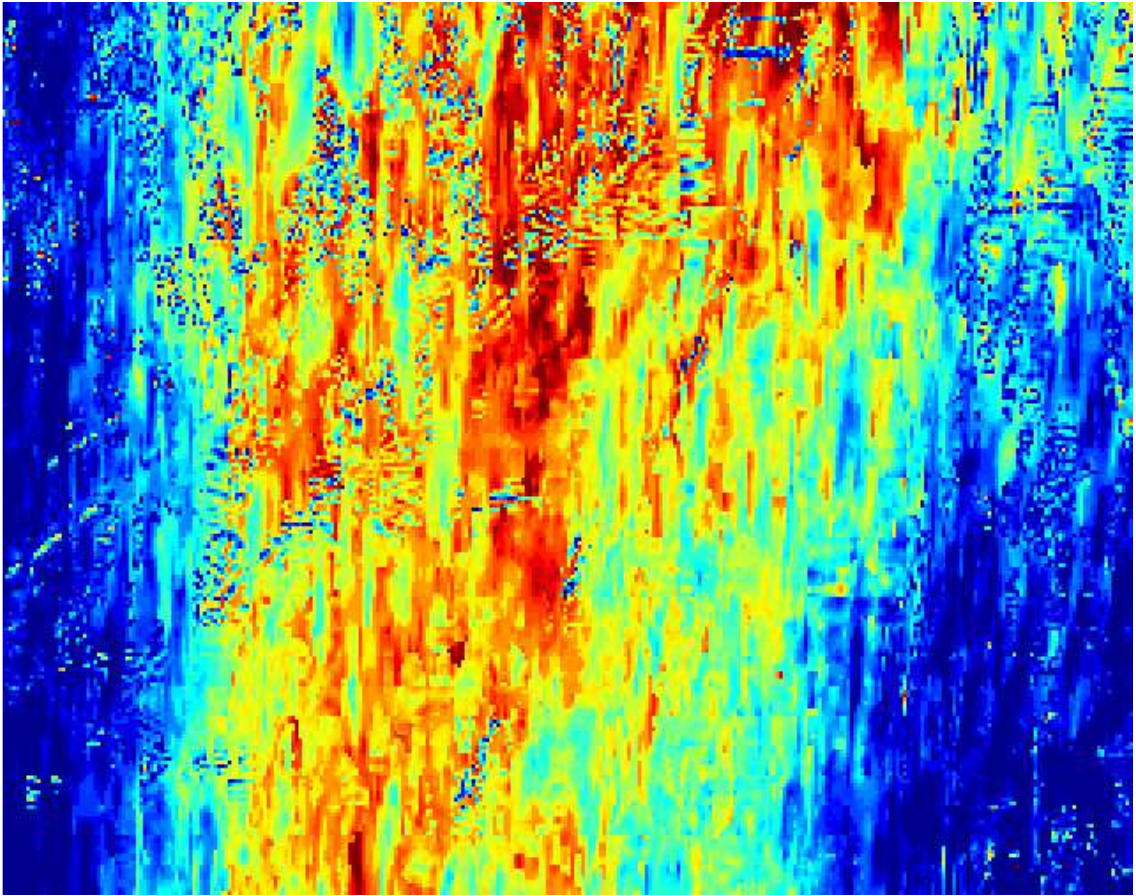}&
\includegraphics[width=0.135\linewidth]{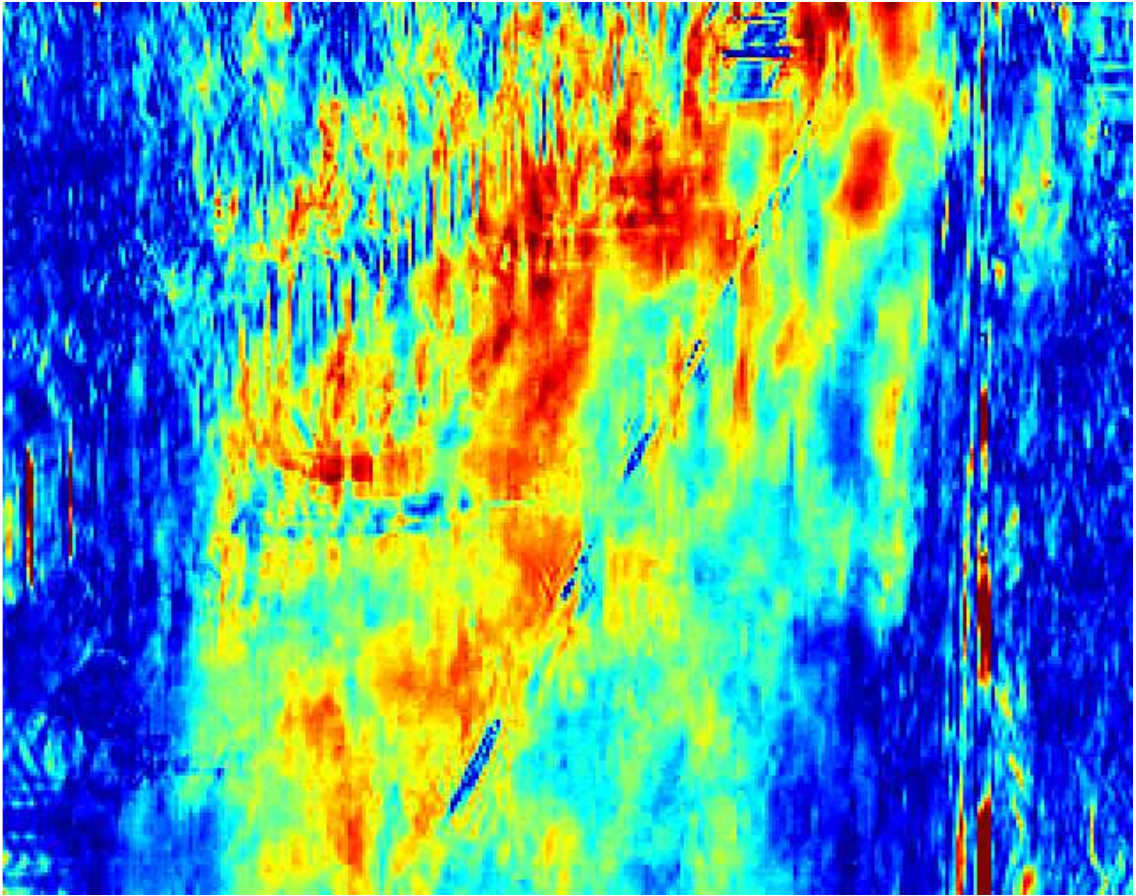}&
\includegraphics[width=0.135\linewidth]{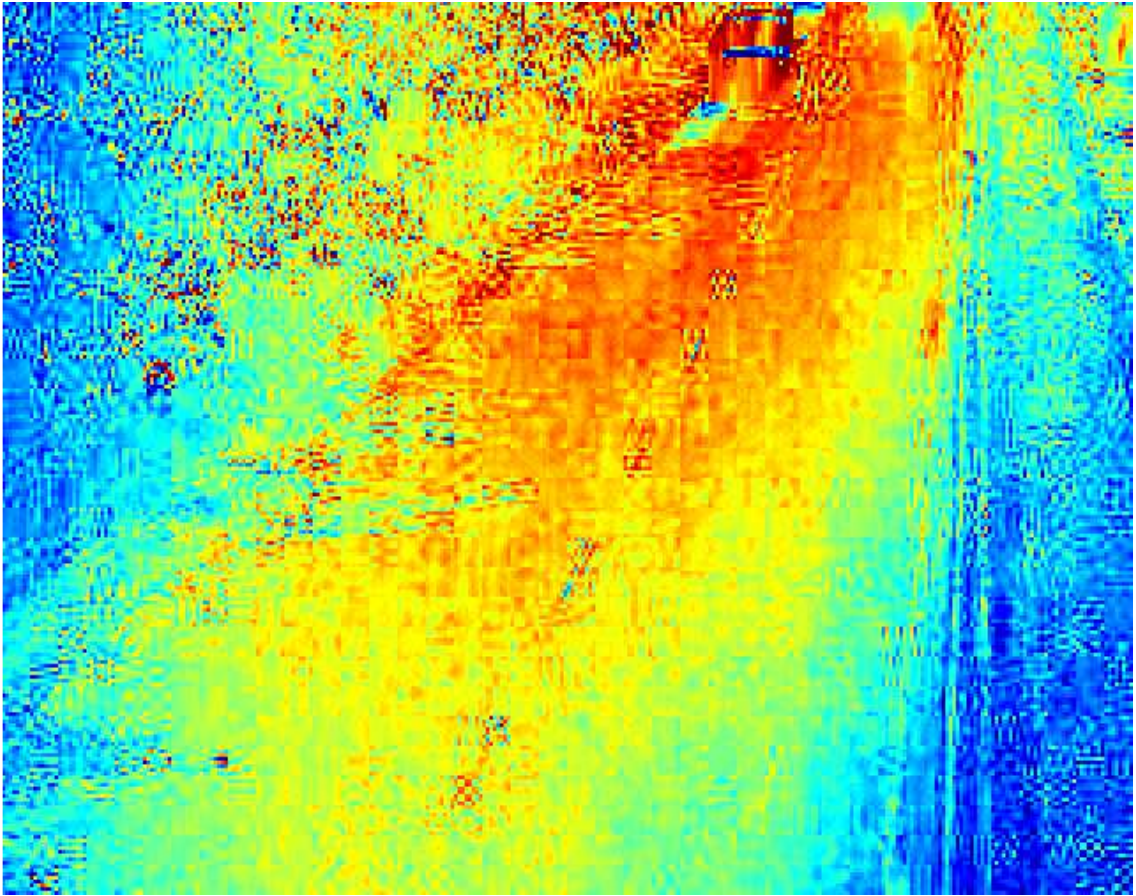}&
\includegraphics[width=0.135\linewidth]{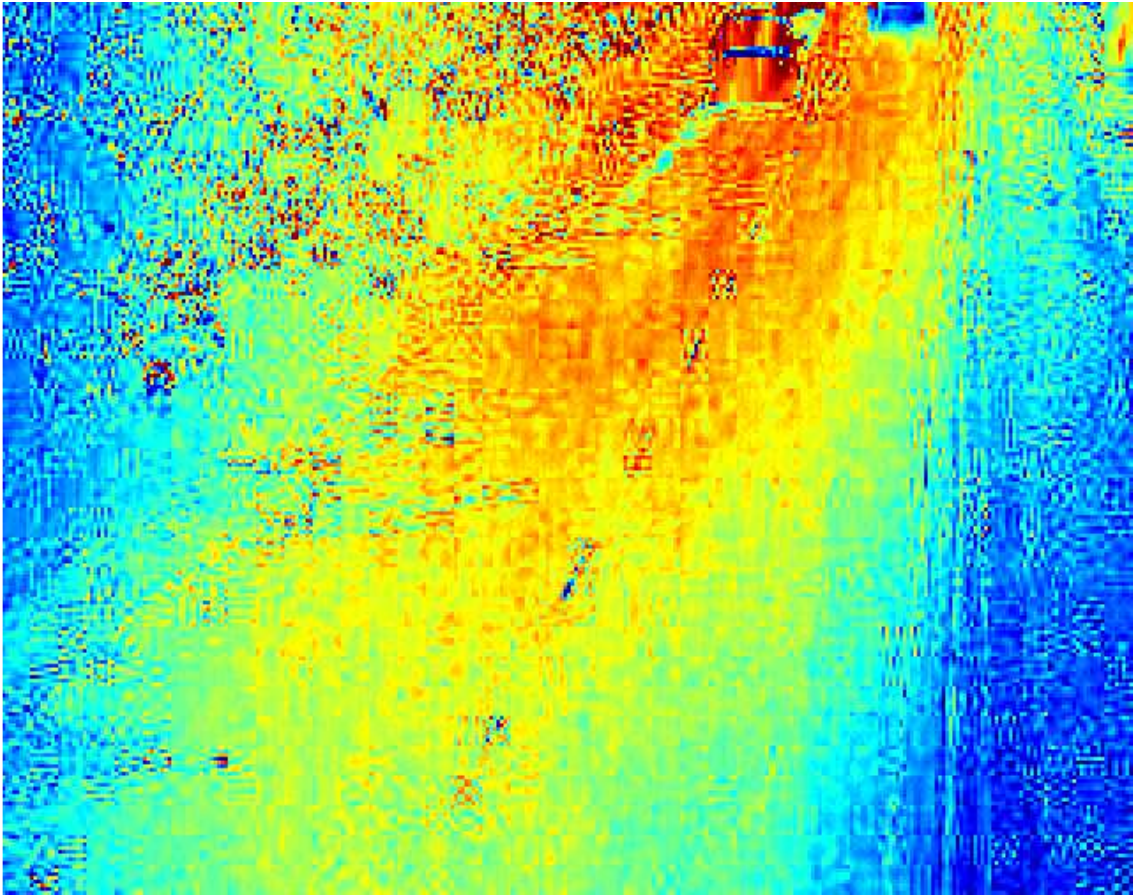}&
\includegraphics[width=0.135\linewidth]{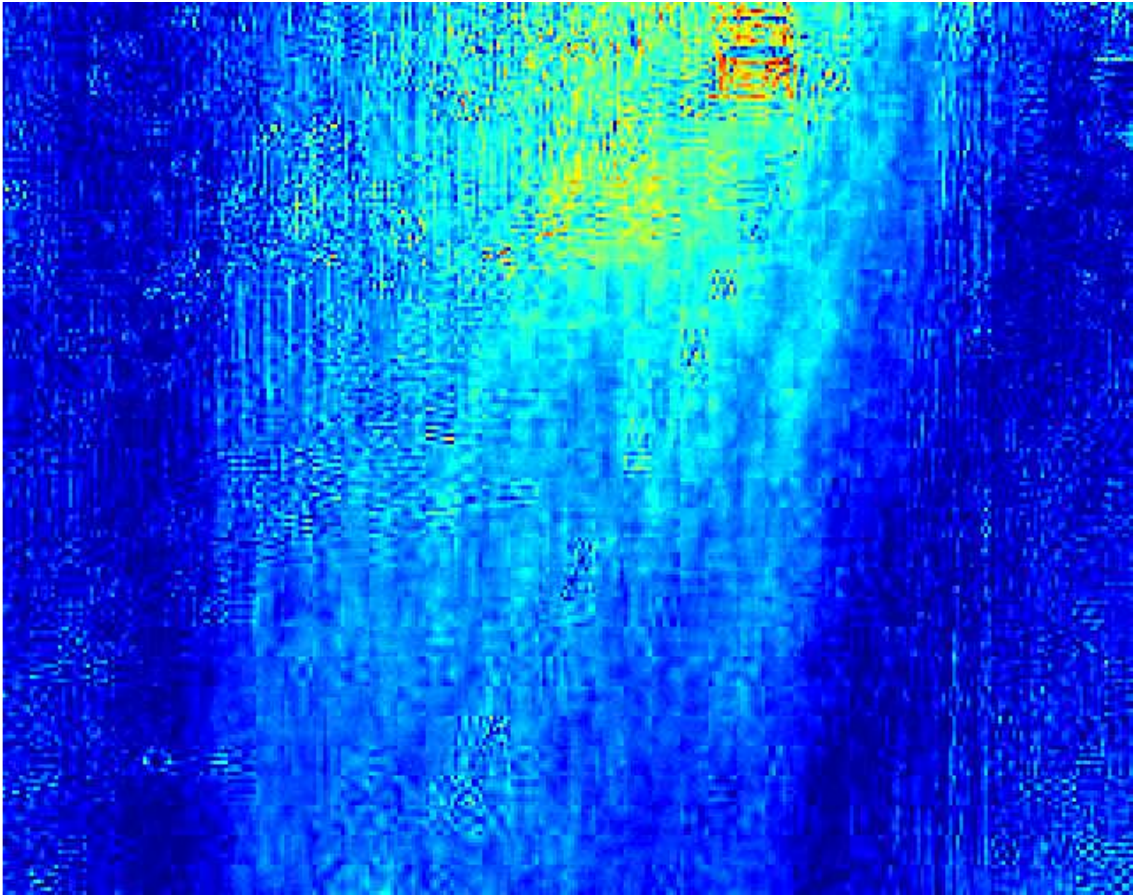} &
\includegraphics[width=0.135\linewidth]{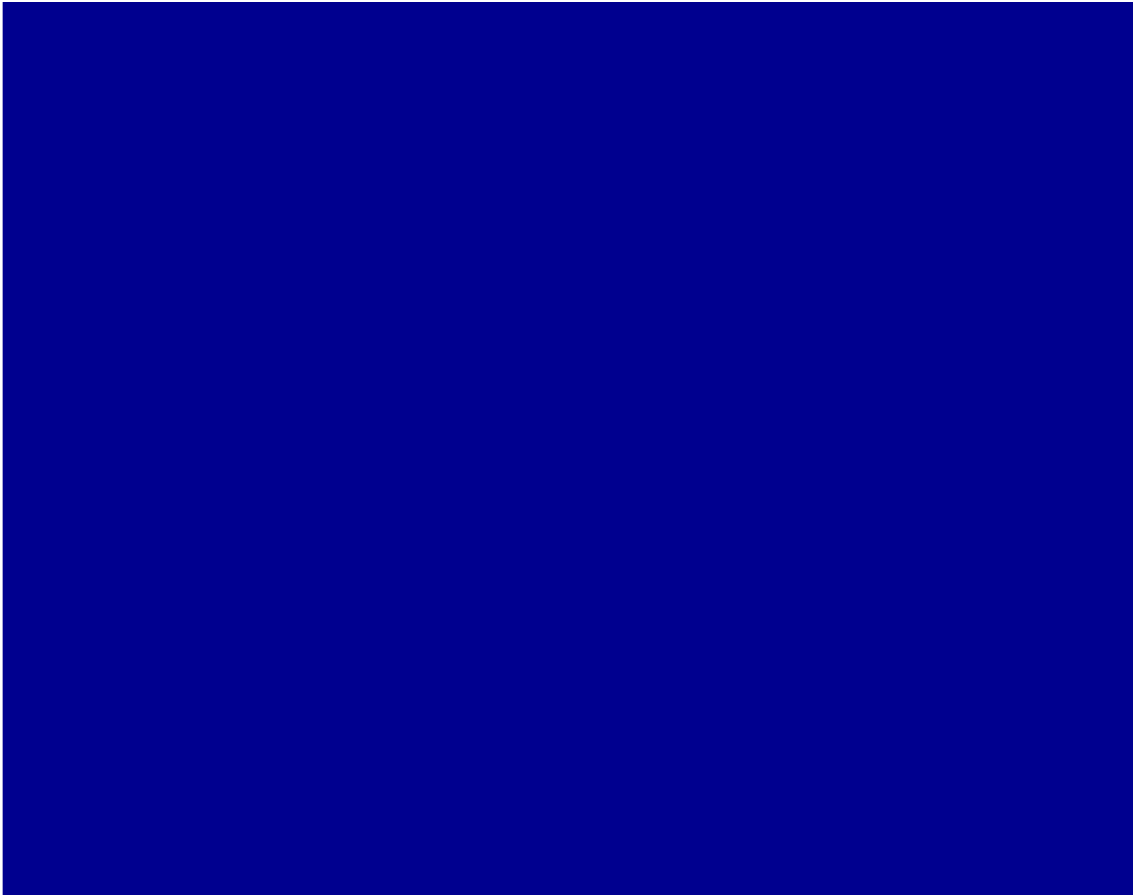} \\

\end{tabular}
\caption{The rainy frame, rain streaks removal results, extracted rain streaks and corresponding error images by different methods with synthetic rain streaks in \textbf{case 3}, respectively. The corresponding videos from top to bottom are the ``'foreman'', ''bus'', ''waterfall'' and ''highway''.
From left to right are: the rainy data (or the color bar), results by TCL \cite{kim2015video}, DDN \cite{fu2017clearing}, (SE \cite{Wei_2017_ICCV}, MS-CSC \cite{li2018video},)  FastDeRain, and the ground truth (GT), respectively.}
\label{fig_case3}
\end{figure}

Four videos are selected as the clean background.
Three videos\footnote{\url{http://trace.eas.asu.edu/yuv/}}, named ``foreman'' with the size of $144\times176\times3\times160$, ``bus'' and ``waterfall'' with the size of $288\times352\times3\times100$, are captured by dynamic cameras, while the other one\footnote{\url{http://www.changedetection.net}}, named `` highway'' with the size of $240\times320\times3\times100$, are recorded by a static camera.

SE \cite{Wei_2017_ICCV} and MS-CSC \cite{li2018video} are designed mainly for the videos captured by static cameras, and directly applying them on the video captured by dynamic camera would result in poor performances (see the gray values Table \ref{QC}).
Therefore, for a fair comparison, the compared methods included DDN \cite{fu2017clearing} and TCL \cite{kim2015video} when dealing with the synthetic rainy data generated on the videos ``foreman'' ``bus'' and ``waterfall''.
When dealing with the rainy data simulated with the video ``highway'', SE \cite{Wei_2017_ICCV} and MS-CSC \cite{li2018video} would be brought into comparison.

\paragraph{Quantitative comparisons}
For quantitative assessment, the peak signal-to-noise ratio (PSNR) of the whole video, and the structural similarity (SSIM) \cite{wang2004image}, the feature similarity (FSIM) \cite{zhang2011fsim}, the visual information fidelity (VIF) \cite{sheikh2006image}, the universal image quality index (UIQI) \cite{wang2002universal}, and the gradient magnitude similarity deviation (GMSD, smaller is better) \cite{xue2014gradient} of each frame are calculated.
The PSNR, the corresponding mean values of SSIM FSIM VIF and UIQI, and the running time are reported in Table \ref{QC}, in which the best quantitative values are in boldface.

As observed in Table \ref{QC}, our method considerably outperformed the other {\color{red}four} state-of-the-art methods in terms of all the selected quality assessment indexes.
Notably, in many cases, the performances of the single-image-based deep learning method DNN \cite{fu2017clearing} surpassed the those of the video-based method TCL \cite{kim2015video}.
This is in agreement with the aforementioned rationality of considering comparisons with the single-image-based method.

The running time of the our FastDeRain is extremely low.
In particular, our method took less than 10 seconds when dealing with all the synthetic data.
Although a tensor system might be expected to be computationally expensive, our algorithm, with closed-form solutions to its sub-problems and a time complexity of approximately $O(mnt\text{log}(mnt))$ for an input video with a resolution of $m\times n$ and $t$ frames, is expected to be efficient.
In the meantime, the aforementioned implementation on the GPU device also largely accelerated our algorithm.

\paragraph{Visual comparisons}\label{sec:vc}
Fig. \ref{fig_case1}, \ref{fig_case2} and \ref{fig_case3} exhibit the results conducted on videos with synthetic rain streaks in case 1,  case 2 and case 3, respectively.
In Fig. \ref{fig_case1}, since the angles of rain streaks in case 1 increase with time, we display the frames at the beginning or end.
Meanwhile, only one frame is exhibited in Fig. \ref{fig_case2}, Fig. \ref{fig_case3} on account of that the rain streaks in every frame are of various directions.

In Fig. \ref{fig_case1}, all the methods removed almost all of the rain streaks and the proposed method maintained the best background.
Many details in the background were incorrectly extracted to the rain streaks by DDN and TCL.
It can be found in the 6-th row of Fig. \ref{fig_case1}, i.e., the error images of the results on the video ``bus'', that little vertical patterns were mistakenly extracted as the rain streaks by the proposed method.

For the rain streaks in case 2, the denser rain streaks imply that it is more difficult than rain streaks in case 1.
For instance, the denser rain streaks visibly degraded the performance of SE.
From Fig. \ref{fig_case2}, we can find that our method preserved the backgrounds well and other four methods erased the details of the backgrounds.

\begin{figure*}[!htbp] 
\centering\footnotesize\setlength{\tabcolsep}{2pt}
\setlength{\tabcolsep}{1pt}
\begin{tabular}{ccccccc}
Rainy&$\alpha_1=10^{-15}$&$\alpha_2=10^{-15}$&$\alpha_3=10^{-15}$&$\alpha_4=10^{-15}$&FastDeRain&Ground truth\\
\includegraphics[width=0.10\linewidth]{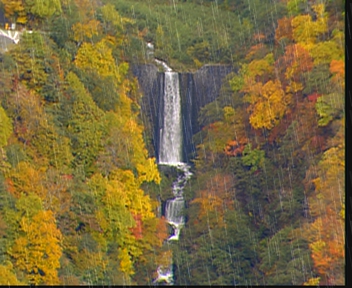}&
\includegraphics[width=0.10\linewidth]{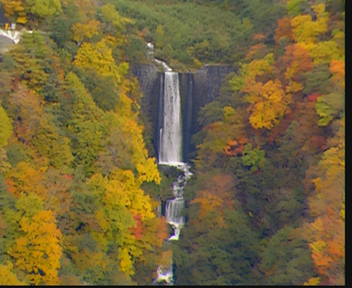}&
\includegraphics[width=0.10\linewidth]{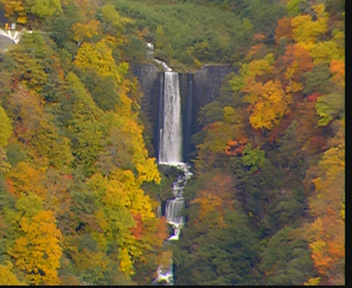}&
\includegraphics[width=0.10\linewidth]{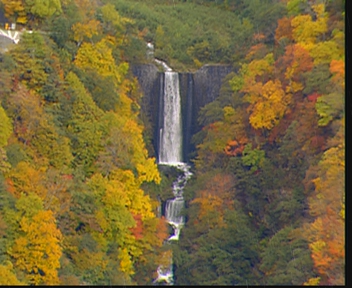}&
\includegraphics[width=0.10\linewidth]{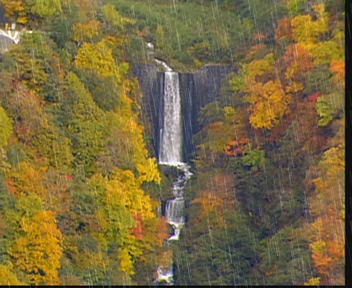}&
\includegraphics[width=0.10\linewidth]{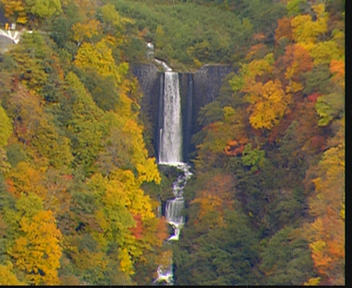}&
\includegraphics[width=0.10\linewidth]{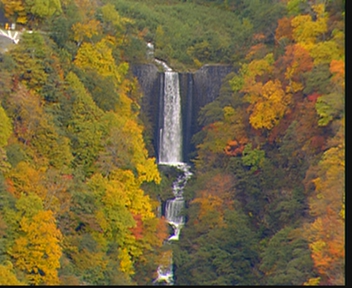}\\

&
\includegraphics[width=0.10\linewidth]{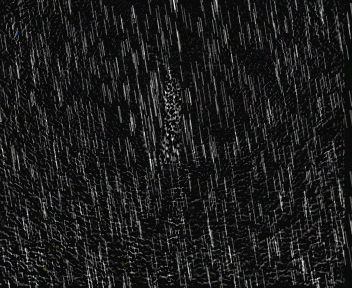}&
\includegraphics[width=0.10\linewidth]{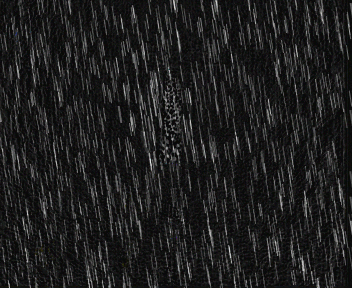}&
\includegraphics[width=0.10\linewidth]{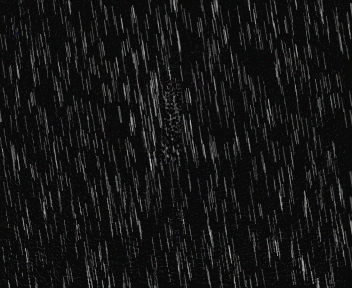}&
\includegraphics[width=0.10\linewidth]{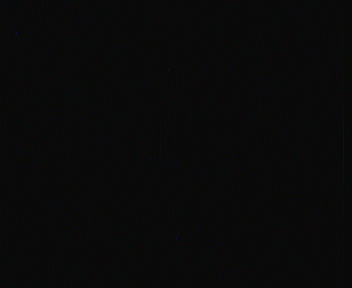}&
\includegraphics[width=0.10\linewidth]{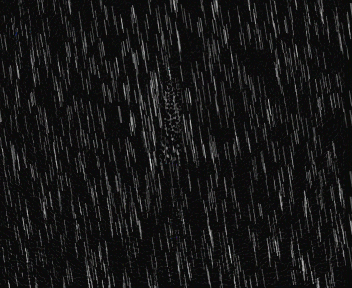}&
\includegraphics[width=0.10\linewidth]{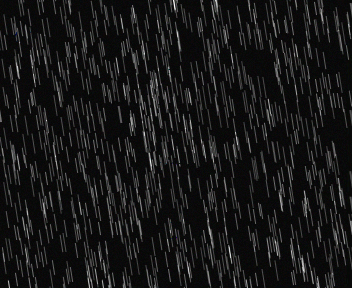}\\

\includegraphics[width=0.10\linewidth]{figs/component/bar_iccv.png} &
\includegraphics[width=0.10\linewidth]{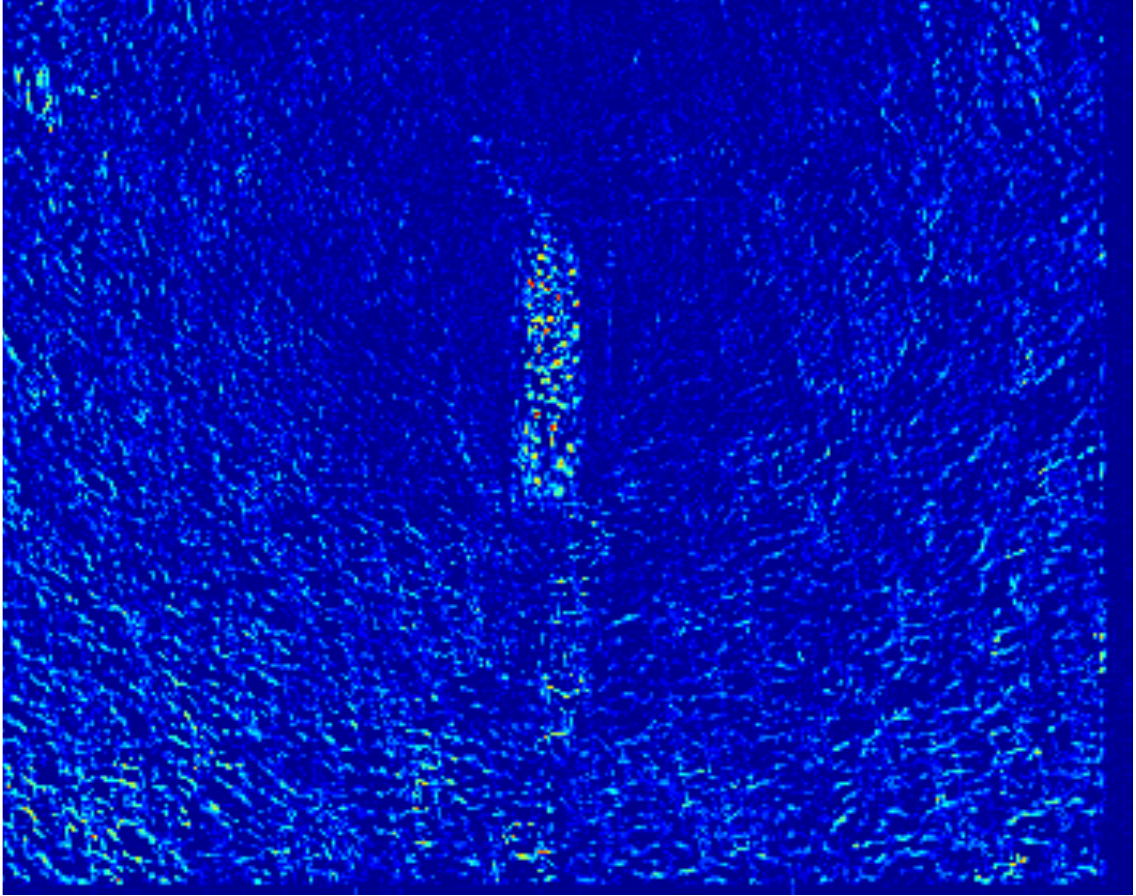}&
\includegraphics[width=0.10\linewidth]{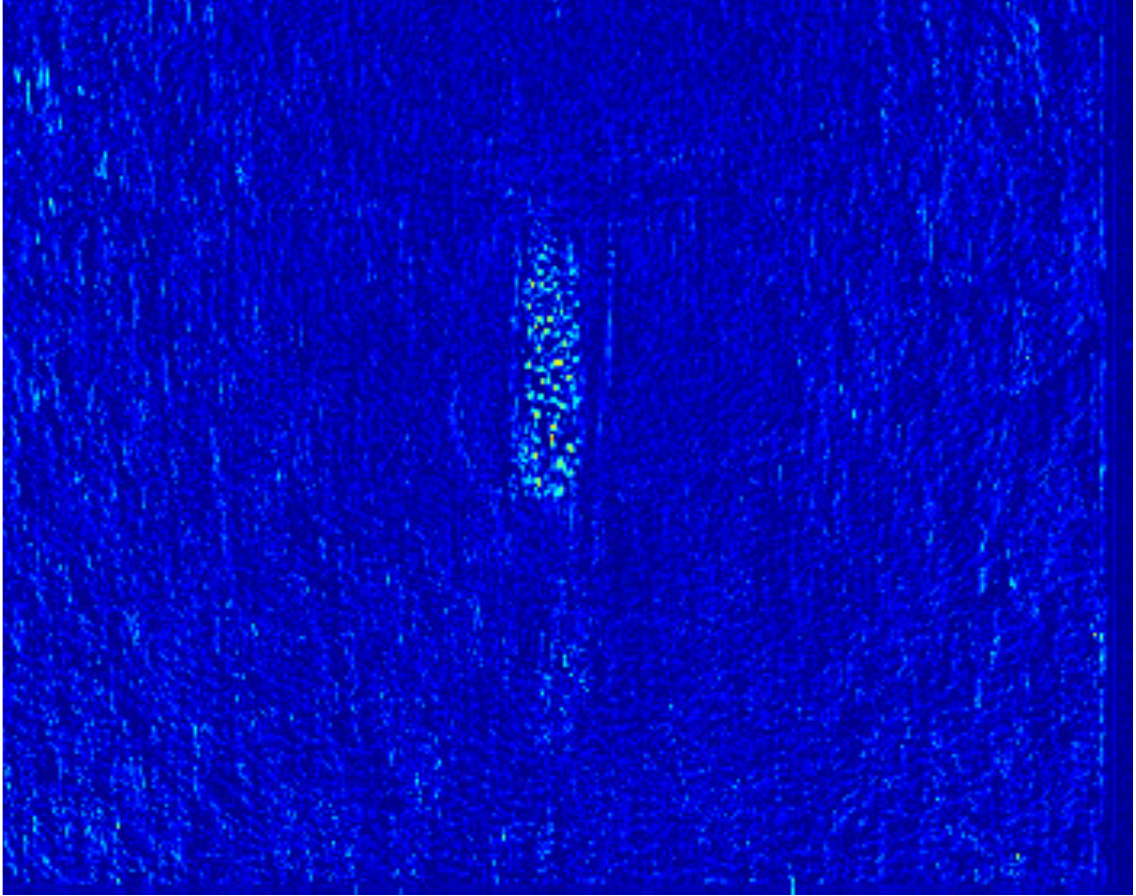}&
\includegraphics[width=0.10\linewidth]{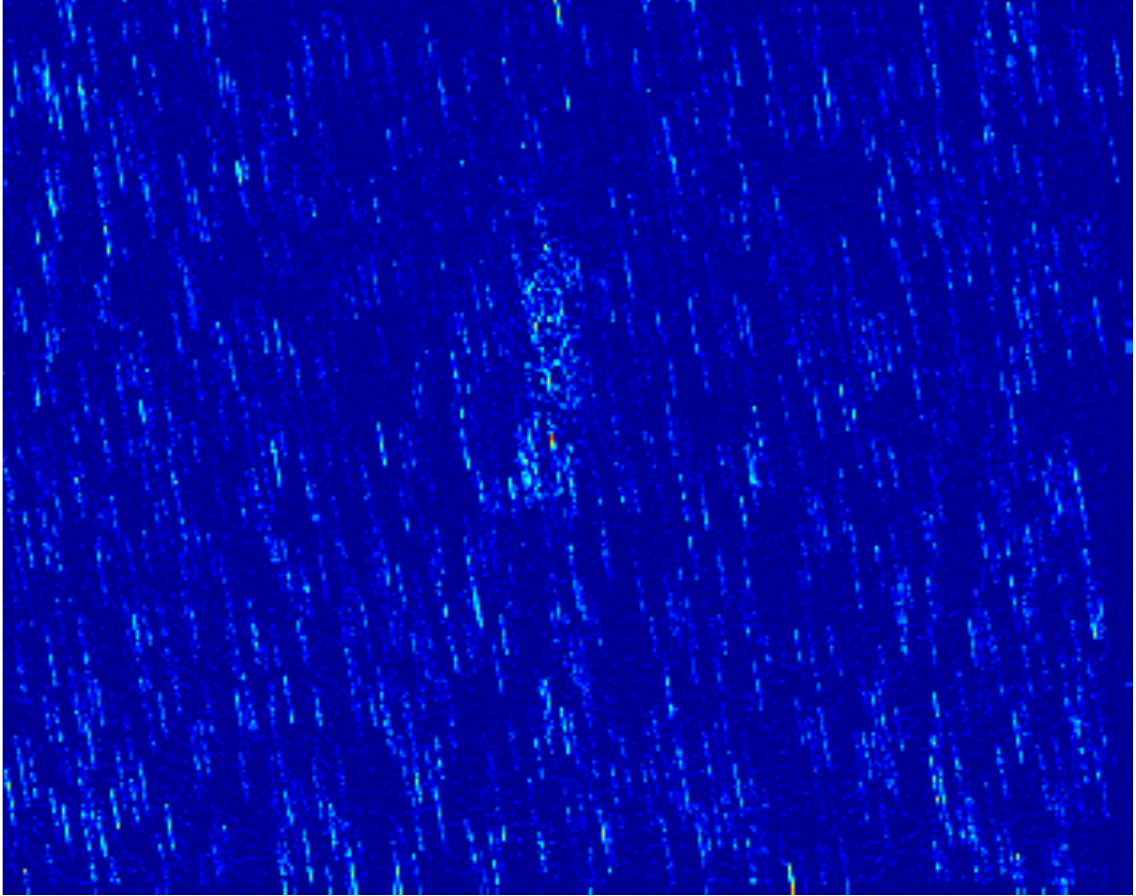}&
\includegraphics[width=0.10\linewidth]{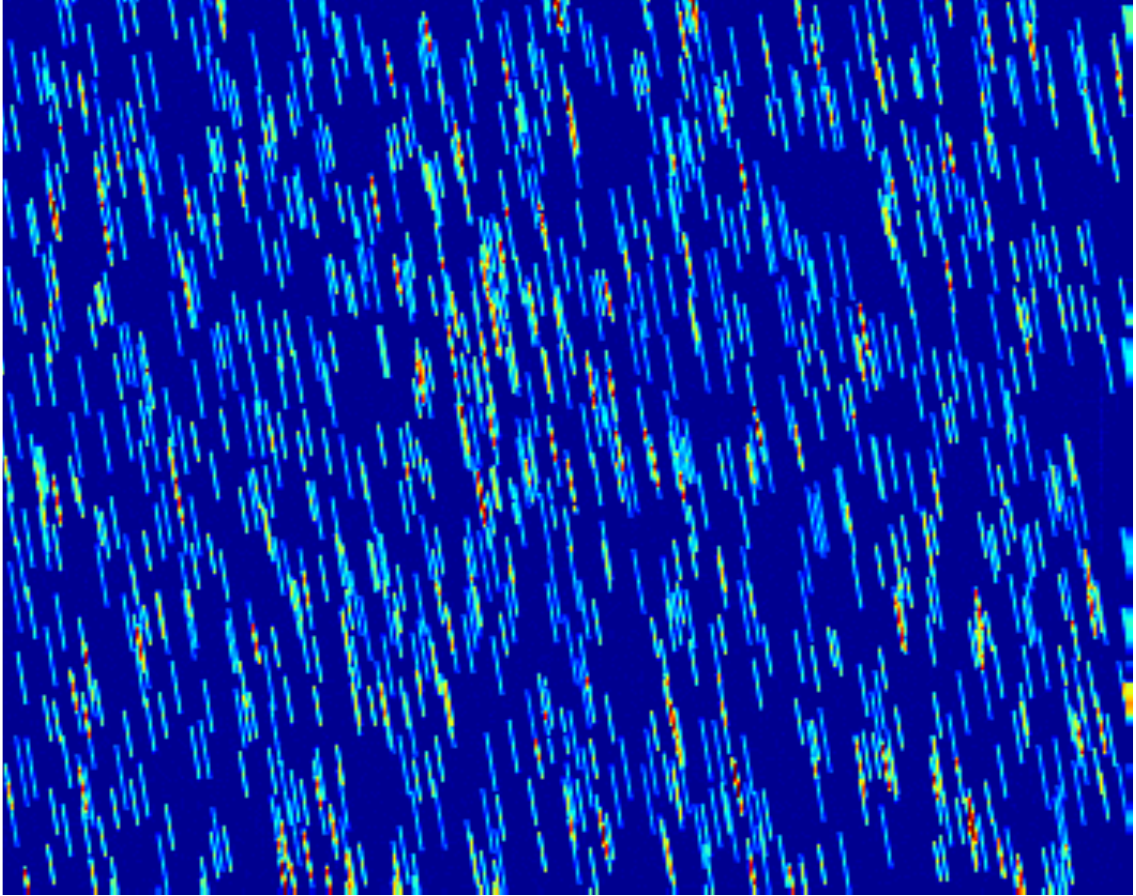}&
\includegraphics[width=0.10\linewidth]{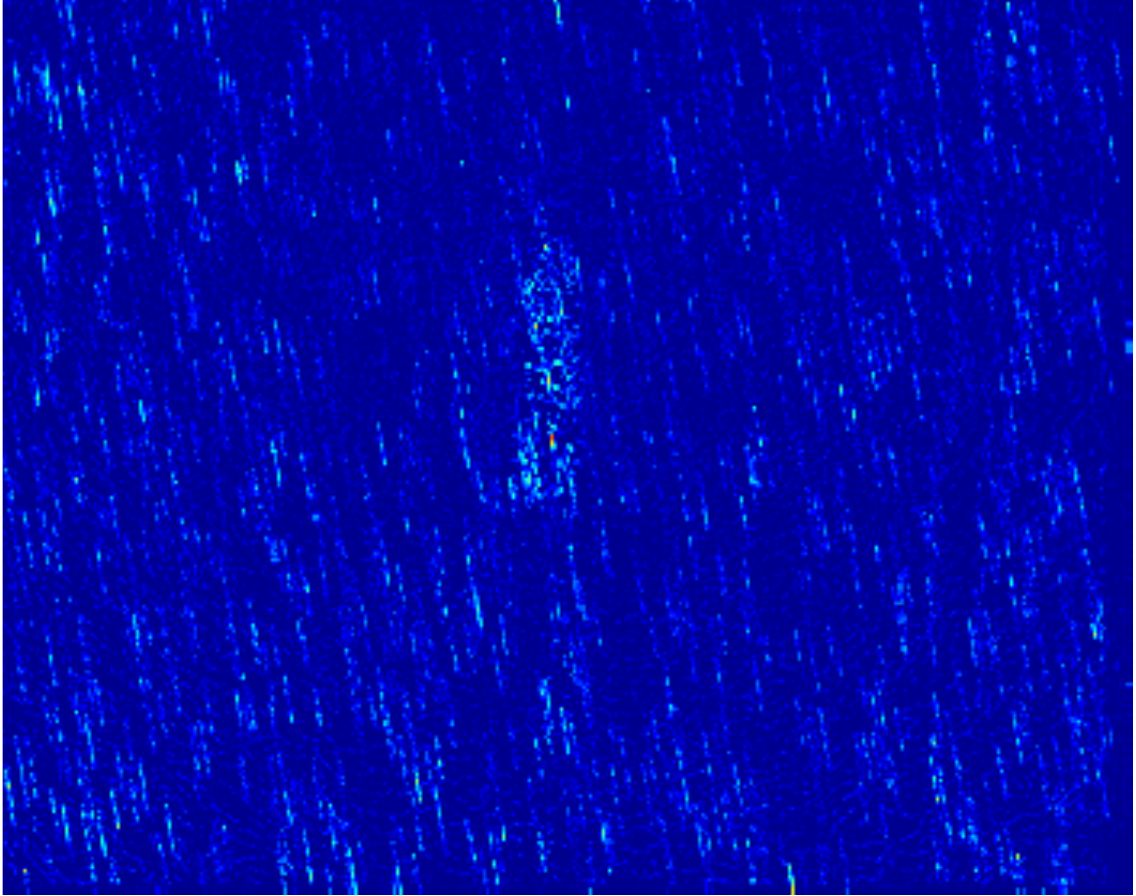}&
\includegraphics[width=0.10\linewidth]{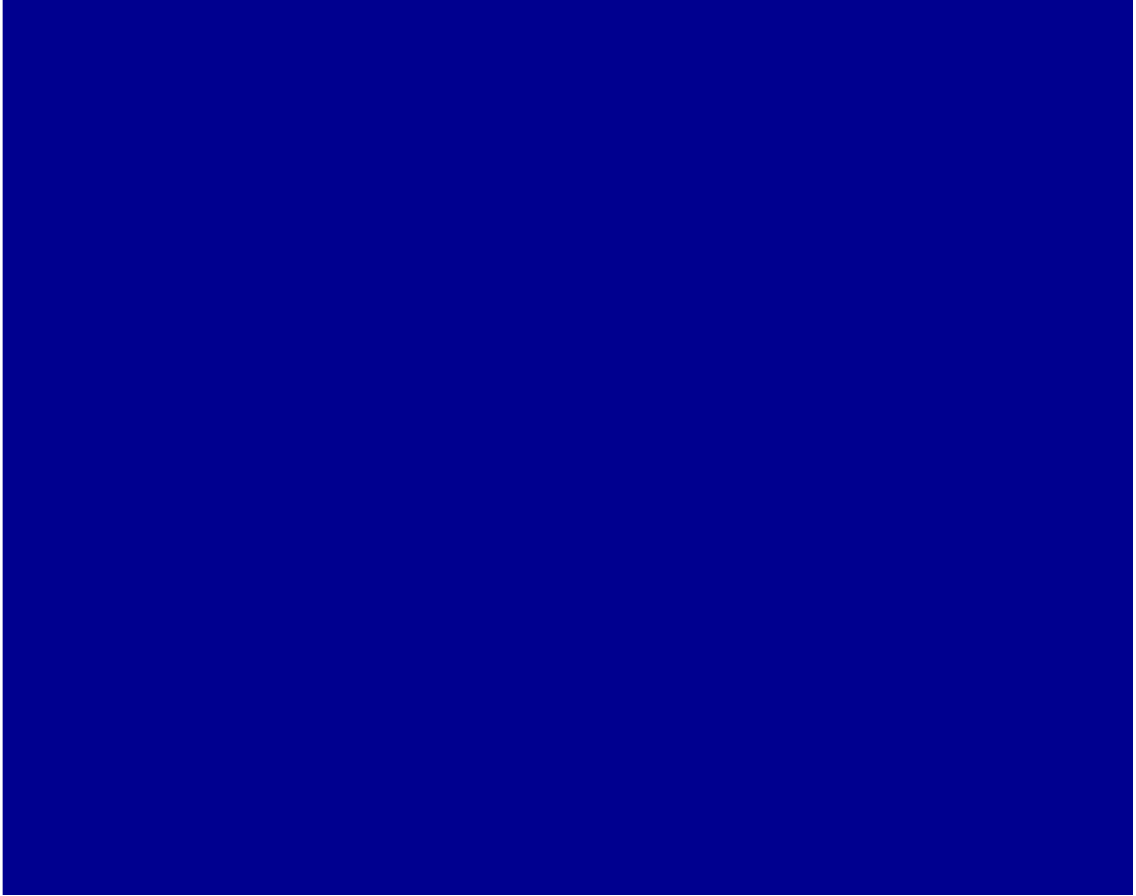}
\end{tabular}
\caption{The top row shows the 80th frame of the rainy video, the results by FastDeRain and its degraded versions, in which the $\alpha_i$s in Eq. (\ref{ADM}) are set as $10^{-15}$ in turn, and the ground truth clean video, respectively.
The middle row presents the extracted rain streaks by FastDeRain and its degraded versions and the ground truth rain streaks, while the color bar and corresponding error images are exhibited in the bottom row.}
\label{component2}
\end{figure*}

\begin{figure*}[!htb] 
\begin{center}
\includegraphics[width=0.9\linewidth]{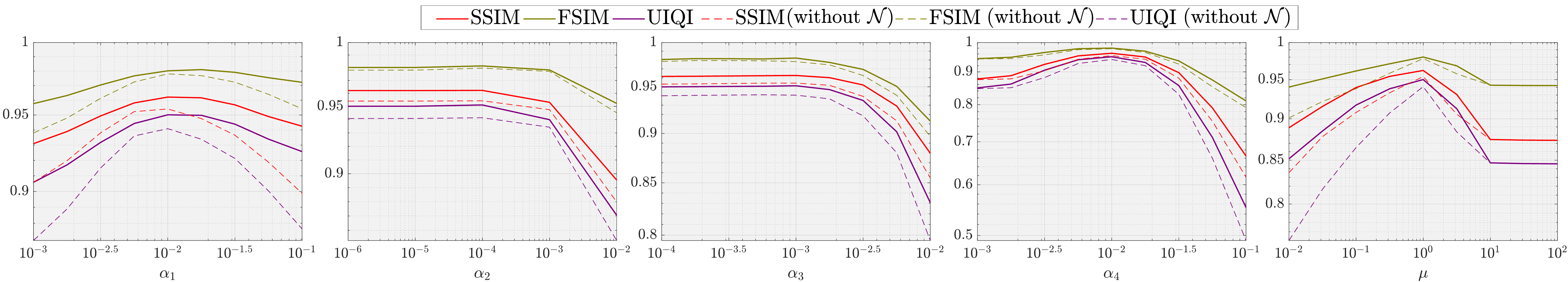}\\
\caption{The mean SSIM FSIM and UIQI values with respect to different values of $\alpha_1$, $\alpha_2$, $\alpha_3$, $\alpha_4$ and $\mu$. The solid lines are corresponding to the results of FastDeRain while the dashed lines are related to the results obtained by our method without the $\mathcal{N}$ in Eq. (\ref{obnoise}). }
\label{para}
\end{center}
\end{figure*}

In Fig. \ref{fig_case3}, the proposed method removed most of the rain streaks and considerably preserves the background.
Other methods tended to obtain over de-rain or under de-rain results.
Considering the similarity of the extract rains streaks to the ground truth rain streaks, our FastDeRain held obvious advantages.

In summary, for these different types of synthetic data, our method can simultaneously remove almost all rain streaks while commendably preserving the details of the underlying clean videos.


\paragraph{Discussion of each component}

\begin{figure}[!htbp] 
\centering
\includegraphics[width=0.85\linewidth]{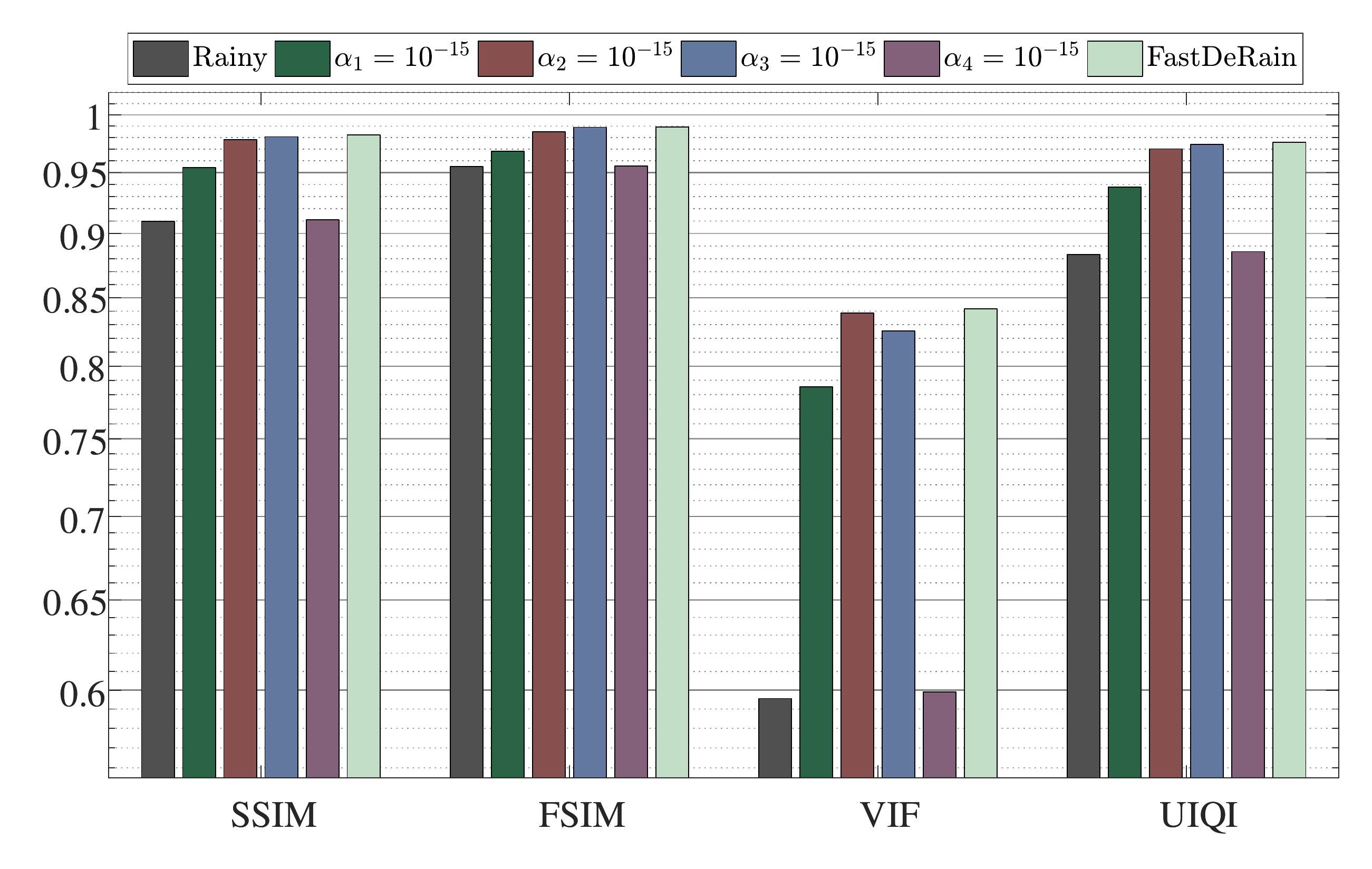}
\caption{The quantitative performances of the proposed method and its degraded versions, in which the $\alpha_i$s in Eq. (\ref{ADM}) are set as $10^{-15}$ in turn.}
\label{component}
\end{figure}

There are four components in our model (\ref{mainModel2}).
To elucidate their distinct effects, we degrade our method by setting each $\alpha_i$ ($i=1,2,3,4$) equal to $10^{-15}$, respectively.
These degraded methods and FastDeRain are tested on the video ``waterfall'' with synthetic rain streaks in case 1.
We present the quantitative assessments in Fig. \ref{component} and the visual results in Fig. \ref{component2}.

From Fig. \ref{component} and Fig. \ref{component2}, we can conclude that all the four components contribute to the removal of rain streaks.
Specifically, (a) when setting $\alpha_1=10^{-15}$, the rain streaks tend to be intermittent along the vertical direction;
(b) the rain streaks are fatter when the sparsity term contributes little;
(c) some rain streaks remain in the background when the horizontal smoothness of the background is not sufficiently enhanced;
(d) the temporal continuity seems overwhelmingly important since that without this regularization term our method nearly failed.


\paragraph{Parameters}
To examine the performance of the proposed FastDeRain with respect to different parameters, we conduct a series of experiments on the rainy data on synthetic video ``waterfall'' with the synthetic rain streaks in case 1 and the Gaussian noise with zero mean and standard deviation 0.02.
In Fig. \ref{para}, a parameter analysis is presented and the SSIM FSIM and MUIQI are selected.
Based on guidance from Fig. \ref{para}, our tuning strategy is as following:
(1) set $\alpha_2$ and $\alpha_3$ as $10^{-5}$ and other $\alpha_i$s to 0.01, and $\mu = 1$,
(2) tune $\alpha_1$ and $\alpha_4$ until the results are barely satisfactory,
(3) and then fix $\alpha_1$ and $\alpha_4$ and enlarge $\alpha_2$ and $\alpha_3$ to further improve the performance.
The tuning principle is as follows: when some of the texture or detail of the clean video is extracted into the estimated rain streaks, we increase $\alpha_2$ and $\alpha_1$ or decrease $\alpha_4$ and $\alpha_3$, and we do the opposite when rain streaks remain in the estimated rain-free content.
Our recommended set of candidate values for $\alpha_1$ through $\alpha_4$ is $\{0.00001, 0.00003, 0.0001, 0.0003, 0.001,0.003, 0.01\}$.
The Lagrange parameter $\mu$ is suggested to be 1.
In practice, the time cost for the empirical tuning of the parameters is not much. 

\begin{figure*}[!htb] 
\scriptsize\renewcommand\arraystretch{1}
\setlength{\tabcolsep}{2pt}
\centering
\begin{tabular}{ccccc}
\subfigure[Video ``waterfall'' (case 1)]{
\begin{tabular}{cccc}
Rainy &\cite{Jiang_2017_CVPR} & FastDeRain&GT\\
\includegraphics[width=0.07\linewidth]{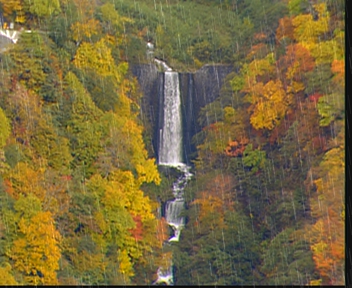} &
\includegraphics[width=0.07\linewidth]{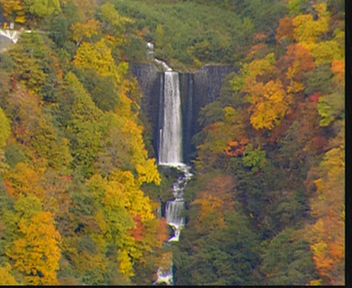} &
\includegraphics[width=0.07\linewidth]{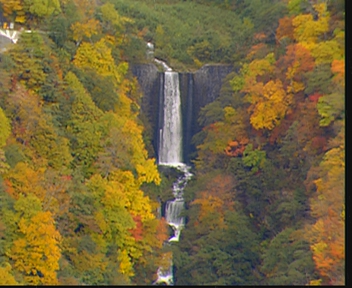} &
\includegraphics[width=0.07\linewidth]{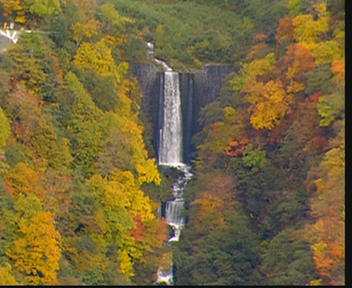} \\
&\includegraphics[width=0.07\linewidth]{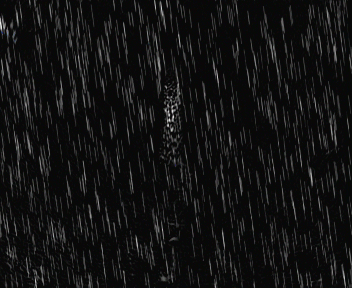} &
\includegraphics[width=0.07\linewidth]{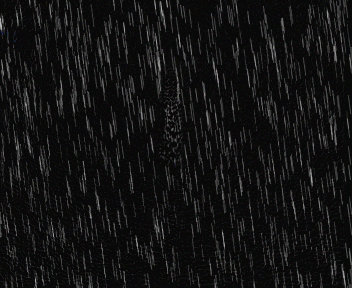} &
\includegraphics[width=0.07\linewidth]{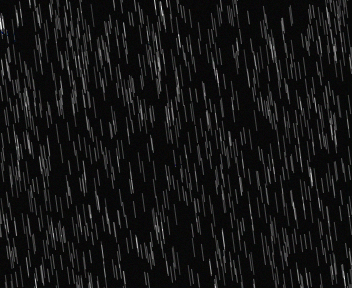} \\
\includegraphics[width=0.07\linewidth]{figs/component/bar_yuv.png}&
\includegraphics[width=0.07\linewidth]{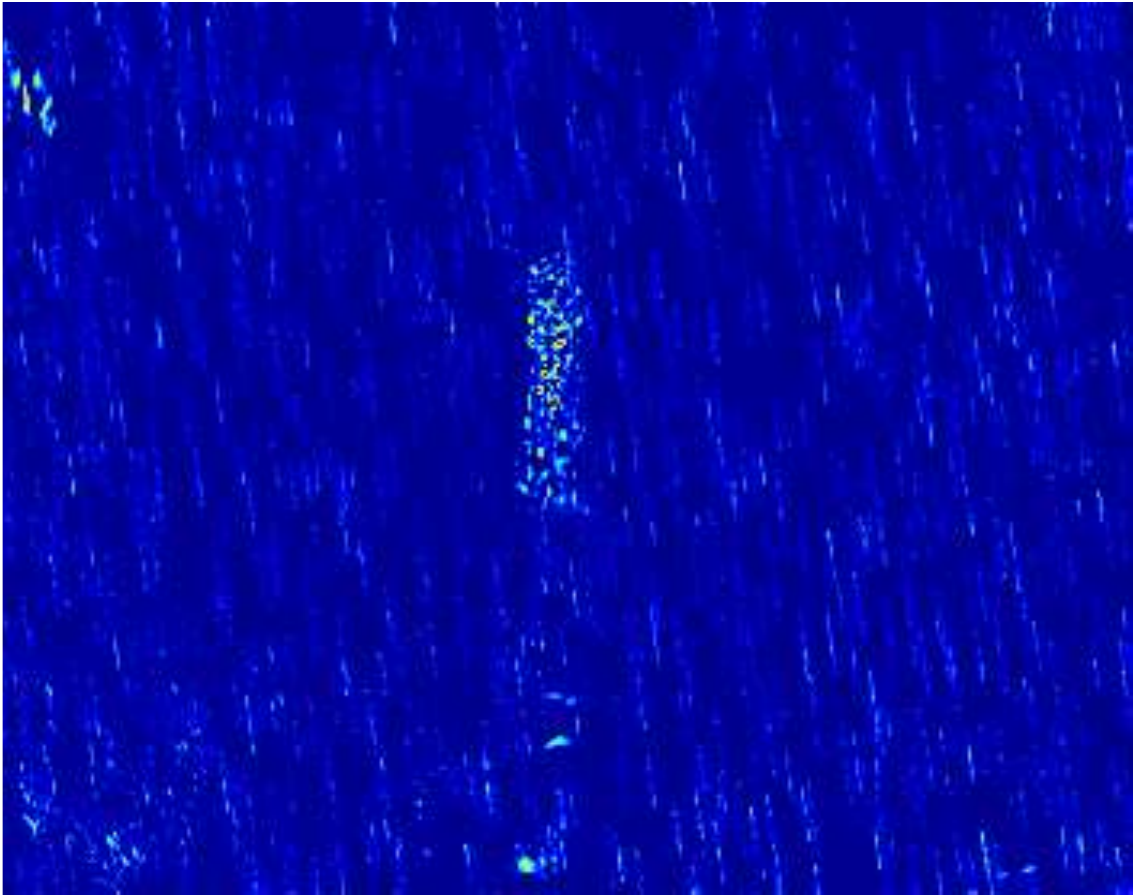}&
\includegraphics[width=0.07\linewidth]{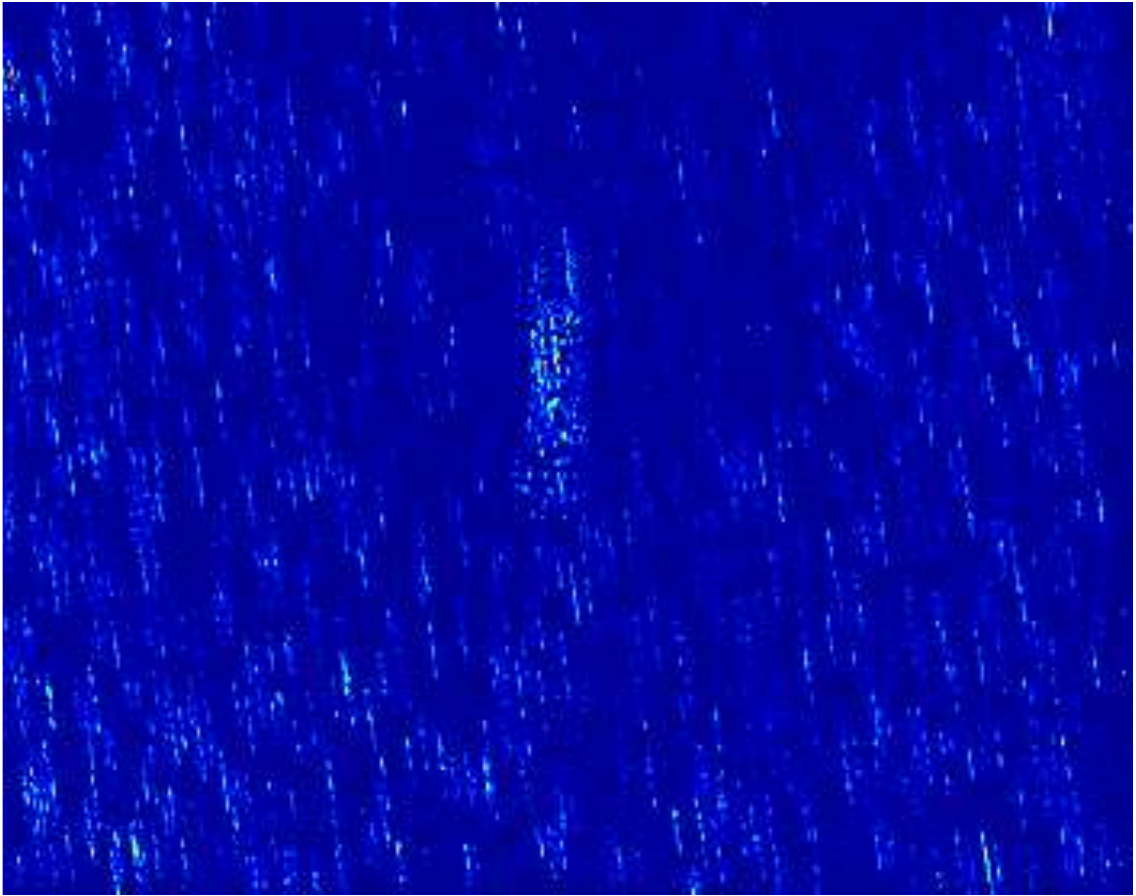} &
\includegraphics[width=0.07\linewidth]{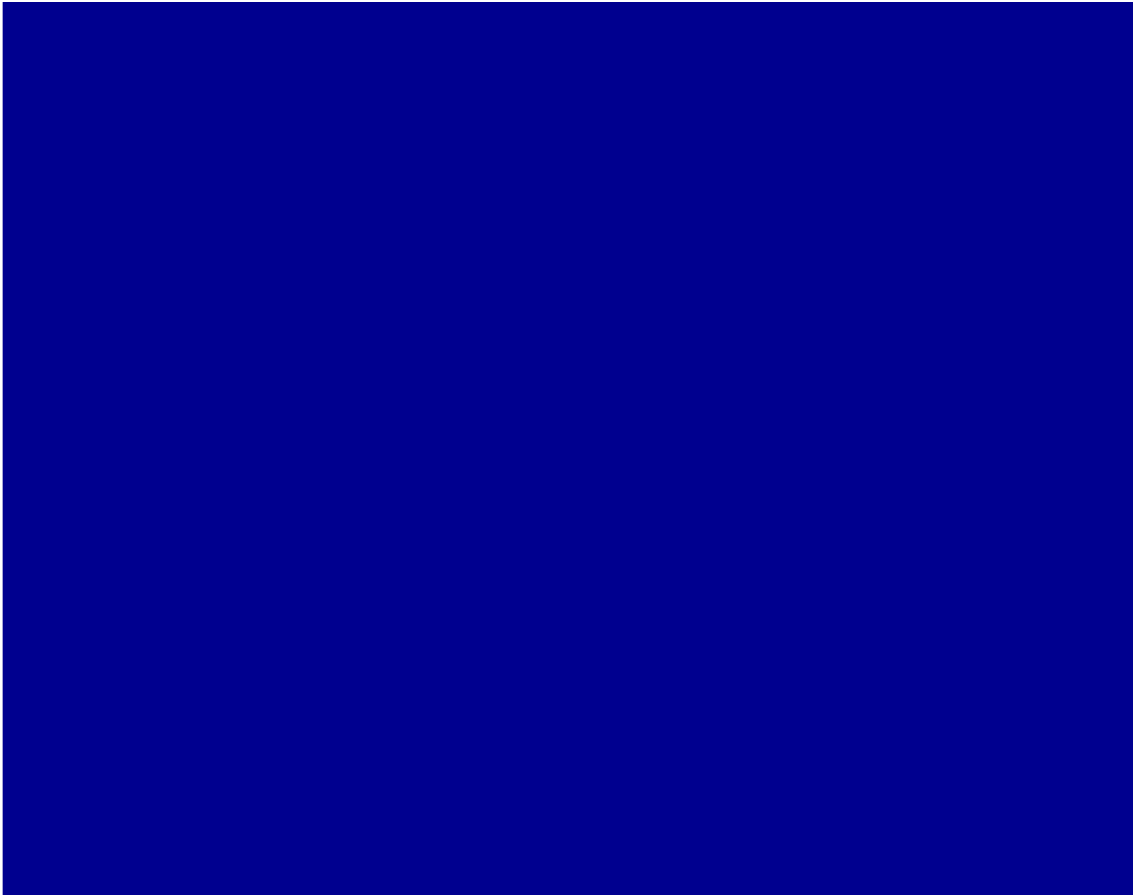} \\
\end{tabular}
}
&&
\subfigure[Video ``bus'' (case 2)]{\begin{tabular}{cccc}
Rainy &\cite{Jiang_2017_CVPR} & FastDeRain&GT\\
\includegraphics[width=0.07\linewidth]{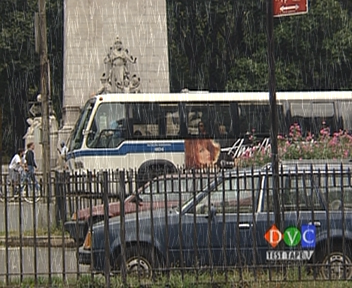} &
\includegraphics[width=0.07\linewidth]{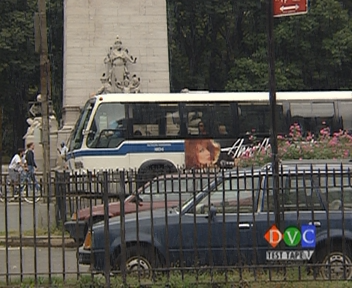} &
\includegraphics[width=0.07\linewidth]{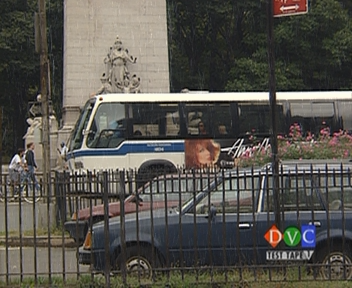} &
\includegraphics[width=0.07\linewidth]{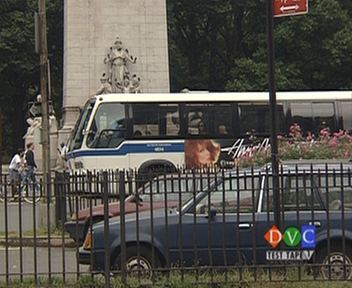} \\
&\includegraphics[width=0.07\linewidth]{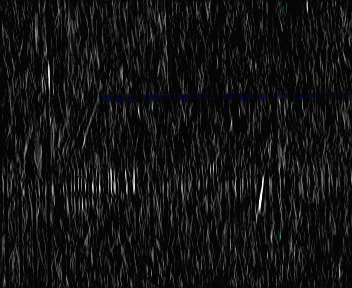} &
\includegraphics[width=0.07\linewidth]{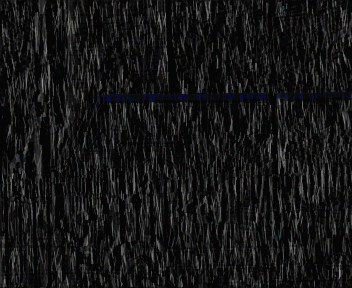} &
\includegraphics[width=0.07\linewidth]{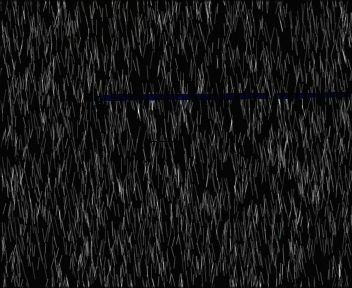} \\
\includegraphics[width=0.07\linewidth]{figs/component/bar_yuv.png}&
\includegraphics[width=0.07\linewidth]{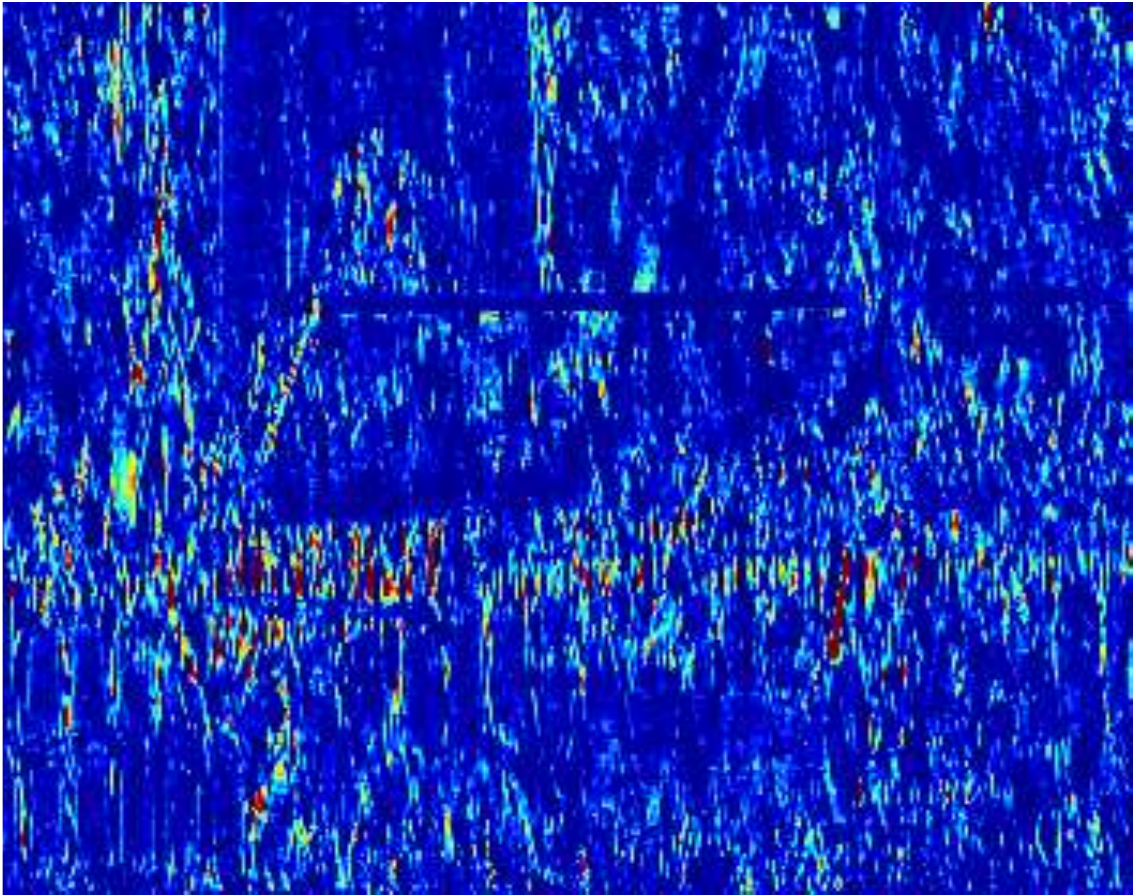}&
\includegraphics[width=0.07\linewidth]{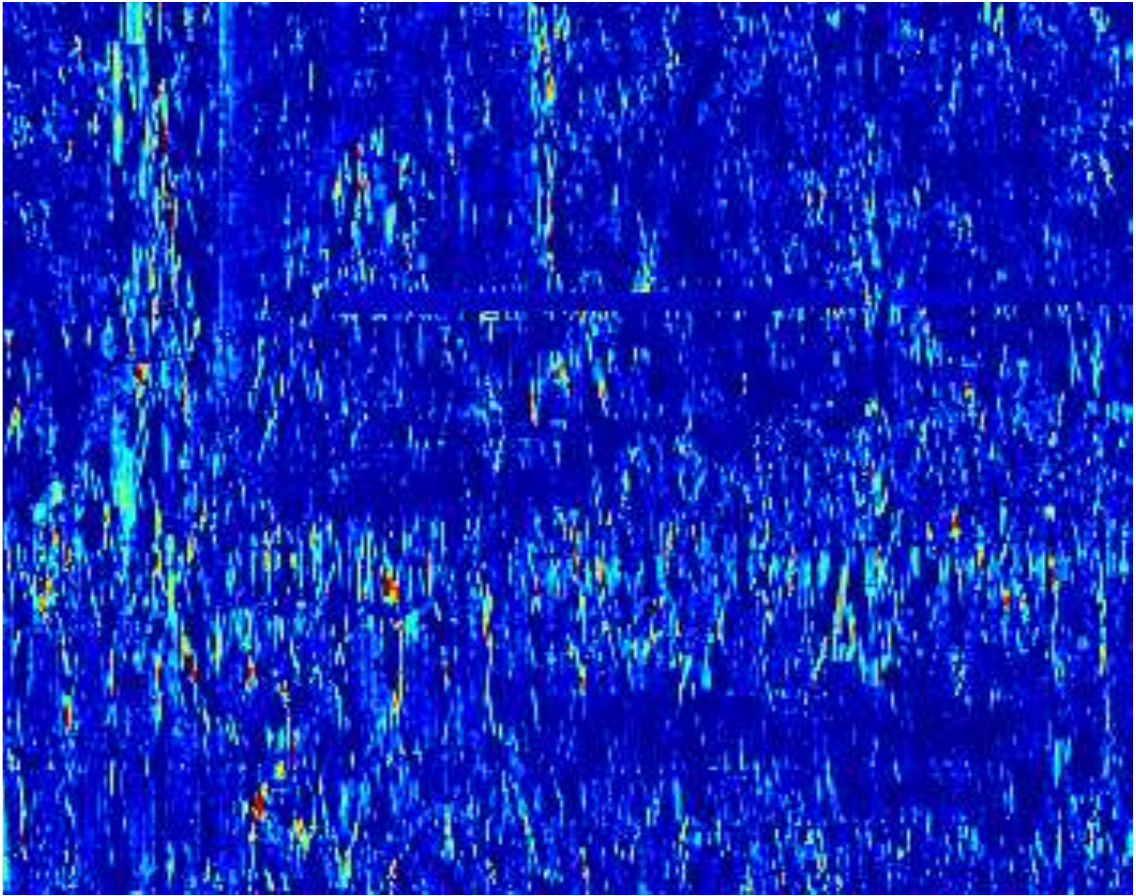} &
\includegraphics[width=0.07\linewidth]{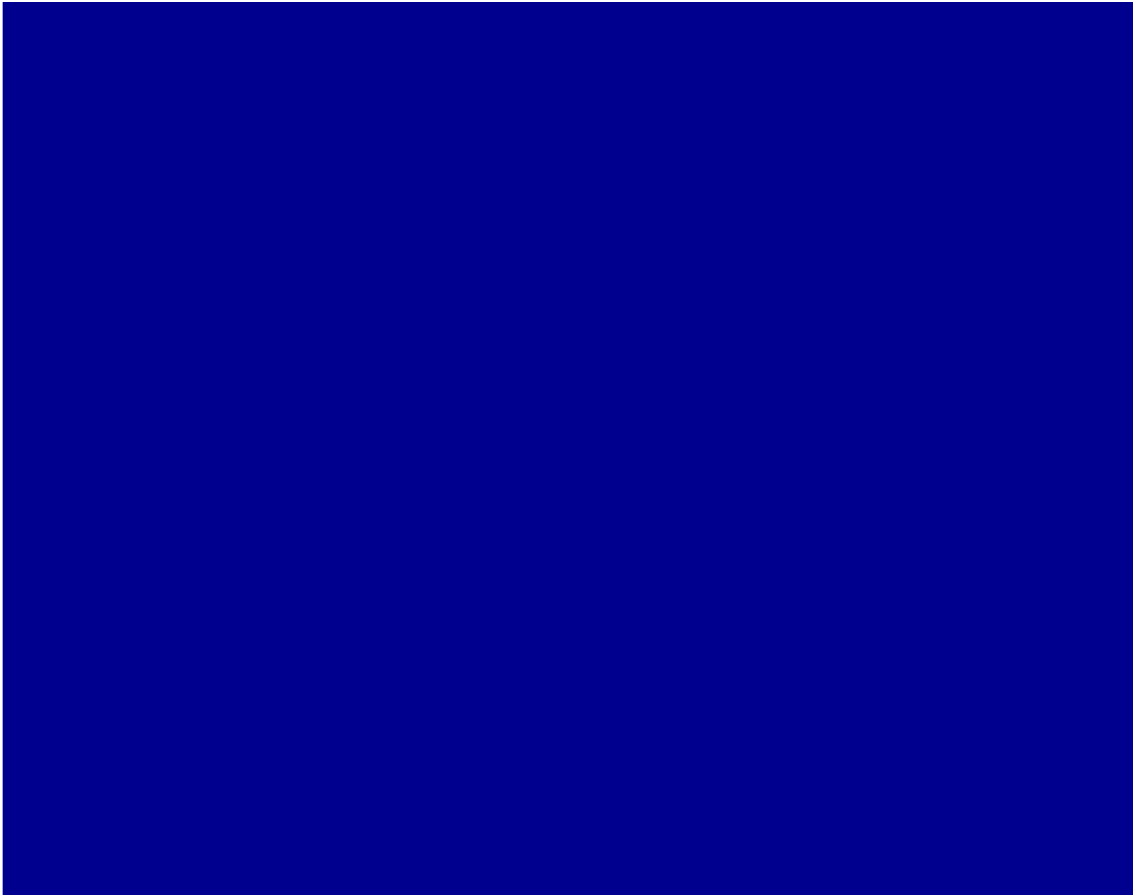} \\

\end{tabular}}
&&
\subfigure[Video ``highway'' (case 3)]{\begin{tabular}{cccc}
Rainy &\cite{Jiang_2017_CVPR} & FastDeRain&GT\\
\includegraphics[width=0.07\linewidth]{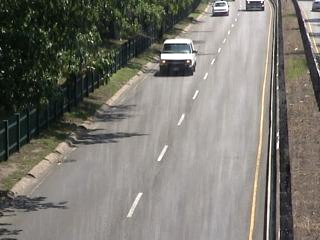} &
\includegraphics[width=0.07\linewidth]{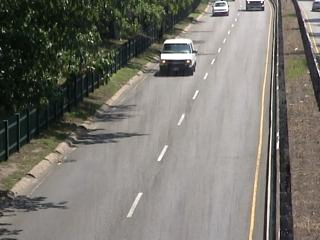} &
\includegraphics[width=0.07\linewidth]{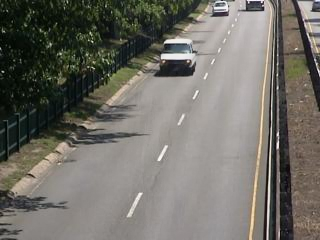} &
\includegraphics[width=0.07\linewidth]{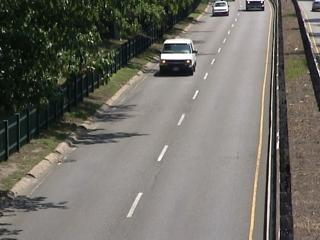} \\
&\includegraphics[width=0.07\linewidth]{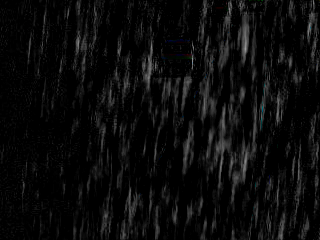} &
\includegraphics[width=0.07\linewidth]{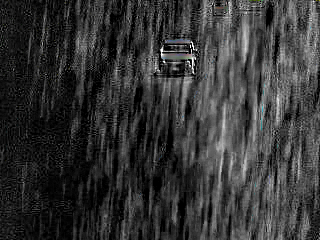} &
\includegraphics[width=0.07\linewidth]{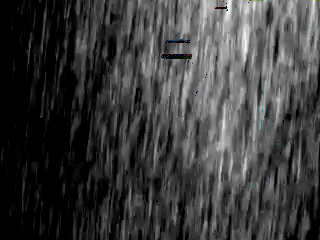} \\
\includegraphics[width=0.07\linewidth]{figs/component/bar_yuv.png}&
\includegraphics[width=0.07\linewidth]{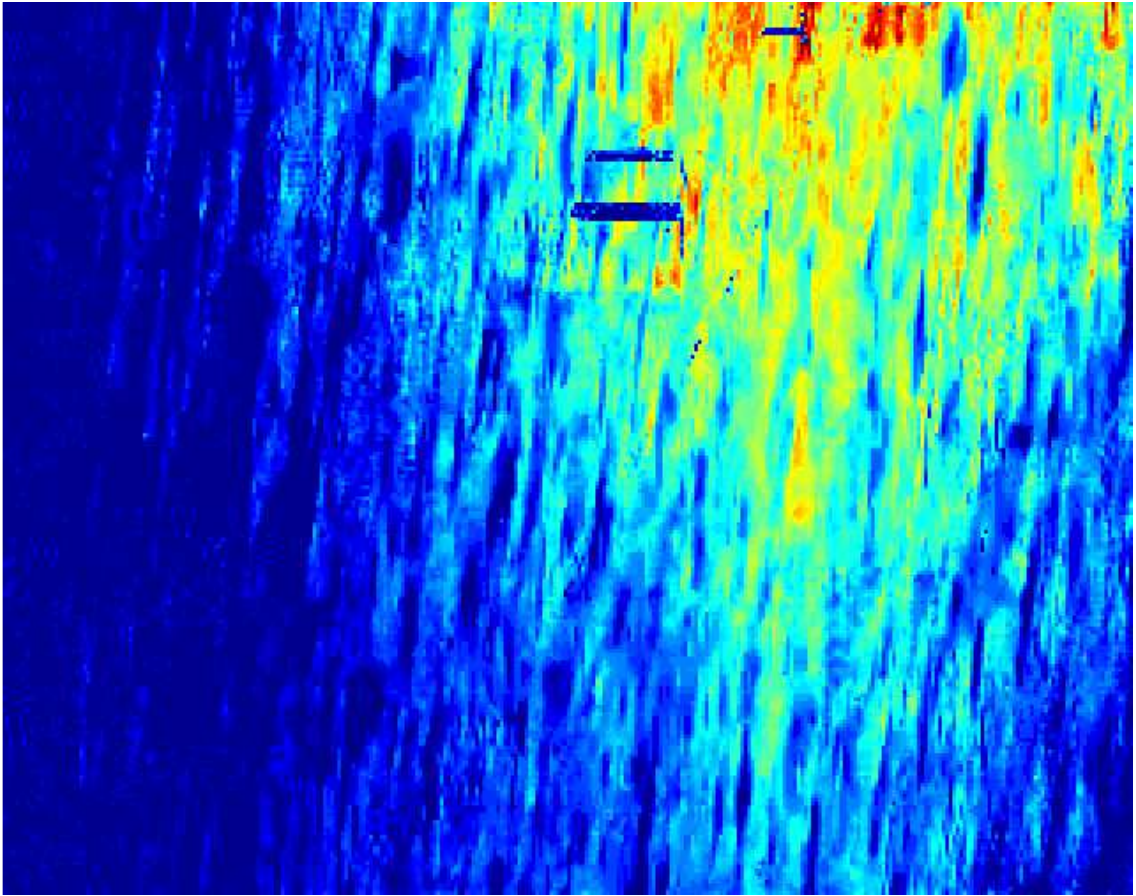}&
\includegraphics[width=0.07\linewidth]{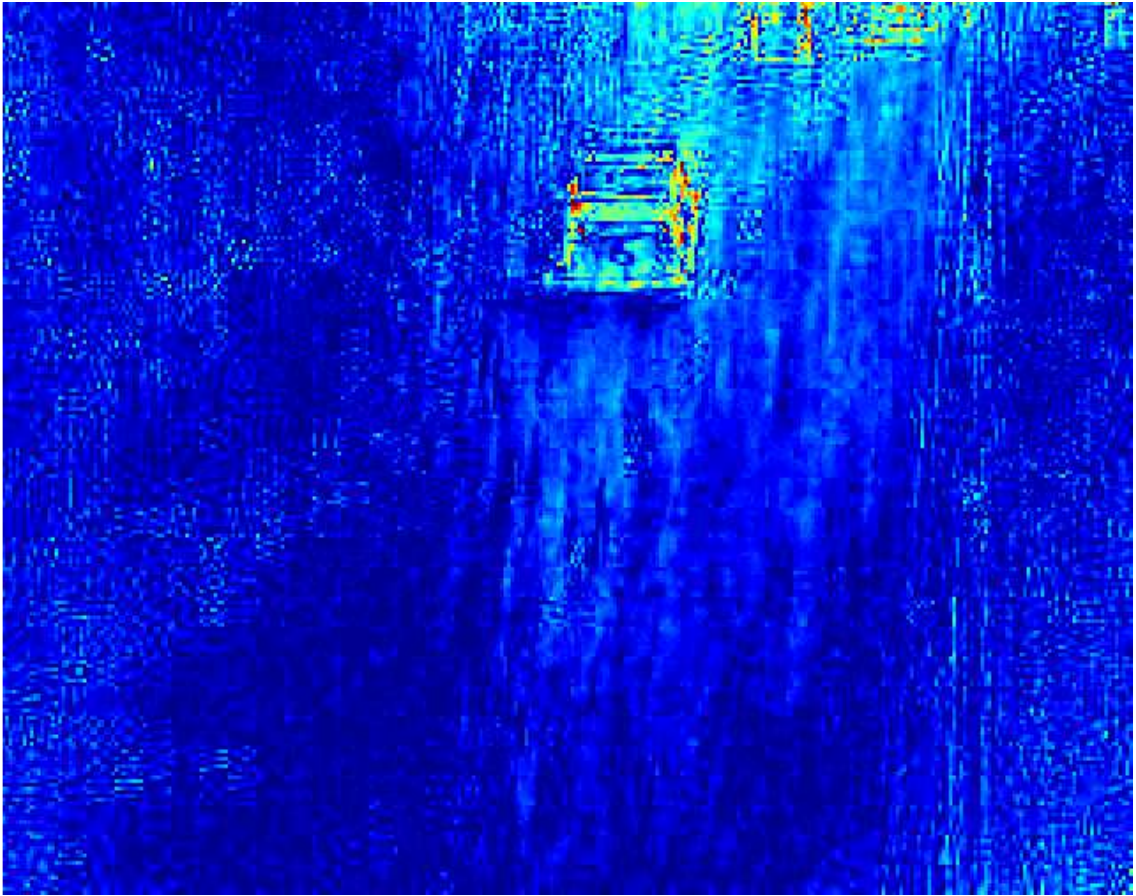} &
\includegraphics[width=0.07\linewidth]{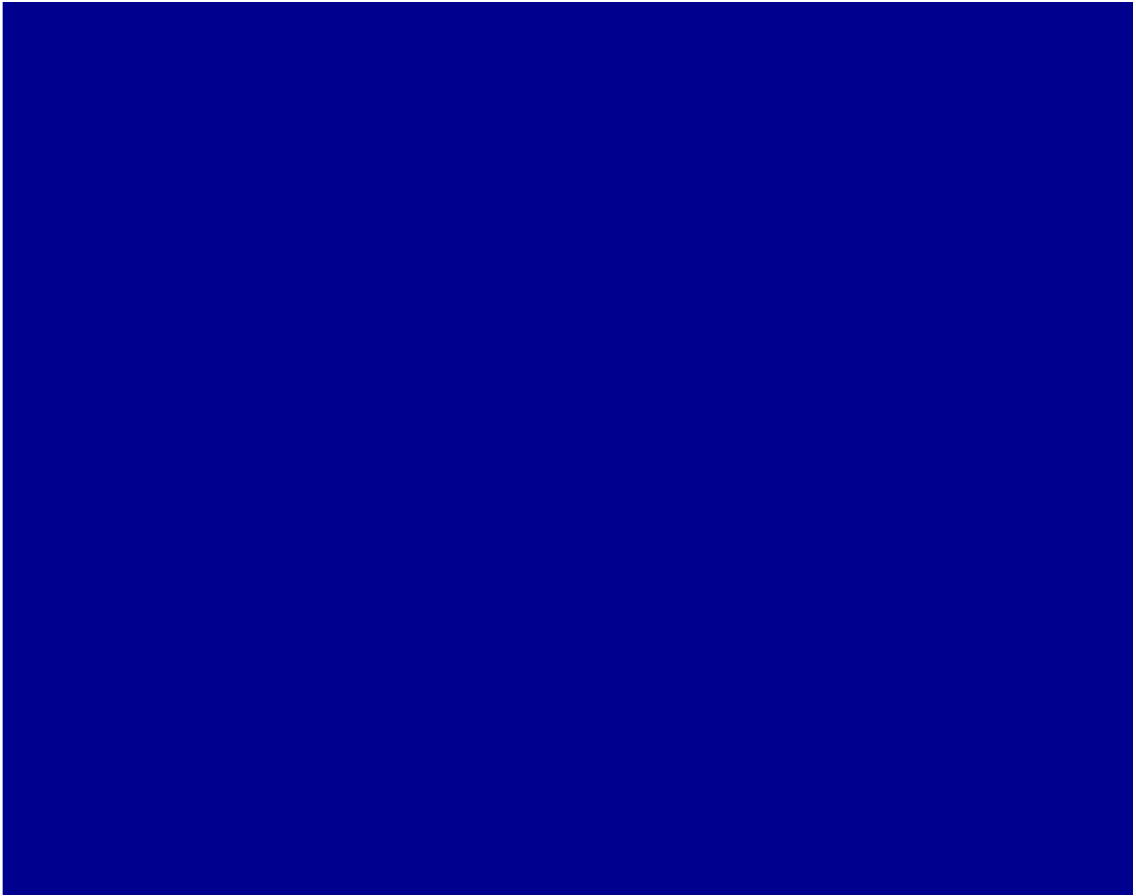} \\

\end{tabular}}
\end{tabular}
\caption{The deraining results by the proposed FastDeRain and the method in \cite{Jiang_2017_CVPR}.}
\label{CONF2}
\end{figure*}

\paragraph{Discussion of the noise term $\mathcal{N} $ in Eq. (\ref{obnoise})}
In this paper, the noise (or error) term ($\mathcal{N}$ in Eq. (\ref{obnoise})) is taken into consideration in the observation model.
To illustrate its effects, we conduct a series of experiments, in which the Gaussian noises of different standard deviations are respectively added to the video ``waterfall'' with synthetic rain streaks in case 1.
The quantitative assessments of the results obtained by the proposed method with and without the noise (or error) term $\mathcal{N}$ taken into consideration (denoted as ``with $\mathcal{N}$'' and ``without $\mathcal{N}$'', respectively ) are reported in Table \ref{noise_robust}.
In addition, we also exhibit the effects of different parameters on the proposed method without $\mathcal{N}$ in Fig. \ref{para}.

\begin{table}[h!] 
\caption{Quantitative comparisons of the rain streak removal results of the proposed FastDeRain with and without the noise term taken into consideration on synthetic video ``waterfall'' with the synthetic rain streaks in case 1. The \textbf{best} quantitative values are in \textbf{boldface}.}
\renewcommand\arraystretch{0.9}\setlength{\tabcolsep}{3.5pt}\scriptsize
\begin{center}
\begin{tabular}{cccccccccc}
\toprule
$\sigma$
&Method    & PSNR    &  SSIM  &  FSIM  &  VIF   &  UIQI  &  GMSD  \\\midrule
\multirow{3}{*}{0}
&Rainy     & 31.63 & 0.9097 & 0.9550 & 0.5956 & 0.8834 & 0.0617 \\
&with $\mathcal{N}$ & 40.52 & 0.9842 & 0.9900 & 0.8588 & 0.9787 & 0.0106 \\
&without $\mathcal{N}$& \bf40.92 & \bf0.9869 & \bf0.9914 & \bf0.8629 & \bf0.9824 & \bf0.0099 \\ \midrule
\multirow{3}{*}{0.01}
&Rainy     & 31.04 & 0.9003 & 0.9516 & 0.5553 & 0.8611 & 0.0622 \\
&with $\mathcal{N}$   & \bf38.22 & \bf0.9764 & \bf0.9869 & \bf0.8373 & \bf0.9685 & \bf0.0111 \\
&without $\mathcal{N}$& 37.95 & 0.9761 & 0.9868 & 0.8324 & 0.9684 & 0.0118 \\ \midrule
\multirow{3}{*}{0.02}
&Rainy     & 29.64 & 0.8735 & 0.9422 & 0.4786 & 0.8042 & 0.0637 \\
&with $\mathcal{N}$   & \bf35.80 & \bf0.9622 & \bf0.9802 & \bf0.7983 & \bf0.9502 & \bf0.0132 \\
&without $\mathcal{N}$& 34.86 & 0.9528 & 0.9764 & 0.7716 & 0.9386 & 0.0163 \\ \midrule
\multirow{3}{*}{0.03}
&Rainy     & 27.99 & 0.8337 & 0.9286 & 0.4059 & 0.7317 & 0.0664 \\
&with $\mathcal{N}$   & \bf34.15 & \bf0.9387 & \bf0.9725 & \bf0.7329 & \bf0.9220 & \bf0.0162 \\
&without $\mathcal{N}$& 33.22 & 0.9210 & 0.9666 & 0.7045 & 0.9017 & 0.0193 \\ \midrule
\multirow{3}{*}{0.04}
&Rainy     & 26.41 & 0.7855 & 0.9125 & 0.3444 & 0.6566 & 0.0704 \\
&with $\mathcal{N}$   & \bf32.52 & \bf0.9038 & \bf0.9613 & \bf0.6593 & \bf0.8824 & \bf0.0211 \\
&without $\mathcal{N}$& 31.63 & 0.8791 & 0.9540 & 0.6343 & 0.8558 & 0.0238 \\ \bottomrule
\end{tabular}
\end{center}
\label{noise_robust}
\end{table}

From Table \ref{noise_robust}, we can conclude our method without $\mathcal{N}$ would acquire a better result when the rainy video is free from the noise.
However, when the video is simultaneously affected by the rain streaks and the noise, which is unavoidable in real data, our method with $\mathcal{N}$ got better results.
Therefore, we adopt the term $\mathcal{N} $ in Eq. (3) which enhances the robustness of our method to the noise.
Meanwhile, the solid lines and the dashed lines in Fig. \ref{para} also demonstrate that taking the noise (or error) term $\mathcal{N}$ into account would contribute to the robustness of the proposed method to different parameters.

\begin{table}[h!] 
\caption{Quantitative comparisons of the rain streak removal results of the proposed FastDeRain and the method in the previous conference paper \cite{Jiang_2017_CVPR} on the synthetic data. The \textbf{best} quantitative values are in \textbf{boldface}.}
\renewcommand\arraystretch{0.9}\setlength{\tabcolsep}{1.5pt}\scriptsize
\begin{center}
\begin{tabular}{cccccccccc}
\toprule
Data & Method   & PSNR    &  SSIM  &  FSIM  &  VIF   &  UIQI  &  GMSD & Time (s) \\\midrule
\multirow{2}{*}{``waterfall''}
&Rainy                  & 31.63 & 0.9097 & 0.9550 & 0.5956 & 0.8834 & 0.0617 & | \\
\multirow{2}{*}{case 1}
&\cite{Jiang_2017_CVPR} & 37.86 & 0.9864 & 0.8397 & 0.9763 & 0.9787 & 0.0164 & 19.9\\
&FastDeRain             & \bf40.52 & \bf0.9842 & \bf0.9900 & \bf0.8588 & \bf0.9787 & \bf0.0106 & \bf9.3 \\ \midrule

\multirow{2}{*}{``bus''}
&Rainy                  & 26.15 & 0.8238 & 0.9300 & 0.4951 & 0.7808 & 0.1150 & | \\
\multirow{2}{*}{case 2}
&\cite{Jiang_2017_CVPR} & 30.07 & 0.9331 & 0.9574 & 0.6369 & 0.8986 & 0.0590 & 23.3\\
&FastDeRain             & \bf32.32 & \bf0.9375 & \bf0.9673 & \bf0.6552 & \bf0.8992 & \bf0.0496 & \bf6.7  \\ \midrule

\multirow{2}{*}{``highway''}
&Rainy                  & 22.90 & 0.9212 & 0.9702 & 0.6611 & 0.7650 & 0.0683 & | \\
\multirow{2}{*}{case 3}
&\cite{Jiang_2017_CVPR} & 24.02 & 0.9487 & 0.9823 & 0.7384 & 0.8312 & 0.0362 & 17.7 \\
&FastDeRain             & \bf30.17 & \bf0.9720 & \bf0.9838 & \bf0.8191 & \bf0.8951 & \bf0.0135 & \bf5.7 \\
 \bottomrule
\end{tabular}
\end{center}
\label{CONF1}
\end{table}

\paragraph{Comparisons with the method in the conference version}
To clarify the improvement of the proposed method from our conference version \cite{Jiang_2017_CVPR}, we compared the performances of our FastDeRain and the method in \cite{Jiang_2017_CVPR}.
To save space, results on the part of the synthetic data, which are listed in the first column of Table \ref{CONF1}, are reported.
The deraining results are exhibited in Fig. \ref{CONF2}, and, to avoid repetition, the numbers of the frames in Fig. \ref{CONF2} are different from those in foregoing figures.
From Table \ref{CONF1} and Fig. \ref{CONF2}, we can conclude our FastDeRain made substantial progress compared with the method in the conference version \cite{Jiang_2017_CVPR}.
These results also accord with the above discussion of the irrationality of the low-rank regularizer.

\subsection{Real data}

\begin{figure*}[htbp] 
\begin{center}\footnotesize\setlength{\tabcolsep}{1pt}
\begin{tabular}{cccccccc}
\renewcommand\arraystretch{0.9}
\multirow{2}{*}{Rainy frames} &TCL \cite{kim2015video}&DDN \cite{fu2017removing}&SE \cite{Wei_2017_ICCV}& MS-CSC \cite{li2018video} & FastDeRain\\
&  (5902.2s)&(75.8s)&(2840.9s) & (495.4s) & ({\bf11.3}s)\\

\includegraphics[width=0.15\linewidth]{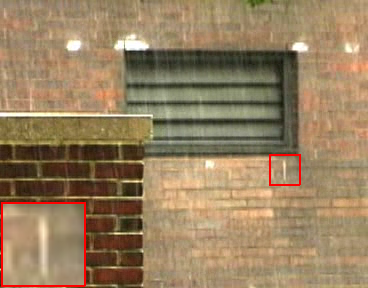} &
\includegraphics[width=0.15\linewidth]{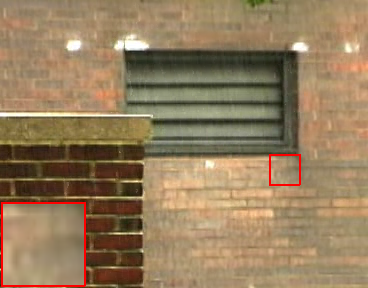} &
\includegraphics[width=0.15\linewidth]{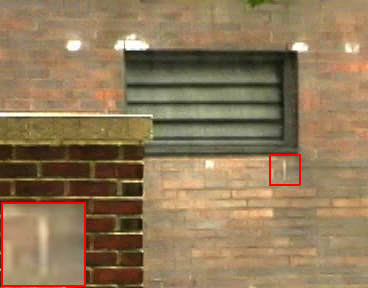} &
\includegraphics[width=0.15\linewidth]{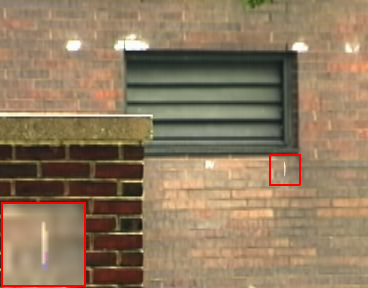} &
\includegraphics[width=0.15\linewidth]{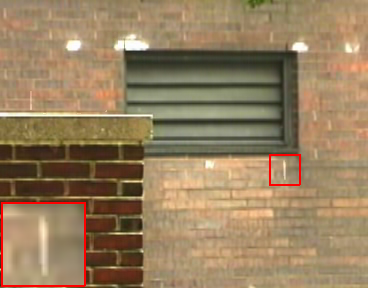} &
\includegraphics[width=0.15\linewidth]{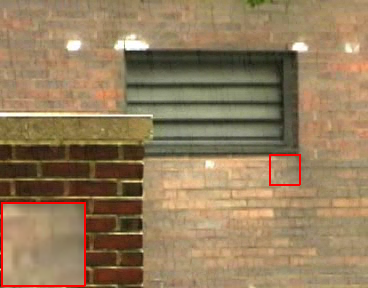} \\
&\includegraphics[width=0.15\linewidth]{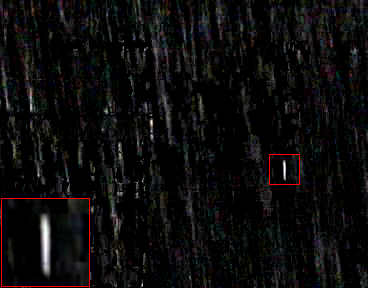} &
\includegraphics[width=0.15\linewidth]{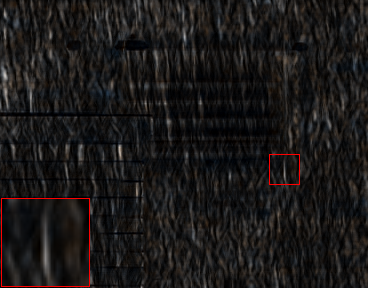} &
\includegraphics[width=0.15\linewidth]{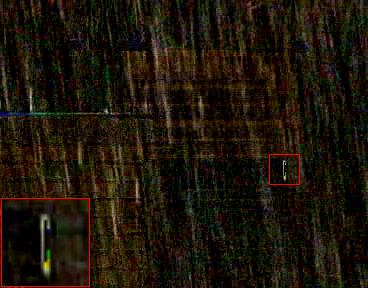} &
\includegraphics[width=0.15\linewidth]{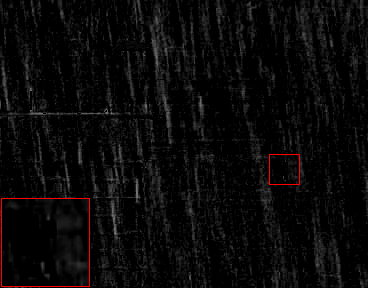} &
\includegraphics[width=0.15\linewidth]{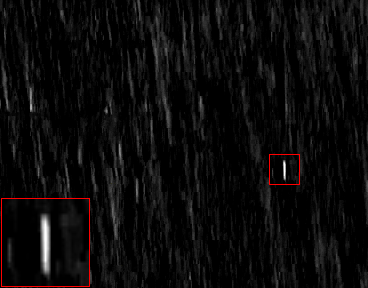} \\
\includegraphics[width=0.15\linewidth]{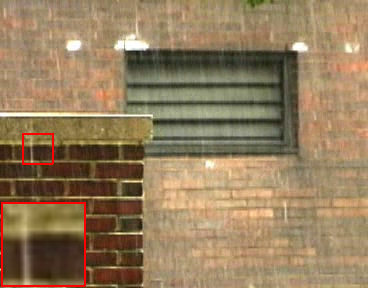} &
\includegraphics[width=0.15\linewidth]{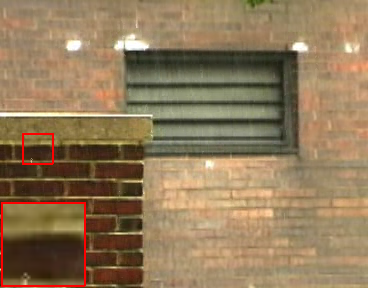} &
\includegraphics[width=0.15\linewidth]{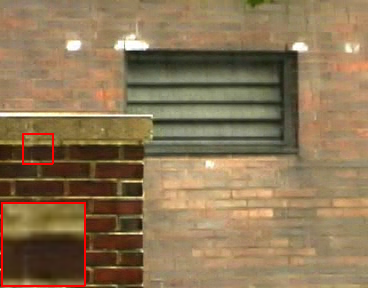} &
\includegraphics[width=0.15\linewidth]{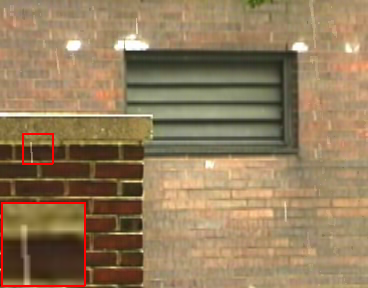} &
\includegraphics[width=0.15\linewidth]{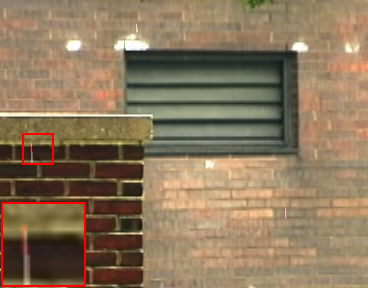} &
\includegraphics[width=0.15\linewidth]{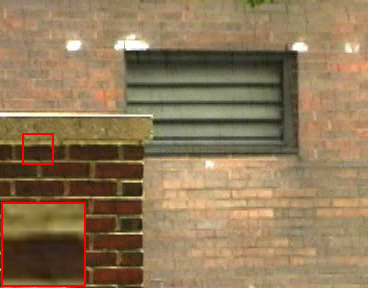} \\
&\includegraphics[width=0.15\linewidth]{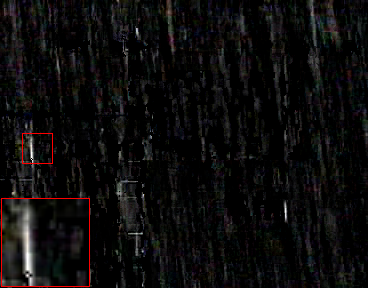} &
\includegraphics[width=0.15\linewidth]{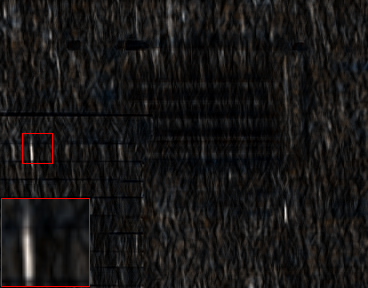} &
\includegraphics[width=0.15\linewidth]{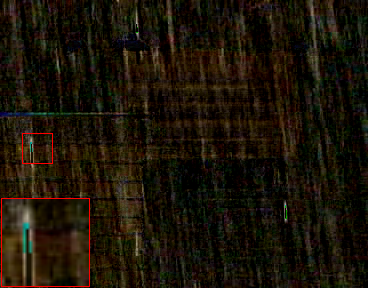} &
\includegraphics[width=0.15\linewidth]{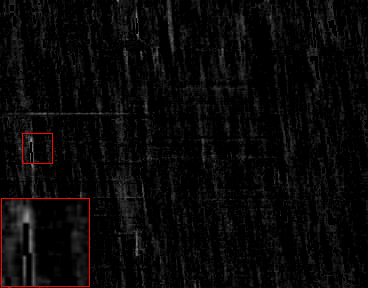} &
\includegraphics[width=0.15\linewidth]{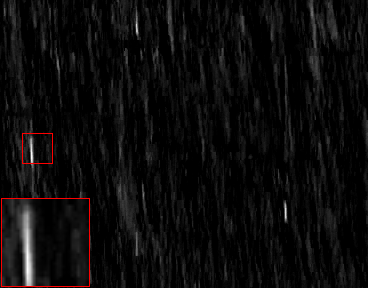} \\

\end{tabular}
\caption{Rain streak removal performance of different methods obtained on the video ``wall''.
From top to bottom, two adjacent frames of the deraining results and corresponding extracted rain streaks are illustrated. From left to right are: the rainy data (or the color bar), results by different methods, and the ground truth.}

\label{wall}
\end{center}
\end{figure*}

\begin{figure*}[htbp]
\begin{center}\setlength{\tabcolsep}{1pt}\footnotesize
\begin{tabular}{ccccccccccc}
\renewcommand\arraystretch{0.9}
Rainy frame&\multicolumn{2}{c}{TCL \cite{kim2015video}} & \multicolumn{2}{c}{DDN \cite{fu2017removing}} & \multicolumn{2}{c}{SE \cite{Wei_2017_ICCV}}& \multicolumn{2}{c}{MS-CSC \cite{li2018video}} & \multicolumn{2}{c}{FastDeRain}\\
&  \multicolumn{2}{c}{(2685.6s)}&\multicolumn{2}{c}{(63.8s)}&\multicolumn{2}{c}{(558.9s)} & \multicolumn{2}{c}{(448.3s)} & \multicolumn{2}{c}{({\bf13.8}s)}\\

\includegraphics[width=0.087\linewidth]{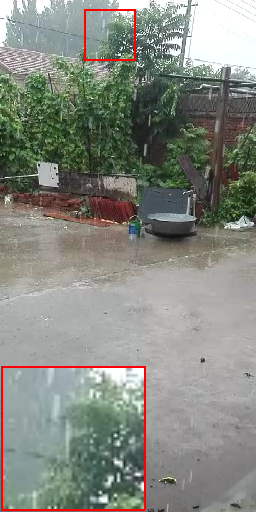} &
\includegraphics[width=0.087\linewidth]{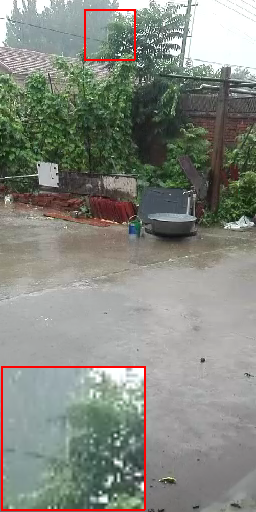} &
\includegraphics[width=0.087\linewidth]{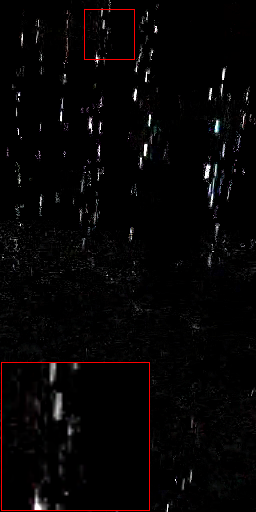} &
\includegraphics[width=0.087\linewidth]{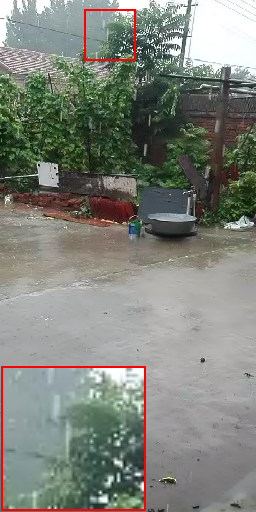} &
\includegraphics[width=0.087\linewidth]{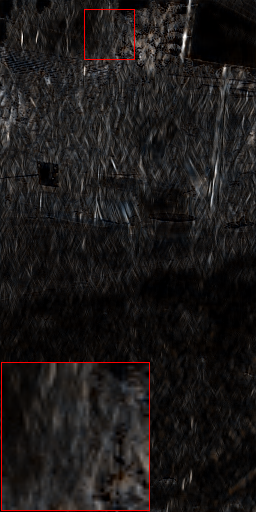} &
\includegraphics[width=0.087 \linewidth]{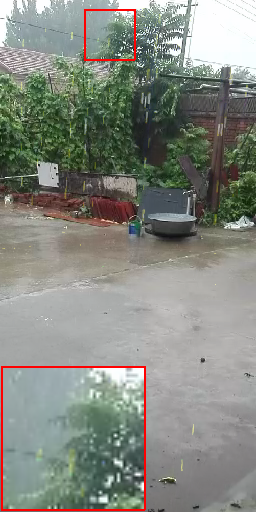} &
\includegraphics[width=0.087\linewidth]{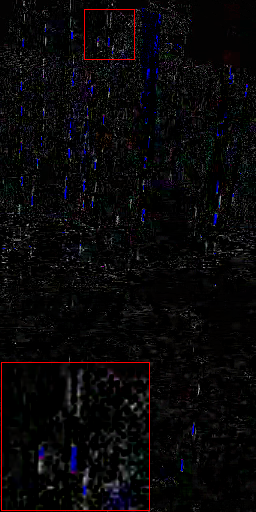} &
\includegraphics[width=0.087\linewidth]{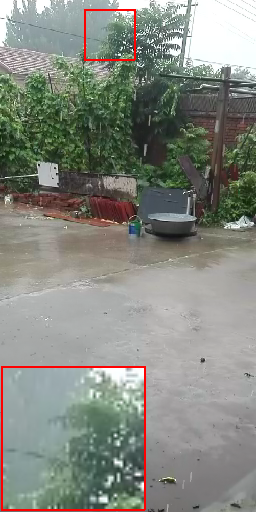} &
\includegraphics[width=0.087\linewidth]{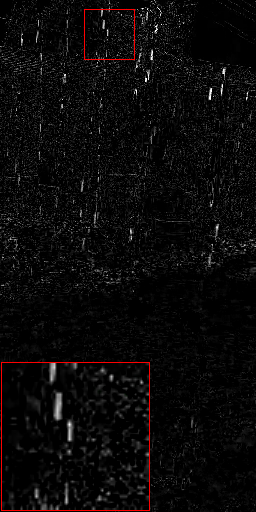} &
\includegraphics[width=0.087\linewidth]{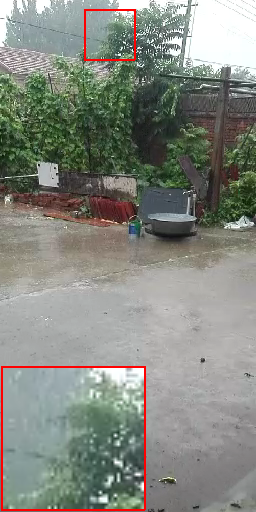}&
\includegraphics[width=0.087\linewidth]{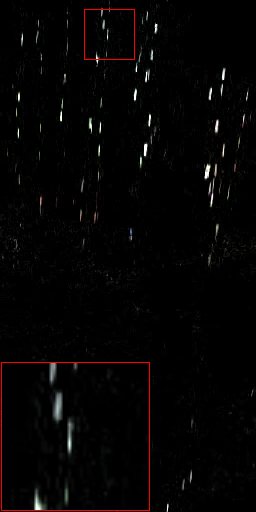} \\
\end{tabular}
\caption{Rain streak removal results on the video ``yard''. From left to right are frames of the rainy video, rain streaks removal results and corresponding extracted rain streaks by different methods, respectively. From left to right are: the rainy data, results by different methods, and the ground truth.}
\label{yard}
\end{center}
\end{figure*}

\begin{figure*}[!htb]
\begin{center}\footnotesize\setlength{\tabcolsep}{2pt}
\begin{tabular}{cccccc}
\renewcommand\arraystretch{0.9}
\multirow{2}{*}{Rainy frames} &TCL \cite{kim2015video}&DDN \cite{fu2017removing}&SE \cite{Wei_2017_ICCV}& MS-CSC \cite{li2018video}& FastDeRain\\
&  (8431.4s)&(80.6s)&(3852.5s) & (484.9s) & ({\bf14.8}s)\\
\includegraphics[width=0.15\linewidth]{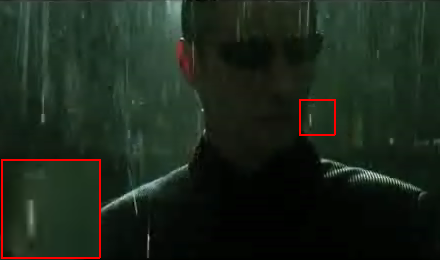} &
\includegraphics[width=0.15\linewidth]{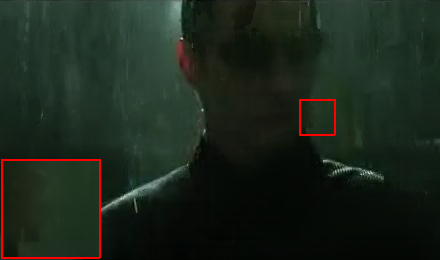} &
\includegraphics[width=0.15\linewidth]{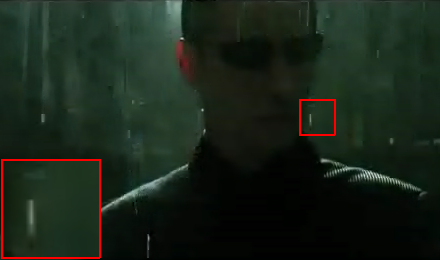} &
\includegraphics[width=0.15\linewidth]{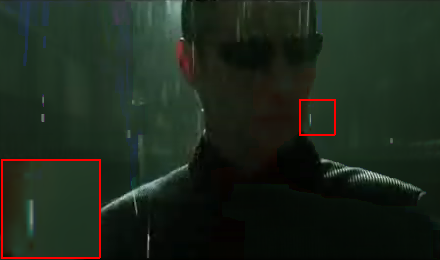} &
\includegraphics[width=0.15\linewidth]{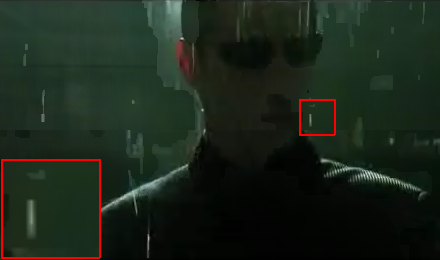} &
\includegraphics[width=0.15\linewidth]{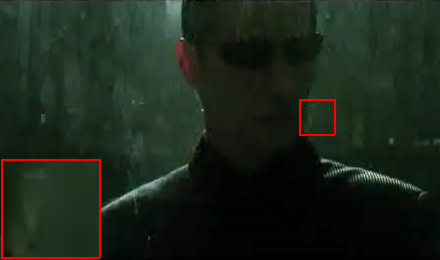} \\
&\includegraphics[width=0.15\linewidth]{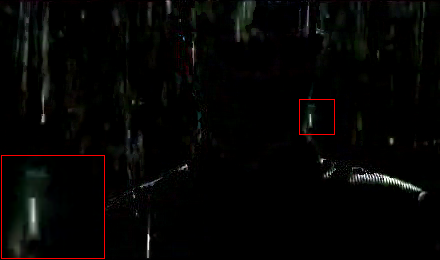} &
\includegraphics[width=0.15\linewidth]{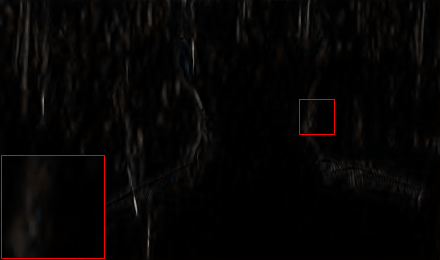} &
\includegraphics[width=0.15\linewidth]{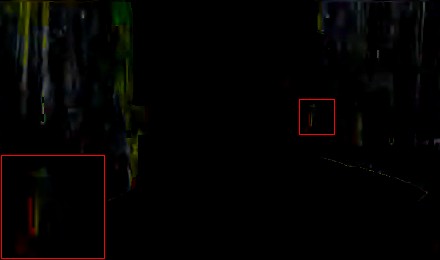} &
\includegraphics[width=0.15\linewidth]{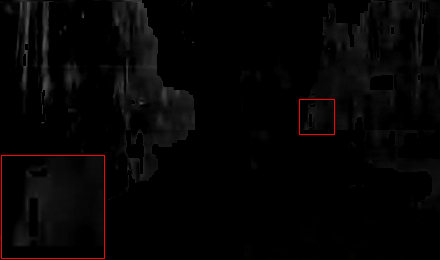} &
\includegraphics[width=0.15\linewidth]{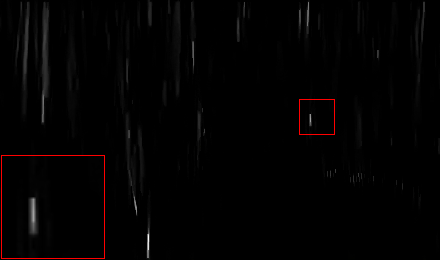} \\
\includegraphics[width=0.15\linewidth]{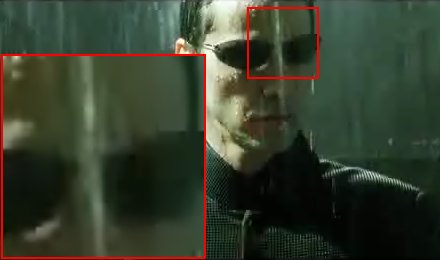} &
\includegraphics[width=0.15\linewidth]{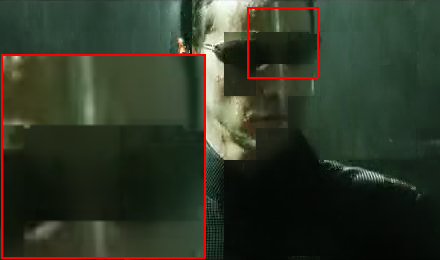} &
\includegraphics[width=0.15\linewidth]{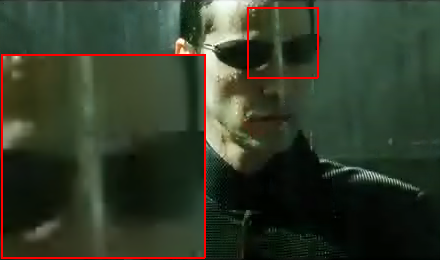} &
\includegraphics[width=0.15\linewidth]{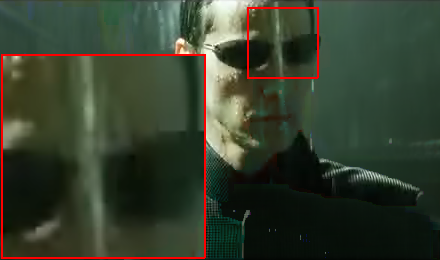} &
\includegraphics[width=0.15\linewidth]{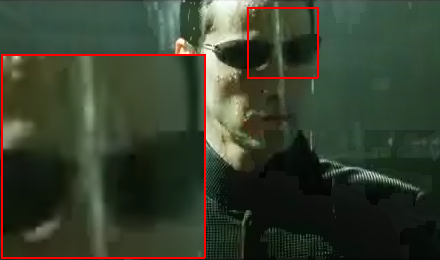} &
\includegraphics[width=0.15\linewidth]{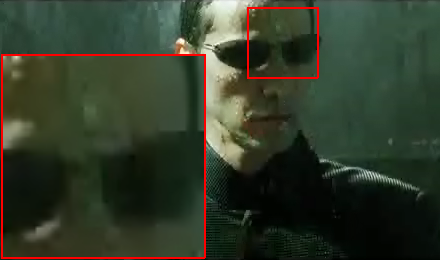} \\
&\includegraphics[width=0.15\linewidth]{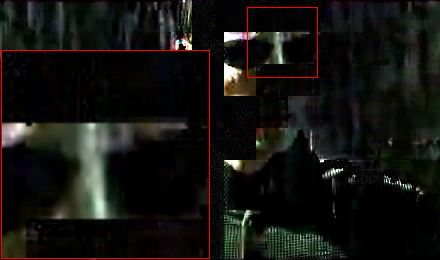} &
\includegraphics[width=0.15\linewidth]{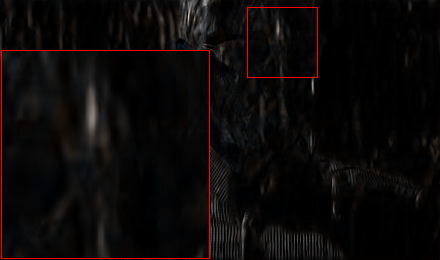} &
\includegraphics[width=0.15\linewidth]{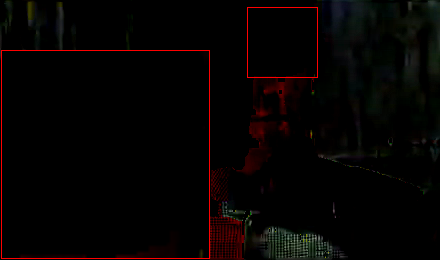} &
\includegraphics[width=0.15\linewidth]{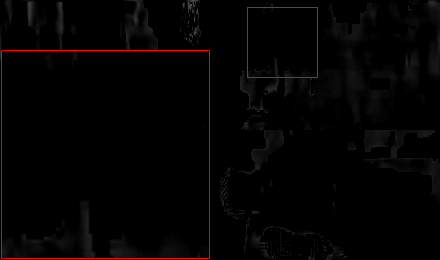} &
\includegraphics[width=0.15\linewidth]{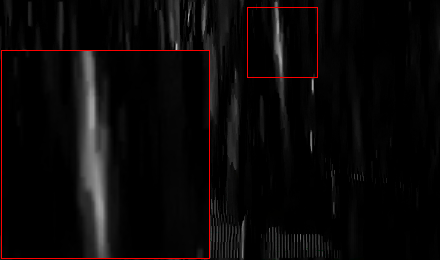} \\

\end{tabular}
\caption{Rain streak removal performance of different methods obtained on the clips of movie ``the Matrix''. From top to bottom, 2 adjacent frames of the rainy video/deraining results and corresponding extracted rain streaks are illustrated. From left to right are: the rainy data, results by different methods, and the ground truth.}
\label{matrix}
\end{center}
\end{figure*}

Four real-world rainy videos are chosen in this subsection.
The first one (denoted as ``wall'') of size $288 \times 368 \times 3 \times 171$ is download from the CAVE dataset\footnote{\url{http://www.cs.columbia.edu/CAVE/projects/camera rain/}} and the second video\footnote{\url{https://github.com/TaiXiangJiang/FastDeRain/blob/master/yard.mp4}}(denoted as ``yard'') of size $512  \times  256  \times 3 \times 126$  was recorded by one of the authors on a rainy day in his backyard.
The background of the video ``wall'' is consist of regular patterns while the background of the video ``yard'' is more complex.
The third video is clipped from the well-known film ``the Matrix''. The scene in this clips changes fast so that it is more difficult to deal with this video.
The last video of size $480 \times  640 \times    3 \times  108$ is denoted as ``crossing''\footnote{\url{https://github.com/hotndy/SPAC-SupplementaryMaterials/blob/master/Dataset_Testing_RealRain/ra4_Rain.rar}}, and it was captured in the crossing with complex traffic conditions.
\begin{figure*}[!htb]
\begin{center}\footnotesize\setlength{\tabcolsep}{1pt}
\begin{tabular}{cccccc}
\renewcommand\arraystretch{0.9}
\multirow{2}{*}{Rainy frame} &TCL \cite{kim2015video}&DDN \cite{fu2017removing}&SE \cite{Wei_2017_ICCV}& MS-CSC \cite{li2018video}& FastDeRain\\
&  (7246.0s)&(54.33s)&(2821.0s) & (484.9s) & ({\bf26.7}s)\\
\includegraphics[width=0.16\linewidth]{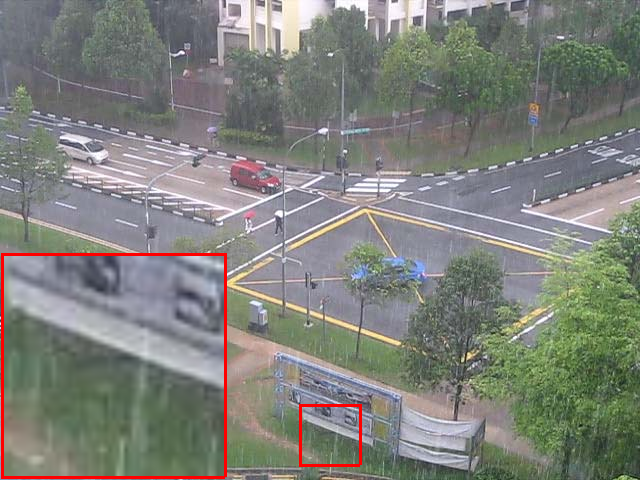} &
\includegraphics[width=0.16\linewidth]{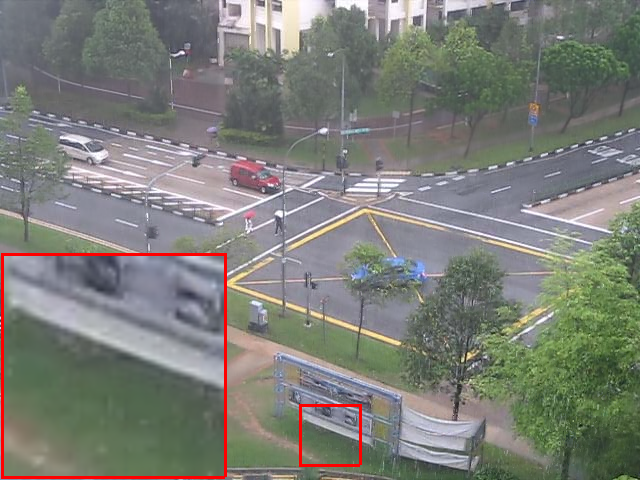} &
\includegraphics[width=0.16\linewidth]{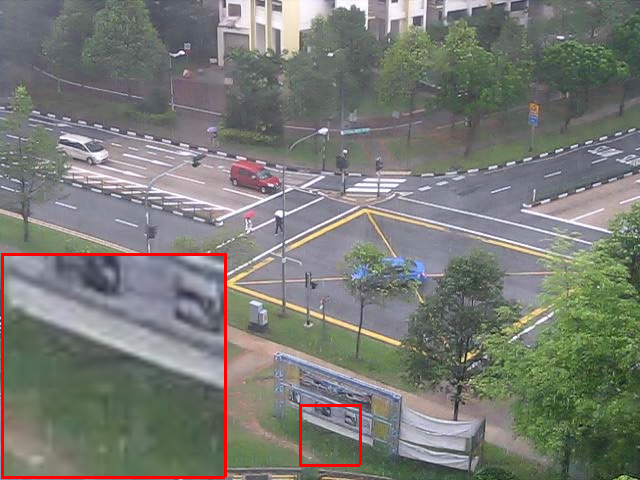} &
\includegraphics[width=0.16\linewidth]{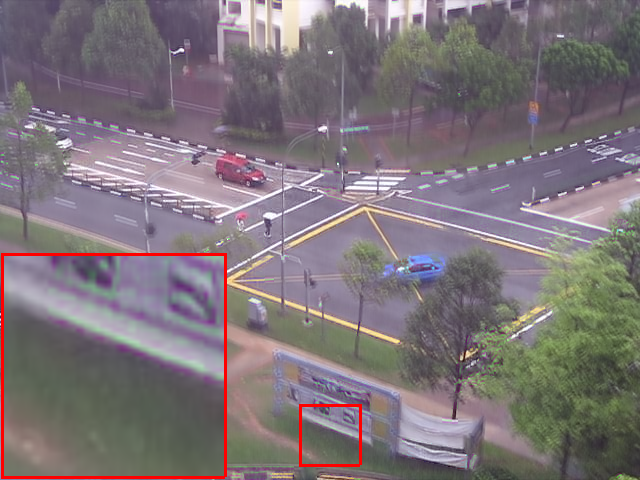} &
\includegraphics[width=0.16\linewidth]{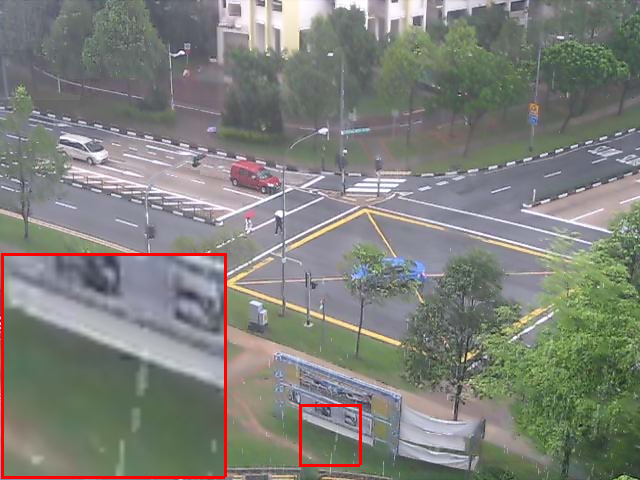} &
\includegraphics[width=0.16\linewidth]{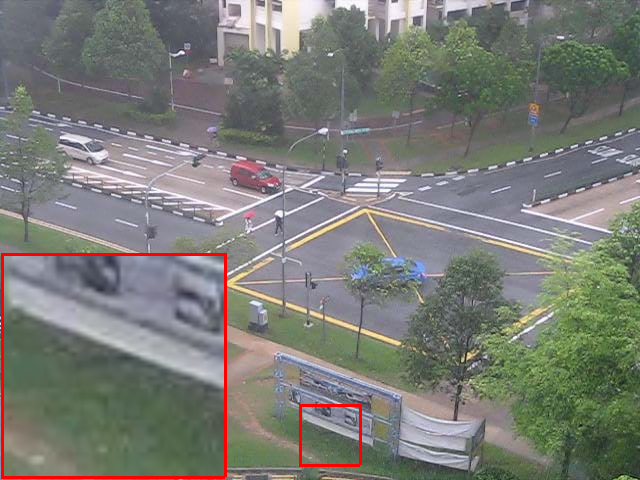} \\
&\includegraphics[width=0.16\linewidth]{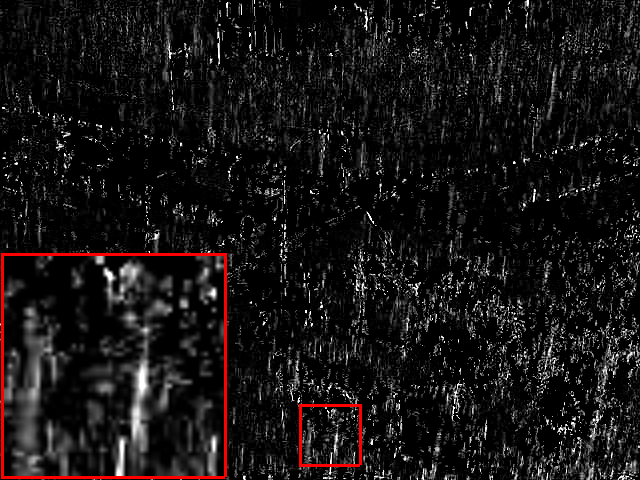} &
\includegraphics[width=0.16\linewidth]{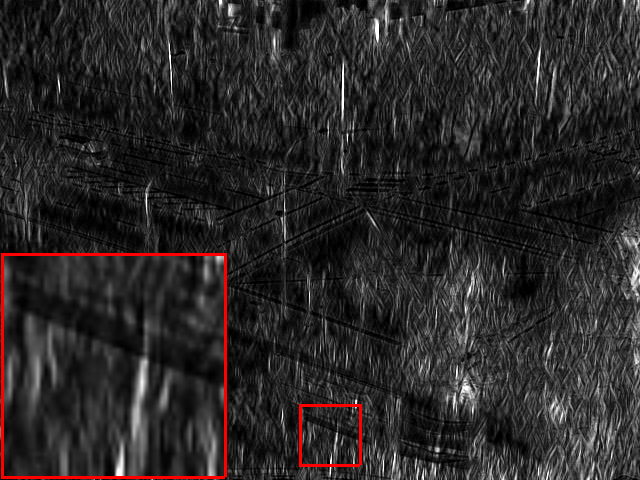} &
\includegraphics[width=0.16\linewidth]{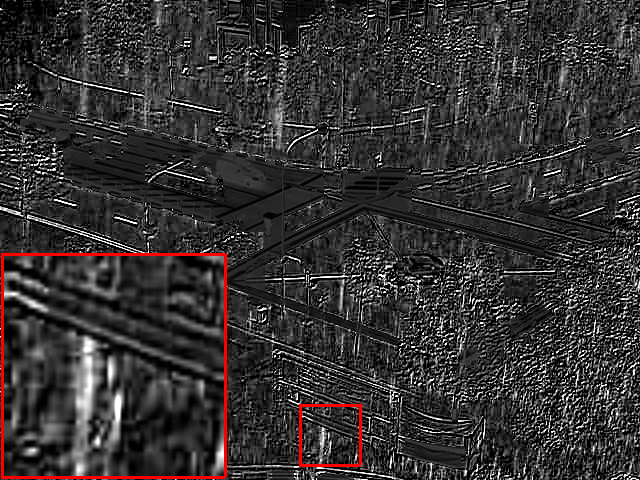} &
\includegraphics[width=0.16\linewidth]{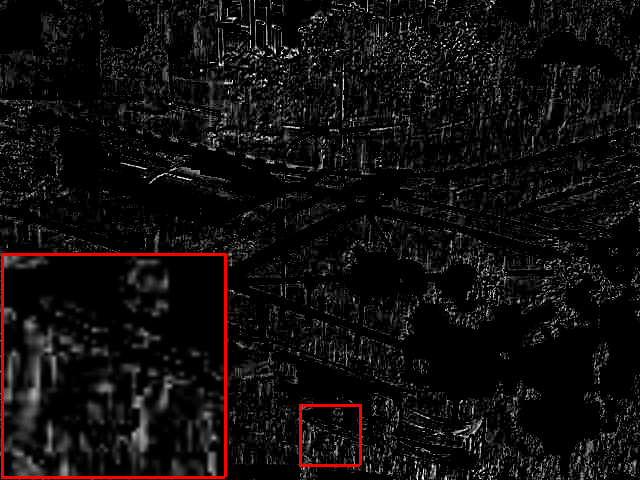} &
\includegraphics[width=0.16\linewidth]{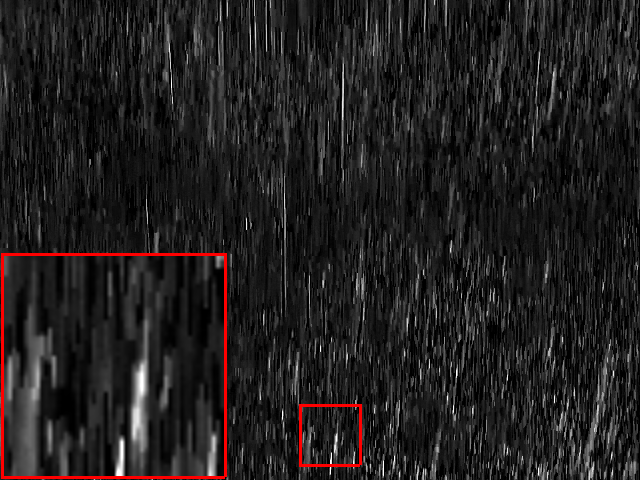} \\
\end{tabular}
\caption{Rain streak removal performance of different methods obtained on the video ``crossing''. From left to right are: the rainy data, deraining results or extracted rain streaks by different methods, and the ground truth.}
\label{crossing}
\end{center}
\end{figure*}

Fig. \ref{wall} shows two adjacent frames of the results obtained on the video ``wall''.
There are many vertical line patterns in the background of this video. Thus, exhibiting two adjacent frames would further help to distinguish the rain streaks from the background.
It can be found in the zoomed in red blocks that this rain streak with high brightness is not handled properly by DNN, SE and MS-CSC.
Our method removes almost all the rain streaks and preserves the background best compared with the results by other three methods.

Since there is little texture or structure similar to rain streaks in the video ``yard'', only one frame is exhibited in Fig. \ref{yard}.
DNN and SE didn't distinguish most of the rain streaks, especially in the zoomed in red blocks.
Although TCL and MS-CSC separated the majority of rain streaks, some fine structures of the background were improperly extracted.
Our FastDeRain removed most of the rain streaks and well preserved the background.

In Fig. \ref{matrix}, two adjacent frames of the rainy video ``the Matrix'' and deraining results by different methods are shown.
The two adjacent rainy frames reveal the rapidly changing of the scene, particularly the luminance.
Once again, our FastDeRain obtained the best result, especially when dealing with the obvious rain streak on the face of Neo.

The results on the rainy video ``crossing'' are exhibited in Fig. \ref{crossing}.
From the zoomed in areas, we can observe that all the methods except MS-CSC entirely removed the rain streaks.
TCL extracted some the structure of the curb line into the rain streaks while DNN tended to remove all the textures with line pattern.
SE erased many structural details.
The extracted rain streaks by the proposed FastDeRain were visually the best among all the results.

The scenarios in these four videos are of large differences.
Our method obtains the best results, both in removing rain streaks and in retaining spatial details.
In addition, the running time of our method is also obviously less than other methods, especially those three video-based methods.


\subsection{Oblique rain streaks}\label{oblique}

\begin{figure*}[!htb]
\begin{center}\setlength{\tabcolsep}{1pt} \footnotesize
\begin{tabular}{ccccccccc}
\renewcommand\arraystretch{0.9}
Rainy&TCL  & DDN &SE &MS-CSC & FastDeRain&GT\\

\includegraphics[width=0.1\linewidth]{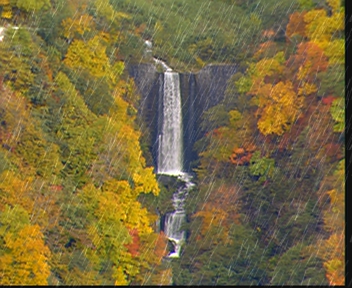} &
\includegraphics[width=0.1\linewidth]{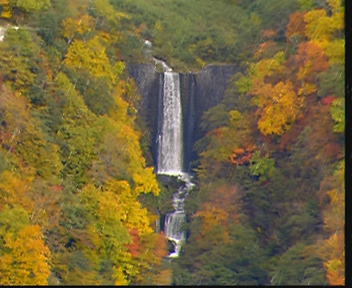} &
\includegraphics[width=0.1\linewidth]{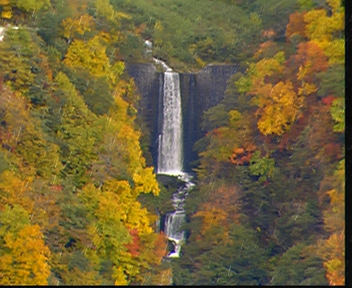} &
\includegraphics[width=0.1\linewidth]{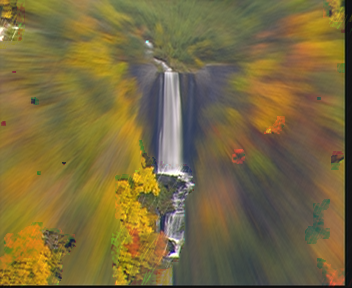} &
\includegraphics[width=0.1\linewidth]{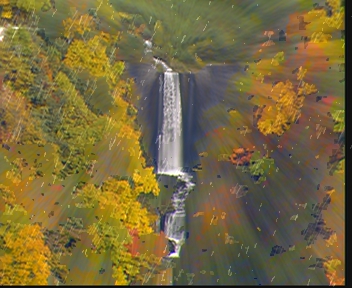} &
\includegraphics[width=0.1\linewidth]{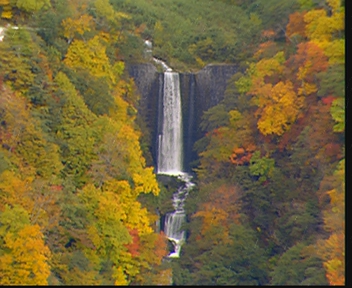} &
\includegraphics[width=0.1\linewidth]{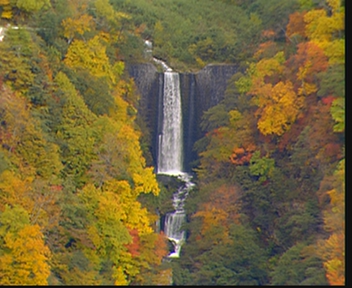} \\
&\includegraphics[width=0.1\linewidth]{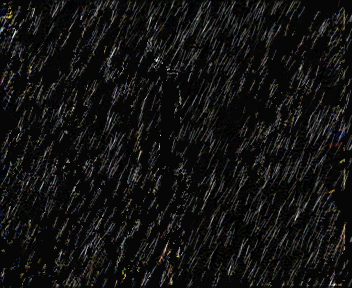} &
\includegraphics[width=0.1\linewidth]{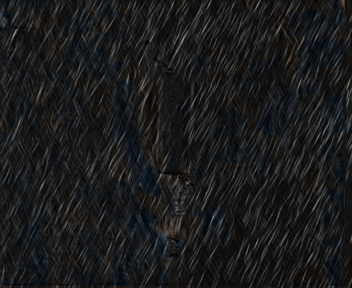} &
\includegraphics[width=0.1\linewidth]{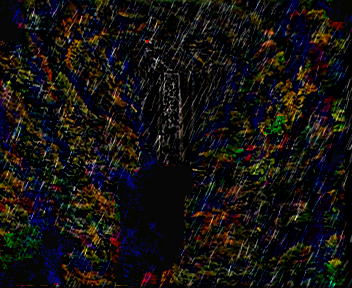} &
\includegraphics[width=0.1\linewidth]{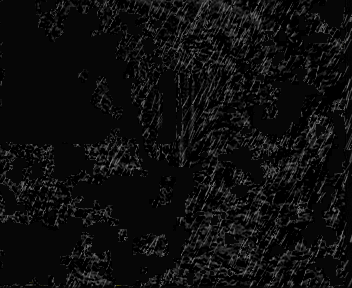} &
\includegraphics[width=0.1\linewidth]{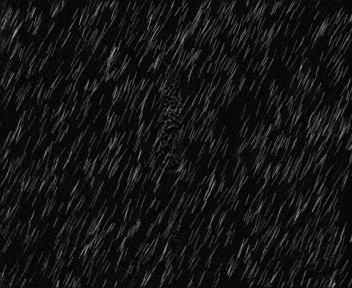} &
\includegraphics[width=0.1\linewidth]{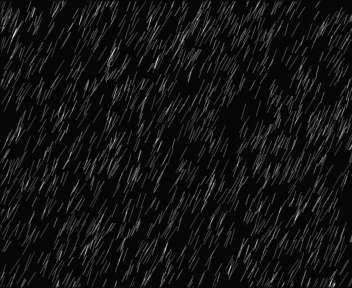} \\
\includegraphics[width=0.1\linewidth]{figs/component/bar_yuv.png}&
\includegraphics[width=0.1\linewidth]{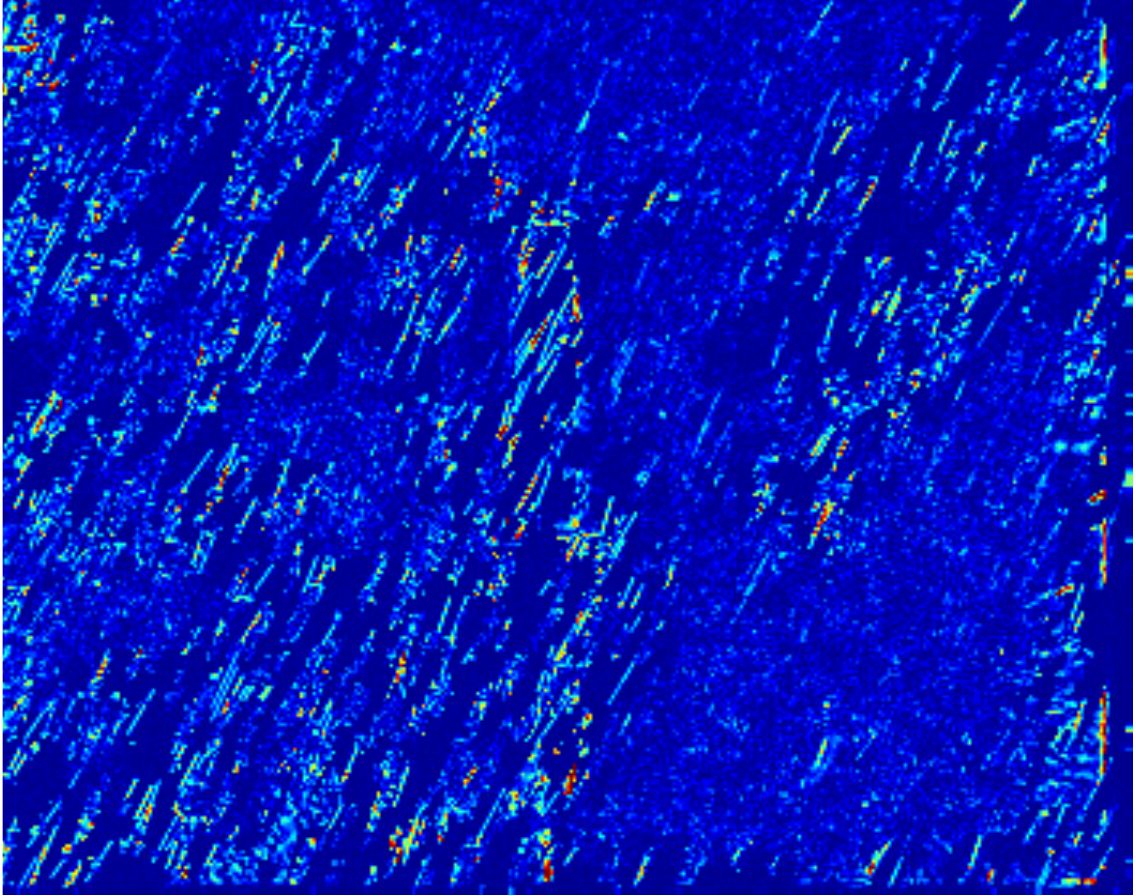}&
\includegraphics[width=0.1\linewidth]{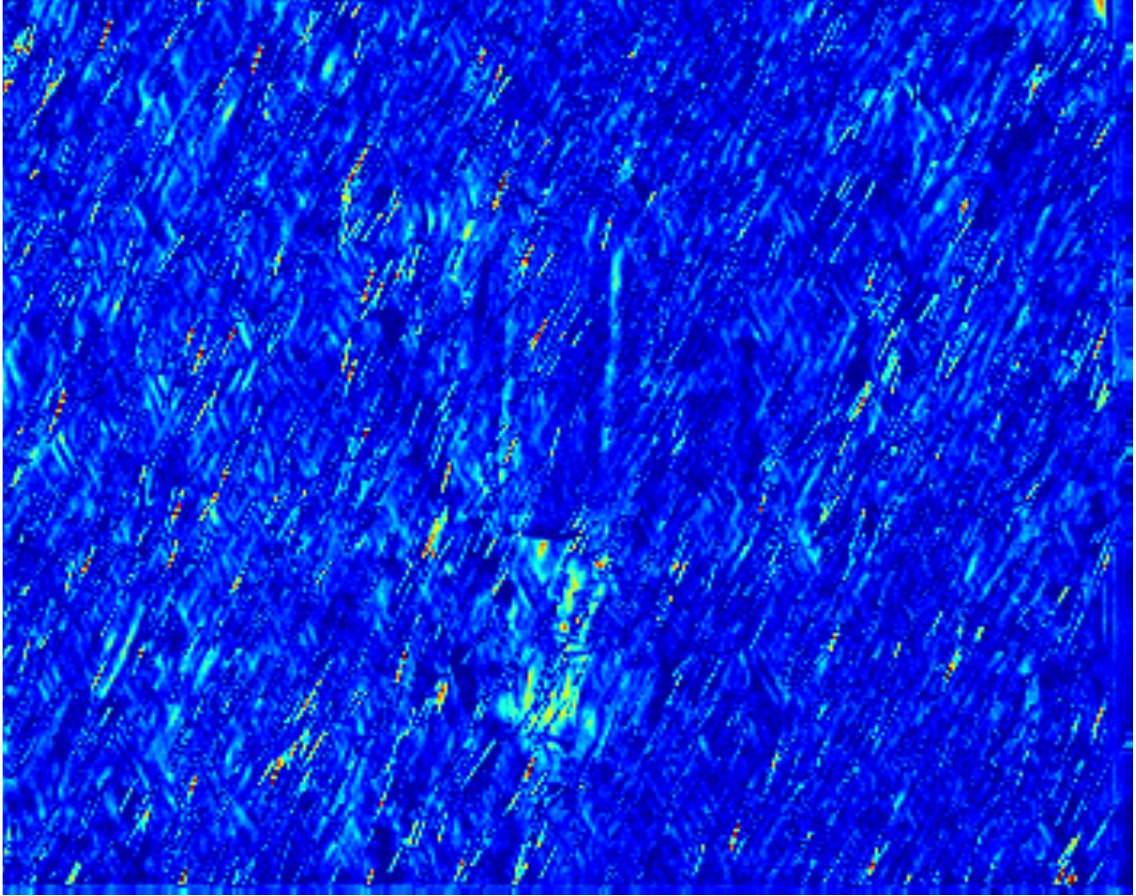}&
\includegraphics[width=0.1\linewidth]{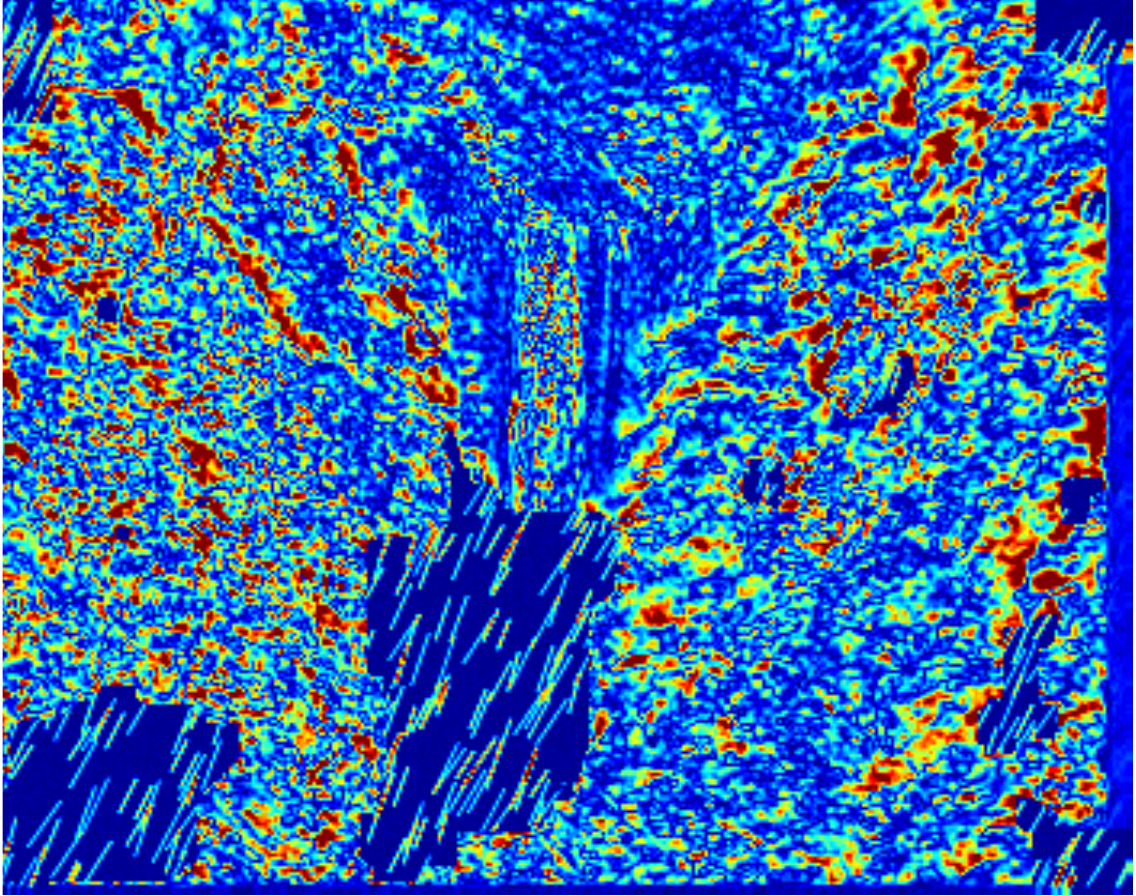}&
\includegraphics[width=0.1\linewidth]{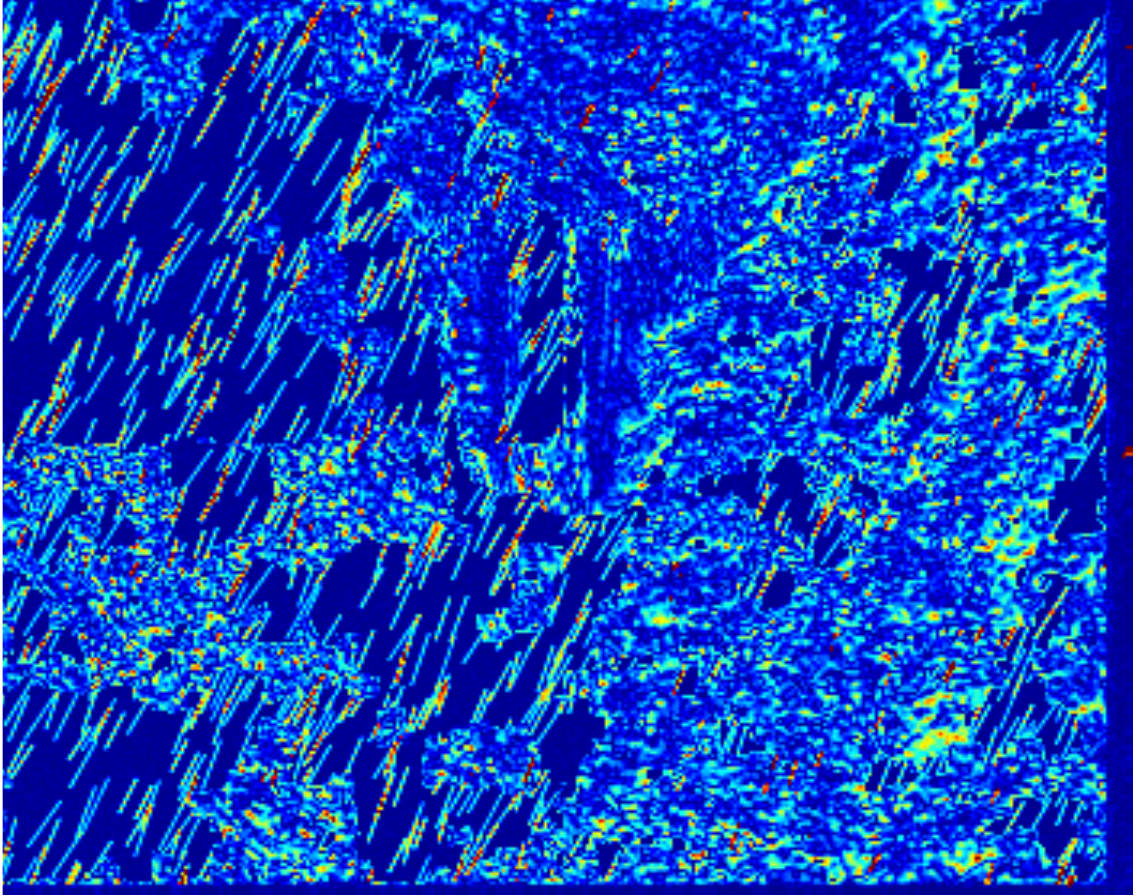}&
\includegraphics[width=0.1\linewidth]{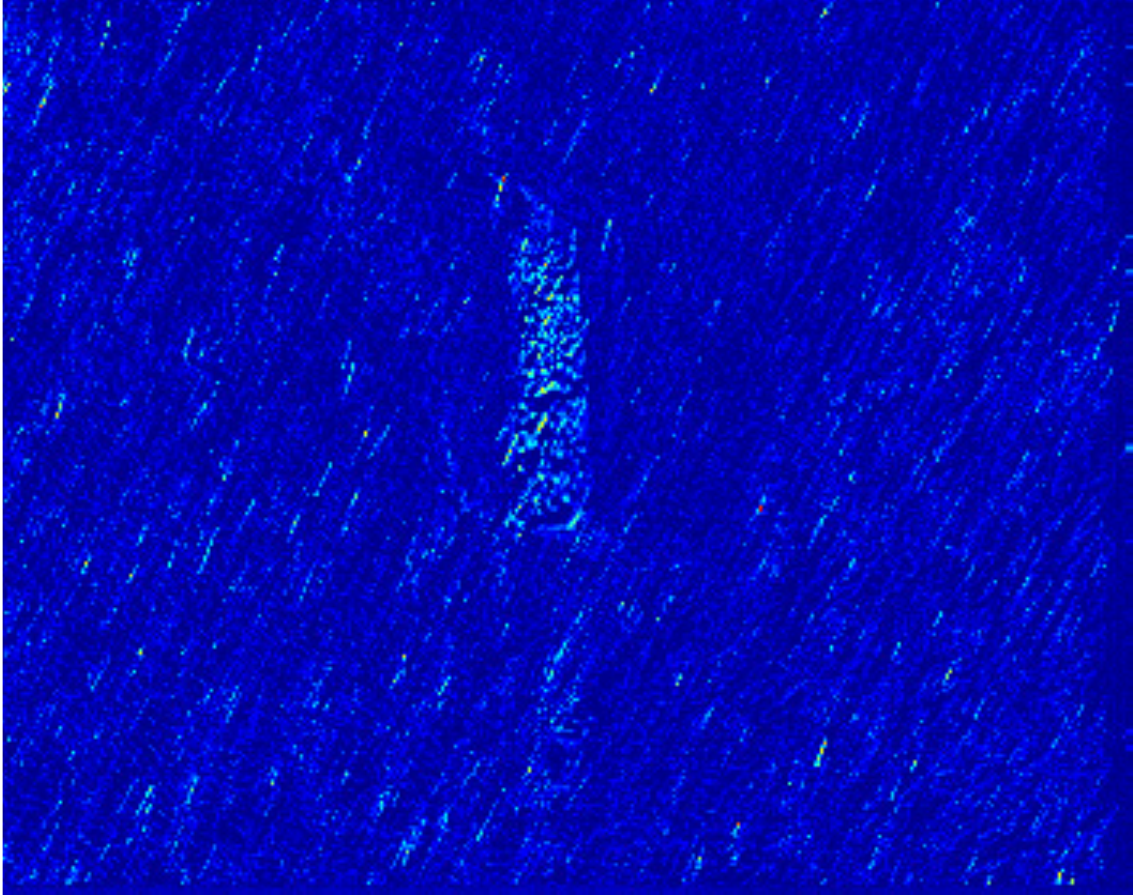} &
\includegraphics[width=0.1\linewidth]{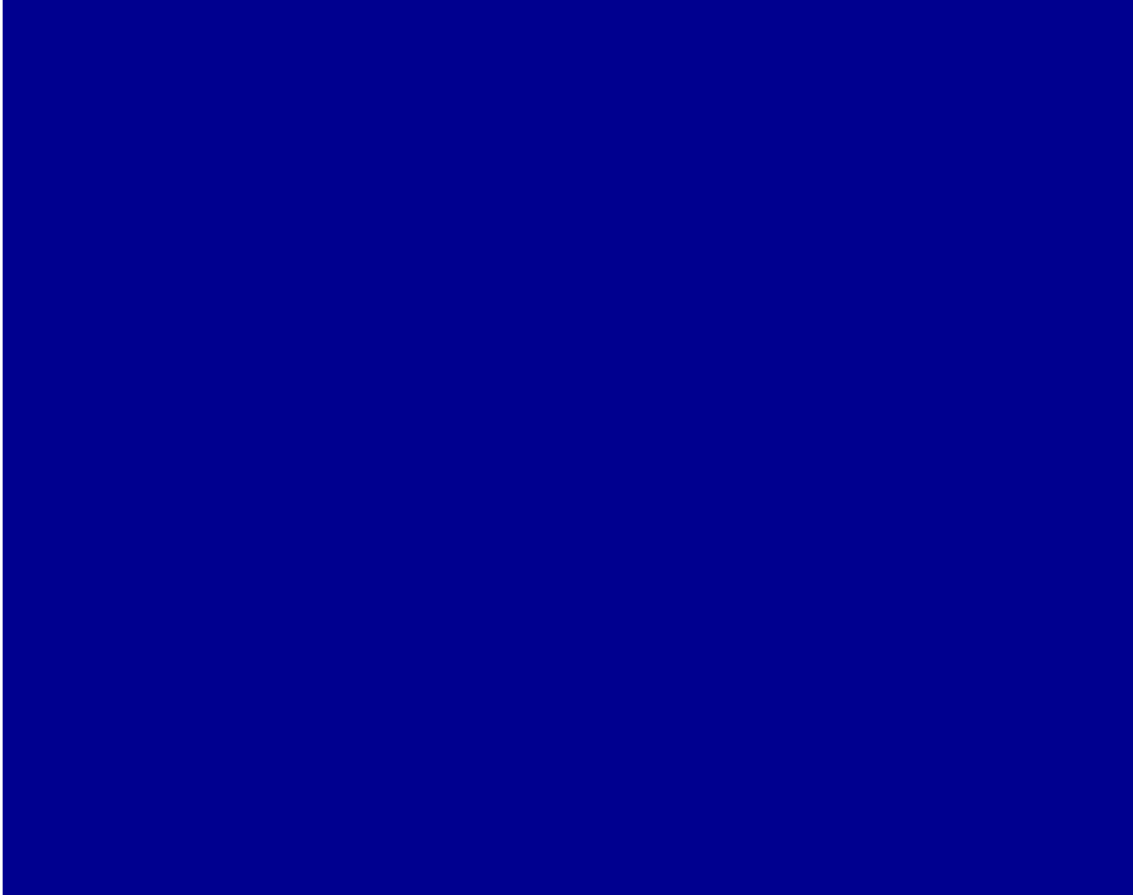} \\

Rainy&TCL  & DDN &SE &MS-CSC & FastDeRain&GT\\

\includegraphics[width=0.1\linewidth]{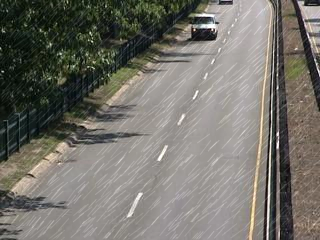} &
\includegraphics[width=0.1\linewidth]{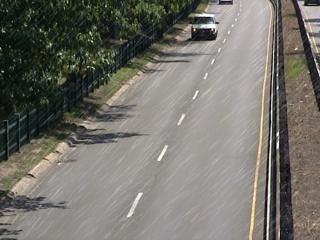} &
\includegraphics[width=0.1\linewidth]{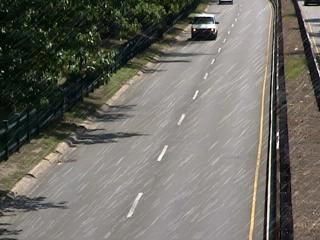} &
\includegraphics[width=0.1\linewidth]{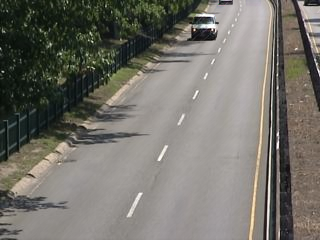} &
\includegraphics[width=0.1\linewidth]{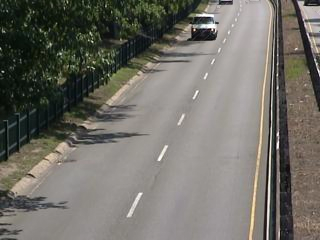} &
\includegraphics[width=0.1\linewidth]{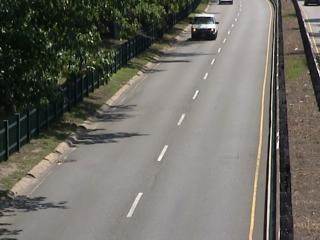} &
\includegraphics[width=0.1\linewidth]{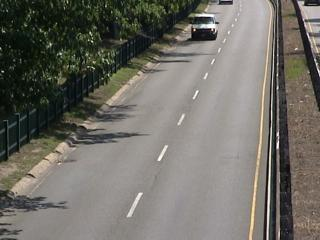} \\
&\includegraphics[width=0.1\linewidth]{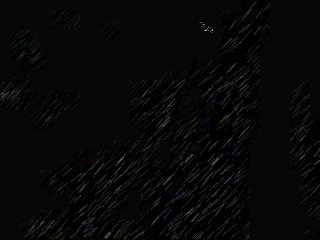} &
\includegraphics[width=0.1\linewidth]{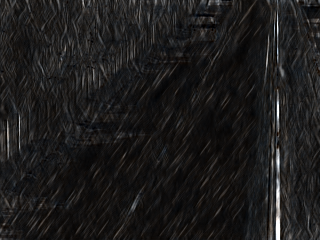} &
\includegraphics[width=0.1\linewidth]{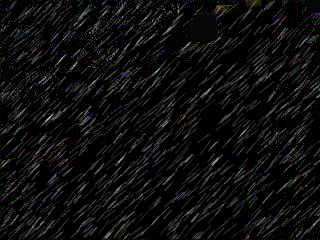} &
\includegraphics[width=0.1\linewidth]{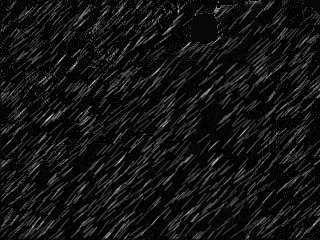} &
\includegraphics[width=0.1\linewidth]{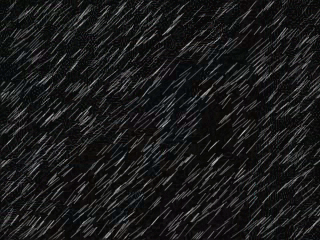} &
\includegraphics[width=0.1\linewidth]{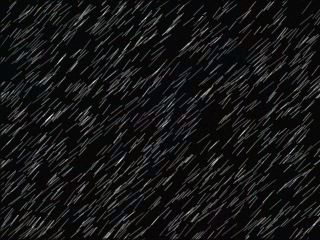} \\
\includegraphics[width=0.1\linewidth]{figs/component/bar_yuv.png}&
\includegraphics[width=0.1\linewidth]{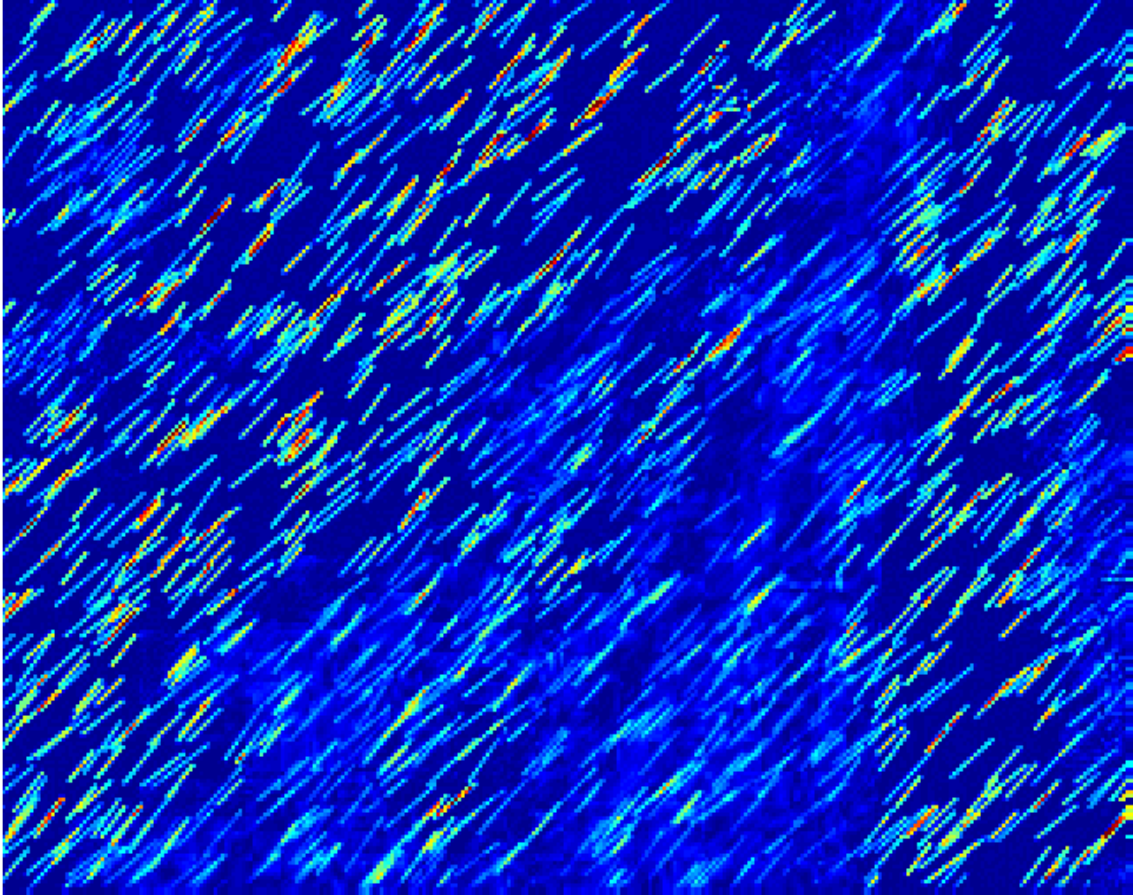}&
\includegraphics[width=0.1\linewidth]{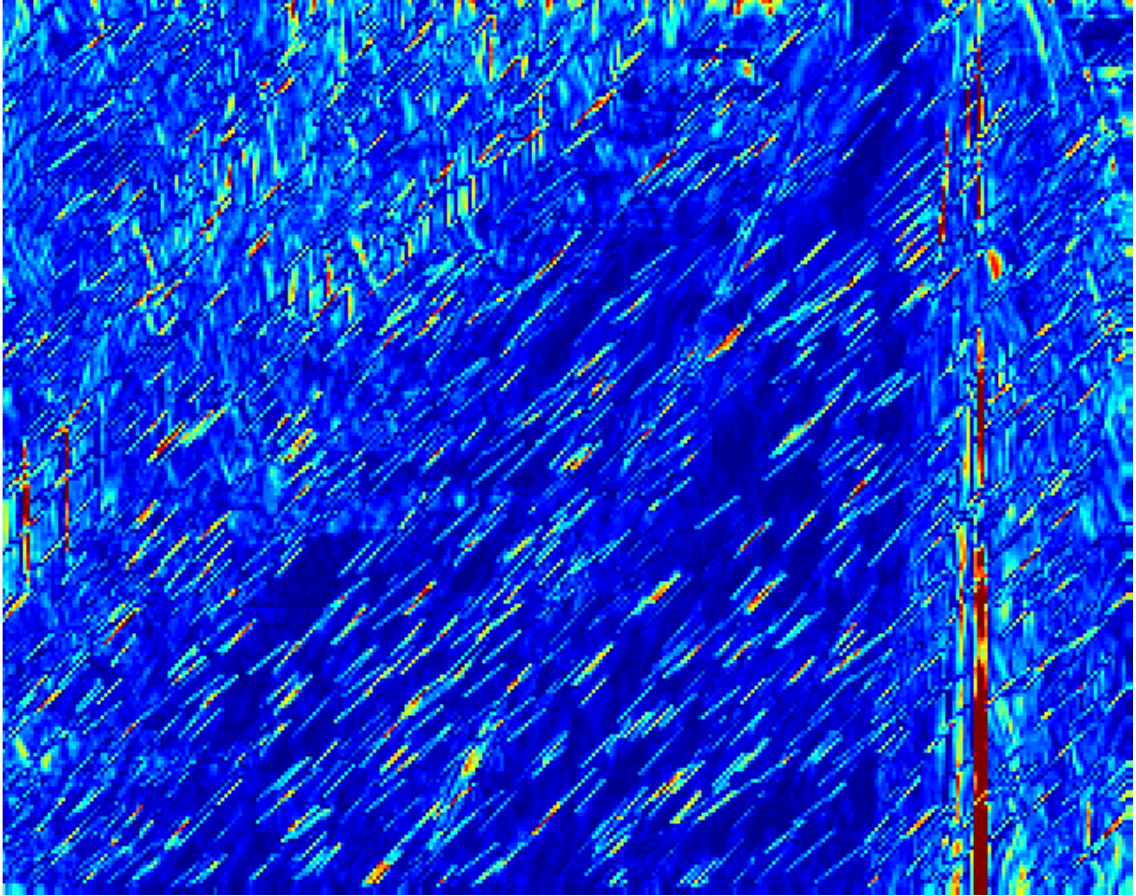}&
\includegraphics[width=0.1\linewidth]{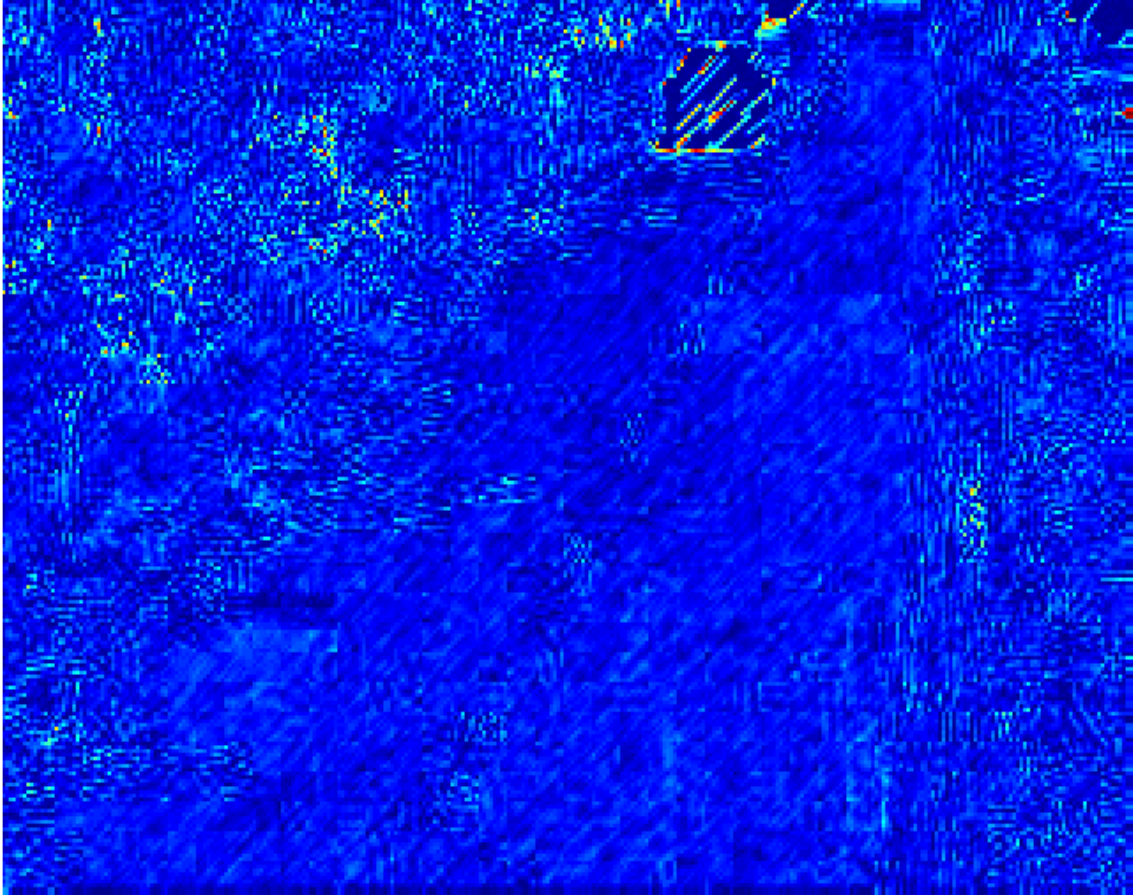}&
\includegraphics[width=0.1\linewidth]{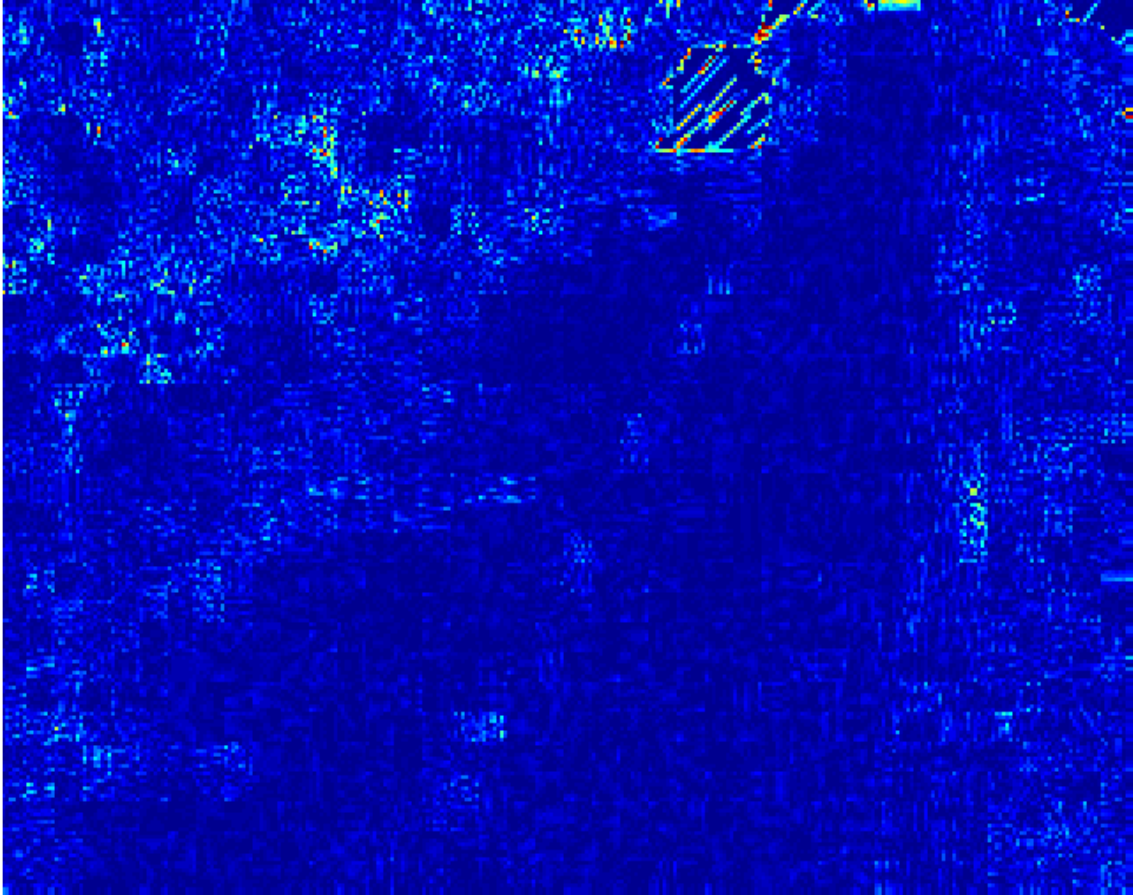}&
\includegraphics[width=0.1\linewidth]{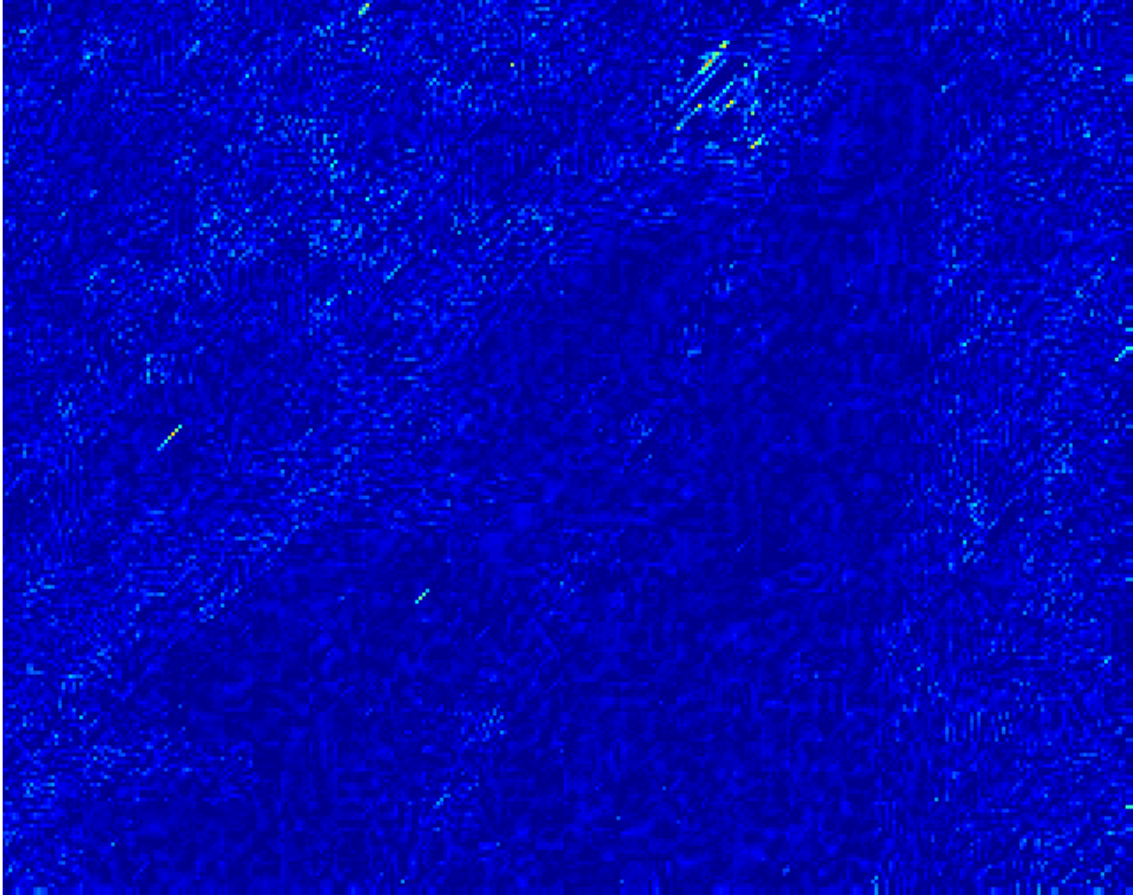} &
\includegraphics[width=0.1\linewidth]{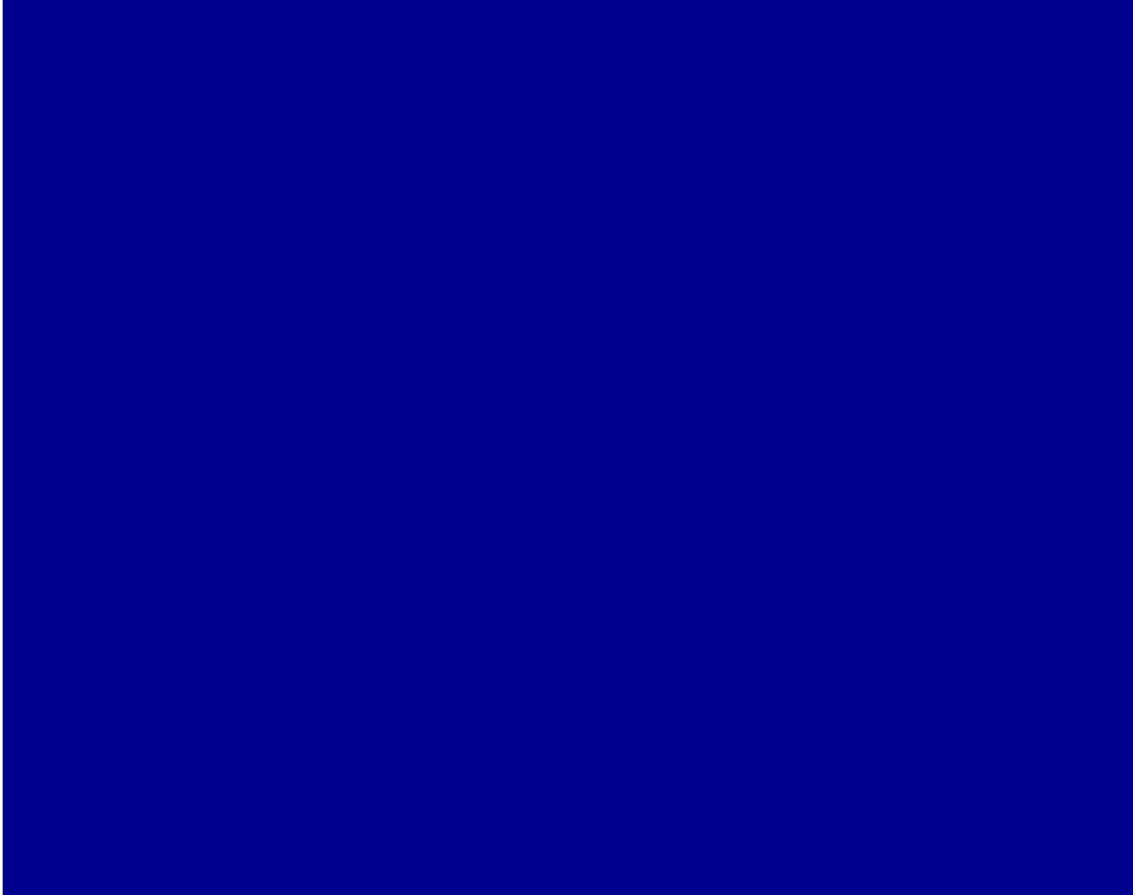} \\
\end{tabular}
\caption{From top to bottom are the rain streaks removal results, extracted rain streaks and corresponding error images by different methods on the video ``highway1'' (top 3 row) and ``highway2'' (bottom 3 row), respectively. From left to right are: the rainy data, results by TCL \cite{kim2015video}, DDN \cite{fu2017removing}, SE \cite{Wei_2017_ICCV}, MS-CSC \cite{li2018video}, FastDeRain with shift strategy and the ground truth.}
\label{rotate}
\end{center}
\end{figure*}

In this subsection, we examine the performance of our method with the shift strategy and other four methods, when the rain streaks are far away from being vertical.
We simulated two rainy videos: one is rain streaks with angles varying in $[15^\circ,35^\circ]$ added to the video ``waterfall'' (captured by a dynamic camera); another one is rain streaks with angles varying in $[35^\circ,55^\circ]$ added to the video ``highway'' (captured by a static camera).
As shown in Table \ref{QCR} and Fig. \ref{rotate},  the shift strategy helped our method to obtains the best results when dealing with the oblique rain streaks.
The superior of the proposed FastDeRain is obvious both quantitatively and visually.

\begin{table}[!htb]
\scriptsize\setlength{\tabcolsep}{1.5pt}
\renewcommand\arraystretch{0.9}
\caption{Quantitative comparisons of the rain streak removal results of \cite{kim2015video}, \cite{fu2017clearing}, \cite{Wei_2017_ICCV}, \cite{li2018video} and the proposed method with the shift strategy when rain streaks are far away from being vertical. The \textbf{best} quantitative values are in \textbf{boldface}.}
\begin{center}
\begin{tabular}{cccccccc}
\toprule
\multicolumn{8}{c}{Video: ``waterfall''\quad\quad \quad Angle: $15^\circ-35^\circ$}\\\midrule

Method &  PSNR    &  SSIM  &  FSIM  &  VIF   &  UIQI  &  GMSD  & time (s) \\\midrule
Rainy                                  & 29.14 & 0.8612 & 0.9323 & 0.5111 & 0.8228 & 0.0754 & | \\
TCL \cite{kim2015video}                & 33.55 & 0.9336 & 0.9602 & 0.6362 & 0.9110 & 0.0363 & 2929.2 \\
DDN \cite{fu2017clearing}              & 32.10 & 0.9283 & 0.9589 & 0.5984 & 0.8993 & 0.0448 & 43.8 \\
SE \cite{Wei_2017_ICCV}                & 25.27 & 0.6219 & 0.7811 & 0.3137 & 0.3844 & 0.1732 & 1028.0 \\
MS-CSC \cite{li2018video}              & 28.44 & 0.7593 & 0.8900 & 0.3876 & 0.6679 & 0.1154 & 264.3 \\
FastDeRain                             & \bf38.01 & \bf0.9701 & \bf0.9838 & \bf0.8224 & \bf0.9597 & \bf0.0138 & \bf31.5 \\
\toprule
\multicolumn{8}{c}{Video: ``highway''\quad\quad \quad Angle: $35^\circ-55^\circ$}\\\midrule
 Method & PSNR    &  SSIM  &  FSIM  &  VIF   &  UIQI  &  GMSD & time (s) \\ \midrule
Rainy                                  & 29.18 & 0.8162 & 0.9197 & 0.4865 & 0.6554 & 0.0957 & | \\
TCL \cite{kim2015video}                & 30.26 & 0.8859 & 0.9399 & 0.5460 & 0.7038 & 0.0603 & 1277.7 \\
DDN \cite{fu2017clearing}              & 28.91 & 0.8208 & 0.9126 & 0.4563 & 0.6510 & 0.0877 & 38.3 \\
SE \cite{Wei_2017_ICCV}                & 33.22 & 0.9703 & 0.9809 & 0.7944 & 0.8974 & 0.0127 & 564.8 \\
MS-CSC \cite{li2018video}              & 36.99 & 0.9747 & 0.9812 & 0.8137 & 0.9177 & 0.0137 & 182.1 \\
FastDeRain                             & \bf39.36 & \bf0.9825 & \bf0.9889 & \bf0.8801 & \bf0.9209 & \bf0.0075 &\bf24.6\\
\bottomrule
\end{tabular}
\end{center}
\label{QCR}
\end{table}

\section{Conclusion}\label{sec:Con}
We have proposed a novel video rain streaks removal approach: FastDeRain.
The proposed method, based on directional gradient priors in combination with sparsity, outperforms a series of state-of-the-art methods both visually and quantitively.
We attribute the outperforming of FastDeRain to our intensive analysis of the characteristic priors of rainy videos, clean videos and rain streaks.
Besides, it notable that our method is markedly faster than the compared methods, even including a every fast single-image-based method.
Our method is not without limitation.
The natural rainy scenario is sometimes mixed with haze, and how to handle the residual rain artifacts remains an open problem.
These issues will be addressed in the future.

\section*{Acknowledgment}
The authors would like to express their sincere thanks to the editor and referees for giving us so many valuable comments and suggestions for revising this paper.
The authors would like to thank Dr. Xueyang Fu, Dr. Wei Wei and Dr. Minghan Li for their generous sharing of their codes.
This research was supported by the National Natural Science Foundation of China (61772003, 61702083), and the Fundamental Research Funds for the Central Universities (ZYGX2016J132, ZYGX2016J129, ZYGX2016KYQD142).

{\footnotesize
\bibliographystyle{ieeetran}
\bibliography{refference}
}
\end{document}